\documentclass{article}

\usepackage{microtype}
\usepackage{graphicx}
\usepackage{subcaption}
\usepackage[noend]{algorithmic}
\usepackage{booktabs} 
\usepackage{multirow}
\usepackage{subcaption}
\usepackage{pifont}
\usepackage{wrapfig}
\usepackage{multicol}
\newcommand\myfontsize{\fontsize{8pt}{11pt}\selectfont}

\newcommand{\cmark}{\ding{51}}%
\newcommand{\xmark}{\ding{55}}%

\usepackage[table, dvipsnames]{xcolor}
\definecolor{greeny}{RGB}{0, 179, 0}
\definecolor{greenyDarky}{RGB}{0, 179, 0}
\definecolor{orangey}{RGB}{230, 115, 0}

\usepackage[backref=page]{hyperref}

\renewcommand{\algorithmiccomment}[1]{\bgroup\hfill~#1\egroup}

\newcommand{\nocontentsline}[3]{} \newcommand{\tocless}[2]{\bgroup\let\addcontentsline=\nocontentsline#1{#2}\egroup} \newcommand{\toclesslab}[3]{\bgroup\let\addcontentsline=\nocontentsline#1{#2\label{#3}}\egroup}

\usepackage[accepted]{icml2024}

\usepackage{amsmath}
\usepackage{amssymb}
\usepackage{mathtools}
\usepackage{amsthm}
\usepackage{bm}
\usepackage{mathrsfs}

\addtocontents{toc}{\protect\setcounter{tocdepth}{0}}

\usepackage[capitalize,noabbrev]{cleveref}

\theoremstyle{plain}
\newtheorem{theorem1}{Theorem}[section]

\newtheorem{definition1}[theorem1]{Definition}

\newtheorem{example1}[theorem1]{Example}
\theoremstyle{remark}

\newenvironment{example}[2]
{
    \begin{example1}\textbf{(#1)}\label{example:#2}
}
{
    \end{example1}
}

\newenvironment{definition}[2]
{
    \begin{definition1}\textbf{(#1)}\label{definition:#2}
}
{
    \end{definition1}
}

\newenvironment{theorem}[2]
{
    \begin{theorem1}\textbf{(#1)}\label{theorem:#2}
}
{
    \end{theorem1}
}

\newsavebox{\bmatrixbox}
\newenvironment{colorbmatrix}
  {\begin{lrbox}{\bmatrixbox}
   \mathsurround=0pt
   $\displaystyle
   \begin{bmatrix}}
  {\end{bmatrix}$%
   \end{lrbox}%
   \usebox{\bmatrixbox}%
   \kern-\wd\bmatrixbox
   \makebox[0pt][l]{$\left[\vphantom{\usebox{\bmatrixbox}}\right.$}%
   \kern\wd\bmatrixbox
}

\usepackage[textsize=tiny]{todonotes}

\icmltitlerunning{A Probabilistic Approach to Learning the Degree of Equivariance in Steerable CNNs}

\begin{document}

\twocolumn[
\icmltitle{A Probabilistic Approach to Learning the Degree of \\ Equivariance in Steerable CNNs}



\icmlsetsymbol{equal}{*}

\begin{icmlauthorlist}
\icmlauthor{Lars Veefkind}{uva,quva}
\icmlauthor{Gabriele Cesa}{uva,qualcomm}
\end{icmlauthorlist}

\icmlaffiliation{uva}{University of Amsterdam, The Netherlands}
\icmlaffiliation{quva}{Work done during internship in QUVA-Lab. All datasets were downloaded/generated in QUVA-Lab}
\icmlaffiliation{qualcomm}{Qualcomm AI Research, an initiative of Qualcomm Technologies, Inc}

\icmlcorrespondingauthor{Lars Veefkind}{larsveefkind@gmail.com}

\icmlkeywords{Machine Learning, ICML, Geometric Deep Learning, Equivariance, Steerable CNNs, Partial Equivariance}

\vskip 0.3in
]



\printAffiliationsAndNotice{} 




\begin{abstract}
Steerable convolutional neural networks (SCNNs) enhance task performance by modelling geometric symmetries through equivariance constraints on weights. Yet, unknown or varying symmetries can lead to overconstrained weights and decreased performance. To address this, this paper introduces a probabilistic method to learn the degree of equivariance in SCNNs. We parameterise the degree of equivariance as a likelihood distribution over the transformation group using Fourier coefficients, offering the option to model layer-wise and shared equivariance. These likelihood distributions are regularised to ensure an interpretable degree of equivariance across the network. Advantages include the applicability to many types of equivariant networks through the flexible framework of SCNNs and the ability to learn equivariance with respect to any subgroup of any compact group without requiring additional layers. Our experiments reveal competitive performance on datasets with mixed symmetries, with learnt likelihood distributions that are representative of the underlying degree of equivariance.

\end{abstract}

\vspace*{-0.5em}
\section{Introduction}
Natural images and structures often exhibit various geometrical symmetries, such as rotations, translations, and reflections. Since~\citet{lecun1995convolutional} introduced translation-equivariant Convolutional Neural Networks (CNNs), integrating \textit{group} equivariance properties as a form of inductive bias into deep learning architectures has been an active area of research~\cite{weiler2021coordinate, bronstein2021geometric}.

\begin{figure}[h!]
    \centering
    \scalebox{1}[-1]{\includegraphics[width=0.30\columnwidth, angle=270, origin=c]{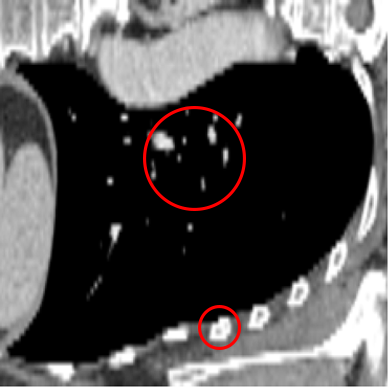}}
    \scalebox{1}[-1]{\includegraphics[width=0.30\columnwidth, angle=270, origin=c]{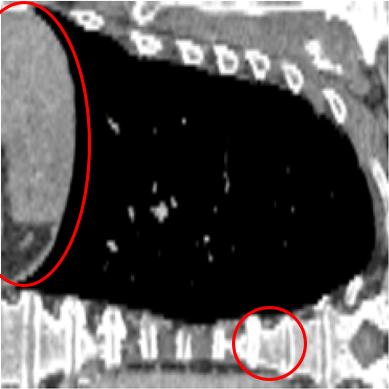}}
    \caption{Coronal CT view of left and right lung~\cite{yang2023medmnist}. Smaller features \textcolor{red}{$\bigcirc$} (i.e., textures/components) are orientation invariant in determining whether the object is a lung, unlike the task of distinguishing between the two lungs as a whole, as they are (approximately) mirrored versions of each other.}
    \label{fig:example_lungs}
\end{figure}

While incorporating higher degrees of equivariance can serve as a powerful inductive bias, over-constraining a model can lead to diminished performance by preventing the use of potentially significant features~\cite{wangApprox, 2much_equivariance_not_group}. Therefore, it is crucial to select an appropriate transformation group that is neither too large nor too small when developing a model. The required degree of equivariance often varies among features, typically correlating with their scale (Fig.~\ref{fig:example_lungs}). In a CNN, with its hierarchical layers processing features of varying sizes (receptive fields), selecting the appropriate transformation group beforehand poses a challenge, demanding substantial knowledge about the dataset and the layers' receptive fields.

To address these complexities, various strategies have been proposed to autonomously learn the relaxation of equivariance constraints on a layer-by-layer basis for Group CNNs~\cite{cohen2016group}. Among these are Residual Pathway Priors (RPPs) \cite{finzi2021residual}, non-stationary GCNNs \cite{van2022relaxing}, and Partial GCNNs \cite{romero2022learning}. We cover these methods in more detail in Sec.~\ref{sec:related_works}. 
Despite the flexibility of the framework of steerable CNNs~\cite{cohen2016steerable} compared to GCNNs, only very few methods are compatible with this framework. 
To the best of our knowledge, RSteer \cite{wangApprox} is the only method directly developed for SCNNs, while RPP provides a generalised method for modulating equivariance regardless of the network type.
However, with no separate parameterisation describing the degree of equivariance, such as a likelihood distribution over the transformation group as proposed by~\citet{romero2022learning}, it is impossible to obtain an interpretable degree of equivariance from these methods without performing computationally expensive analyses. 
Furthermore, RPPs achieve this modulation through residual connections that are non-equivariant, or equivariant with respect to a smaller subgroup of the large group.
Still, accurately choosing the relevant equivariance subgroups can be challenging: for instance, there are infinitely many $SO(2)$ subgroups in $SO(3)$. We cover these shortcomings in more details in Sec.~\ref{sec:related_works}.

\vspace*{-0.3em}
\looseness=-1
\subsection{Contributions and Outline}
Building upon the SCNN framework developed by~\citet{cesa2022ae(n)}, we propose a general solution to obtain a learnable degree of equivariance in SCNNs. Parameterised by an interpretable likelihood distribution over the transformation group, our solution models partial equivariance by using connections that are not available in an equivariant setting.

In Sec.~\ref{sec:leveraging_averaging}, we exploit the interpretation of the kernel projection as an implicit uniform averaging over the group. We show that this uniform distribution can be modelled using \textit{Fourier coefficients} in Sec.~\ref{sec:show_avg}. By updating these coefficients through backpropagation, we enable the learning of a likelihood distribution over any compact group in $n$D, including continuous groups. The learnt likelihood corresponds to the model's inherent degree of equivariance and is easily obtained with the inverse Fourier transform without forward passes, achieving an \textbf{interpretable} degree of equivariance. Moreover, to prevent a layer from erroneously regaining previously lost equivariance, we propose employing \textit{KL-divergence}~\cite{kullback1951information} between subsequent layers. Additionally, our approach allows the user to tune the \textit{band-limiting} of the Fourier coefficients as means of regularisation and improved computational performance.

To evaluate our approaches, we conduct an extensive series of experiments in Sec.~\ref{sec:experiments}, comparing our method with various baselines. These include regular SCNNs with pre-tuned degrees of equivariance through group restrictions~\cite{cesae(2)}, RPPs~\cite{finzi2021residual} and Relaxed SCNNs~\cite{wangApprox}, as these are, to our knowledge, the only approaches directly applicable to SCNNs. Our experiments assess the performance impact on various computer vision datasets. Additionally, we analyse the interpretability and correctness of the learnt likelihood distributions and study the influence of bandlimiting and KL-divergence on both performance and interpretability. Finally, we compare data-efficiency and generalisation capabilities between our approach and the baselines. Through these experiments, we offer the following contributions:
\vspace{-12pt}
\begin{itemize}
    \setlength\itemsep{-4pt}
    \item  We develop a \textit{probabilistic} approach to model partial equivariances in SCNNs and show that it achieves competitive performance on partially symmetric datasets.
    \item Our findings reveal that the learnt likelihood distributions accurately represent the degree of equivariance, enhancing interpretability.
    \item We demonstrate that bandlimiting serving as a useful tool to regularise the degree of equivariance.
    \item We show that it outperforms other methods in generalising from discrete to continuous group symmetries.
    \item We show that our approach achieves high data-efficiency on fairly symmetric datasets. 
\end{itemize}

\section{Background on Steerable CNNs}
This section briefly covers the framework of SCNNs. For a comprehensive overview of all mathematical preliminaries we refer to Apx.~\ref{chap:preliminaries}. We restrict ourselves to convolutional networks acting on Euclidean spaces $X=\mathbb{R}^n$ and compact groups $H\leq O(n)$, resulting in networks equivariant to transformations in $G = (\mathbb{R}^n, +) \rtimes H$, i.e. the \textit{semidirect product} between translations $(\mathbb{R}^n, +)$ and $H$.

\subsection{Steerable CNNs}
As the name suggests, each layer $l$ in a $G$-equivariant neural network satisfies the \textit{equivariance constraint}:
\begin{equation}
    \forall x \in X, \forall g \in G\ \! \colon \quad l(g\cdot x) = g\cdot l(x),
\end{equation}
meaning that these networks have the property to commute with the action of a symmetry group $G$ on their input, intermediate features and output. In this paper, we focus on SCNNs. These networks operate on \textit{feature fields}, which are comparable to feature maps, except that they are associated with a well defined action of $G$ and a geometric \textit{type}. This type is described by an orthogonal group \textit{representation} $\rho\! \colon H \! \to \!\mathbb{R}^{d_\rho \times d_{\rho}}$ that determines the transformation law of the $d_\rho$-dimensional feature vectors in the feature field. Specifically, a feature field of type $\rho$ is a map $f \! \colon \mathbb{R}^n \! \to \!\mathbb{R}^{d_\rho}$ that, given a group action $g=(t,h) \! \in\! (\mathbb{R}^n, +) \rtimes H$, transforms by spatially moving feature vectors from $h^{-1}(x-t)$ to $x$ and uses $\rho(h)$ to transform each individual feature\footnote{This is modelled via the \textit{induced representation} $\left(\left [ \operatorname{Ind}_H^{(\mathbb{R}^n, +) \rtimes H} \rho \right ] (th) f \right)(x) = \rho(h)f(h^{-1}(x-t))$.}.



The simplest example of feature fields are scalar fields, used in conventional CNNs. These fields transform according to the trivial representation $\psi_0(h)=1$. GCNNs are special cases and transform according to the \textit{regular representation} of $G$. As neural networks are often composed of multiple feature fields $f_i$, we often describe the entire feature field as a \textit{direct sum} $f=\bigoplus_i f_i$ of the individual feature fields, i.e. by stacking them along the channel dimension. The resulting feature space consequently transforms under the \textit{direct sum representation} $\rho = \bigoplus_i \rho_i$; see Def.~\ref{definition:directsum}.

The goal of SCNNs is to form a general framework to obtain equivariance, regardless of the types of the feature fields. To this end, it has been shown that the most general equivariant maps between feature spaces are convolutions with $H$-steerable kernels~\citet{weiler20183d}. Given input and output feature field types $\rho_{\operatorname{in}}\! \colon H \! \to \!\mathbb{R}^{d_{\operatorname{in}}\times d_{\operatorname{in}}}$ and $\rho_{\operatorname{out}}\! \colon H \! \to \!\mathbb{R}^{d_{\operatorname{out}}\times d_{\operatorname{out}}}$ respectively, an $H$-steerable kernel is a convolution kernel $K\! \colon \mathbb{R}^n \! \to \!\mathbb{R}^{d_{\operatorname{out}}\times d_{\operatorname{in}}}$ that satisfies the \textit{steerability constraint} $\forall h \in H, x \in \mathbb{R}^n$:
\begin{equation}\label{eq:kernel_constraint}
    K(x) = \rho_{\operatorname{out}}(h)K(h^{-1}x)\rho_{\operatorname{in}}(h)^{-1}.
\end{equation}
Therefore, in order to build SCNNs we need to find a basis of the vector space of such $H$-steerable kernels that can be used to parameterise the convolutions.

\subsection{Generating Steerable Bases}\label{sec:show_avg}
\citet{cesa2022ae(n)} derives a general parameterisation of $H$-steerable kernels. In this section, we offer a summary of an \emph{alternative derivation} that arrives at the same result, but with a probabilistic interpretation; see Apx.~\ref{ap:eq_constraints} for a detailed proof. In our derivation, we consider a general Euclidean space $\mathbb{R}^n$, and compact group $H \leq O(n)$ in $G=(\mathbb{R}^n, +) \rtimes H$. As such, we can assume that all representations are \textit{orthogonal}, and thus $\rho(g^{-1})=\rho(g)^{-1}=\rho(g)^\top$. 

\paragraph{Irrep Decomposition} While $\rho_{\operatorname{in}}$ and $\rho_{\operatorname{out}}$ can be any type of representation, through \textit{irrep decomposition} (Thm.~\ref{theorem:irrep_decomp}) we can decompose any orthogonal representation into a direct sum of convenient \textit{irreducible representations} (or \textit{irreps}): $\rho(h) = Q \left [\bigoplus_{{\textcolor{MidnightBlue}{i}}\in \textit{I}} {\textcolor{MidnightBlue}{\psi_i}}(h)\right]Q^T$, where ${\textcolor{MidnightBlue}{\psi_i}}$ are irreps in the set of irreps $\widehat{H}$ of $H$, $I$ is an index set ranging over $\widehat{G}$, and $Q$ is a change of basis. \citet{cesae(2)} show that the solution for any $\rho_{\operatorname{in}}$ and $\rho_{\operatorname{out}}$ can be obtained from the solution of their irreps decomposition. Therefore, we can assume the input and output representations to be irreps $\psi_l$ and $\psi_J$. Furthermore, it is convenient to use the vectorised form of our kernel: $\kappa(\cdot)=\operatorname{vec}\left(K(\cdot)\right)\!\colon X\!\to\!\mathbb{R}^{d_{\operatorname{in}} \cdot d_{\operatorname{out}}}$. Finally, as the constraint is linear, it can be solved by projecting unconstrained square integrable kernel $\widehat{\kappa}\in L^2(\mathbb{R}^n)$ via integration; a similar idea was used in~\cite{van2020mdp}. As this is essentially performing a uniform averaging over the group $H$, we can safely inject a uniform likelihood distribution $\mu(h)=1 \ \forall h\in H$, assuming $H$ is a compact subgroup of a group with a finite Haar measure. Thus\footnote{$\operatorname{vec}(ABC)=(C^\top \otimes A)\operatorname{vec}(B)$, with Kronecker product~$\otimes$.}:
\begin{equation}
    \kappa(x) = \int_{h\in H} \mu(h) (\psi_l \otimes \psi_J)(h) \widehat{\kappa}(h^{-1}x) dh.\label{eq:constraint_1}
\end{equation}\
\paragraph{Tensor Product}\! $(\psi_l \otimes \psi_J)$ is a \emph{tensor product representation} (Def. \ref{definition:tensor_rep}). It can be decomposed into a sum of irreps ${\textcolor{violet}{\psi_{j'}}}$ via \textit{Clebsch-Gordan decomposition} (Thm.~\ref{theorem:clebsch_gordan}):
\begin{equation}
    \psi_l \otimes \psi_J = \sum_{{\textcolor{violet}{j'}}}\sum_s^{[{\textcolor{violet}{j'}}(Jl)]} \left [ {\textcolor{violet}{\operatorname{CG}_s^{{\textcolor{violet}{j'}}(Jl)}}} \right]^\top {\textcolor{violet}{\psi_{j'}}} {\textcolor{violet}{\operatorname{CG}_s^{{\textcolor{violet}{j'}}(Jl)}}}.\label{eq:tp}
\end{equation}$[{\textcolor{violet}{j'}}(Jl)]$ is the \textit{multiplicity} of irrep ${\textcolor{violet}{\psi_{j'}}}$. ${\textcolor{violet}{\operatorname{CG}_s^{{\textcolor{violet}{j'}}(Jl)}}}$ are the \textit{Clebsch-Gordan coefficients} for the $s$-th occurrence of ${\textcolor{violet}{\psi_{j'}}}$.

\paragraph{Steerable Basis} An \emph{unconstrained} kernel $\widehat{\kappa}$ can be parameterised via 
weights $W_{\textcolor{red}{j},\textcolor{red}{k}} \in \mathbb{R}^{d_Jd_{l}\times d_{{\textcolor{red}{j}}}}$ and
an $H$-steerable basis~(Def.~\ref{definition:steerable_basis}) $\mathcal{B}=\left\{\textcolor{red}{Y}_{\textcolor{red}{j}}^{\textcolor{red}{k}}: \mathbb{R}^n \to \mathbb{R}^{d_{{\textcolor{red}{j}}}} \mid {\textcolor{red}{\psi_j}} \in \widehat{H}, \textcolor{red}{k}\right\}$:
\begin{equation}
    \widehat{\kappa}(x) = \sum_{\textcolor{red}{j},\textcolor{red}{k}} W_{\textcolor{red}{j},\textcolor{red}{k}} \textcolor{red}{Y}_{\textcolor{red}{j}}^{\textcolor{red}{k}}(x).
\end{equation}
Let $\operatorname{unvec}(\cdot)$ be the reverse of the vectorisation.
We apply this parameterisation and the \textit{Clebsch-Gordan decomposition} in Eq.~\ref{eq:constraint_1}, and substitute $W_{\textcolor{red}{j}, {\textcolor{violet}{j'}}, \textcolor{red}{k}, s} = \operatorname{vec}\left( {\textcolor{violet}{\operatorname{CG}_s^{{\textcolor{violet}{j'}}(Jl)}}} W_{\textcolor{red}{j}\textcolor{red}{k}}\right)$:
\begin{equation}
\begin{split}
    \kappa(&x) = \sum_{\textcolor{red}{j}, \textcolor{red}{k}}\sum_{{\textcolor{violet}{j'}}} \sum_s^{[{\textcolor{violet}{j'}}(Jl)]}  \left [ {\textcolor{violet}{\operatorname{CG}_s^{{\textcolor{violet}{j'}}(Jl)}}} \right]^\top \\
    &\operatorname{unvec}\! \left [ \int\! \mu(h) \left({\textcolor{red}{\psi_j}}\! \otimes\! {\textcolor{violet}{\psi_{j'}}}\right)\!(h) dh \ W_{\textcolor{red}{j}, {\textcolor{violet}{j'}}, \textcolor{red}{k}, s} \right]\! \textcolor{red}{Y}_{\textcolor{red}{j}}^{\textcolor{red}{k}}(x)\label{eq:cnn_int_to_replace}
\end{split}            
\end{equation}
Here the integral contains a tensor product, and can thus be decomposed through the \textit{Clebsch-Gordan decomposition}:
\begin{equation}
\begin{split}
    \int \mu(h) & \left({\textcolor{red}{\psi_j}} \otimes  {\textcolor{violet}{\psi_{j'}}}\right)(h)  dh =\\  &Q \bigoplus_{{\textcolor{MidnightBlue}{i}}} \bigoplus_{r}^{[{\textcolor{MidnightBlue}{i}}(\textcolor{red}{j}{\textcolor{violet}{j'}})]}\left (\int \mu(h) {\textcolor{MidnightBlue}{\psi_i}}(h) dh\right)Q^\top\label{eq:after_decompose_GG}
\end{split}
\end{equation}

\paragraph{Peter Weyl and inverse Fourier Transform}
The \textit{Peter-Weyl theorem}~\cite{peter1927vollstandigkeit} (Thm.~\ref{theorem:peter_weyl}) states that the entries of \textit{complex} irreps of a compact group $H$ span the vector space $L^2(H)$. 
However, in this work we consider \textit{real} irreps, which may contain redundant entries.
A proper basis can be constructed by selecting an independent subset of $n_{\textcolor{MidnightBlue}{i}}=\frac{d_{\textcolor{MidnightBlue}{i}}}{[0({\textcolor{MidnightBlue}{ii}})]}$ columns of each irrep ${\textcolor{MidnightBlue}{\psi_i}}$
\footnote{
This construction only works under a proper choice of basis for ${\textcolor{MidnightBlue}{\psi_i}}$; $[0({\textcolor{MidnightBlue}{ii}})]\in \{1, 2, 4\}$ depending on the type of irrep ${\textcolor{MidnightBlue}{\psi_i}}$. See \cite{cesa2022ae(n)} Apx. C.3 for more details.
}, where $[0({\textcolor{MidnightBlue}{ii}})]$ is the multiplicity of the trival irrep $\psi_0$ in ${\textcolor{MidnightBlue}{\psi_i}} \otimes {\textcolor{MidnightBlue}{\psi_i}}$, as in Eq.~\ref{eq:tp}.
We denote the non-redundant columns of ${\textcolor{MidnightBlue}{\psi_i}}(h)$ with $\overline{\textcolor{MidnightBlue}{\psi}}_{\textcolor{MidnightBlue}{i}}(h)\in\mathbb{R}^{d_{\textcolor{MidnightBlue}{i}}\times n_{\textcolor{MidnightBlue}{i}}}$ and define the map  ${\textcolor{MidnightBlue}{\mathcal{R}_{\psi_i}}}: \mathbb{R}^{d_{\textcolor{MidnightBlue}{i}}\times n_{\textcolor{MidnightBlue}{i}}}\to\mathbb{R}^{d_{\textcolor{MidnightBlue}{i}}\times d_{\textcolor{MidnightBlue}{i}}}$ that performs the inverse projection.

Using ${\textcolor{MidnightBlue}{\mathcal{R}_{\psi_i}}}$, we can describe $\mu(h)$ through its inverse Fourier Transform~(IFT):
\begin{equation}
    \mu(h) = \sum_{{\textcolor{MidnightBlue}{\psi_i}} \in \widehat{H}} \sqrt{d_{\textcolor{MidnightBlue}{i}}}\text{Tr}\left ({\textcolor{MidnightBlue}{\psi_i}}(h)^\top {\textcolor{MidnightBlue}{\mathcal{R}_{\psi_i}}}\left(\widehat{\mu}(\overline{\textcolor{MidnightBlue}{\psi}}_{\textcolor{MidnightBlue}{i}})\right) \right)
\end{equation}
Due to the orthogonality of the Fourier coefficients, by expanding $\mu$ via its IFT, the integral in Eq.~\ref{eq:after_decompose_GG} simplifies to:
\small
\begin{equation}
        \int\! \mu(h)\! \left({\textcolor{red}{\psi_j}}\! \otimes\! {\textcolor{violet}{\psi_{j'}}}\right)(h) dh\! =\! Q \bigoplus_{{\textcolor{MidnightBlue}{i}}}\! \bigoplus_{r}^{[{\textcolor{MidnightBlue}{i}}(\textcolor{red}{j}{\textcolor{violet}{j'}})]}\frac{{\textcolor{MidnightBlue}{\mathcal{R}_{\psi_i}}}\left(\widehat{\mu}(\overline{\textcolor{MidnightBlue}{\psi}}_{\textcolor{MidnightBlue}{i}})\right)}{\sqrt{d_{\textcolor{MidnightBlue}{i}}}}Q^\top\label{eq:start_substitute}
\end{equation}
\normalsize
As $\mu(h)$ is uniform, only the Fourier coefficients $\widehat{\mu}(\psi_0)=1$ corresponding to the \textit{trivial} representation $\psi_0$ with size $d_{\psi_0}=1$ are non-zero. 
As such, the summation selects only certain relevant columns $\{Q_r = \operatorname{CG}_r^{0(\textcolor{red}{j}{\textcolor{violet}{j'}})} \}_r^{[0(\textcolor{red}{j}{\textcolor{violet}{j'}})]}$ from $Q$:
\begin{equation}
        \int\! \mu(h)\! \left({\textcolor{red}{\psi_j}}\! \otimes\! {\textcolor{violet}{\psi_{j'}}}\right)(h) dh = \sum_{r}^{[0(\textcolor{red}{j}{\textcolor{violet}{j'}})]} Q_rQ_r^\top.\label{eq:int_final}
\end{equation}
One can verify\footnote{
    Eq.~\ref{eq:int_final} essentially defines the orthogonal projection of a (vectorised) matrix into the space of (vectorised) matrices commuting with ${\textcolor{red}{\psi_j}}$ and ${\textcolor{violet}{\psi_{j'}}}$.
} that the set $\{\operatorname{unvec}(Q_r)\}_r^{[0(\textcolor{red}{j}{\textcolor{violet}{j'}})]}$ forms a basis for all equivariant maps from ${\textcolor{red}{\psi_j}}$ to ${\textcolor{violet}{\psi_{j'}}}$.
Hence, whenever ${\textcolor{red}{\psi_j}} \neq {\textcolor{violet}{\psi_{j'}}}$, $[0({\textcolor{red}{j}}{\textcolor{violet}{j'}})] = 0$  (\emph{Schur's Lemma} Apx.~\ref{theorem:schur}); in this case, Eq.~\ref{eq:int_final} results in a null-map and the corresponding weights can be discarded.
Then, Eq.~\ref{eq:cnn_int_to_replace} turns into:
\small
\begin{equation}
\begin{split}
    \kappa(x)\! =\! \sum_{\textcolor{red}{j}, \textcolor{red}{k}}\! \sum_s^{[\textcolor{red}{j}(Jl)]}\! \left [ {\textcolor{red}{\operatorname{CG}_s^{j(Jl)}}} \right]^\top\! \operatorname{unvec} \left [ \sum_{r}^{[0(\textcolor{red}{j}\textcolor{red}{j})]} c_r^{\textcolor{red}{j}} W_{\textcolor{red}{j}, \textcolor{red}{k}, s,r} \right]\!  \textcolor{red}{Y}_{\textcolor{red}{j}}^{\textcolor{red}{k}}(x),\label{eq:eq_final}
\end{split}            
\end{equation}
\normalsize
where $Q_r^\top \in \mathbb{R}^{1\times d_{\textcolor{red}{j}} d_{{\textcolor{violet}{j'}}}}$ and $W_{\textcolor{red}{j}, {\textcolor{violet}{j'}}, \textcolor{red}{k}, s}$ have been merged into a single weight $W_{\textcolor{red}{j}, \textcolor{red}{k}, s, r}$ to reduce parameters. 

\section{Learning Equivariance}
This section details how we obtain an interpretable and learnable degree of equivariance in SCNNs.
\subsection{Leveraging the Averaging Operator}\label{sec:leveraging_averaging}
In SCNNs, equivariance stems from the uniform averaging over the group $H$ performed during the projection of the unconstrained kernels, ensuring that each element $h\in H$ is equally weighted. To learn the degree of equivariance, we propose to replace the uniform likelihood distribution $\mu(h)$ with a learnable likelihood distribution ${\textcolor{YellowOrange}{\lambda}}(h)$. As a result, we substitute $\mu(h)$ with ${\textcolor{YellowOrange}{\lambda}}(h)$ in Eq.~\ref{eq:start_substitute}:
\small
\begin{equation}\label{eq:projection_1}
        \int\! {\textcolor{YellowOrange}{\lambda}}(h)\! \left({\textcolor{red}{\psi_j}}\! \otimes\! {\textcolor{violet}{\psi_{j'}}}\right)(h) dh\! =\! Q \bigoplus_{{\textcolor{MidnightBlue}{i}}}\! \bigoplus_{r}^{[{\textcolor{MidnightBlue}{i}}(\textcolor{red}{j}{\textcolor{violet}{j'}})]}\frac{{\textcolor{MidnightBlue}{\mathcal{R}_{{\textcolor{MidnightBlue}{\psi_i}}}}}\left(\widehat{{\textcolor{YellowOrange}{\lambda}}}(\overline{\textcolor{MidnightBlue}{\psi}}_{\textcolor{MidnightBlue}{i}})\right)}{\sqrt{d_{\textcolor{MidnightBlue}{i}}}}Q^\top
\end{equation}
\normalsize
While the uniformity of $\mu(h)$ meant only the Fourier coefficient of the trivial representation had to be considered, ${\textcolor{YellowOrange}{\lambda}}(h)$ is not necessarily uniform. As such, the Fourier coefficient of any representation ${\textcolor{MidnightBlue}{\psi_i}}$ can be nonzero, and the previous simplifications can no longer be performed. Thus, using weights $W_{\textcolor{red}{j}, {\textcolor{violet}{j'}}, \textcolor{red}{\textcolor{red}{k}}, s}\in \mathbb{R}^{d_{\textcolor{red}{j}}d_{{\textcolor{violet}{j'}}}}$ the final projection becomes\footnote{Note here that the summation over ${\textcolor{violet}{j'}}$ has returned, as the projection no longer amounts to a nullmap if ${\textcolor{red}{\psi_j}}\neq{\textcolor{violet}{\psi_{j'}}}$.}:
\small
\begin{equation}
   \kappa(x) = \sum_{\textcolor{red}{j}, \textcolor{red}{\textcolor{red}{k}}} \sum_{{\textcolor{violet}{j'}}, s}  \left [ {\textcolor{violet}{\operatorname{CG}_s^{{\textcolor{violet}{j'}}(Jl)}}} \right]^\top \operatorname{unvec} \left [ c^{\textcolor{red}{j}{\textcolor{violet}{j'}}} W_{\textcolor{red}{j}, {\textcolor{violet}{j'}}, \textcolor{red}{\textcolor{red}{k}}, s} \right] {\textcolor{red}{Y}}_{\textcolor{red}{j}}^{\textcolor{red}{\textcolor{red}{k}}}(x),\label{eq:prob_cnn_final}
\end{equation}
\normalsize
where $c^{\textcolor{red}{j}{\textcolor{violet}{j'}}}$ is defined using the Fourier coefficients:
\begin{equation}
    c^{\textcolor{red}{j}{\textcolor{violet}{j'}}} = Q \left (\bigoplus_{{\textcolor{MidnightBlue}{i}}}\! \bigoplus_{r}^{[{\textcolor{MidnightBlue}{i}}(\textcolor{red}{j}{\textcolor{violet}{j'}})]}\frac{{\textcolor{MidnightBlue}{\mathcal{R}_{{\textcolor{MidnightBlue}{\psi_i}}}}}\left(\widehat{{\textcolor{YellowOrange}{\lambda}}}(\overline{\textcolor{MidnightBlue}{\psi}}_{\textcolor{MidnightBlue}{i}})\right)}{\sqrt{d_{\textcolor{MidnightBlue}{i}}}}\right)\label{eq:projection}.
\end{equation}

To learn the degree of equivariance we propose to initialise the Fourier coefficients such that ${\textcolor{YellowOrange}{\lambda}}(h)$ is a uniform distribution, and the projection is initially fully $H$-equivariant:
\begin{equation}\label{eq:initialise_fourier}
        \widehat{{\textcolor{YellowOrange}{\lambda}}}({\textcolor{MidnightBlue}{\psi_i}}) = \begin{cases}
            1 & {\textcolor{MidnightBlue}{i}}=0\\
            \bm{0}_{d_{\textcolor{MidnightBlue}{i}} \times {n_{\textcolor{MidnightBlue}{i}}}} & {\textcolor{MidnightBlue}{i}}\neq 0
    \end{cases}.
\end{equation}
These Fourier coefficients $\widehat{{\textcolor{YellowOrange}{\lambda}}}({\textcolor{MidnightBlue}{\psi_i}})$ are stored as learnable parameters and updated through back-propagation during training, permitting a partially $H$-equivariant projection\footnote{In practice, only the Fourier coefficients of irreps appearing in one of the Clebsch-Gordan decompositions need to be considered.}. 

\subsection{Ensuring Interpretable Equivariance}\label{sec:ensure_eq}
Through IFT it is possible to construct the underlying likelihood distribution ${{\textcolor{YellowOrange}{\lambda}}}(h)$ from the corresponding learnt Fourier coefficients $\widehat{{{\textcolor{YellowOrange}{\lambda}}}}({\textcolor{MidnightBlue}{\psi_i}})$. However, a few steps are required to ensure interpretable likelihoods.
\paragraph{Normalisation} To ensure consistency in the magnitude of the likelihood distribution we normalise the learnt likelihood distribution ${{\textcolor{YellowOrange}{\lambda}}}$ to a PDF. This improves interpretability, and also ensures that any potential weight regularisation cannot simply uniformly increase the scale of the learnt likelihood in favour of increasing the kernel weights. To achieve this, we apply the Softmax non-linearity to ensure ${{\textcolor{YellowOrange}{\lambda}}}$'s non-negativity and normalisation $\int {{\textcolor{YellowOrange}{\lambda}}}(h) dh = 1$:
\begin{equation}\label{eq:normalise}
    \bm{\widehat{{{\textcolor{YellowOrange}{\lambda}}}}} = \operatorname{FT} \left(\sigma\left(\operatorname{IFT} \left(\bm{\widehat{{{\textcolor{YellowOrange}{\lambda}}}'}}\right)\right)\right),
\end{equation}
where $\bm{\widehat{{{\textcolor{YellowOrange}{\lambda}}}}}$ is the stack of Fourier coefficients of the density ${{\textcolor{YellowOrange}{\lambda}}}$, $\bm{\widehat{{{\textcolor{YellowOrange}{\lambda}}}'}}$ the stack of learnable Fourier coefficients of its logits, and $\sigma(\cdot)$ is the Softmax non-linearity. As such, the following is computed for each $h$ in a discrete sampling set $\mathcal{H} \subset H$ of $N$ elements, using $z_{{\textcolor{YellowOrange}{\lambda}}} = \frac{1}{N} \sum_{n=1}^{N} e^{{{\textcolor{YellowOrange}{\lambda}}}'(n) - \max({{\textcolor{YellowOrange}{\lambda}}}')}$\footnote{We subtract the max value to make softmax numerically stable.}:
\begin{equation}\label{eq:normalisationn}
    {{\textcolor{YellowOrange}{\lambda}}}(h) = \sigma({{\textcolor{YellowOrange}{\lambda}}}'(h)) = \frac{e^{{{\textcolor{YellowOrange}{\lambda}}}'(h) - \max({{\textcolor{YellowOrange}{\lambda}}}')}}{ z_{{\textcolor{YellowOrange}{\lambda}}}}.
\end{equation}
While other normalisation methods are also valid~\footnote{E.g., the use Parseval's identity might be a straightforward method to normalise the Fourier coefficients without sampling.}, we opted for Softmax normalisation as the $z$-term and logits can be used directly to compute the KL-divergence described later in this section.
\paragraph{Likelihood Alignment} While the Fourier parameterisation enables the reconstruction of the likelihood distribution ${{\textcolor{YellowOrange}{\lambda}}}$, there is no guarantee that the weights are properly aligned with the learnt likelihood distribution. We show in Apx.~\ref{ap:alignment_loss} that it is possible for the model to freely shift the likelihood distribution by shifting the weights accordingly, diminishing the interpretability of the distributions.

To address this concern, we propose constraining the Fourier coefficients by aligning the maximum likelihood with the identity element $e \in H$. The theoretical justification lies in the intrinsic impossibility of violating equivariance when the identity element is applied. 
Hence, we desire:
\begin{equation} \label{eq:align_constraint}
\forall h \in H \colon {{\textcolor{YellowOrange}{\lambda}}}(h) \leq {{\textcolor{YellowOrange}{\lambda}}}(e)
\end{equation}

Since directly enforcing this constraint on the Fourier coefficients is hard, we opt for a soft alignment constraint:
\begin{equation}
        D_{\text{align}}({{\textcolor{YellowOrange}{\lambda}}}) = \operatorname{max}({{\textcolor{YellowOrange}{\lambda}}}) - {{\textcolor{YellowOrange}{\lambda}}}(e). \label{eq:align_loss}
\end{equation}
This  error can be added as a loss term during training. Provided that the likelihood ${{\textcolor{YellowOrange}{\lambda}}}$ has been sampled using a sufficient number of samples $N$, this error term is equal to zero if Eq.~\ref{eq:align_constraint} holds, and greater than zero otherwise.

While this regularisation is rather sparse, as only the largest likelihood contributes to the loss, we found that it is sufficient in ensuring proper alignment and therefore did not consider less sparse approaches.

\paragraph{Regaining Equivariance} Once a layer breaks the equivariance, this equivariance cannot be accurately regained in subsequent layers. Since our approach allows independent parameterisations of the distributions in subsequent layers, the learnt distributions do not necessarily describe such a monotonically decreasing equivariance. To this end, we propose to employ KL-divergence as regularisation between the layers. As the goal is to ensure that layer $n+1$ does not regain equivariance previously lost by layer $n$, we treat layer $n$ as reference distribution for layer $n+1$. To prevent layer $n+1$ from influencing layer $n$, we detach the likelihood of layer $n$ and treat it as constant (see Apx~\ref{sec:monotonically} for more details on the potential advantage of this detaching, and on monotonically decreasing equivariance in general). Given the likelihoods ${{\textcolor{YellowOrange}{\lambda}}}_n$ and ${{\textcolor{YellowOrange}{\lambda}}}_{n+1}$, we show in Apx.~\ref{ap:kl_divergence} that KL-divergence can be computed using the $z$-terms and Fourier coefficients of the logits and normalised distribution:
\begin{equation}
\begin{split}
     &D_{KL}({{\textcolor{YellowOrange}{\lambda}}}_{n+1}  \ || \  {{\textcolor{YellowOrange}{\lambda}}}_n)  =\bm{\widehat{{{\textcolor{YellowOrange}{\lambda}}}_{n+1}}}^\top  \bm{\widehat{{{\textcolor{YellowOrange}{\lambda}}}_{n+1}'}} - \operatorname{max}({{\textcolor{YellowOrange}{\lambda}}}_{n+1}')  \\ & - \log z_{{\textcolor{YellowOrange}{\lambda}}_{n+1}} - \bm{\widehat{{{\textcolor{YellowOrange}{\lambda}}}_{n+1}}}^\top \  \bm{\widehat{{{\textcolor{YellowOrange}{\lambda}}}_n'}} + \operatorname{max}({{\textcolor{YellowOrange}{\lambda}}}_n') + \log z_{{\textcolor{YellowOrange}{\lambda}}_n}. \label{eq:final_kl}
\end{split}
\end{equation} 

\subsection{Bandlimiting}\label{sec:bandlimit}
While our approach allows for a flexible and interpretable degree of equivariance, it also significantly increases the number of learnable parameters. In the equivariant setting in~Eq.~\ref{eq:eq_final}, for each irrep ${\textcolor{red}{\psi_j}}$ from the steerable basis, the kernel is parameterised by $[0({\textcolor{red}{jj}})]$ single weights $W_{{\textcolor{red}{j}},\textcolor{red}{k},s,r}$. Instead, our projection in Eq.~\ref{eq:prob_cnn_final} is parameterised by a vector of size $d_{\textcolor{red}{j}}d_{\textcolor{violet}{j'}} =\sum_{\textcolor{MidnightBlue}{i}}[{\textcolor{MidnightBlue}{i}}({\textcolor{red}{j}}{\textcolor{violet}{j'}})] \cdot d_{\textcolor{MidnightBlue}{i}}$ for each pair of ${{\textcolor{red}{\psi_j}}}$ and ${\textcolor{violet}{\psi_{j'}}}$. 

Recall the composition of the partially equivariant projection matrix $c^{{\textcolor{red}{j}}{\textcolor{violet}{j'}}}$ in Eq.~\ref{eq:projection}. The size of the matrix is dependant on the number of irreps ${\textcolor{MidnightBlue}{\psi_i}}$ appearing in the tensor product ${\textcolor{red}{\psi_j}}\otimes{\textcolor{violet}{\psi_{j'}}}$. 
For example, if we consider the group $SO(3)$, whose irreps are indexed by their rotational frequency, ${\textcolor{red}{\psi_j}}\otimes{\textcolor{violet}{\psi_{j'}}}$ contains all irreps ${\textcolor{MidnightBlue}{\psi_i}}$ with frequencies in the range ${\textcolor{MidnightBlue}{i}} \in |{\textcolor{red}{j}} - {\textcolor{violet}{j'}}|, \dots, {\textcolor{red}{j}} + {\textcolor{violet}{j'}}$.
As such, we propose to reduce the number of parameters through a tunable bandlimit $L$ on the maximum frequency of irreps ${\textcolor{MidnightBlue}{\psi_i}}$, reducing the parameter space and the flexibility of the distributions. 

We note that band-limiting is \textit{not strictly required} to obtain a reasonable number of parameters compared to other partially equivariant approaches. Instead, it is an \textit{optional tool} to reduce computational complexity and filter higher frequencies as a form of regularisation (see Apx.~\ref{sec:result_bandlimit})

\subsection{Implementation} 
Our implementation\footnote{We will publish the implementation at \url{https://github.com/QUVA-Lab/partial-escnn} and integrate it into the \texttt{escnn} library \url{https://github.com/QUVA-Lab/escnn}.} relies on the \texttt{escnn} library~\cite{cesa2022ae(n)}, providing an intuitive environment for configuring SCNNs with compact groups $H$ and obtaining the corresponding $H$-steerable basis $\mathcal{B}$. For our solution, we adapt Algorithm 1 in \cite{cesa2022ae(n)}, which we present in Algorithm~\ref{algo:algo}. An important consideration is that the \texttt{escnn} library is optimised for equivariant SCNNs. As such, our implementation on top of this library is likely sub-optimal. E.g., unlike in regular SCNNs acting on a static space $X$, our current implementation resamples the steerable basis after each forward pass, which may not be necessary.


\begin{algorithm*}[t!]
   \caption{Generate a learnable equivariant $H$-steerable basis on space $X$}
   \label{algo:algo}
   \centering
\begin{algorithmic}[1]
\footnotesize
   \REQUIRE $\rho_{\text{in}}=\psi_l$ and $\rho_{\text{out}}=\psi_J$, $H$-steerable basis $\mathcal{B}=\left \{ {\textcolor{red}{Y_{j}^{\textcolor{red}{k}}}}\right\}_{{\textcolor{red}{j}}}^{\textcolor{red}{k}}$,
   $\widehat{H}=\left\{{\textcolor{red}{\psi_j}} \right\}_{\textcolor{red}{j}},
   \ x\in X, L\in\mathbb{N}^+$, stack of logit coefficients $\bm{\widehat{{\textcolor{YellowOrange}{\lambda}}'}}$.
   \STATE $\textcolor{violet}{\operatorname{CG}^{lJ}}$, $\left\{[{\textcolor{violet}{j'}}(Jl)]\right\}_{{\textcolor{violet}{j'}}} \leftarrow$ decompose$(\psi_l \otimes \psi_J)$ \COMMENT{Eq.~\ref{eq:tp} (Thm.~\ref{theorem:clebsch_gordan})}
   \FORALL{${\textcolor{red}{Y_{j}^{\textcolor{red}{k}}}} \in \mathcal{B}$}
   \FORALL{ ${\textcolor{violet}{j'}} \in \widehat{H}: [{\textcolor{violet}{j'}}(Jl)] > 0, s \leq [{\textcolor{violet}{j'}}(Jl)]$}
   \STATE  $ \left\{ \widehat{{\textcolor{YellowOrange}{\lambda}}}(\overline{\textcolor{MidnightBlue}{\psi}}_{\textcolor{MidnightBlue}{i}}) \right\}_{\textcolor{MidnightBlue}{i}} \leftarrow$ unstack$\left(\operatorname{FT} \left(\sigma\left(\operatorname{IFT} \left(\bm{\widehat{{\textcolor{YellowOrange}{\lambda}}'}}\right)\right)\right)\right)$ \COMMENT{Eq.~\ref{eq:normalisationn}}
   \STATE $Q, \{[{\textcolor{MidnightBlue}{i}}({\textcolor{red}{j}}{\textcolor{violet}{j'}})]\}_i \leftarrow \operatorname{decompose}({\textcolor{red}{\psi_j}} \otimes {\textcolor{violet}{\psi_{j'}}})$ \COMMENT{Eq. \ref{eq:after_decompose_GG} (Thm. \ref{theorem:clebsch_gordan})}
   \STATE $Q \leftarrow$ bandlimit$(Q, L)$ \COMMENT{Sec.~\ref{sec:bandlimit}}
    \STATE $c^{\textcolor{red}{j}{\textcolor{violet}{j'}}} \leftarrow Q \cdot$ diagonal$\left(\operatorname{repeat}\left(\left\{ \widehat{{\textcolor{YellowOrange}{\lambda}}}(\overline{\textcolor{MidnightBlue}{\psi}}_{\textcolor{MidnightBlue}{i}}) \right\}_i\right)\right)$ \COMMENT{Repeats $[{\textcolor{MidnightBlue}{i}}({\textcolor{red}{j}}{\textcolor{violet}{j'}})]$ times each ${\textcolor{MidnightBlue}{\mathcal{R}_{{\textcolor{MidnightBlue}{\psi_i}}}}}\left(\widehat{\textcolor{YellowOrange}{\lambda}}(\overline{\textcolor{MidnightBlue}{\psi}}_{\textcolor{MidnightBlue}{i}})\right)$ and diagonalises result (Eq.~\ref{eq:projection})}
   \STATE$K_{{\textcolor{red}{j}}{\textcolor{red}{k}}{\textcolor{violet}{j'}}s}(x) \leftarrow \operatorname{unvec}\left(\left[\textcolor{violet}{\operatorname{CG}_s^{j'(Jl)}}\right]^\top \cdot c^{{\textcolor{red}{j}}{\textcolor{violet}{j'}}} \cdot {\textcolor{red}{Y_{{\textcolor{red}{j}}}^{\textcolor{red}{k}}}}(x)\right)$ \COMMENT{Eq.~\ref{eq:prob_cnn_final}}  
   \ENDFOR
   \ENDFOR
\end{algorithmic}
\end{algorithm*}

\section{Related Work}\label{sec:related_works}

Group CNNs (GCNNs)~\cite{cohen2016group}, extend the application of the regular convolution operator to the domain of a group to obtain group equivariance using discrete groups. Steerable CNNs (SCNNs)~\cite{cohen2016steerable} provide a more general theory of equivariance, offering more flexibility, the ability to model continuous groups and some computational advantages for large groups.
Previous works proposed other strategies to construct equivariance, see for example \cite{worrall2017harmonic,kondor2018generalization,bekkers2019b,finzi2020generalizing}

Several methods have recently been proposed to obtain varying degrees of equivariance in equivariant NNs. \citet{cesae(2)} offer a manual solution via group restrictions. Other works allow the degree of equivariance to be learnt by the network. Partial GCNNs~\cite{romero2022learning} parameterise a learnable probability distribution over the elements of the subgroup \( H \) of $G$. Rather than uniformly sampling group elements as in traditional GCNNs, Partial GCNNs use this learnt probability distribution to guide the sampling of group elements, achieving a form of partial \( H \)-equivariance as a result. Non-stationary GCNNs~\cite{van2022relaxing} utilised non-stationary kernels to achieve partial $G$-equivariance. Proposed by \citet{finzi2021residual}, Residual Pathway Priors (RPP) offer a more versatile strategy that can be applied across a variety of equivariant networks. In RPPs, each intermediate feature field comprises (at least) two layers: one that maintains $G$-equivariance and another with softer constraints, with a predetermined prior controlling the contribution. Based on non-stationary GCNNs and RPPs, \citet{new_equivariance} took a more unified approach and employed Bayesian model selection using differentiable Laplace approximations. They learnt layer-wise equivariances as hyper-parameters, offering a framework for discovering symmetries at different scales. \citet{dehmamy2021automatic} define group convolution in terms of the Lie-algebra: by learning the algebra basis, their model is able to learn the required symmetries. \citet{wang2022equivariantQ} use lift expansion to achieve partial equivariance for $Q$ learning by concatenating encoded non-equivariant features to encoded equivariant features in an encoder-decoder setting. \citet{wangApprox} introduce RSteer and RGroup, SCNN and GCNN based approaches that break equivariance through a regularised function mapping between group element and partially equivariant weights. Finally, \citet{maile2023equivarianceaware} propose an equivariance-aware neural architecture search method to discover equivariance during training.

To learn the equivariance with respect to a specific subgroup $S$ of $H$, RPPs require a residual connection for each of these subgroups. This is computationally infeasible, as it would require an infinite number of residual connections for continuous groups. Furthermore, enumerating such subgroups is complicated for 3D rotation groups. Specifically, the collection of rotations around any arbitrary 3D axis constitutes a subgroup of $SO(3)$. However, sampling uniformly over the space of 3D axes is not straightforward and requires a large number of samples and, therefore, additional residual layers. Thus, RPPs are unfeasible for learning equivariance to an unknown subgroup $S$ of a large or continuous group~$H$. 

Like RPPs, our approach models partial equivariance as a combination of equivariant and non-equivariant mappings. 
Indeed, RPP is comparable to our method under a \textit{specific and non-learnable} choice of distribution $\lambda$.
Furthermore, similar to Partial GCNNs, our approach uses a learnt likelihood over the group as a sampling mechanism. Whereas Partial GCNNs directly sample this likelihood to perform group convolutions, our solution implicitly integrates the learnt likelihood within the SCNN framework.
Through this likelihood, we can model equivariance with respect to any subgroup $S$ of $H$, or even subsets $S$ that are not proper groups, without additional layers. 
See Apx.~\ref{sec:comp_approaches} for an in-depth comparison between these approaches.

\section{Experiments and Evaluation}\label{sec:experiments}
\begin{table*}[t!]
\centering
\resizebox{0.71\textwidth}{!}{%
\begin{tabular}{ll|cccccc}
\toprule
\multirow{2}{*}{\begin{tabular}[l]{@{}c@{}}\textbf{Network}\\ \textbf{Group}\end{tabular}} &
  \multirow{2}{*}{\begin{tabular}[c]{@{}c@{}}\textbf{Partial}\\ \textbf{Equivariance}\end{tabular}} &
  \multicolumn{6}{c}{\textbf{Symmetries for individual digits}} \\
 &   &  $\bm{C_1}$ &  $\bm{C_4}$ &  $\bm{SO(2)}$ &  $\bm{D_1}$ &  $\bm{D_4}$ &  $\bm{O(2)}$ \\ \midrule \midrule
CNN &  N/A &  $\bm{\underline{0.962}} \ $\myfontsize{$( 0.002)$} &  $\underline{0.868} \ $\myfontsize{$( 0.011)$} &  $\underline{0.807} \ $\myfontsize{$( 0.007)$} &  $\underline{0.919} \ $\myfontsize{$( 0.006)$} &  $\underline{0.711} \ $\myfontsize{$( 0.009)$} &  $\underline{0.649} \ $\myfontsize{$( 0.019)$} \\ \midrule
\multirow{4}{*}{$C_4$} &  None&  $0.933 \ $\myfontsize{$( 0.002)$} &  $0.484 \ $\myfontsize{$( 0.008)$} &  $0.459 \ $\myfontsize{$( 0.007)$} &  $0.893 \ $\myfontsize{$( 0.005)$} &  $0.427 \ $\myfontsize{$( 0.008)$} &  $0.405 \ $\myfontsize{$( 0.008)$} \\
 &
  Restriction&  $\underline{0.954} \ $\myfontsize{$( 0.003)$} &  $0.911 \ $\myfontsize{$( 0.006)$} &  $0.877 \ $\myfontsize{$( 0.009)$} &  $\underline{0.928} \ $\myfontsize{$( 0.006)$} &  $0.827 \ $\myfontsize{$( 0.013)$} &  $0.776 \ $\myfontsize{$( 0.018)$} \\
 &
  RPP &  $0.937 \ $\myfontsize{$( 0.006)$} &  $0.901 \ $\myfontsize{$( 0.012)$} &  $0.867 \ $\myfontsize{$( 0.025)$} &  $0.899 \ $\myfontsize{$( 0.014)$} &  $0.821 \ $\myfontsize{$( 0.023)$} &  $0.772 \ $\myfontsize{$( 0.009)$} \\
 &
  Ours &  $0.947 \ $\myfontsize{$( 0.006)$} &  $\underline{0.916} \ $\myfontsize{$( 0.005)$} &  $\underline{0.891} \ $\myfontsize{$( 0.006)$} &  $0.923 \ $\myfontsize{$( 0.007)$} &  $\underline{0.848} \ $\myfontsize{$( 0.007)$} &  $\underline{0.795} \ $\myfontsize{$( 0.011)$} \\ \midrule
\multirow{4}{*}{$D_4$} &
  None&  $0.895 \ $\myfontsize{$( 0.005)$} &  $0.439 \ $\myfontsize{$( 0.010)$} &  $0.396 \ $\myfontsize{$( 0.009)$} &  $0.473 \ $\myfontsize{$( 0.010)$} &  $0.431 \ $\myfontsize{$( 0.010)$} &  $0.394 \ $\myfontsize{$( 0.008)$} \\
 &
  Restriction&  $\underline{0.953} \ $\myfontsize{$( 0.004)$} &  $0.912 \ $\myfontsize{$( 0.007)$} &  $\underline{0.887} \ $\myfontsize{$( 0.003)$} &  $\underline{0.930} \ $\myfontsize{$( 0.007)$} &  $0.827 \ $\myfontsize{$( 0.007)$} &  $0.773 \ $\myfontsize{$( 0.009)$} \\
 &
  RPP  &  $0.934 \ $\myfontsize{$( 0.007)$} &  $0.888 \ $\myfontsize{$( 0.014)$} &  $0.867 \ $\myfontsize{$( 0.014)$} &  $0.895 \ $\myfontsize{$( 0.007)$} &  $0.821 \ $\myfontsize{$( 0.020)$} &  $0.775 \ $\myfontsize{$( 0.013)$} \\
 &
  Ours &  $0.949 \ $\myfontsize{$( 0.005)$} &  $\underline{\bm{0.922}} \ $\myfontsize{$( 0.007)$} &  $0.885 \ $\myfontsize{$( 0.012)$} &  $0.921 \ $\myfontsize{$( 0.008)$} &  $\underline{0.848} \ $\myfontsize{$( 0.011)$} &  $\underline{0.801} \ $\myfontsize{$( 0.009)$} \\ \midrule
\multirow{4}{*}{SO(2)} &
  None&  $0.936 \ $\myfontsize{$( 0.005)$} &  $0.485 \ $\myfontsize{$( 0.010)$} &  $0.474 \ $\myfontsize{$( 0.016)$} &  $0.890 \ $\myfontsize{$( 0.006)$} &  $0.430 \ $\myfontsize{$( 0.010)$} &  $0.403 \ $\myfontsize{$( 0.021)$} \\
 &
  Restriction&
  $0.949 \ $\myfontsize{$( 0.002)$} &  $0.911 \ $\myfontsize{$( 0.010)$} &  $0.893 \ $\myfontsize{$( 0.009)$} &  $0.928 \ $\myfontsize{$( 0.003)$} &  $0.841 \ $\myfontsize{$( 0.012)$} &  $0.796 \ $\myfontsize{$( 0.011)$} \\
 &
  RPP  &  $0.935 \ $\myfontsize{$( 0.008)$} &  $0.890 \ $\myfontsize{$( 0.005)$} &  $0.870 \ $\myfontsize{$( 0.011)$} &  $0.899 \ $\myfontsize{$( 0.009)$} &  $0.821 \ $\myfontsize{$( 0.022)$} &  $0.779 \ $\myfontsize{$( 0.021)$} \\
 &
  Ours &  $\underline{0.953} \ $\myfontsize{$( 0.004)$} &  $\underline{\bm{0.922}} \ $\myfontsize{$( 0.005)$} &  $\underline{\bm{0.901}} \ $\myfontsize{$( 0.005)$} &  $\bm{\underline{0.932}} \ $\myfontsize{$( 0.005)$} &  $\underline{\bm{0.863}} \ $\myfontsize{$( 0.009)$} &  $\underline{\bm{0.823}} \ $\myfontsize{$( 0.005)$} \\ \midrule
\multirow{4}{*}{O(2)} &
  None&  $0.881 \ $\myfontsize{$( 0.005)$} &  $0.415 \ $\myfontsize{$( 0.008)$} &  $0.391 \ $\myfontsize{$( 0.012)$} &  $0.461 \ $\myfontsize{$( 0.012)$} &  $0.424 \ $\myfontsize{$( 0.009)$} &  $0.399 \ $\myfontsize{$( 0.014)$} \\
 &
  Restriction&
  $0.953 \ $\myfontsize{$( 0.005)$} &  $0.914 \ $\myfontsize{$( 0.005)$} &  $\underline{0.894} \ $\myfontsize{$( 0.005)$} &  $\underline{0.928} \ $\myfontsize{$( 0.005)$} &  $0.845 \ $\myfontsize{$( 0.011)$} &  $0.799 \ $\myfontsize{$( 0.008)$} \\
 &
  RPP  &  $0.931 \ $\myfontsize{$( 0.005)$} &  $0.891 \ $\myfontsize{$( 0.003)$} &  $0.861 \ $\myfontsize{$( 0.013)$} &  $0.891 \ $\myfontsize{$( 0.004)$} &  $0.824 \ $\myfontsize{$( 0.009)$} &  $0.772 \ $\myfontsize{$( 0.019)$} \\
 &
  Ours &  $\underline{0.958} \ $\myfontsize{$( 0.003)$} &  $\underline{0.919} \ $\myfontsize{$( 0.006)$} &  $\underline{0.894} \ $\myfontsize{$( 0.004)$} &  $0.927 \ $\myfontsize{$( 0.004)$} &  $\underline{0.859} \ $\myfontsize{$( 0.011)$} &  $\underline{0.819} \ $\myfontsize{$( 0.010)$}\\
  \bottomrule
\end{tabular}}
    \caption{Test accuracies on various \texttt{DDMNIST} data symmetries. The first column indicates the equivariance group, where CNN is a regular CNN. The second column indicates the method used to break the equivariance. For each symmetry, the highest accuracy for each network group is \underline{underlined}, with the highest overall accuracy in \textbf{bold}. Standard deviations over $5$ runs are in parentheses.}
    \label{tab:double_mnist}
\end{table*}
We explore the capabilities of our approach in terms of overall performance, generalisability, data-efficiency and interpretability on 2D and 3D classification tasks using various groups $H\leq O(n)$ (See Apx.~\ref{sec:model_architectures}). In our 2D experiments we use \texttt{DDMNIST}, our adapted double-digit version of \texttt{MNIST}~\cite{mnist}. Here, the individual digits in each sample can be randomly and independently transformed during training according to a chosen subgroup; see Apx.~\ref{ap:ddmnist}. For our 3D experiments we use the \texttt{Organ}, \texttt{Synapse} and \texttt{Nodule} subsets from the biomedical collection \texttt{MedMNIST}~\cite{yang2023medmnist}, as these datasets vary in their symmetries. In these experiments, we use 7-layer CNNs, SCNNs, SCNNs with a \textit{group restriction} to no equivariance in the last two layers, RPPs and our Partial SCNNs (PSCNNs). Each (partially) equivariant model outputs \textit{trivial} irreps in the final convolution layer to obtain structural invariance. See Apx.~\ref{sec:model_architectures} for more details.


To provide a broader comparison between our approach and alternative approaches, we perform experiments on the \texttt{JetFlow} and \texttt{Smoke} 2D simulation datasets with partial rotational symmetries as employed by \citet{wang2023general}, where they compare their approach to a collection of partial, fully and non-equivariant approaches. For our experiments we simply port their best RSteer configuration as reported by their repository, to our approach without additional fine-tuning. See~\cite{wang2023general} and Apx.~\ref{sec:simulation datasets}~\&~\ref{sec:model_architectures} for more details on the setup and alternative approaches.

All our models are trained with the following objective:
\begin{equation}
    \mathcal{L}_\text{total} = \mathcal{L}_{\text{task}} + \alpha_{\text{align}} \mathcal{L}_{\text{align}} + \alpha_{\text{KL}} \mathcal{L}_{\text{KL}}.
\end{equation}
$\mathcal{L}_{\text{KL}}$ and $\mathcal{L}_{\operatorname{align}}$ are the mean average KL-divergence between consecutive layer pairs, and alignment loss for each layer, with tunable weights $\alpha_{\text{KL}}$ and $\alpha_{\text{align}}$. Apx.~\ref{ap:additional_experiments} provides additional experiments and \textit{ablation studies} showing that these two terms are necessary for interpretable likelihoods.
\subsection{Benchmarks}\label{sec:benchmarks_main}

\begin{figure}[h!]
    \centering
    \begin{subfigure}[t]{0.325\columnwidth}
        \includegraphics[width=\textwidth, height=\textwidth]{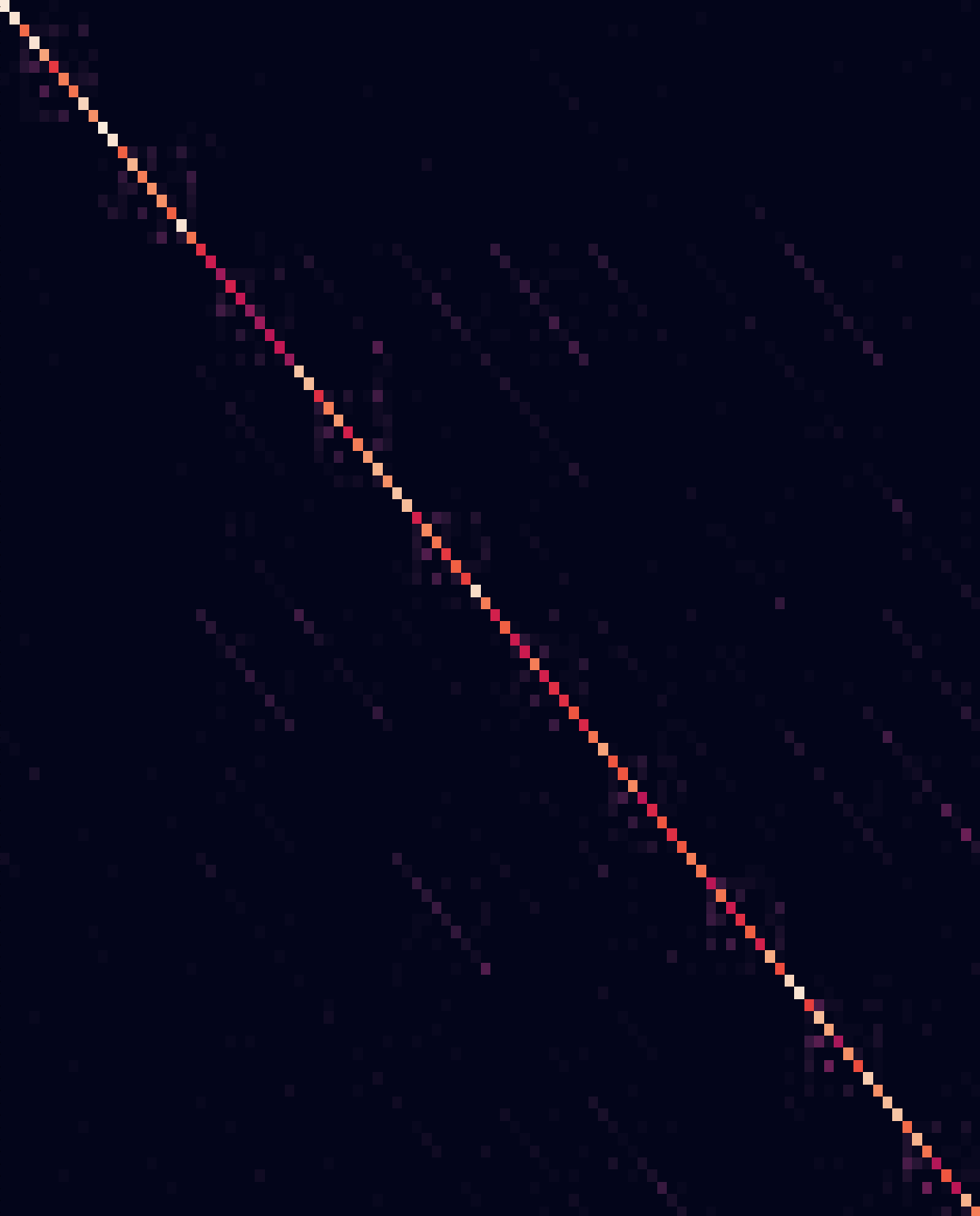}
    \caption{CNN}
    \label{fig:confuse_cnn}
    
    \end{subfigure}
    \begin{subfigure}[t]{0.325\columnwidth}
        \includegraphics[width=\textwidth, height=\textwidth]{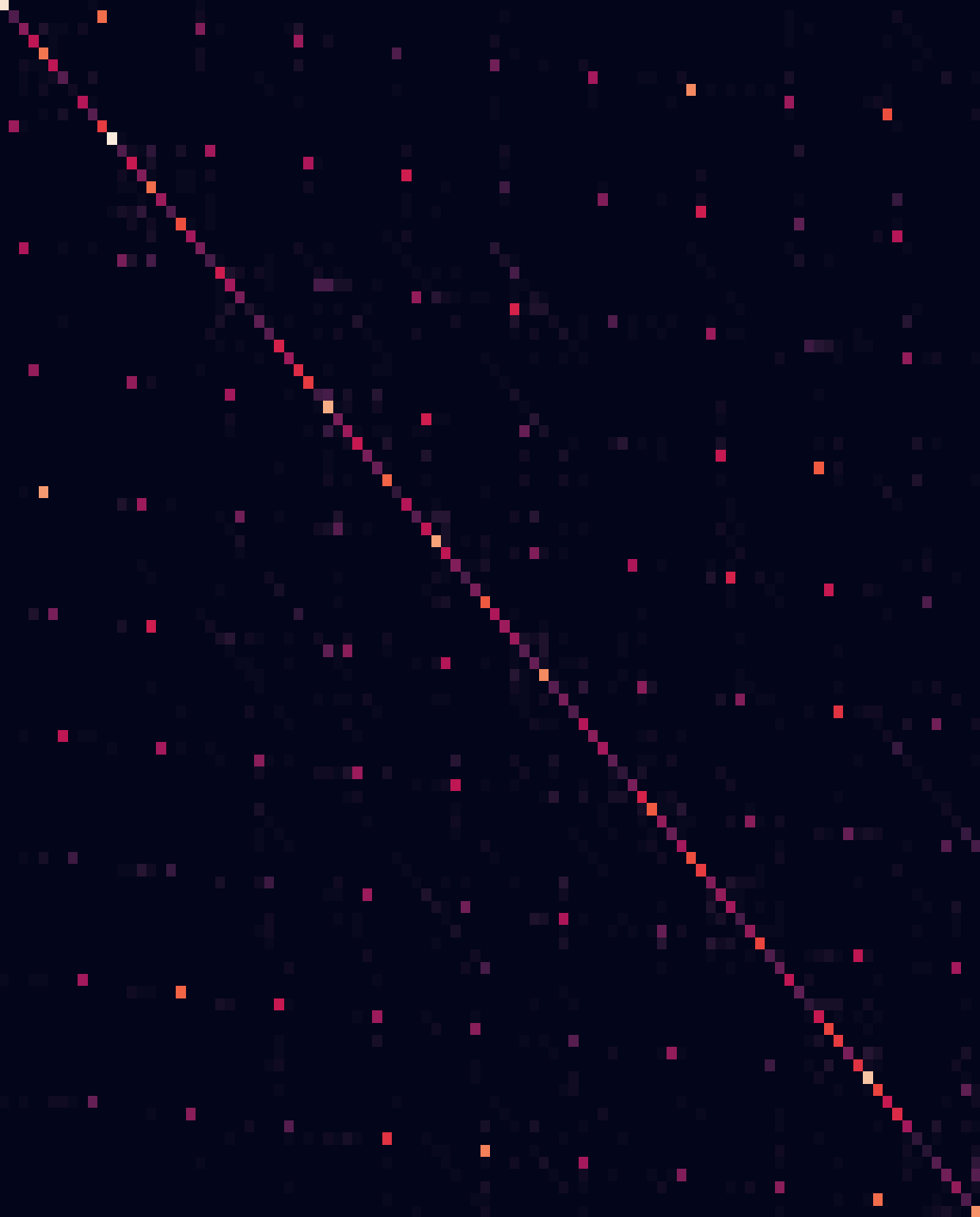}
    \caption{O(2) SCNN}
    \label{fig:confuse_o2}
    \end{subfigure}
    \begin{subfigure}[t]{0.325\columnwidth}
        \includegraphics[width=\textwidth, height=\textwidth]{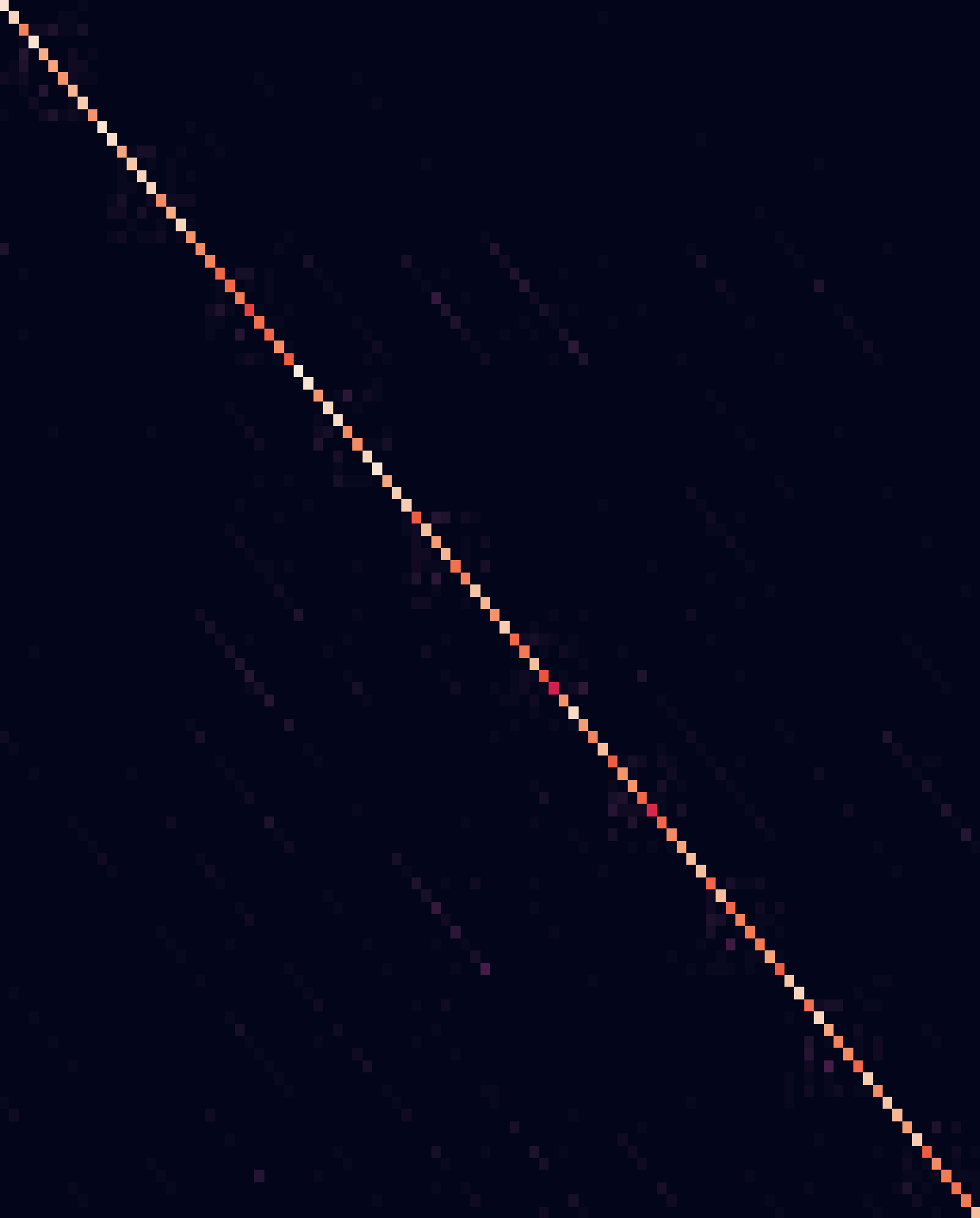}
        \caption{Our~O(2)~PSCNN}
        \label{fig:confuse_ours}
    
    \end{subfigure}    
    \caption{Confusion matrices for \texttt{DDMNIST} with $O(2)$ symmetries. Labelled 0-99 from top to bottom and left to right.}
    \label{fig:confusion_double}
    \vspace{-0.3cm}
\end{figure}

\paragraph{DDMNIST}
Tab.~\ref{tab:double_mnist} contains the classification accuracies for \texttt{DDMNIST}. For all symmetries and choices of $H$, non-equivariant CNNs outperform fully $H$-invariant SCNNs. The performance gap is relatively small for $C_1$ symmetries, but significantly larger for other symmetries. However, employing any of the partially equivariant methods results in improved performance compared to the regular CNN for all symmetries except \(C_1\). The confusion matrices in Fig.~\ref{fig:confuse_cnn} show that the invariant $O(2)$ SCNN is unable to distinguish between digit reversals (e.g. $37$ and $73$). It seems that our $O(2)$ PSCNN manages to eliminate these confusions by breaking equivariance. Furthermore, our approach, particularly the $SO(2)$ variant, outperforms the \textit{restricted} SCNN and RPP methods across each data symmetry in most cases. Finally, taking a group that is too large --e.g. taking $H=O(2)$ rather than $SO(2)$ for $D_1$ data symmetries--  results in a higher drop in performance for the fully equivariant SCNN than for the partially equivariant approaches. 

\begin{table}[t!]
\centering
\resizebox{0.9\columnwidth}{!}{%
\begin{tabular}{ll|ccc}
\toprule
\begin{tabular}[c]{@{}c@{}}\textbf{Network}\\ \textbf{Group}\end{tabular} & \begin{tabular}[c]{@{}c@{}}\textbf{Partial}\\ \textbf{Equivariance}\end{tabular} & \textbf{Nodule} & \textbf{Synapse} & \textbf{Organ} \\ \midrule\midrule
CNN                     & N/A                           & $\bm{\underline{0.873}} \ $\myfontsize{$( 0.005)$} & $\underline{0.716} \ $\myfontsize{$( 0.008)$}      & $\underline{0.920} \ $\myfontsize{$( 0.003)$} \\ \midrule
\multirow{3}{*}{$SO(3)$}& None     & $\bm{\underline{0.873}} \ $\myfontsize{$( 0.002)$} & $0.738 \ $\myfontsize{$( 0.009)$}                  & $0.607 \ $\myfontsize{$( 0.006)$} \\
                        & RPP   & $0.801 \ $\myfontsize{$( 0.003)$}                  & $0.695 \ $\myfontsize{$( 0.037)$}                  & $\underline{0.936} \ $\myfontsize{$( 0.002)$} \\
                        & Ours                          & $0.871 \ $\myfontsize{$( 0.001)$}                  & $\bm{\underline{0.770}} \ $\myfontsize{$( 0.030)$} & $0.902 \ $\myfontsize{$( 0.006)$} \\ \midrule
\multirow{3}{*}{$O(3)$} & None     & $0.868 \ $\myfontsize{$( 0.009)$}                  & $0.743 \ $\myfontsize{$( 0.004)$}                  & $0.592 \ $\myfontsize{$( 0.008)$} \\
                        & RPP   & $0.810 \ $\myfontsize{$( 0.013)$}                  & $0.722 \ $\myfontsize{$( 0.023)$}                  & $\bm{\underline{0.940}} \ $\myfontsize{$( 0.006)$} \\
                        & Ours                          & $\bm{\underline{0.873}} \ $\myfontsize{$( 0.008)$} & $\underline{0.769} \ $\myfontsize{$( 0.005)$}      & $0.905 \ $\myfontsize{$( 0.004)$}\\
 \bottomrule
\end{tabular}}
\caption{Test accuracies on \texttt{MedMNIST} datasets.}
\label{tab:med_results}
\vspace{-0.5cm}
\end{table}

\paragraph{MedMNIST}
Tab.~\ref{tab:med_results} contains the classification accuracies for \texttt{MedMNIST}. On \textbf{Nodule}, the performance differences are negligible between CNNs, SCNNs and our PSCNNs, with RPPs lagging behind. On \textbf{Synapse} the performance differences are larger, with PSCNNs significantly outperforming all models, while regular SCNNs outperform both CNNs and RPPs. Finally, on \textbf{Organ} the invariant SCNN is significantly outperformed by all other models, with RPPs achieving the highest performance. In Apx.~\ref{ap:med_result} we show that the poor SCNN performance is largely due to confusions between left and right versions of some organs. Overall, these results suggest that \textbf{Nodule} is highly symmetric, \textbf{Synapse} is fairly symmetric --while still containing relevant non-invariant features-- and \textbf{Organ} is highly non-symmetric.
\paragraph{Smoke and JetFlow}
Tab. \ref{tab:smoke_jetflow_results} presents parameter counts and test RMSE scores for \texttt{JetFlow} and \texttt{Smoke} using two methodologies; \textit{Future} denotes testing on the same simulation as the trainset, but in different time-steps in the simulation while \textit{domain} corresponds with testing at the same time-step but in different simulations. All equivariant models use $C4$-equivariance, requiring no band-limiting for our approach (See Apx.~\ref{sec:result_bandlimit} for band-limiting experiments). Our approach uses fewer parameters than MLP and the partially equivariant RSteer, Lift, and RPP approaches, with more parameters compared to CNN, Combo, RGroup and \texttt{e2cnn}. Furthermore, our approach outperforms all other approaches on the \texttt{Smoke} dataset, particularly using the \textit{domain} testset. On \texttt{JetFlow} our approach performs on-par with RSteer, while outperforming Lift and \texttt{e2cnn}.

\begin{table}[h!]
\centering
\resizebox{\columnwidth}{!}{%
\begin{tabular}{l|ccc|ccc}
\toprule
 & \multicolumn{3}{c|}{\textbf{Smoke}} & \multicolumn{3}{c}{\textbf{JetFlow}} \\
\textbf{Model} & Future & Domain & Params (M) &  Future & Domain & Params (k) \\ \midrule\midrule
MLP & 1.38 \myfontsize{$( 0.06)$}  & 1.34 \myfontsize{$( 0.03)$} & 8.33* & -  & - & 510* \\
CNN & 1.21 \myfontsize{$( 0.01)$}  & 1.10 \myfontsize{$( 0.05)$} & 0.25* & -  & - & 10* \\
\texttt{e2cnn} & 1.05 \myfontsize{$( 0.06)$}  & 0.76 \myfontsize{$( 0.02)$} & 0.62* & 0.21 \myfontsize{$( 0.02)$}  & 0.27 \myfontsize{$( 0.03)$} & 21* \\
\midrule
RPP & 0.96 \myfontsize{$( 0.10)$}  & 0.82 \myfontsize{$( 0.01)$} & 4.36* & 0.16* \myfontsize{$( 0.01)$}  & 0.19* \myfontsize{$( 0.01)$} & 145* \\
Combo & 1.07 \myfontsize{$( 0.00)$}  & 0.82 \myfontsize{$( 0.02)$} & 0.53* & -  & -  & 19* \\
CLCNN & 0.96 \myfontsize{$( 0.05)$}  & 0.84 \myfontsize{$( 0.10)$} & - & -  & -  & - \\
Lift & 0.82 \myfontsize{$( 0.01)$}  & 0.73 \myfontsize{$( 0.02)$} & 3.32* & 0.18 \myfontsize{$( 0.02)$}  & 0.21 \myfontsize{$( 0.04)$} & 479* \\
RGroup & 0.82 \myfontsize{$( 0.01)$}  & 0.73 \myfontsize{$( 0.02)$} & 1.88* & -  & - & 63* \\
RSteer & 0.80 \myfontsize{$( 0.00)$}  & 0.67 \myfontsize{$( 0.01)$} & 5.60* & 0.17 \myfontsize{$( 0.01)$}  & \textbf{0.16} \myfontsize{$( 0.01)$} & 185* \\ \midrule
Ours & \textbf{0.77} \myfontsize{$( 0.01)$}  & \textbf{0.57} \myfontsize{$( 0.00)$} & 3.12 & \textbf{0.15} \myfontsize{$( 0.00)$}  & 0.17 \myfontsize{$( 0.01)$} & 105 \\ \bottomrule
\end{tabular}}
\caption{RMSE \textit{future} and \textit{domain} test scores on \texttt{Smoke} and \texttt{JetFlow} datasets. Adapted from Tab.~1~and~2 from~\cite{wangApprox}. * Indicates our replication attempt, see Apx.~\ref{sec:model_architectures}.}
\label{tab:smoke_jetflow_results}
\vspace{-0.5cm}
\end{table}




\subsection{Inspecting the Learnt Likelihood Distributions} Following sections are best viewed in colour. To differentiate between elements in $O(2)$ we use colour coding: \textcolor{blue}{blue} signifies a basic rotation, and \textcolor{red}{red} a reflection and rotation.

\begin{figure}[h!]
    \centering
    \begin{subfigure}{\columnwidth}
    \includegraphics[width=0.89\linewidth, trim={0cm 0cm 21cm 15.5cm},clip]{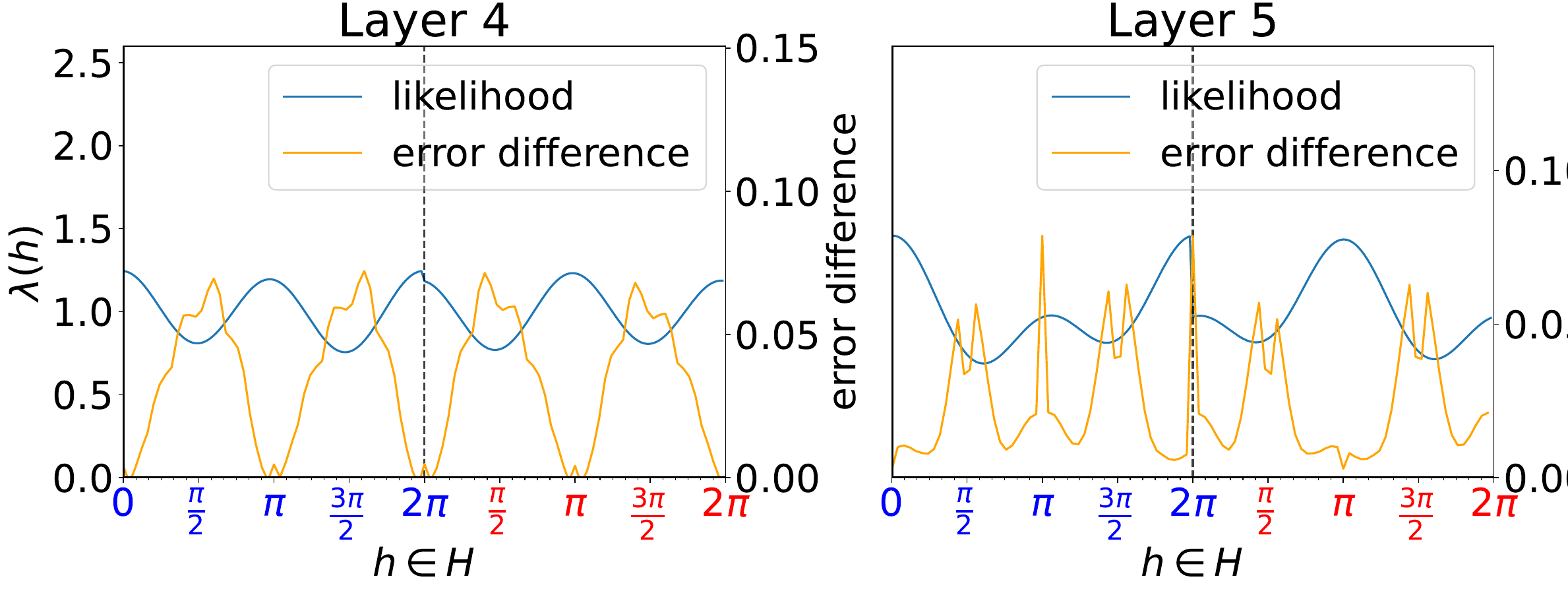}
    \caption{$O(2)$ symmetries}
    \label{fig:o2_on_o2}        
    \end{subfigure}
    \begin{subfigure}{\columnwidth}
    \includegraphics[width=0.89\linewidth, trim={0 0cm 21cm 15.5cm},clip]{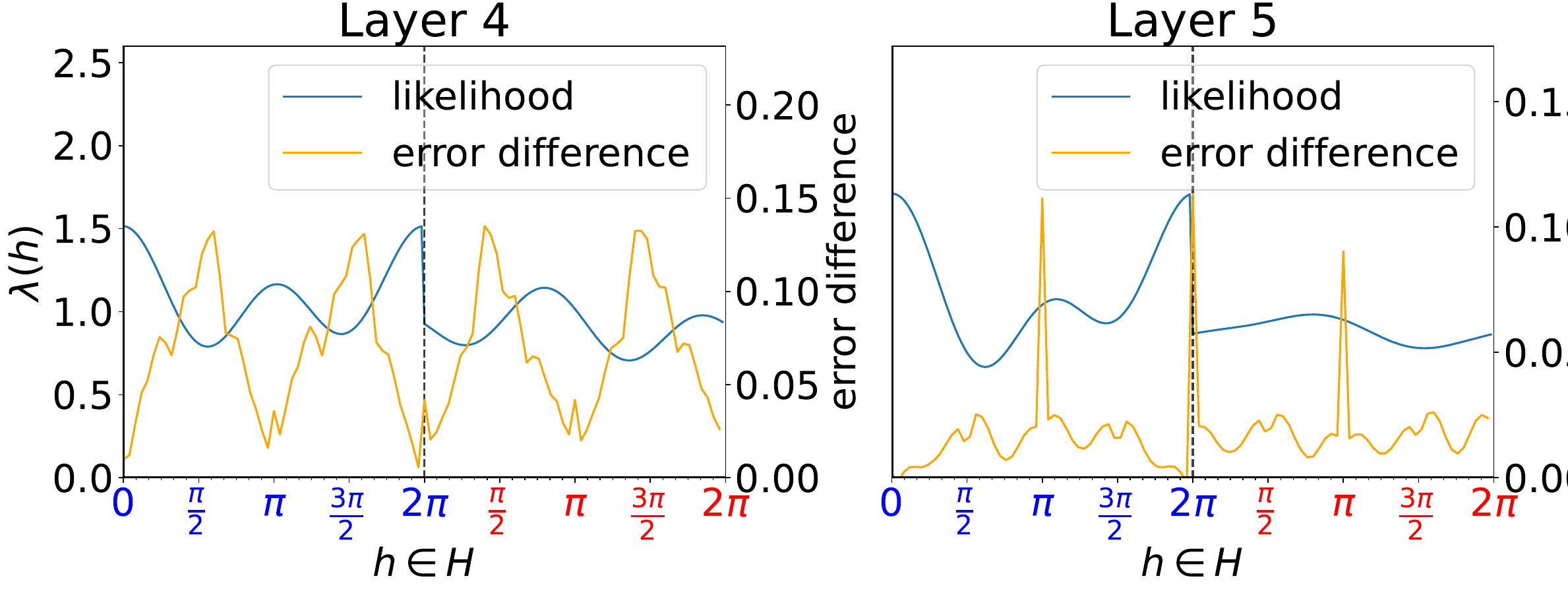}
    \caption{$C_1$ symmetries (i.e. no symmetries)}
    \label{fig:o2_on_C_1}   
    \end{subfigure}
    \caption{Learnt likelihood $\lambda$ and error difference for layers 4 and 5 of our $O(2)$ PSCNN trained on \texttt{DDMNIST} with $O(2)$ and $C_1$ symmetries. Dotted line marks the $O(2)$ reflection domain transition. Note the scale of the error between the plots.}\label{fig:double_error_vs_learnt}
    \vspace{-0.3cm}
\end{figure}

Fig.~\ref{fig:double_error_vs_learnt} shows the resulting likelihoods and corresponding equivariance error\footnote{The error is computed as the difference between the error vector norms between our model and a fully invariant $O(2)$-SCNN.} for layers 4 and 5 of our $O(2)$ PSCNN trained on \texttt{DDMNIST} with $O(2)$ and $C_1$ symmetries. Firstly, we observe that a lower likelihood results in a higher equivariance error. Next, in layer 4 equivariance is maintained for \textcolor{blue}{$\pi$} for $O(2)$ symmetries, but not $C_1$ symmetries. Most notably, on $O(2)$ symmetries a high likelihood appears at \textcolor{red}{$\pi$} in layer 5. As this is one of the latest layers in the model, this corresponds to a global transformation. Indeed, a global rotation and reflection effectively swap the digits twice, so this is an example of correct global equivariance.
\begin{figure}[h!]
    \centering
    \includegraphics[trim={0 0 19cm 0}, clip, height=3cm]{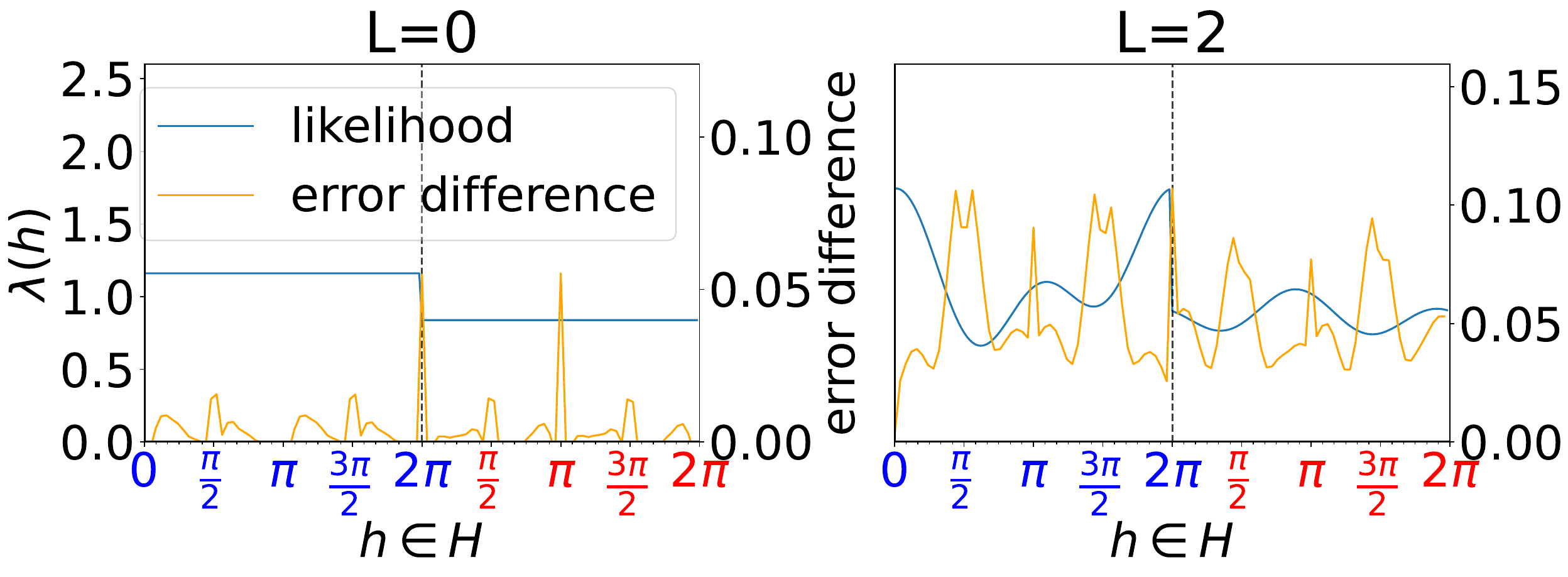}
    \includegraphics[trim={0 0 57.7cm 0}, clip, height=3cm]{figures/bandlimit.pdf}
    \includegraphics[trim={42.6cm 0 0 0}, clip, height=3cm]{figures/bandlimit.pdf}
    \vspace{-0.2cm}
    \caption{Likelihoods and errors of the fifth $O(2)$ PSCNN layer trained on $SO(2)$ \texttt{DDMNIST} under various bandlimits $L$.}
    \label{fig:bandlimit}
    \vspace{-0.5cm}
\end{figure}
\subsection{Investigating the Effect of Bandlimiting}
Tab.~\ref{tab:bandlimit} shows the classification accuracies on \texttt{DDMNIST} of our $O(2)$ PSCNN with various levels of bandlimitting compared to an invariant SCNN. Here, we observe that while all PSCNN models outperform the SCNN models, higher values for $L$ achieve higher results. However, there seems to be diminishing returns on this dataset for $L>2$. Notably, a bandlimiting of $L=0$ already yields a significant performance improvement compared to the SCNN on $SO(2)$ symmetries, as this allows the model to reduce equivariance for reflections that do not appear in $SO(2)$. Fig.~\ref{fig:bandlimit} shows the likelihoods for the fifth layer of our $O(2)$ PSCNN trained on $SO(2)$ symmetries using $L\in\{0, 2, 3\}$. Apx.~\ref{sec:result_bandlimit} contains more  band-limiting results, including the effects on the parameter space, showing that band-limiting is not necessarily required to obtain a reasonable number of parameters.

\begin{table}[h!]
    \centering
    \resizebox{\columnwidth}{!}{%
        \begin{tabular}{l|c|ccccc}
            \toprule
            & \multicolumn{5}{c}{\textbf{Bandlimit $L$}} \\
            $G$ & \textbf{SCNN} & \textbf{0} & \textbf{1} & \textbf{2} & \textbf{3} & \\ \midrule\midrule
            $SO(2)$ & $0.391 \ $\myfontsize{$( 0.012)$} & $0.469 \ $\myfontsize{$( 0.010)$} & $\underline{0.894} \ $\myfontsize{$( 0.011)$} & $\underline{0.894} \ $\myfontsize{$( 0.004)$} & $0.889 \ $\myfontsize{$( 0.013)$}  \\
            $O(2)$ & $0.399 \ $\myfontsize{$( 0.014)$} & $0.402 \ $\myfontsize{$( 0.003)$} & $0.780 \ $\myfontsize{$( 0.009)$} & $\underline{0.819} \ $\myfontsize{$( 0.010)$} & $0.817 \ $\myfontsize{$( 0.007)$}  \\ \bottomrule
        \end{tabular}
    }
    \caption{$G$-\texttt{DDMNIST} test accuracies using our $O(2)$ PSCNN with various bandlimits $L$ and SCNN without bandlimiting.}
    \label{tab:bandlimit} 
    \vspace{-0.5cm}
\end{table}

\subsection{Data Ablation Study} 
\begin{figure}[h!]
        \centering
    \begin{subfigure}{0.49\columnwidth}
        \includegraphics[width=\linewidth]{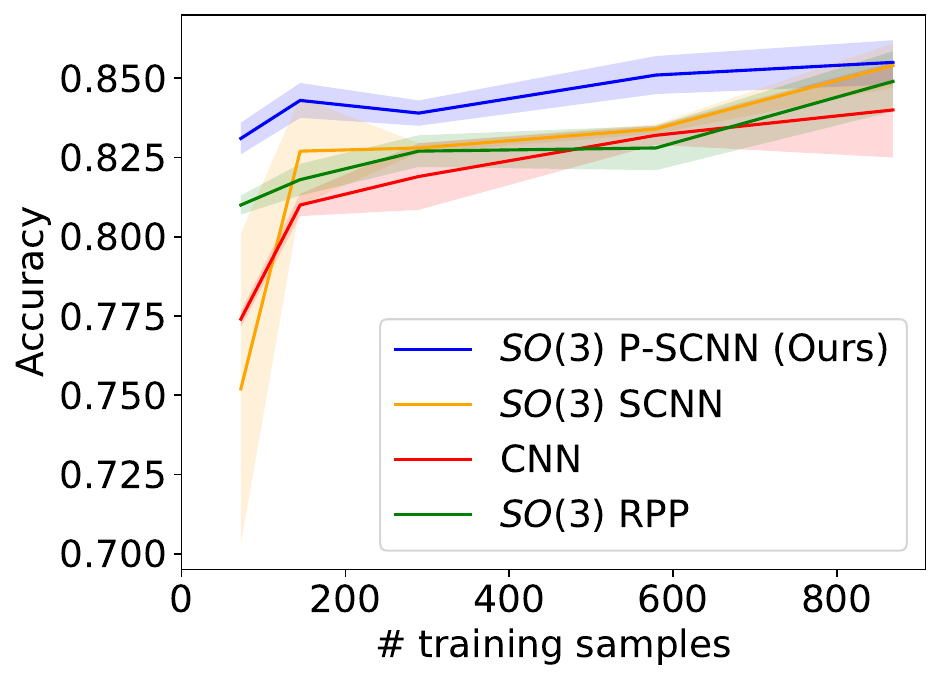}
    \caption{\textbf{Nodule}}\label{fig:ablation_nodule}    
    \end{subfigure}
        \begin{subfigure}{0.455\columnwidth}
        \includegraphics[width=\linewidth]{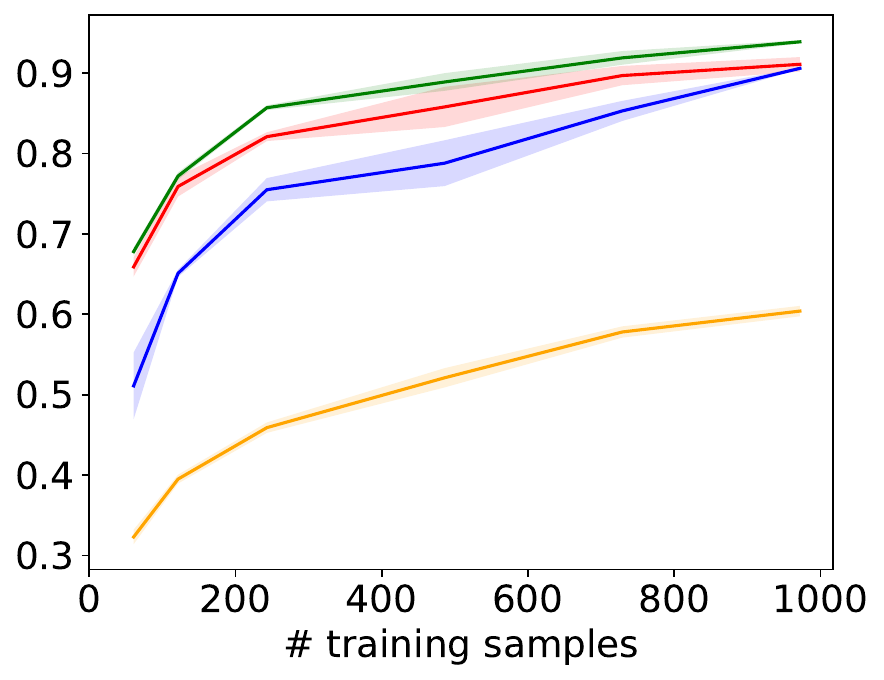}
        \caption{\textbf{Organ}}\label{fig:ablation_organ}
    \end{subfigure}
    \caption{Data ablation study on \textbf{Organ} and \textbf{Nodule}.}\label{fig:data_ablation}
    \vspace{-0.3cm}
    \end{figure}

One of the advantages of equivariant neural networks is their increased data-efficiency. To investigate the data-efficiency of the various approaches, we perform a data ablation study using the non-symmetric \textbf{Organ} and highly symmetric \textbf{Nodule} datasets. See Fig.~\ref{fig:data_ablation} for the results. On the symmetric \textbf{Nodule} subset our approach consistently achieves the highest performance, particularly on smaller train sets. Conversely, on the non-symmetric \textbf{Organ} dataset both RPP and CNN outperform our PSCNN, especially on smaller train sets. This is likely due to the fact that RPPs and CNNs, unlike our approach, do not initialise as equivariant due to the regular convolutions. When little data is available our approach therefore struggles to break equivariance. On highly symmetric datasets, the reverse applies.

    \begin{table}[h!]
\centering
\resizebox{0.8\columnwidth}{!}{%
\begin{tabular}{lc|cc|c}
    \toprule
    &  & \multicolumn{2}{c|}{\textbf{Test Digit Symmetries}} \\
    \begin{tabular}[c]{@{}c@{}}\textbf{Network}\\ \textbf{Group}\end{tabular} & \begin{tabular}[c]{@{}c@{}}\textbf{Partial}\\ \textbf{Equivariance}\end{tabular} & $\bm{D_4}$  & $\bm{O(2)}$ & $\bm{\Delta}$ \\ \midrule\midrule
CNN & N/A & $\underline{0.711} \ $\myfontsize{$( 0.009)$}  & $\underline{0.474} \ $\myfontsize{$( 0.037)$} & \underline{-0.237} \\ \midrule
\multirow{3}{*}{$SO(2)$} & Restriction & \multicolumn{1}{c}{$0.842 \ $\myfontsize{$( 0.015)$}}  & \multicolumn{1}{c|}{$0.624 \ $\myfontsize{$( 0.012)$}} & \underline{-0.218} \\
 & RPP & $0.821 \ $\myfontsize{$( 0.022)$}  & $0.569 \ $\myfontsize{$( 0.030)$} & -0.252 \\
 & Ours & $\bm{\underline{0.863}} \ $\myfontsize{$( 0.009)$}  & $\bm{\underline{0.637}} \ $\myfontsize{$( 0.012)$} & -0.226 \\ \midrule
\multirow{3}{*}{$O(2)$} & \multicolumn{1}{c|}{Restriction} & \multicolumn{1}{c}{$0.848 \ $\myfontsize{$( 0.007)$}}  & \multicolumn{1}{c|}{$\underline{0.632} \ $\myfontsize{$( 0.007)$}} & $\bm{\underline{-0.216}}$ \\
 & RPP & $0.824 \ $\myfontsize{$( 0.009)$}  & $0.566 \ $\myfontsize{$( 0.029)$} & -0.258 \\
 & Ours & $\underline{0.857} \ $\myfontsize{$( 0.011)$}  & $0.615 \ $\myfontsize{$( 0.023)$} & -0.242 \\
 \bottomrule
\end{tabular}}
 \caption{Generalisability results for models trained on discrete $D_4$ symmetries. All models are evaluated on $D_4$ and $O(2)$ symmetries. $\Delta$ shows the difference in performance between the two cases.}\label{tab:generalisability}
\vspace{-0.5cm}
\end{table}

\subsection{Generalisability}
Another major appeal of equivariant neural networks is their improvement in terms of generalisability to data outside the training data. To investigate the influence of our approach on the generalisability we train the models on \texttt{DDMNIST} with discrete $D_4$ symmetries and compare their $D_4$ and $O(2)$ performance (Tab.~\ref{tab:generalisability}). Our models consistently outperform RPPs, both in terms of performance and performance regression. However, \textit{restricted} SCNN shows a smaller performance regression, likely due to the fixed equivariance.

\section{Limitations and Future Work}
Although our approach offers an interpretable solution to learn the degree of equivariance in SCNNs, it does have a few limitations. Most notable, our approach is limited in breaking equivariances for compact groups, and thus cannot model partial translation equivariance. Furthermore, compared to regular SCNNs, our approach uses significantly more parameters, even with band-limiting. Moreover, the steerable basis needs to be recomputed after each update, unlike regular SCNNs applied fixed grids, increasing computational cost. Besides, our approach requires an alignment and, optional, KL-divergence term, resulting in additional hyper-parameter tuning. Nonetheless, we have observed our approach to be fairly robust to the choice of these terms. 

While we include a comparison of parameter space, due to the current sub-optimal implementation we did not perform a more detailed computational analysis. Similarly, we do not provide a theoretical proof that the learnt equivariance corresponds to the true equivariance. Both would be interesting directions for future research. Furthermore, we have only considered \textit{reducing} the degree of equivariance during training by initialising a uniform likelihood, however, it would be interesting to initialise a non-uniform likelihood distribution as a prior. Finally, we only consider band-limiting to filter high frequencies, but it could also be employed to ensure a lower-limit to the learnable equivariance.

\section{Conclusion}
This work presents a solution to generate partially $H$-steerable kernels from a likelihood distribution over $H$ and proposes an algorithm to parameterise this likelihood distribution in terms of its Fourier coefficient within the flexible framework of SCNNs. Furthermore, through the employment of normalisation, bandlimiting, KL-divergence, and our alignment loss, we ensure an interpretable likelihood distribution with user-tunable flexibility describing the degree of equivariance. Our experiments show that such a learnable degree of equivariance can result in performance improvements compared to fully equivariant approaches. Finally, in addition to theoretical benefits, our approach shows competitive performance compared to alternative partially equivariant approaches as well as \textit{restricted} SCNNs in task-specific performance, data-efficiency and generalisation.

\section*{Acknowledgements}
We express our gratitude towards the anonymous reviewers and Dr. Erik Bekkers for their valuable feedback and discussions. We also want to thank SURF for the use of the National Supercomputer Snellius. Finally, the presentation of this paper at the conference was financially supported by the Amsterdam ELLIS Unit and Qualcomm.

\section*{Impact Statement}
This paper presents work whose goal is to advance the field of Machine Learning. There are many potential societal consequences of our work, none which we feel must be specifically highlighted here.

\bibliography{references}
\bibliographystyle{icml2024}

\newpage

\appendix
\onecolumn

\renewcommand*\contentsname{Table of Contents}
\addtocontents{toc}{\protect\setcounter{tocdepth}{3}}
\tableofcontents
\newpage
{\LARGE\sc Appendix \par}
\section{Overview of Appendix}
These sections cover the supplementary materials of our work. First, in Appendix~\ref{chap:preliminaries} we provide an extensive mathematical background. Next, Appendix~\ref{ap:additional_derivs} covers additional and more in-depth derivations from concepts and solutions in the paper. In Appendix~\ref{ap:implement_details} we provide additional and more in-depth implementation details. Appendix~\ref{ap:experiment_details} contains an extensive overview of the experimental details. Finally, in Appendix~\ref{ap:additional_experiments} we present additional experiments.

\section{Mathematical Preliminaries}\label{chap:preliminaries}
In this section, we cover the mathematical preliminaries required for the work in this paper. We will start with the basics of Group Theory and Representation Theory, work our way through \textit{equivariance} and \textit{invariance} after which we close with \textit{Peter Weyl Theorem} and \textit{Fourier Transforms}.

\subsection{Group Theory}
\begin{definition}{Group}{group}
A \textit{group} is a pair $(G,\cdot)$, containing a set $G$ and a binary operation $\cdot$, also called the \textit{group law}:\\ 
$$\cdot \ : G\times G \to G, (h, g) \mapsto h \cdot g$$\\ 
satisfying the following axioms of group theory:\\
\begin{itemize}
    \item{\makebox[5cm]{Associativity: $\forall a, b, c \in G:$\hfill}$\quad a \cdot (b \cdot c) = (a \cdot b) \cdot c$}
    \item{\makebox[5cm]{Identity: $\forall e \in G, \forall g \in G:$\hfill}$\quad g \cdot e = e \cdot g = g$.}
    \item{\makebox[5cm]{Inverse: $\forall g \in G, \exists g^{-1} \in G:$\hfill}$\quad g \cdot g^{-1} = g^{-1} \cdot g = e$.}
\end{itemize}
It can be proven that the inverse element $g^{-1}$ of an element $g$, and the identity element $e$ are unique.
\end{definition}
To reduce notation it is common to write $hg$ instead of $h\cdot g$ and to refer to the whole group with $G$ instead of $(G, \cdot)$ as long as this is not ambiguous. It is also common to use the power notation to abbreviate the combination of the element $g$ with itself $n$ times:
$$g^n=\underbrace{g\cdot g \cdot g  \dotsm  g}_{n\text{ times}}$$

\begin{example}{General Linear Group}{GLG}
The \textit{general linear group} $\left(GL\left(\mathbb{R}^n\right), \cdot\right)$ is the group with the matrix multiplication $\cdot$ as group law and the set $GL(\mathbb{R}^n)$ of all invertible $n\times n$ matrices:
\begin{equation}
    GL\left(\mathbb{R}^n\right) = \left\{O\in \mathbb{R}^{n\times n} \mid O^{-1}O=\textbf{I}_n\right\}
\end{equation}
The identity element $e$ is the identity matrix $\textbf{I}_n$ of size $n\times n$. Since the matrix multiplication is associative, this definition satisfies the axioms from Definition~\ref{definition:group}.
    
\end{example}
Beyond the structure of a group itself, it is useful to consider the cardinality of the set $G$ within a group, formally termed as the order of a group.
\begin{definition}{Order of a Group}{order}
The \textit{order} of a group $G$ is the \textit{cardinality} of its set $G$ and it is indicated by $|G|$.
    
\end{definition}
Given the order, groups naturally fall into two categories: finite and infinite groups.
\begin{definition}{Finite Group}{finite}
A \textit{finite group} is a group with a finite number of elements, and therefore a finite cardinality. 
\end{definition}
Therefore, an infinite group is a group $G$ with cardinality $|G|=\infty$. An example of a finite group is the set of integers module $n$.

\begin{example}{$\mathbb{Z}/n\mathbb{Z}$}{integ_mod}
The set $\left\{e=0, 1, 2,\dots,n-1\right\}$ with the associative group law sum modulo $n$:
\begin{equation}
    +:(a,b)\mapsto (a+ b) \quad \mod{n}
\end{equation}
forms the group $\mathbb{Z}/n\mathbb{Z}$ of integers modulo $n$.

\begin{enumerate}
    \item Identity: The element $0$ serves as the identity element for this group. For any integer $a$ in $\mathbb{Z}/n\mathbb{Z}$, $a + 0 \equiv a \mod{n}$. Thus, adding 0 to any element does not change its value in this set.
    
    \item Inverse Element: Every element $a$ in $\mathbb{Z}/n\mathbb{Z}$ has an inverse $b$ such that $a + b \equiv 0 \mod{n}$. For instance, when $n = 10$, the element $4$ has the inverse $6$ because $(4 + 6) \mod{10} \equiv 0 \mod{10}$.
\end{enumerate}

The group has order $|\mathbb{Z}/n\mathbb{Z}|=n\in \mathbb{N}^+$, and is therefore a \textit{finite group}. 
\end{example}

Now that we have introduced the basic concepts of group theory, we move onto more advanced concepts. These involve relationships between groups, and transformations of groups.
\begin{definition}{Group Homomorphism}{homomorhism}
Given two groups $(G,\cdot)$ and $(H, *)$, a map $f: G \to H$, mapping an element in group $G$ to an element in group $H$, is a \textit{group homomorphism} from $G$ to $H$ if:
$$\forall a, b \in G\colon \quad f(a \cdot b) = f(a) * f(b)$$

\end{definition}
A special case of a homomorphism is an \textit{isomorphism}, which adds the condition of \textit{bijectivity}. 
\begin{definition}{Group Isomorphism}{isomorphism}
    A group homomorphism $f$ from $G$ to $H$ is a group \textit{isomorphism} if it is bijective (surjective and injective), i.e. if and only if:
$$\forall h \in H, \ \exists! \ g \in G: \quad f(g) = h.$$
\end{definition}
If we have two groups $G$ and $H$ and there exists an isomorphism between the two, then the two groups are considered to be isomorphic.

\begin{definition}{Group Automorphism}{automorphism}
A group homomorphism $f$ from $G$ to $G$ itself is called a group endomorphism.
A group endomorphism which is also bijective (i.e., an isomorphism) is called a group automorphism.
    
\end{definition}
Groups can be useful for describing space-based symmetries. In this work we focus on describing or negating symmetries in $\mathbb{R}^2$ and $\mathbb{R}^3$. Mathematically, these symmetries are described as group actions acting on a specific space.
\begin{definition}{Group Action and $G$-space}{gspace}
Given a group $G$, a (left) $G$-space $X$ is a set equipped with a group action $G \times X \to X$, $(g,x) \to g \cdot x$, i.e. a map satisfying the following axioms:
\begin{itemize}
    \item Identity: $\forall x \in X\colon \quad e\cdot x = x$
    \item Compatibility: $\forall a,b \in G, \forall x \in X\colon \quad a\cdot(b\cdot x) = (a\cdot b)\cdot x$

\end{itemize}
If these axioms hold, $G$ acts on the $G$-space $X$.
\end{definition}
For any group $(G, \cdot)$, its group law $\cdot: G\times G \to G$ trivially defines a group action of the group on itself.


\begin{definition}{Homogeneous Space}{hom_space}
    A homogeneous space is a $G$-space with a transitive action on $G$, meaning that for every pair of points $x, y \in X$, there exists an element $g \in G$ such that the action of $g$ on $x$ moves $x$ to $y$. Formally, this is expressed as:

\begin{equation}
    \forall x, y \in X, \exists g \in G \text{ s.t. } g \cdot x = y,
\end{equation}

where $\cdot$ denotes the group action from Def \ref{definition:gspace}.
\end{definition}
One of the main symmetries we are interested in are rotational symmetries. The Cyclic Group can be used to model rotations, but it is limited to a fixed number of discrete rotations.
\begin{example}{Cyclic Group}{cyclic}
The collection of all complex $N$-th roots of unity, $\left\{e^{ik\frac{2\pi}{N}}|0\leq k\leq n\right\}$, forms a group under multiplication. This group is isomorphic to the group $\mathbb{Z}/n\mathbb{Z}$, as depicted in Ex.~\ref{example:integ_mod}. A homomorphism $f$ can be established between these two groups and can be defined as
\begin{equation}
    f: e^{ik\frac{2\pi}{n}} \mapsto k.
\end{equation}

This finite group, generally denoted as $C_N$, is referred to as the Cyclic Group of order $|C_N| = N$. On a more abstract level, this group can be described as 
\begin{align}
    C_N = \left\{e=g^0, g, g^2, \dots, g^{n-1} \mid g^m= g^n \iff m\equiv n \text{ mod } N\right\}.
\end{align}
Here, the elements of the group are indicated using power notation.\\
The Cyclic Group appears frequently in this work, manifesting itself as discrete rotations when acting on the plane $\mathbb{R}^2$, or as a reflection when acting on rotation groups.
\end{example}\noindent
In contrast to the Cyclic Group, the Special Orthogonal Group contains continuous rotations.
\begin{example}{Special Orthogonal Group}{SO}
The Special Orthogonal Group is the group of all continuous rotations. This infinite group is denoted as $SO(n)$, where $n$ denotes the dimensionality of the space on which the group acts. The most notable examples are $SO(2)$ and $SO(3)$, denoting planar and volumetric rotations respectively. The action of a rotation $r_{\Theta} \in SO(n)$ with the rotation angles (or singular angle in the case of $SO(2)$) parameterized by the set $\Theta$ on the respective space can be defined as:

$$\forall x \in X = \mathbb{R}^n\colon \quad r_\Theta,x \mapsto r_\Theta \cdot x = \psi(\Theta)x$$

Where $\psi(\Theta)$ is the rotation matrix, i.e., for rotations in $SO(2)$ with $\Theta = \{\theta\}$:

\begin{align}
    \psi(\Theta) &= \begin{bmatrix}\cos{\theta} & - \sin{\theta}\\ \sin{\theta}& \phantom{-}\cos{\theta} \end{bmatrix},
\end{align}
and $\psi(\Theta)x$ is the regular matrix-vector product.\\

The $SO(n)$ groups are therefore the group of all $n\times n$ orthogonal real matrices with positive determinant
\begin{align}
    SO(n) &= \left\{O \in \mathbb{R}^{n \times n} \mid O^\top O = \mathbf{I}_n\text{ and }\det(O) = 1 \right\}.
\end{align}    
\end{example}\noindent

In this work, we often consider groups whose elements are completely included in another larger group. These groups are called subgroups.
\begin{definition}{Subgroup}{subgroup}
Given a group $(G, \cdot)$, a non-empty subset $H \subseteq G$ is a subgroup of $G$ if it forms a group $(H,*)$ under the same group law, restricted to $H$.\\
For $H$ to be a subgroup of $G$, it is required and sufficient that the restricted group law and the inverse are closed in $H$: 
\begin{itemize}
    \item $\forall a,b \in H\colon \quad a * b \in H$
    \item $\forall h \in H\colon \quad h^{-1} \in H$
\end{itemize}
In this case, it is common to write $H \leq G$.
\label{eq:subgroup}
\end{definition}

\begin{definition}{Coset}{coset}
    Consider the group $G$ and subgroup $H \leq G$. A \textit{left coset} of $H$ in $G$ is $gH = \{gh \mid h\in H\}$ for an element $g \in G$. Likewise, a \textit{right coset} of $H$ in $G$ is $Hg = \{hg
     \mid h\in H\}$ for an element $g\in G$.
\end{definition}
Therefore, a coset of an element $g$ is the set of all $g'\in G$ which are obtainable through the respective left or right action of an element $h \in H<G$ on the element $g$. 

\begin{example}{Left Coset of \(SO(2)\) in \(O(2)\)}{coset-example}
Let us consider the \textit{left coset} of \(SO(2)\) in \(O(2)\). 

Given \(SO(2)\) as our subgroup, consider a reflection $g \in O(2)$ which acts as a pure reflection along the x-axis. The matrix representation of this reflection is:
\[
g = \begin{bmatrix} -1 & 0 \\ 0 & 1 \end{bmatrix}
\]

The left coset generated by this reflection with respect to \(SO(2)\) will be:
\[ gSO(2) = \left\{g \cdot r_{\Theta} \mid r_{\Theta} \in SO(2)\right\} \]

Given our representation of rotations in \(SO(2)\) from Ex.~\ref{example:SO}, for any rotation \(\psi(\Theta)\in SO(2)\), the result of the matrix multiplication \(g \cdot \psi(\Theta)\) will be a reflection followed by a rotation by \(\theta\). Since $\text{det}(g) = -1$ and $\text{det}(\psi(\Theta)) = 1$, the resulting coset is as follows:
\begin{align}
    gH = gSO(2) &= \left\{O \in \mathbb{R}^{2 \times 2} \mid O^\top O = \mathbf{I}_n\text{ and }\det(O) = -1 \right\}.
\end{align}
\end{example}

\begin{definition}{Normal Subgroup}{normal}
    Given a group $G$ with a subgroup $H \leq G$, $H$ is a normal subgroup of $G$ iff:
    \begin{equation}
        \forall g \in G\colon \quad gH = Hg
    \end{equation}
    If this holds, the fact that $H$ is a normal subgroup of $G$ can be denoted as $H \unlhd G$.
\end{definition}
To illustrate the concept of a subgroup, let us consider the cyclic group $C_N$ and its relation to the group of all continuous rotations in a plane, $SO(2)$.
\begin{example}{$C_N$ as a Subgroup of $SO(2)$}{cyclic_sub_so}
    In Examples \ref{example:SO} and \ref{example:cyclic} we defined the group of all continuous rotations in a plane, $SO(2)$, and the cyclic group $C_N$ respectively. The cyclic group $C_N$, being a finite group of rotations, is actually a subgroup of the group $SO(2)$ of infinite rotations. It represents the group of $N$ rotations by integer multiples of an angle equal to $\frac{2\pi}{N}$.\\
    From Ex.~\ref{example:cyclic}, we know that $C_N$ is isomorphic to the group of $N$-th roots of unity under multiplication, which can be viewed as rotations in the complex plane, denoted by $\mathbb{C}$. 
    We can relate the elements of $C_N$ to $SO(2)$ through the following mapping function:
    \begin{equation}
        C_N \mapsto SO(2), \quad g^k \mapsto r_{k \frac{2\pi}{N}}.
    \end{equation}
    For example, if we take the cyclic group $C_4$, together with the element $g^3\in C_4$, then this element corresponds to $r_{3\frac{2\pi}{4}}=r_{\frac{3\pi}{2}}$, or a rotation of $270$ degrees.
\end{example}

Occasionally we want to describe certain groups as combinations of two or more smaller groups. There are two types of group products allowing for such decompositions; the \textit{direct product} and \textit{semi-direct product}. First, lets consider the \textit{direct product}.
\begin{definition}{Direct Product}{direct_product}
Given two groups $(K, *)$ and $(H, +)$, the direct product group $(K \times H, \cdot)$ is defined as the Cartesian product $K\times H$ of the two sets, along with the group law:
\begin{equation}
    (k_1,h_1)\cdot (k_2,h_2) = (k_1*k_2,h_1+h_2).
\end{equation}
$K\times H$ is often used as the shorthand notation for the direct product between $H$ and $K$. 
\end{definition}
Given a direct product set $K \times H$, we find that the subsets $\{(e_K, h) | h \in H\}$ and $\{(k, e_H) | k \in K\}$ form distinct normal subgroups. Furthermore, these subsets are isomorphic to $H$ and $K$, respectively. This structure enables us to uniquely decompose any element $(k, h)$ of the set $K \times H$ as a product of an element from $K$ and an element from $H$. For instance, we can represent $(k, h)$ as either $(e_K, h) \cdot (k, e_H)$ or $(k, e_H) \cdot (e_K, h)$. A notable characteristic of these sets is that all elements of $K$ commute with the elements of $H$. We can observe the structure of a direct-product in the decomposition of groups of translations.

    

\begin{example}{Decomposition of Translations}{}
    The group of translations on the plane, denoted as $(\mathbb{R}^2, +)$, can be decomposed into its constituent vertical and horizontal translations. As a result, we can say that the group $(\mathbb{R}^2, +)$ is isomorphic to the direct product $(\mathbb{R}, +) \times (\mathbb{R}, +)$, which represents two instances of the group $(\mathbb{R}, +)$, signifying translations along a line. Following the same line of thought, we can use the direct product between more than two groups. For example, for translations in three-dimensional space, represented by the group $(\mathbb{R}^3, +)$. This group is isomorphic to the direct product $(\mathbb{R}, +) \times (\mathbb{R}, +) \times (\mathbb{R}, +)$, which can be viewed as three instances of the group $(\mathbb{R}, +)$.
\end{example}
While the direct product decomposes a group in the product of two normal subgroups whose elements commute with each other, the semi-direct product only requires one of the subgroups to be normal subgroups.

\begin{definition}{Semi-Direct Product}{semi-direct}
    Considering two groups $(N,*)$ and $(H,+)$ and a group action $\phi: H \times N \to N$ of $H$ on $N$, the semi-direct product group $N \rtimes_\phi H$ is defined as the Cartesian product $N\times H$ with the following binary operation:
    \begin{equation}
        (n_1,h_1)\cdot (n_2,h_2) = (n_1 * \phi(h_1,n_2),h_1 + h_2)
    \end{equation}
    It is important to note here that the resulting group depends on the map $\phi$ and that changing the map leads to a change in the group.\\
    Like in a direct product, any element of a semi-direct product can be uniquely identified by a pair of elements on the two subgroups.
\end{definition}
Here, only the group $N$ is required to be a normal subgroup of $G$. Additionally, when $\phi$ is the identity map on $N$ for any $h\in H$, we obtain the more specific definition for the previously defined direct product. The semi-direct product plays a role in the decomposition of the special euclidean- and orthogonal groups.

\begin{example}{Orthogonal Group $O(n)$}{orthogonal}
    In Ex.~\ref{example:SO} we introduced the Special Euclidean Group $SO(n)$, the group of continuous rotations in $n$-dimensional space. Another common group in this work is the Orthogonal Group $O(n)$. In addition to rotations, this group also contains reflections in the $n$-dimensional space. Like in $SO(n)$, the action of a rotation $r_\Theta \text{ in } O(n)$ is defined as:
    \begin{equation}
        \forall x \in X = \mathbb{R}^n\colon \quad r_\Theta,x \mapsto r_\Theta \cdot x = \psi(\Theta)x
    \end{equation}
    Where $\psi(\Theta)$ is the rotation matrix.


    A potential reflection may or may not reflect the point $x$ around the first axis. This operation can be represented by inverting the first coordinate of a point, depending on the value of a parameter $p$ that dictates whether or not a reflection occurs. For $p=-1$, a reflection occurs, while for $p=1$, the original point is preserved. Thus we have that for a given a potential reflection operator $f_p$ where $p\in\{-1, 1\}$, the mapping is defined as follows:
    \begin{equation}
    \forall x \in X = \mathbb{R}^n\colon \quad f_p, x \mapsto f_p \cdot x  = \begin{bmatrix} p & 0\\ 0 & \textbf{I}_{n-1}
    \end{bmatrix}x
    \end{equation}

    Here, $p=-1$ results in a reflection, whereas $p=1$ leaves the point unchanged and $\textbf{I}_{n-1}$ is the $(n-1)\times(n-1)$ identity matrix.
    Therefore, this group is the group of all $n\times n$ orthogonal real matrices:
    \begin{equation}
        O(n) = \left\{O\in \mathbb{R}^{n\times n} \mid O^\top O = \textbf{I}_n\right\}
    \end{equation}
    With $\forall O \in O(n) \quad \det(O) = \pm 1$.\\

    Let us consider the planar case $n=2$, hence $G=O(2)$. $O(2)$ can be decomposed as a semi-direct product of $SO(2)$ and the cyclic group $C_2$. Any element of $O(2)$ can be identified by a pair $(r_\Theta, f_p)$, where $r_\Theta\in SO(2)$ and $f_p \in C_2$. The resulting product between these two elements is:
    \begin{align}
        (r_{\Theta_1}, f_{p_1}) \cdot (r_{\Theta_2}, f_{p_2}) &= r_{\Theta_1}f_{p_1}r_{\Theta_2}f_{p_2}\\
        &= r_{\Theta_1}r_{\phi(f_{p_1})\Theta_2}f_{p_1}f_{p_2}\\
        &= (r_{\Theta_1}r_{\phi(f_{p_1})\Theta_2}, f_{p_1}f_{p_2}) \\
        &= (r_{\Theta_1}r_{\phi(f_{p_1})\Theta_2}, f_{p_1}f_{p_2})
    \end{align}
    The action $\phi$ can be identified as:
    \begin{equation}
        \phi: C_2 \times SO(2) \to  (r_{\Theta_1}, f_{p_{2}}), SO(2) \mapsto r_{\phi(f_{p_2})\Theta_1}
    \end{equation}
    Hence, $O(n)$ can be decomposed as a semi-direct product: \begin{equation}
        O(2) = C_2 \rtimes_\phi SO(2)
    \end{equation}
    Here, the action of the $C_2$ group can be interpreted as a rotation of $0$ or $\pi$ around the $x$-axis in the coordinate system, where the $2$-dimensional space is embedded in a $3$-dimensional space.
\end{example}

\begin{example}{Special Euclidean Group $SE(n)$}{SE}
    The Special Euclidean Group, denoted as $SE(n)$, is the group of all translations and rotations in the $n$ dimensional space. Groups $SO(n)$ (Ex.~\ref{example:SO}) and $(\mathbb{R}^n, +)$ can be chosen as the respective subgroup and normal subgroups of $SE(n)$. Any element of $SE(n)$ can be identified by a pair $(t_v,r_\Theta)$, where $t_v\in \mathbb{R}^n$ and $r_\Theta \in SO(n)$. The resulting product between these two elements is:
    \begin{align}
        (t_{v_1}, r_{\Theta_1}) \cdot (t_{v_2}, r_{\Theta_2}) &= t_{v_1}r_{\Theta_1}t_{v_2}r_{\Theta_2}\\
        &= t_{v_1}t_{\psi(\Theta_1)v_2}r_{\Theta_1}r_{\Theta_2}\\
        &= (t_{v_1}t_{\psi(\Theta_1)v_2}, r_{\Theta_1}r_{\Theta_2}) \\
        &= (t_{v_1 + \psi(\Theta_1)v_2}, r_{\Theta_1}r_{\Theta_2})
    \end{align}
    The action $\phi$ can be identified as:
    \begin{equation}
        \phi: (\mathbb{R}^n, +) \times SO(n) \to (\mathbb{R}^n, +), (t_{v_2}, r_{\Theta_1}) \mapsto t_{\psi(\Theta_1)v_2}
    \end{equation}
    Hence the $SE(n)$ can be decomposed as a semi-direct product:
    \begin{equation}
        SE(n) = (\mathbb{R}^n, +) \rtimes_\phi SO(n)
    \end{equation}
\end{example}

\newpage

\subsection{Group Representation Theory}
In this section, we will focus on \textit{linear group representations}. These group actions model group elements $G$ as matrices, making them suitable for acting on $G$-spaces that are vector spaces. This is particularly useful for deep learning, since both the data and intermediate features are generally represented as vectors, and therefore form a vector $G$-space.

\begin{definition}{Linear Group Representation}{LGP}
A Linear Group Representation $\rho$ of a group $G$ on a vector space (representation space) $V$ is a group homomorphism from $G$ to the general linear group $GL(V)$, i.e., it is a map
\begin{equation}
    \rho: G \to GL(V) \quad \text{s.t.} \quad \forall g_1, g_2 \in G\colon \quad \rho(g_1g_2) = \rho(g_1)\rho(g_2)
\end{equation}    
\end{definition}
In the rest of this work, the shorthand \textit{representation} or \textit{group representation} will be used for \textit{linear group representation}. Perhaps the simplest example of a \textit{representation} is the \textit{trivial representation}.

\begin{example}{Trivial Representation}{trivial}
    The \textit{trivial representation} $\rho: G \to GL(
    \mathbb{R})$ maps any group element to the identity. Therefore,
    \begin{equation}
        \forall g \in G\colon \quad \rho(g)=1.
    \end{equation}
\end{example}
The groups used in this work often contain rotations, which are modelled as rotation matrices.

\begin{example}{Rotation Matrices}{rotmat}
    As shown in Example~\ref{example:SO}, an example of a representation of the group $SO(2)$ are the two-dimensional rotation matrices:
        \begin{equation}
        \psi : SO(2) \to GL(\mathbb{R}^2), r_{\theta} \mapsto \psi(r_{\theta}) = \begin{bmatrix*}[r]\cos{\theta} & -\sin{\theta}\\ \sin{\theta} & \cos{\theta}
        \end{bmatrix*}.
    \end{equation}
\end{example}
Representations that act similarly to one another are called \textit{equivalent representations}.
\begin{definition}{Equivalent Representations}{eq_rep}
    Two representations $\rho $ and $\rho'$ on a vector space $V$ are called equivalent, or isomorphic if and only if they are related by a change of basis $Q\in GL(V)$, that is: \begin{equation} \forall g \in G\colon \quad \rho'(g) = Q\rho(g)Q^{-1}
    \end{equation}
\end{definition}
It is straightforward that \textit{equivalent representations} behave similarly, since their composition is independent of the basis:
\begin{equation}
    \rho'(g_1)\rho'(g_2) = Q\rho(g_1)Q^{-1}Q\rho(g_2)Q^{-1} = Q\rho(g_1)\rho(g_2)Q^{-1}
\end{equation}
A collection of representations can be combined into a single representation through the \textit{direct sum}.
\begin{definition}{Direct Sum}{directsum}
    Given two representations $\rho_1: G \to GL(V_1)$ and $\rho_2:G\to GL(V_2)$, their direct sum $\rho_1 \oplus \rho_2 : G \to GL(V_1 \oplus V_2) $ is defined as:
    \begin{equation}
        (\rho_1 \oplus \rho_2)(g) = \begin{bmatrix}\rho_1(g) & 0 \\ 0 & \rho_2(g)\end{bmatrix}.
    \end{equation}
    Here, the corresponding matrices of the two representations are combined by simply taking the direct sum between them. Therefore, the action of the direct sum is given by the independent actions of $\rho_1$ and $\rho_2$ on the orthogonal subspaces $V_1$ and $V_2$, respectively, in $V_1 \oplus V_2$.
\end{definition}
When consecutively taking the \textit{direct sum} between multiple representations, we use $\bigoplus$, e.g.:
\begin{equation}
    \bigoplus_i \rho_i(g) = \rho_1(g) \oplus \rho_2(g) \oplus \rho_3(g) \oplus \dots.
\end{equation}

Suppose $\rho: G \rightarrow GL(V)$ is a group representation and $V_1$ is a subspace of $V$ such that $\forall g \in G:\rho(g)(V_1) \subseteq V_1$. Then $V_1$ is an invariant subspace under the action of $\rho$. We can find an orthogonal complement $V_2$ of $V_1$ such that $V = V_1 \oplus V_2$. If we choose a basis of $V$ consistent with this decomposition, then the matrix of $\rho(g)$ with respect to this basis has a block-diagonal form:
\begin{equation}
\rho(g) = \begin{bmatrix}\rho_1(g) & 0 \\ 0 & \rho_2(g)\end{bmatrix},
\end{equation}
where $\rho_1: G \rightarrow GL(V_1)$ and $\rho_2: G \rightarrow GL(V_2)$ are the restrictions of $\rho$ to $V_1$ and $V_2$ respectively. In this case, we say that $\rho$ is the direct sum of the representations $\rho_1$ and $\rho_2$. Representations that do not leave a subspace invariant are called \textit{irreducible representations}, which turn out to be useful building blocks for equivariant networks.

\begin{definition}{Irreducible Representation}{irrep}
    A representation is irreducible if it does not contain any non-trivial invariant subspaces. \textit{Irreps} is often used as shorthand for irreducible representations. We often denote irreps as $\psi$ rather than $\rho$.
\end{definition}
Given a group $G$, the set of irreducible representations of $G$ is often denoted as $\hat{G}$. The \textit{trivial representation} (Ex.~\ref{example:trivial}) is always in the set of irreducible representations $\hat{G}$.

In this work, in addition to translations, we will mostly focus on the rotation and reflection groups, both the discrete and continuous settings. These groups are all subsets of $O(n)$. Let us provide an overview of these subgroups in the $2$D case $H\leq O(2)$:

\begin{example}{Irreducible Representations of Subgroups $H\leq O(2)$}{irreps_groups}
Here we denote the irreps of the Cyclic group $C_N$, Dihedral group $D_N$, Special Orthogonal group $SO(2)$ and Orthogonal group $O(n)$ in the real case. For any of these groups, we will denote the irreps as $\psi^{H}_{i}(p)$, where $H$ is the considered subgroup, $i$ denotes some (group-specific and potentially multi-) index, and $p$ is the parametrisation of the irrep.

\paragraph{Special Orthogonal Group $SO(2)$} The Special Orthogonal group (Ex.~\ref{example:SO}) is the group of all continuous planar rotations. $SO(2)$ has one $1$-dimensional representation (the \textit{trivial representation}), and infinitely many $2\times 2$ rotation matrices as irreps.
\begin{itemize}
    \item trivial representation: $\psi_0^{SO(2)}(r_\theta) = 1$
    \item non-trivial irreps: $\psi_k^{SO(2)}(r_\theta) = \begin{bmatrix}
        \cos{(k\theta)} & -\sin{(k\theta)} \\
        \sin{(k\theta)} & \phantom{-}\cos{(k\theta)}
    \end{bmatrix} = \psi(k\theta),\quad k\in \mathbb{N}^+$ 
\end{itemize}
where $\psi(k\theta)$ is the rotation matrix from Ex.~\ref{example:SO}. 

\paragraph{Orthogonal Group $O(2)$} The Orthogonal group (Ex.~\ref{example:orthogonal}) is the group of planar reflections and continuous rotations with index $i=(r,k)$, where $r\in \{0,1\}$ denotes a possible reflection. In addition to the trivial representation, it has another $1$-dimensional irreducible representation that performs the reflection.
\begin{itemize}
    \item trivial representation: $\psi_{0,0}^{O(2)}(r_\theta p) = 1$
    \item reflection irrep: $\psi_{1,0}^{O(2)}(r_\theta p) = f, \quad  p\in \{-1, 1\}$
    \item other irreps: $\psi_{1,k}^{O(2)}(r_\theta p) = \begin{bmatrix}
        \cos{(k\theta)} & -\sin{(k\theta)} \\
        \sin{(k\theta)} & \phantom{-}\cos{(k\theta)}
    \end{bmatrix}\begin{bmatrix}
        f & 0\\
        0 & 1 
    \end{bmatrix} = \psi(k\theta)f_p, \ k\in \mathbb{N}^+,  p\in \{-1, 1\}$
\end{itemize}

Note that irreps $\psi^{O(2)}_{0, k}, k>0, k\in \mathbb{N}^+$ do not exist.  

\paragraph{Cyclic Group $C_N$} The Cyclic group (Ex.~\ref{example:cyclic}) is the group of $N$ discrete planar rotations. Its irreps are similar to the irreps of the special orthogonal group, only limited to $k \leq \lfloor \frac{N}{2} \rfloor$, where $\lfloor \cdot \rfloor$ is the flooring operation.

\begin{itemize}
    \item trivial representation: $\psi_{0}^{C_N}(r_\theta p) = 1$
    \item non-trivial irreps: $\psi_k^{SO(2)}(r_\theta) = \begin{bmatrix}
        \cos{(k\theta)} & -\sin{(k\theta)} \\
        \sin{(k\theta)} & \phantom{-}\cos{(k\theta)}
    \end{bmatrix} = \psi(k\theta),\quad k\in \left\{1, \dots, \lfloor \frac{N-1}{2} \rfloor\right\}$ 
\end{itemize}
In the case that $N$ is even, there is an additional $1$-dimensional irrep:
\begin{itemize}
    \item $N$-even irrep: $\psi_{N/2}^{SO(2)}(r_\theta) = \cos{(\frac{N}{2}\theta)} = \pm 1$
\end{itemize}

\paragraph{Dihedral Group $D_N$} The Dihehedral group is the group of planar reflections and $N$ discrete planar rotations. Like the $O(2)$ group, it always has two $1$-dimensional representations, with an additional two $1$-dimensional irreps if $N$ is even:
\begin{itemize}
    \item trivial representation: $\psi_{0,0}^{D_N}(r_\theta p) = 1$
    \item reflection irrep: $\psi_{1,0}^{D_N}(r_\theta p) = p, \quad  p\in \{-1, 1\}$
    \item other irreps: $\psi_{1,k}^{D_N}(r_\theta p) = \begin{bmatrix}
        \cos{(k\theta)} & -\sin{(k\theta)} \\
        \sin{(k\theta)} & \phantom{-}\cos{(k\theta)}
    \end{bmatrix}\begin{bmatrix}
        f & 0\\
        0 & 1 
    \end{bmatrix} = \psi(k\theta)f_p$, \\
    \vspace{-0.2cm}
    \begin{flushright}
        $k\in \left\{1, \dots, \lfloor \frac{N-1}{2} \rfloor\right\},  p\in \{-1, 1\}$
    \end{flushright}
\end{itemize}
    And the $N$-even irreps:
\begin{itemize}
    \item $N$-even irrep: $\psi_{0, N/2}^{SO(2)}(r_\theta p) = \cos{(\frac{N}{2}\theta)} = \pm 1$
    \item $N$-even reflection irrep: $\psi_{1, N/2}^{SO(2)}(r_\theta p) = f_p\psi_{0, N/2}^{SO(2)}(r_\theta p) = p\cos{(\frac{N}{2}\theta)} = \pm 1$
\end{itemize}

\end{example}
Upon closer inspection, it is notable that the irreps in the example above correlate with some form of frequency. For each of these groups, the irrep with index $k$ evaluated at a rotation of $\theta$ results, up to a possible reflection, in a rotation matrix corresponding to a rotation of $k\theta$. Therefore, we often refer to these irreps as \textit{irreps of frequency} $k$. A similar intuition applies for the irreps of subgroups $H=O(3)$ and $H=SO(3)$, since the irreps for these groups are similar to the irreps in this example, except that $O(3)$ also has non-reflective irreps $\psi^{O(3)}_{0, k}, k \in \mathbb{N}^+$. These irreps are therefore indiced by two separate independent frequencies $r\in \{0,1\}$ and $k \in \mathbb{N}^+$ respectively. The concept of viewing these irreps as frequencies is relevant for future sections.

It turns out that any reducible representation can be decomposed into a direct sum of \textit{irreps}.

\begin{theorem}{Irreps Decomposition (Peter-Weyl theorem part 1)}{irrep_decomp}
    Any unitary (or orthogonal) representation $\rho:G\to V$ of a compact group $G$ over a field with characteristic zero (e.g. the complex $\mathbb{C}^n$ and real $\mathbb{R}^n$ fields) is a direct sum of irreducible representations. Each irrep corresponds to an invariant subspace of the vector space $V$ with respect to the action of $\rho$. 
    In particular, any real linear representation $\rho: G \to \mathbb{R}^n$ of a compact group $G$ can be decomposed as
    \begin{equation}
        \rho(g) = Q \left [\bigoplus_{i\in \textit{I}} \psi_i(g)\right]Q^{-1},
    \end{equation}
    where $\textit{I}$ is an index set that specifies the irreducible representations $\psi_i$ contained in $\rho$ (with possible repetitions) and $Q$ is a change of basis. Alternatively, we can write this equation more explicitly using the multiplicity $s$ for each irrep $\psi_j \in \widehat{G}$:
        \begin{equation}
        \rho(g) = Q \left [\bigoplus_{j\in \widehat{G}}\bigoplus_{s}^{[j]} \psi_j(g)\right]Q^{-1},
    \end{equation}
    where $[j]$ denotes the multiplicity of $j$ in the decomposition.
\end{theorem}
For proofs, it is often only required to consider the irreps. A more specific method of irreps decomposition will be introduced in Section~\ref{sec:tp_gg}. 
\\
\\
A type of representation that is particularly useful for finite groups is the \textit{ regular representation}.


\begin{definition}{Regular Representation}{reg_rep}
Let $G$ be a finite group, and let $\mathbb{R}^{|G|}$ be a vector space. Each basis vector $\textbf{e}_g$ of $\mathbb{R}^{|G|}$ is associated with an element $g \in G$.

The regular representation of $G$, denoted as $\rho^G_{\text{reg}}: G \rightarrow GL(\mathbb{R}^{|G|})$, is defined as the map that takes an element $\widetilde{g} \in G$ and associates it with a permutation matrix $\rho^G_{\text{reg}}(\widetilde{g})$ in $GL(\mathbb{R}^{|G|})$.

The action of $\rho^G_{\text{reg}}(\widetilde{g})$ on a basis vector $\textbf{e}_g$ is defined as:

\begin{equation}
\rho^G_{\text{reg}}(\widetilde{g}) \cdot \textbf{e}_g = \textbf{e}_{\widetilde{g}g}.
\end{equation}

This means that $\rho^G_{\text{reg}}(\widetilde{g})$ permutes the basis vector $\textbf{e}_g$ to the basis vector $\textbf{e}_{\widetilde{g}g}$. 

\end{definition}

\begin{example}{Regular Representation of $D_2$}{reg_d2}
For the Dihedral group $D_2$, we represent each element of the group by a $4\times4$ permutation matrix. The complete regular representation of $D_2$ is as follows:

\begin{center}
\begin{tabular}{ c | c | c | c | c }
$g$ & $e$ & $r_{\pi}$ & $s_x$ & $s_y$ \\
\hline
$\rho^{D_2}_{\text{reg}}$ & $\begin{bmatrix} 1 & 0 & 0 & 0 \\ 0 & 1 & 0 & 0 \\ 0 & 0 & 1 & 0 \\ 0 & 0 & 0 & 1 \end{bmatrix}$ & $\begin{bmatrix} 0 & 1 & 0 & 0 \\ 1 & 0 & 0 & 0 \\ 0 & 0 & 0 & 1 \\ 0 & 0 & 1 & 0 \end{bmatrix}$ & $\begin{bmatrix} 0 & 0 & 1 & 0 \\ 0 & 0 & 0 & 1 \\ 1 & 0 & 0 & 0 \\ 0 & 1 & 0 & 0 \end{bmatrix}$ & $\begin{bmatrix} 1 & 0 & 0 & 0 \\ 0 & 1 & 0 & 0 \\ 0 & 0 & 0 & 1 \\ 0 & 0 & 1 & 0 \end{bmatrix}$ \\
\end{tabular}
\end{center}

In this representation, the $i$-th basis vector of the four-dimensional vector space $\mathbb{R}^4$ is associated with the $i$-th element in the sequence ${e,\ r_{\pi},\ s,\ sr_{\pi}}$ of $D_2$. Specifically, the enumeration is defined as $e \rightarrow 1,\ r_{\pi} \rightarrow 2,\ s \rightarrow 3,\ sr_{\pi} \rightarrow 4$. Thus, the $i$-th axis of $\mathbb{R}^4$ corresponds to the transformation associated with the $i$-th element of $D_2$.
\end{example}

The \textit{regular representations} directly map group elements to a matrix that performs the action of the group on a vector space. However, deep learning models are comprised of mapping functions. Therefore, it can be useful to apply a group directly to a function by transforming the domain. The \textit{left-regular representation} and \textit{right-regular representation} define such mappings.

\begin{definition}{Left and Right-Regular Representations}{left_reg_rep}
Given a group $G$ and a function $f: G \to Y$, where $Y$ is any field, the \textit{left-regular representation} $\mathscr{L}$ acts on $f$ by ``shifting" its domain from the left. For a group element $g$, the action $\mathscr{L}_g$ on $f$ results in a new function $\mathscr{L}_gf: G \to Y$ defined by:

\begin{equation}
(\mathscr{L}_gf)(h) = f(g^{-1}h) \quad \forall h,g \in G,
\end{equation}

Similarly, the \textit{right-regular representation} $\mathscr{R}$ acts on $f$ by shifting its domain from the right, thus:
\begin{equation}
    (\mathscr{R}_gf)(h) = f(hg) \quad \forall h,g \in G,
\end{equation}
\end{definition}

In the context of deep learning, it can be useful to vary the considered group throughout the network. To achieve this, the \textit{restricted representation} can be used.
\begin{definition}{Restricted Representation}{restrict_repr}
    Any representation $\rho: G \to GL(\mathbb{R}^n)$ can be uniquely restricted to a representation of a subgroup $H$ of $G$ by restricting its domain to $H$:
    \begin{equation}
        \text{Res}^G_H(\rho): H \to GL(\mathbb{R}^n),\quad h \mapsto \rho|_H(h)
    \end{equation}
\end{definition}
If the representation $\rho$ of $G$ is an irrep, the resulting \textit{restricted representation} $\rho|_H$ is not necessarily an irrep, but can be decomposed into a direct-sum of $H$ irreps (Thm. \ref{theorem:irrep_decomp}). 

\newpage
\subsubsection{Tensor Product Representation and its Decomposition}\label{sec:tp_gg}
The tensor product and its decomposition in irreps will both show to be important for regular steerable networks, as well as for the extension towards learnable equivariance.
\begin{definition}{Tensor Product of Representations}{tensor_rep}
Given two representations $\rho_1: G \to GL(V_1)$ and $\rho_2:G\to GL(V_2)$, the \textit{tensor product of representations}, denoted as $\rho_1 \otimes \rho_2 : G \to GL(V_1 \otimes V_2)$, is defined as:
\begin{equation}
(\rho_1 \otimes \rho_2)(g) = \rho_1(g) \otimes \rho_2(g)
\end{equation}
for every $g \in G$. Here, the tensor product of the two representations is obtained by taking the Kronecker product $\otimes$ of their corresponding matrices:
\begin{definition}{Kronecker Product}{kron_product}
    Given a matrix $A\in \mathbb{R}^{m\times n}$ and $B\in \mathbb{R}^{p\times q}$, the resulting Kronecker product block matrix $(A \otimes B) \in \mathbb{R}^{pm \times qn}$ is constructed as follows:
    \begin{equation}
        A\otimes B = \begin{bmatrix}
            a_{11}B & a_{12}B & \dots & a_{1n}B\\
            a_{21}B & a_{22}B  &\dots & a_{2n}B\\
            \vdots & \vdots & \ddots & \vdots \\
            a_{m1}B & a_{m2}B & \dots & a_{mn}B
        \end{bmatrix}
        \end{equation}

    or, more precisely:

    \begin{equation}
        A\otimes B = \begin{bmatrix}
        a_{11}b_{11} &  a_{11}b_{12} & \dots & a_{11}b_{1q} & \dots& \dots & a_{1n}b_{11} & a_{1n}b_{12} & \dots & a_{1n}b_{1q}\\
        a_{11}b_{21} &  a_{11}b_{22} & \dots & a_{11}b_{2q} & \dots & \dots & a_{1n}b_{21} & a_{1n}b_{22} & \dots & a_{1n}b_{2q}\\
        \vdots & \vdots & \ddots & \vdots & & & \vdots & \vdots & \ddots & \vdots \\
        a_{11}b_{p1} &  a_{11}b_{p2} & \dots & a_{11}b_{pq} & \dots& \dots & a_{1n}b_{p1} & a_{1n}b_{p2} & \dots & a_{1n}b_{pq}\\
        \vdots & \vdots &  & \vdots & \ddots &  & \vdots & \vdots &  & \vdots \\
        \vdots & \vdots &  & \vdots &  & \ddots & \vdots & \vdots &  & \vdots \\
        a_{m1}b_{11} &  a_{m1}b_{12} & \dots & a_{m1}b_{1q} & \dots& \dots & a_{mn}b_{11} & a_{mn}b_{12} & \dots & a_{mn}b_{1q}\\
        a_{m1}b_{21} &  a_{m1}b_{22} & \dots & a_{m1}b_{2q} & \dots& \dots & a_{mn}b_{21} & a_{mn}b_{22} & \dots & a_{mn}b_{2q}\\
        \vdots & \vdots & \ddots & \vdots & & & \vdots & \vdots & \ddots & \vdots \\
        a_{m1}b_{p1} &  a_{m1}b_{p2} & \dots & a_{m1}b_{pq} & \dots& \dots & a_{mn}b_{p1} & a_{mn}b_{p2} & \dots & a_{mn}b_{pq}\\
        \end{bmatrix}
        \end{equation}
    
This product is a special case of the regular tensor product.
\end{definition}

Thus, the tensor product representation maps the space of matrices to the space of the Kronecker product $ V_1 \otimes V_2$.
\end{definition}

A tensor product between two irreps does not necessarily result in an irrep. However, through the \textit{Clebsch-Gordan decomposition} they can be decomposed into a direct-sum of irreps.

\begin{theorem}{Clebsch-Gordan Decomposition}{clebsch_gordan}
Given two irreducible representations $\rho_l: G \to GL(V_1)$ and  $\rho_J: G \to GL(V_2)$ of a compact group $G$, their tensor product representation $\rho_l \otimes \rho_J$ is not necessarily irreducible. However, as stated in Theorem \ref{theorem:irrep_decomp}, any representation can be decomposed into a \textit{direct sum} of irreducible representations. The \textit{Clebsch-Gordan decomposition} is the irreps decomposition for such tensor products of irreducible representations: 
\begin{equation}
(\rho_l \otimes \rho_J)(g) = \left [ C^{lJ}\right ]^\top\left(\bigoplus_j\bigoplus_s^{[j(Jl)]}\psi_j(g) \right) C^{lJ} 
\end{equation}
Here $C^{lJ}$ is the change of basis, $\psi_j: G \to GL(V_j)$ are the irreps in the decomposition, and $[j(Jl)]$ denotes the \textit{multiplicity} of the irrep $\psi_j$. \\

Visually, this decomposition looks as follows in block-matrix form:
\begin{equation}(\rho_l \otimes \rho_J)(g) = 
    \underbrace{\begin{colorbmatrix}
        \rowcolor{blue!20}
        [C^{lJ}_{j_{1}}]^\top\\ \rowcolor{green!20}
        [C^{lJ}_{j_{2}}]^\top\\
        \vdots \\ \rowcolor{red!30}
        [C^{lJ}_{j_{n}}]^\top\\
    \end{colorbmatrix}}_{\left[ C^{lJ} \right]^\top}    \underbrace{\begin{colorbmatrix}
        \cellcolor{blue!20}\psi_{j_1}(g) & 0 & \cdots & 0 \\
        0 & \cellcolor{green!20}\psi_{j_2}(g) & \cdots & 0 \\
        \vdots & \vdots & \ddots & \vdots \\
        0 & 0 & \cdots & \cellcolor{red!30}\psi_{j_n}(g)
    \end{colorbmatrix}}_{\left(\bigoplus_j\bigoplus_s^{[j(Jl)]}\psi_j(g) \right)} 
    \underbrace{\begin{colorbmatrix}
        \cellcolor{blue!20}C^{lJ}_{j_1} &
        \cellcolor{green!20}C^{lJ}_{j_2}&
        \dots &
        \cellcolor{red!30}C^{lJ}_{j_n}
    \end{colorbmatrix}}_{C^{lJ}}
\end{equation}
Here, $C^{lJ}_{j_{1}}$ through $C^{lJ}_{j_{n}}$ are blocks of columns in $C^{lJ}$. Note that, due to the block-diagonal nature, these blocks act only on one specific irrep (as visualised with the colours). Therefore, we can re-write this equation replacing the direct-sums with regular summations: 

\begin{equation}
(\rho_l \otimes \rho_J)(g) = \sum_j\sum_s^{[j(Jl)]} \left[\operatorname{CG}_{s}^{j(Jl)}\right]^\top \psi_j(g)\operatorname{CG}_{s}^{j(Jl)}
\end{equation}
Where $\operatorname{CG}_{s}^{j(Jl)}\in \mathbb{R}^{d_{j} \times d_{l}d_{J}}$ are the resulting \textit{Clebsch-Gordan coefficients}.
\end{theorem}
In the context of the group \( SO(2) \), the \textit{Clebsch-Gordan decomposition} of the tensor product between two irreps with frequencies \( k_1 \) and \( k_2 \) results in a decomposition of up to two irreps of the frequencies \( |k_1 - k_2| \) and \( k_1 + k_2 \) respectively. Notably, the trivial representation (where all frequencies are zero) appears only in the tensor product between equal irreps, \( k_1 = k_2 \), since only in that case \( |k_1 - k_2| = 0 \). For the group \( SO(3) \), the decomposition includes not only the lowest and highest frequencies but also all frequencies in between \( |k_1 - k_2| \) and \( k_1 + k_2 \). It is important to note that for other groups, the decomposition pattern might differ. However, a similar pattern can be observed for groups such as \( O(n) \), \( C_n \), and \( D_n \).

\newpage

\subsection{Equivariance and Schur's Lemma}
The aim of Steerable Networks is to become equivariant, or perhaps invariant, to a specific set of transformations. Therefore, in this section, we mathematically define the concepts of \textit{equivariance} and the more special case of \textit{invariance} in terms of representation- and group theory. Furthermore, we use these definitions to introduce the concept of intertwiners, and \textit{Schur's Representation Lemma}.
\begin{definition}{Equivariance}{equivariance}
    Given a group $G$ and two sets $X$ and $Y$ that are acted on by $G$, a map $f:X \to Y$ is equivariant iff
    \begin{equation}
        \forall x \in X, \forall g \in G\colon \quad f(g\cdot x) = g\cdot f(x)
    \end{equation}
\end{definition}
In other words, when an operation is equivariant with respect to group $G$, or \textit{$G$-equivariant}, when the input transforms the out transforms predictably. Equivariance is therefore a generalisation of the more specific \textit{invariance}. 

\begin{definition}{Invariance}{inv}
    Given a group $G$ and two sets $X$ and $Y$ that are acted on by $G$, a map $f:X \to Y$ is invariant iff
    \begin{equation}
        \forall x \in X, \forall g \in G \colon \quad f(g\cdot x) = f(x).
    \end{equation}
    Therefore, invariance is a special case of equivariance, where the action of $G$ on the set $Y$ is the trivial action:
    \begin{equation}
        \forall y \in Y, \forall g \in G: \quad g\cdot y = y.
    \end{equation}
\end{definition}
Functions that perform an equivariant mapping between representations are called \textit{intertwiners}. These are particularly relevant for equivariant deep learning, as the mapping between features must be equivariant.
\begin{definition}{Intertwiner}{intertwiner}
    Consider the group $G$ and the two representations $\rho: G \to GL(V_1)$ and $\rho_2 : G \to GL(V_2)$. A linear map $W$ from $V_1$ to $V_2$ is an intertwiner between $\rho_1$ and $\rho_2$ if it is an equivariant map:
    \begin{equation}
        \forall \bm{v} \in V_1,\forall g\in G \colon \quad W\rho_1(g)\bm{v}=\rho_2(g)W\bm{v}
    \end{equation}
    from which follows that:
    \begin{equation}
        \forall g\in G\colon \quad W\rho_1(g) = \rho_2(g)W.
    \end{equation}
\end{definition} 
In fact, it turns out that two irreducible representations $\rho_1$ and $\rho_2$ must be equivalent representations if $W$ is not a null map.
\begin{theorem}{Schur's Representation Lemma}{schur}
    Consider group $G$ along with two of its irreps $\psi_1: G \to V_1$ and $\psi_2: G \to V_2$, and a linear map $W: V_1 \to V_2$ satisfying $\forall g\in G \colon \quad \psi_2(g)W = W\psi_1(g)$. Then either one of the following must hold:
    \begin{itemize}
        \item $W$ is the null map
        \item $W$ is an isomorphism, and therefore $\psi_1$ and $\psi_2$ are equivalent representations (Def.~\ref{definition:eq_rep}) where $W$ is the change of basis ($Q$ from Def.~\ref{definition:eq_rep}) between $\psi_1$ and $\psi_2$.
    \end{itemize}
\end{theorem}

\newpage

\subsection{Peter-Weyl Theorem: Fourier Transform and Steerable Basis}\label{app:peter_weyl}

In this section we discuss the \textit{Peter-Weyl theorem}, along with two relevant applications. While we have already seen in Ex.~\ref{example:irreps_groups} that the irreps of certain groups represent frequencies, through the\textit{ Peter-Weyl theorem} we build upon this notion by relating irreps to the Fourier transform; a popular technique in signal processing. This allows us to describe a continuous signal in terms of its frequencies. Additionally, for steerable CNNs we require a way to parameterise the kernels. Using the \textit{Peter-Weyl theorem}, a convenient parameterisation of kernels can be built.

\begin{theorem}{Peter-Weyl Theorem}{peter_weyl}
Let $G$ be a compact group and $\rho: G \to GL(V)$ a unitary representation. Then $\rho$ decomposes into a direct sum of finite-dimensional and unitary irreducible representations of $G$ (Thm. \ref{theorem:irrep_decomp}).

Moreover, let $L^2(G$) be the vector space of square integrable functions over $G$.
Consider the regular representation of $G$ on $L^2(G)$ given by its left-action (Def. \ref{definition:left_reg_rep}):
\begin{equation}
    f \mapsto \rho(g) : \left[\mathscr{L}_g f\right](h) = f(g^{-1} h)
\end{equation}
This representation is unitary, and therefore decomposes according to Thm. \ref{theorem:irrep_decomp}.
In particular, the matrix coefficients of the irreps of $G$ span the vector space $L^2(G)$.
If complex valued irreps are considered, then their matrix coefficients form a basis for this space.

Finally, an orthonormal basis for \textit{complex} square integrable functions in $L^2(G)$ can be explicitly built as
$\left\{\sqrt{d_\psi} [\psi(g)]_{ij} \mid \psi \in \widehat{G}, j \leq d_\psi\right\}$.
\end{theorem}
It is important to note that this theorem only holds as such for the complex case. In the real case, there can be some redundancy in the irreps. Consider, for example, the $2$-dimensional irreps for $SO(2)$ (Ex.~\ref{example:irreps_groups}). Here, the second column of an irrep $\psi_k^{SO(2)}$ can be obtained by multiplying the first column with the anti-symmetric matrix $\begin{bmatrix}0&-1\\
1&\phantom{-}0\end{bmatrix}$: 
\begin{equation}
    \begin{bmatrix}-\sin{(k\theta)}\\ \phantom{-}\cos(k\theta)\end{bmatrix} = \begin{bmatrix}0&-1\\
1&\phantom{-}0\end{bmatrix}\begin{bmatrix}\cos{(k\theta)}\\ \sin(k\theta)\end{bmatrix}.
\end{equation} 
The relation between the irreps and their non-redundant columns can be modelled by the \textit{endomorphism basis}.

\begin{definition}{Endomorphism Basis}{end_basis}
    Let \( G \) be a compact group and \( \psi_i: G \to GL(V) \) be a   real irrep of \( G \).
    The \textit{endomorphism space} \( \operatorname{End}_{\psi_i} \) consists of all linear endomorphisms \( c: V \to V \) such that \( c \circ \psi_i(g) = \psi_i(g) \circ c \) for all \( g \in G \).
    
\end{definition}

Now, define \( \overline{\psi}_i : G \to \mathbb{R}^{d_{i}\times n_i} \) to contain the non-redundant columns of \( \psi_i \). Here, \( n_i = \frac{d_{i}}{m_i} \), where \( m_i=[0(ii)] \) is the multiplicity of the trivial representation in the Clebsch-Gordan decomposition of the tensor product \( \psi_i \otimes \psi_i \) (Thm. \ref{theorem:clebsch_gordan}).

A particular basis \( \mathcal{C}_{\psi_i} = \{c^i_r \mid c^i_r \in \mathbb{R}^{d_{i} \times d_{i}}, r \leq m_i\} \ \forall \psi_i \in \widehat{G} \) spanning the endormorphism space \( \operatorname{End}_{\psi_i} \) can be constructed such that the elements \( c^i_r \) describe how the non-redundant columns \( \widetilde{\psi}_i \) relate to the entire irrep \( \psi_i \). See \cite{cesa2022ae(n)} Appendix C for more details.\\

To transform the non-redundant columns \( \overline{\psi}_i \) into the full irrep \( \psi_i \), we introduce the function \( \mathcal{R}_{\psi_i} : \mathbb{R}^{d_{i} \times n_i} \to \mathbb{R}^{d_{i} \times d_{i}} \). This function accepts \( \overline{\psi}_i(g) \) as its input. Leveraging the endomorphism basis \( \mathcal{C}_{\psi_i} \), it produces the entire irrep \( \psi_i(g) \) by sequentially left-multiplying the elements \( c_r^i \) with the non-redundant columns \( \overline{\psi}_i(g) \):
\begin{equation}
\psi_i(g) = \left [
\begin{array}{c|c|c|c}
c^i_0 \overline{\psi}_i(g) & c^i_1 \overline{\psi}_i(g) & \dots & c^i_{m_i} \overline{\psi}_i(g)
\end{array} \right ]
\end{equation}

We can also construct an inverse \( \mathcal{R}_{\psi_i}^{-1} \), which performs the opposite operation.

For example, in the case of irreps $\psi_{0}^{SO(2)}$ and $\psi_{k}^{SO(2)}$, the endomorphism bases are defined as follows: 
\begin{align}
     \mathcal{C}_{\psi_{0}^{SO(2)}} &= \left \{\begin{bmatrix}
        1
    \end{bmatrix}\right \} \\
    \mathcal{C}_{\psi_{k}^{SO(2)}} &= \left \{\begin{bmatrix}
    1 & 0\\
    0 & 1
\end{bmatrix}, \begin{bmatrix}0&-1\\
1&\phantom{-}0\end{bmatrix}\right \}.\\
\end{align}
On the contrary, the endomorphism basis for any irrep $\psi^{O(2)}$ of $O(2)$ only contains the identity matrix, since these irreps do not contain redundant columns.
\begin{equation}
    \mathcal{C}_{\psi^{O(2)}} =  \left \{ I_{d_{\psi^{O(2)}} \times d_{\psi^{O(2)}}}\right \}.
\end{equation}

\citet[Appendix C.3]{cesa2022ae(n)} have shown that in such cases we only need to consider the first $n$ non-redundant columns of the irreps $\psi$ to build an orthonormal basis. Therefore, using real irreps, an orthonormal basis can be explicitly built as follows:
\begin{equation}
    \left\{\sqrt{d_\psi} \psi_i(g)_{j} \mid \psi_i \in \widehat{G}, 1 \leq i, j \leq n_i\right\},
\end{equation}
replacing $d_\psi$ with $n_i$, for example $n_{\psi_k^{SO(2)}}=\frac{2}{2}=1$ or $n_{\psi_{1,k}^{O(2)}}=\frac{2}{1}=2$ for $SO(2)$ and $O(2)$ respectively.

The Peter-Weyl theorem plays a vital role in the Fourier transform and the steerable basis.
\paragraph{Fourier Transform}
    Since, for a compact group $G$, the matrix entries of the irreps both span the space of square integrable functions $L^2(G)$ and constitute an orthonormal basis, these irreps can be used to express a function $f:G \to \mathbb{R}$ through the \textit{inverse Fourier transform}.
    \begin{definition}{Inverse Fourier Transform}{inv_fourier}
    Let $G$ be a group and $f:G\to \mathbb{R}$ be a function defined over $G$. The \textit{inverse Fourier transform} allows us to express $f$ as a linear combination of the irreps of $G$. This can be formally stated as:
    \begin{equation}
    f(g) = \sum_{\psi_j \in \widehat{G}} \sqrt{d_j}\text{Tr}\left (\psi_j(g)^\top \mathcal{R}_{\psi_j}\left(\widehat{f}(\psi_j)\right) \right),
    \end{equation}
    where $g$ is an element of $G$, $\widehat{G}$ is the set of irreducible representations of $G$, $d_j$ is the dimension of the irrep $\psi_j$, and $\widehat{f}(\psi_j)$ denotes the Fourier coefficient corresponding to the irrep $\psi_j$ which characterises the contribution of $\psi_j$ to the function $f$.
    \end{definition}
    Similarly, we can also express the Fourier coefficients $\widehat{f}(\psi_j)$ with the \textit{Fourier transform}.
    
    \begin{definition}{Fourier Transform}{fourier}
    Given a group $G$ and a function $f:G\to \mathbb{R}$, the \textit{Fourier transform} of $f$ is given as:
    \begin{equation}
    \widehat{f}(\psi_j) = \sqrt{d_j}\int_G f(g) \mathcal{R}^{-1}_{\psi_j}\left({\psi}_j(g)\right)dg
    \end{equation}
    if $G$ is an infinite group, and
    \begin{equation}
    \widehat{f}(\psi_j) = \sqrt{d_j}\frac{1}{|G|}\sum_{g\in G} f(g) \mathcal{R}^{-1}_{\psi_j}\left({\psi}_j(g)\right)
    \end{equation}
    if $G$ is a finite group. Here, $\widehat{f}(\psi_j)$ denotes the Fourier coefficient corresponding to the irrep $\psi_j$.
    \end{definition}
    The \textit{Fourier transform} and its inverse of a signal $x$ and Fourier coefficients $\widehat{x}$ are often denoted as $\mathcal{F}(x)$ and $\mathcal{F}^{-1}(x)$ respectively, where $\mathcal{F}^{-1}(\mathcal{F}(x)) = x$.
\paragraph{Steerable Basis} Convolution kernels, which are square integrable functions, belong to the space \( L^2(\mathbb{R}^n) \). Parameterising such functions is conveniently achieved using steerable bases \cite{cesa2022ae(n), lang2020wigner}.

\begin{definition}{$H$-Steerable Basis}{steerable_basis}
    Given a compact group $H$ with a unitary action on $\mathbb{R}^n$, an $H$-steerable basis for $L^2(\mathbb{R}^n)$ is a collection of orthogonal functions:
    \begin{equation}
        \left\{Y_{j}^{km}\colon \mathbb{R}^n \to \mathbb{R}\ \mid \ \psi_j \in \widehat{H},m\leq d_{j}\right\}.
    \end{equation}
    This is an application of the \textit{Peter-Weyl theorem} (Thm. \ref{theorem:peter_weyl}), by considering the action of $H$ on $L^2(\mathbb{R}^n)$. Therefore, the stack $\left\{Y_{j}^{km}\right\}^{d_{j}}_{m=1}$, denoted by $Y_{j}^k: \mathbb{R}^n \to \mathbb{R}^{d_{j}}$, has the following defining property:
    \begin{equation}
        Y_{j}^k(g\cdot x) = \psi_j(g)Y_{j}^k(x) \quad \forall g\in G, \ x \in \mathbb{R}^n
    \end{equation}
\end{definition}
For groups $G=(\mathbb{R}^n, +) \rtimes H$, with compact subgroup $H$, it has been shown that the well-known \textit{Wigner-Eckart theorem} can be generalised to build such $G$-steerable bases \cite{lang2020wigner}.

\section{Additional Derivations}\label{ap:additional_derivs}
In this section we cover additional and more in-depth derivations. In \ref{ap:irrep_decomp} we discuss the irreps decomposition. In Section~\ref{ap:eq_constraints} we solve the equivariant constraint in more detail. In Section~\ref{ap:alignment_loss} we cover the alignment problem in more detail. Finally, in Section~\ref{ap:kl_divergence} we show a derivation for our solution in Eq.~\ref{eq:final_kl} regarding the computation of KL-divergence on the Fourier domain.
\subsection{Irreps Decomposition}\label{ap:irrep_decomp}
The steerability constraint depends on the input and output representations $\rho_{\text{in}}$ and $\rho_{\text{out}}$, and are therefore dependent on the type of feature field. This requires the constraint to be solved independently for each pair of input and output types, which can result in a large computational overhead for mixed feature fields. However,~\cite{cesa2022ae(n), cesae(2)} propose to decompose $\rho_{\text{in}}$ and $\rho_{\text{out}}$ into a direct sum of irreps, which is possible for any representation of a compact group $H$ (Thm.~\ref{theorem:irrep_decomp}). Therefore, the kernel constraint from~\ref{eq:kernel_constraint} can be written as:

    \begin{align}
        k(hx) &= Q^{-1}_{\text{out}} \left [\bigoplus_{i\in I_{\text{out}}} \psi_i(h)\right ]Q_{\text{out}} k(x ) Q^{-1}_{\text{in}}\left [\bigoplus_{j\in I_{\text{in}}} \psi_j(h)^{-1}\right ]Q_{\text{in}} \quad \forall h \in H, x \in \mathbb{R}^n.\\
        \intertext{Applying a change of variable: $\widetilde{k} = Q_{\text{out}}k Q^{-1}_{\text{in}}$, and noting that compact representations are orthogonal, hence $\forall h \in H\colon \quad \psi(h)^{-1} = \psi(h)^\top$:}\nonumber\\
        \widetilde{k}(hx) &=\left [\bigoplus_{i\in I_{\text{out}}} \psi_i(h)\right ] \widetilde{k}(x ) \left [\bigoplus_{j\in I_{\text{in}}} \psi_j(h)^\top\right ] \quad \forall h \in H, x \in \mathbb{R}^n.
    \end{align}

    Due to the block-diagonal nature of the direct-sum, we can visualise this equation as follows:
    \begin{equation}
        \underbrace{\begin{bmatrix} \widetilde{k}^{i_1j_1}(hx) & \widetilde{k}^{i_1j_2}(hx) & \dots\\
        \widetilde{k}^{i_2j_1}(hx) & \widetilde{k}^{i_2j_2}(hx) &  \dots\\
        \vdots & \vdots  & \ddots
        \end{bmatrix}}_{\widetilde{k}(hx)} = \underbrace{\begin{bmatrix}
            \psi_{i_1}(h) &  &  \\
             & \psi_{i_2}(h) &   \\
             &   & \ddots \\
        \end{bmatrix}}_{\left [\bigoplus_{i\in I_{\text{out}}} \psi_i(h)\right ]} \underbrace{\begin{bmatrix} \widetilde{k}^{i_1j_1}(x) & \widetilde{k}^{i_1j_2}(x) &  \dots\\
        \widetilde{k}^{i_2j_1}(x) & \widetilde{k}^{i_2j_2}(x) &  \dots\\
        \vdots & \vdots & \ddots
        \end{bmatrix}}_{\widetilde{k}(x)}
        \underbrace{\begin{bmatrix}
            \psi_{j_1}(h)^\top &   &  \\
             & \psi_{j_2}(h)^\top &   \\
             &   & \ddots \\
        \end{bmatrix}}_{\left [\bigoplus_{j\in I_{\text{in}}} \psi_j(h)^\top\right ]}.
    \end{equation}
   Thus, this constraint decomposes into multiple independent constraints:
    \begin{align}
        \widetilde{k}^{ij}(hx) &= \psi_i(h)\widetilde{k}^{ij}(x)\psi_j(h)^\top \quad \forall h \in H, x \in \mathbb{R}^n,
    \end{align}
    where the index $i$ and $j$ denote $i\in I_{\text{out}}$ and $j\in I_{\text{in}}$ respectively.

    The same strategy can be applied to the MLP kernel constraint. Therefore, regardless of chosen input- and output-representation types, the kernel constraint can always be decomposed into multiple kernel constraints using irreducible representations and a change of basis. Consequently, as suggested by~\cite{cesae(2)}, when further analysing the kernel constraints, we assume irreducible representations as input and output representations. To remove clutter, we simply use $k$ and $W$ rather than $\widetilde{k}^{ij}$ and $\widetilde{W}^{ij}$ respectively.
\subsection{Solving Equivariant Kernel Constraint}\label{ap:eq_constraints}

Here we provide a more complete derivation for the kernel constraint for fully equivariant steerable convolutions. Consider the kernel constraint:
    \begin{equation}
        k(x) = \psi_J(h)k(h^{-1}x)\psi_l(h)^\top \quad \forall h \in H, x\in \mathbb{R}^n
    \end{equation}

    We insert a uniform likelihood distribution $\mu(h) = 1 \quad \forall h\in H$ and use a linear projection $\Pi_H(\widehat{K})$, which projects the unconstrained square integrable kernel $\widehat{K}\in L^2(\mathbb{R}^n)$ to an $H$-equivariant kernel $k$:
    \begin{align}        
        k(x) &= \Pi_H(\widehat{K})(x) = \int_{h\in H} \mu(h)\psi_J \widehat{K}(h^{-1}x)\psi_l(g)^\top dg\label{eq:cnn_averaging}\\
        \intertext{Applying columnwise vectorisation using $\kappa(\cdot) = \operatorname{vec}(k(\cdot))$:}
        \kappa(x)&=  \operatorname{vec}(\Pi_H(\widehat{K})(x))= \int_{h\in H} \mu(h) (\psi_l \otimes \psi_J)(h) \widehat{\kappa}(h^{-1}x) dh\label{eq:col_vectorisation}
    \end{align}

    To parameterise the vectorised unconstrained kernel $\widehat{\kappa}:\mathbb{R}^n \to \mathbb{R}^{d_{j}d_{l}}$, we use an $H$-steerable basis $\mathcal{B}=\left\{Y_{j}^k: \mathbb{R}^n \to \mathbb{R}^{d_{{j}}} \mid \psi_{j} \in \widehat{H}, k\right\}$ (Def.~\ref{definition:steerable_basis}), and a weight matrix $W_{j,k}$ of shape $d_{j}d_{l}\times d_{{j}}$:
    \begin{equation}
        \widehat{\kappa(x)} = \sum_{j,k} W_{j,k} Y^k_{j}(x).
    \end{equation}

    Noting that $Y_{j}^k(h\cdot x) = \psi_{j}(h)Y_{j}^k(x)$ (Def.~\ref{definition:steerable_basis}), we re-write Eq.~\ref{eq:col_vectorisation} using this steerable basis parameterisation:

    \begin{align}
        \kappa(x) &=  \sum_{j, k}\left [ \int_{h\in H} \mu(h) \left( \psi_l \otimes \psi_J \right )(h) W_{jk} \psi_{j}(h)^\top dh \right ] Y^k_{j}(x). \label{eq:kernel_constraint_decomp_2}\\
    \end{align}
    We can decompose the tensor product using the \textit{Clebsch-Gordan decomposition} (Thm.~\ref{theorem:clebsch_gordan}):

    \begin{equation}
        (\psi_l \otimes \psi_J)(h) = \sum_{j'}\sum_s^{[j'(Jl)]} \left [ \operatorname{CG}_s^{j'(Jl)} \right]^\top \psi_{j'}(h) \operatorname{CG}_s^{j'(Jl)}.
    \end{equation}
   Using this decomposition, Eq.~\ref{eq:kernel_constraint_decomp_2} becomes:
    \begin{align}
        \kappa(x) &=  \sum_{j, k}\left [ \int_{h\in H} \mu(h) \sum_{j'}\sum_s^{[j'(Jl)]} \left [ \operatorname{CG}_s^{j'(Jl)} \right]^\top \psi_{j'}(h) \operatorname{CG}_s^{j'(Jl)} W_{jk} \psi_{j}(h)^\top dh \right ] Y^k_{j}(x).\\
    \end{align}

    As the summations and the Clebsch-Gordan coefficients do not depend on $h$:
    \begin{align}
        \kappa(x) &=  \sum_{j, k}\sum_{j'}\sum_s^{[j'(Jl)]} \left [ \operatorname{CG}_s^{j'(Jl)} \right]^\top \left [ \int_{h\in H} \mu(h) \psi_{j'}(h) \operatorname{CG}_s^{j'(Jl)} W_{jk} \psi_{j}(h)^\top dh \right ] Y^k_{j}(x).\label{eq:additional_steps}\\
    \end{align}

    Subsequently, we define $W_{j, j', k, s} = \operatorname{vec}\left( \operatorname{CG}_s^{j'(Jl)} W_{jk}\right) \in \mathbb{R}^{d_jd_{j'}}$ and $\operatorname{unvec}(\cdot)$, which reverses the columnwise vectorisation given by $\operatorname{vec}(\cdot)$. Using the property $\operatorname{vec}\left(ABC\right) = \left(C^\top \otimes A\right)\operatorname{vec}\left(B\right)$, we re-write Eq.~\ref{eq:additional_steps} to:    
    \begin{equation}
        \kappa(x) = \sum_{j, k} \sum_{j'}\sum_s^{[j'(Jl)]}  \left [ \operatorname{CG}_s^{j'(Jl)} \right]^\top \operatorname{unvec} \left [ \int_{h\in H} \mu(h) \left(\psi_{j} \otimes \psi_{j'}\right)(h) W_{j, j', k, s} dh  \right] Y_{j}^k(x)\label{eq:cnn_int_to_replace_ap}
    \end{equation}
    Consequently, the integral contains a tensor product and can thus be decomposed through the \textit{Clebsch-Gordan} decomposition (Thm.~\ref{theorem:clebsch_gordan})::
    \begin{align}
        \int_{h\in H} \mu(h) \left(\psi_{j} \otimes \psi_{j'}\right)(h) W_{j, j', k, s} dh &= \int_{h\in H} \mu(h) \left(Q\left(\bigoplus_{i} \bigoplus_{r}^{[i(jj')]} \psi_i(h)\right) Q^\top\right) W_{j, j', k, s} dh \\
        &= Q \bigoplus_{i} \bigoplus_{r}^{[i(jj')]}\left (  \int_{h\in H} \mu(h) \psi_i(h) dh\right)Q^\top W_{j, j', k, s}\label{eq:cnn_sim_mlp}\\
    \end{align}

    If we consider the irreps $\psi_i$ as basis-functions, as described in Theorem~\ref{theorem:peter_weyl}, we can describe the function $\mu(h)$ in terms of its Fourier coefficients (Def.~\ref{definition:fourier}):
    \begin{equation}\label{eq:mu_ft}
        \widehat{\mu}(\psi_i) = \sqrt{d_{{i}}} \int_{h \in H} \mu(h) \mathcal{R}^{-1}_{\psi_i}\left({\psi}_i(h)\right) dh \quad \forall \psi_i \in \widehat{H},
    \end{equation}
    where $d_{{i}}$ is the dimensionality of irrep $\psi_i$, $\mathcal{R}^{-1}_{\psi_i}$ converts an irrep to only its non-redundant columns (Def.~\ref{definition:end_basis}). Therefore, we reformulate the integral in Equation~\ref{eq:cnn_sim_mlp} in terms of its Fourier coefficients:

    \begin{equation}
         \int_{h\in H} \mu(h) \left(\psi_{j} \otimes \psi_{j'}\right)(h) W_{j, j', k, s} dh = Q \bigoplus_{i} \bigoplus_{r}^{[i(jj')]}\left (  \frac{\mathcal{R}_{\psi_i}\left(\widehat{\mu}(\overline{\psi}_i)\right)}{\sqrt{d_{i}}}\right)Q^\top W_{j, j', k, s}
    \end{equation} 

    It should be noted that, due to the uniformity of the likelihood distribution, the Fourier coefficient corresponding to the trivial representation (frequency zero), denoted as $\widehat{\mu}(\psi_{0})$, is equal to $1$. All other Fourier coefficients corresponding to different irreps, and therefore frequencies, are set to zeros. Thus, since the endomorphism basis $\mathcal{C}_{\psi_0} = \left\{\begin{bmatrix}
    1
    \end{bmatrix}\right\}$,
    \begin{align}
        \int_{h\in H} \mu(h) \left(\psi_{j} \otimes \psi_{j'}\right)(h) W_{j, j', k, s} dh &= Q\underbrace{\begin{bmatrix}
    I_{[0(jj')] \times [0(jj')]} & \begin{array}{ccc}
                        0 & \cdots & 0
                     \end{array} \\
    \begin{array}{c}
    0 \\
    \vdots \\
    0
    \end{array}
    &
    \begin{array}{ccc}
    0 & \cdots & 0 \\
    \vdots & \ddots & \vdots \\
    0 & \cdots & 0
    \end{array}
    \end{bmatrix}}_{\left (\bigoplus_{i}\bigoplus_{r}^{[i(jj')]} \frac{\widehat{\mu}(\psi_i)}{\sqrt{d_{i}}} \right)}Q^\top W_{j, j', k, s},\label{eq:mlp_final_2}
    \end{align}
    \begin{wrapfigure}[20]{r}{0.3\textwidth}
    \begin{center} 
    \includegraphics[width=0.7\linewidth, trim={0cm 1.5cm 0cm 0cm}, clip]{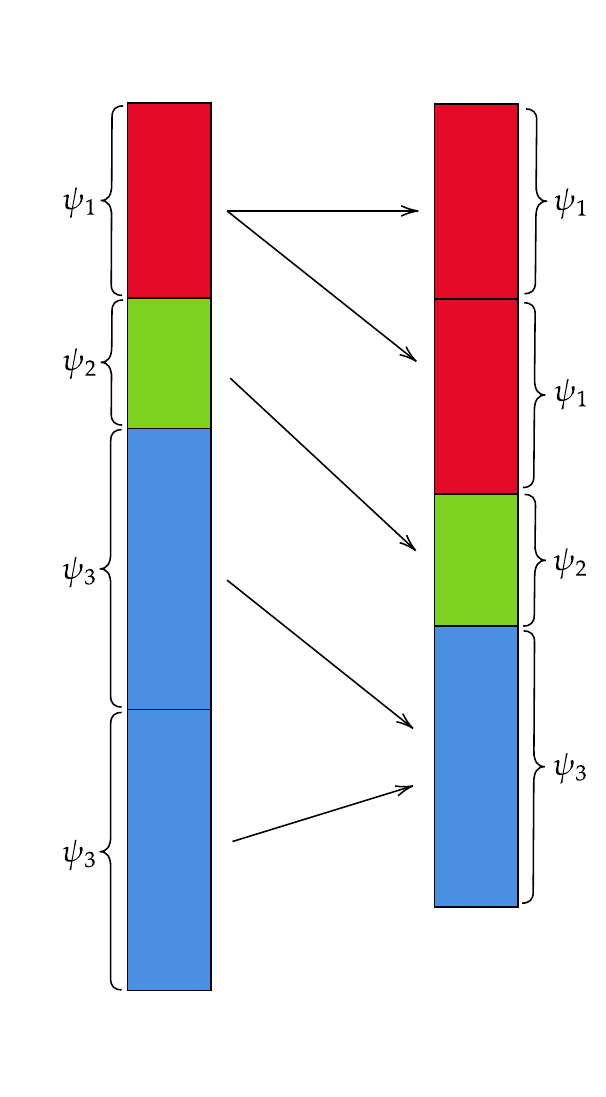}
    \end{center}
    \caption{Mapping between different input and output representations. Representations are colour-coded. A mapping only (and always) exists between equal irreps.}
    \label{fig:mapping}
    \end{wrapfigure}
    where $[0(jj')]$ is the multiplicity of the trivial representation in the decomposition of $\psi_j \otimes \psi_{j'}$. The resulting sparse block-diagonal matrix selects the relevant columns from the change of basis $Q$. Equivalently, using a sum, we have:
    \begin{align}
        \int_{h\in H} \mu(h) \left(\psi_{j} \otimes \psi_{j'}\right)(h) W_{j, j' k, s} dh &= \sum_r^{[0(jj')]}Q_rQ_r^\top W_{j, k, s, r},\label{eq:mlp_final_3}
    \end{align}
        
    where $Q_r$ is the $r$-th column of $Q$. To reduce the number of parameters, we absorb $Q_r^\top\in \mathbb{R}^{1\times d_{l}d_{j}}$ into the weights $W_{j, k, s, r}$, resulting in a single weight $W_{j, k, s, r} \in \mathbb{R}$.

    When the representations, $\psi_j$ and $\psi_{j'}$, are identical, the multiplicity $[0(jj')]$ of the trivial representation in the Clebsch-Gordan decomposition of $\psi_j \otimes \psi_{j'}$ is equal to the size of the endomorphism basis $|\operatorname{End}_{\psi_{j}}|=c_{\psi_{j}}$. In this case, it turns out that the $Q_r$ vectors span the same space as the vectorised endomorphism basis $\mathcal{C}_{\psi_{j}} = \left \{ c^{j}_r \in \mathbb{R}^{d_{j} \times d_{j}}\right \}$. Thus, Eq~\ref{eq:mlp_final_3} can be re-written as follows:
\begin{equation}
    \int_{h\in H} \mu(h) \left(\psi_{j} \otimes \psi_{j'}\right)(h) W_{j, k, s, r} dh =  \sum_r^{[0(jj')]} \operatorname{vec}(c^J_r) W_{j, k, s, r}\label{eq:cnn_int}
\end{equation}
 Conversely, if $\psi_j \neq \psi_{j'}$, the Clebsch-Gordan decomposition lacks a trivial representation and, therefore, multiplicity $[0(lJ)]=0$. This leads to a null-map as all emerging Fourier coefficients are zero. Consequently, the weight matrix is also projected to zeros, rendering it a null-map and making these weights redundant. See Figure~\ref{fig:mapping} for a visualisation.

    Replacing the integral in Eq.~\ref{eq:cnn_int_to_replace_ap} with Eq.~\ref{eq:cnn_int}, and replacing $j'$ with $j$ gives:
    \begin{align}\label{eq:cnn_final_yes}
        \kappa(x) = \Pi_H(\widehat{K})(x) &= \sum_{j, k} \sum_s^{[j'(Jl)]}  \left [ \operatorname{CG}_s^{j(Jl)} \right]^\top \operatorname{unvec} \left [ \sum_r^{0[jj]} \operatorname{vec}(c_r^j) W_{j, k, s, r}  \right] Y_{j}^k(x)\\
        &= \sum_{j, k} \sum_s^{[j'(Jl)]}  \left [ \operatorname{CG}_s^{j(Jl)} \right]^\top  \left [ \sum_r^{0[jj]} c_r^j W_{j, k, s, r}  \right] Y_{j}^k(x) \label{eq:cnn_final_reduced}
    \end{align}
    Thus, removing the summation over $j'$, and therefore all linear maps when $\psi_j \neq \psi_{j'}$. This equation is analogous to Eq. 4 in~\cite{cesa2022ae(n)}, where the summations have been removed in favour of additional indexing.

\subsection{Alignment Problem}\label{ap:alignment_loss}
While the Fourier parameterisation enables the reconstruction of the likelihood distribution $\lambda$, there is no guarantee that the weights are properly aligned with the learnt likelihood distribution. Recall the learnable projection integral for the partially equivariant SCNN from Eq.~\ref{eq:projection_1}, where we insert the constrained weight matrices $W_{j, j', k, s}$.:
        \begin{equation}
            \int\! \lambda(h)\! \left(\psi_{j}\! \otimes\! \psi_{j'}\right)(h) dh W_{j, j', k, s} =\! Q \sum_{i}\! \sum_{r}^{[i(jj')]}\frac{\mathcal{R}_{\psi_i}\left(\widehat{\lambda}(\overline{\psi_i})\right)}{\sqrt{d_{i}}}Q^\top W_{j, j', k, s}.
        \end{equation}
        Note here that the change of basis $Q^\top$ is not absorbed in the weights $W_{j, j', k, s}$. To illustrate the problem, we perform the projection given by $\xi_{\mathscr{R}_h \lambda}(W_{j, j', k, s}')$, where $W_{j, j', k, s}' = (\psi_j \otimes \psi_{j'})(h)W_{j, j', k, s}$. Here, the likelihood distribution undergoes a transformation through the right-action of an element $h\in H$ using the right-regular representation $\mathscr{R}_h$ (Def.~\ref{definition:left_reg_rep}).

        \begin{align}
             \int\! \lambda(h)\! \left(\psi_{j}\! \otimes\! \psi_{j'}\right)(h) dh W_{j, j', k, s}' &=Q\left ( \bigoplus_{i}\bigoplus_{r}^{[i(jj')]}  \frac{\mathcal{R}_{\psi_i}\left(\widehat{\left [\mathscr{R}_h \cdot \lambda\right ]}(\overline{\psi_i})\right)}{\sqrt{d_{i}}} \right ) Q^\top \widehat{W}'\\
            &= Q\left ( \bigoplus_{i}\bigoplus_{r}^{[i(jj')]}  \frac{\mathcal{R}_{\psi_i}\left(\widehat{\lambda}(\overline{\psi_i})\right)\psi_i(h)^\top}{\sqrt{d_{i}}} \right ) Q^\top W_{j, j', k, s}'\\
            \intertext{Owing to the block-diagonal structure, the equation can be formulated as:}
             &= Q\left ( \bigoplus_{i}\bigoplus_{r}^{[i(jj')]}  \frac{\mathcal{R}_{\psi_i}\left(\widehat{\lambda}(\overline{\psi_i})\right)}{\sqrt{d_{i}}} \right ) Q^\top Q\left(\bigoplus_i \bigoplus_r^{[i(jj')]} \psi_i(h)^\top\right)Q^\top W_{j, j', k, s}'\\
            \intertext{Recall that the second direct-sum over $\psi_i$ is simply equal to the Clebsch-Gordan decomposition of the tensor product $(\psi_j \otimes \psi_{j'})(h)$. Therefore:}
             &= Q\left ( \bigoplus_{i}\bigoplus_{r}^{[i(jj')]}  \frac{\mathcal{R}_{\psi_i}\left(\widehat{\lambda}(\overline{\psi_i})\right)}{\sqrt{d_{i}}} \right ) Q^\top (\psi_j \otimes \psi_{j'})(h)^\top  \widehat{W}'\\
            \intertext{since $ W_{j, j', k, s}' = (\psi_j \otimes \psi_{j'})(h) W_{j, j', k, s}$}
            \operatorname{vec}\left(\xi_{\mathscr{R}_h  \lambda }\left(\widehat{W}'\right) \right) &= Q\left ( \bigoplus_{i}\bigoplus_{r}^{[i(jj')]}  \frac{\mathcal{R}_{\psi_i}\left(\widehat{\lambda}(\overline{\psi_i})\right)}{\sqrt{d_{i}}} \right ) Q^\top (\psi_j \otimes \psi_{j'})(h)^\top(\psi_j \otimes \psi_{j'})(h)\widehat{W}\\
            &= Q\left ( \bigoplus_{i}\bigoplus_{r}^{[i(jj')]}  \frac{\mathcal{R}_{\psi_i}\left(\widehat{\lambda}(\overline{\psi_i})\right)}{\sqrt{d_{i}}} \right ) Q^\top\widehat{W}\\
            &=\int\! \lambda(h)\! \left(\psi_{j}\! \otimes\! \psi_{j'}\right)(h) dh W_{j, j', k, s}
        \end{align}
        As a result, performing these transformations on the likelihood $\lambda$ and the weights $W_{j, j', k, s}$ has no effect on the resulting constrained weights $W$. Consequently, the model is able to freely shift the learnt likelihood distribution by shifting the weights accordingly, thus diminishing the interpretability of the learnt likelihood distribution.

\subsection{KL-Divergence on Fourier domain}\label{ap:kl_divergence}
In equivariant neural networks, global equivariance is obtained by ensuring that each individual operation is equivariant. Once a particular operation breaks the equivariance, this equivariance cannot be regained in subsequent operations. Since our method permits an independent parameterisation of the likelihood distributions of subsequent layers, the resulting likelihood distributions do not necessarily reflect this permanent loss of equivariance. For instance, when the equivariance with respect to element $h\in H$ is reduced in layer $n$, layer $m>n$ is able to model a likelihood distribution exhibiting an increased level of equivariance for element $h$, which is not representative of the actual state of equivariance in layer $m$.

To achieve representative likelihood distributions, we employ Kullback-Leibler divergence (KL divergence) \cite{kullback1951information}. KL divergence is a non-negative statistical distance measure between two \textit{probability} distributions. It is commonly used as a regularisation term in deep learning, most notably for variational autoencoders \cite{kingma2013auto}. Given a distribution $P$ along with a reference distribution $Q$, KL-divergence is defined as:        

\begin{align}
    D_{KL}(P \ || \ Q) &= \int_{-\infty}^{\infty} p(x) \log \left ( \frac{p(x)}{q(x)} \right) dx. \label{kl_def}
\end{align}
Or, in a discretised setting:
\begin{align}
    D_{KL}(P \ || \ Q) &= \sum_{x \in X } p(x) \log \left ( \frac{p(x)}{q(x)} \right).\label{eq:kl_def}
\end{align}
Note that, due to the weighting of the logarithmic term by $p(x)$, the KL divergence is not a symmetrical measure, thus $D_{KL}(P \ || \ Q) \neq D_{KL}(Q \ || \ P)$; a larger value of $p(x)$ at a given point results in a stronger weight of the term $\log \left ( \frac{p(x)}{q(x)} \right)$. 
Therefore, taking the equivariance of layer $n$ as distribution $Q$ and the equivariance from layer $n+1$ as distribution $P$ aligns with our objectives. This approach assigns a higher weight when layer $n+1$ models a high degree of equivariance, particularly when it models the reacquisition of previously lost equivariance.

The definition of Eq.~\ref{eq:kl_def} requires a sampled likelihood distribution. As a result, the accuracy of the KL-divergence is dependant on the number and the uniformity of the samples used by the inverse Fourier transform. Therefore, we show that KL-divergence over a reference distribution $\lambda_0$ and a distribution $\lambda_1$ can be computed directly on the Fourier coefficients by first re-writing Eq.~\ref{eq:kl_def}:
\begin{align}
    D_{KL}(\lambda_1 \ || \ \lambda_0) &= \sum_{x \in X } \lambda_1(x) \log  \lambda_1(x) - \sum_{x \in X }\lambda_1(x)  \log \lambda_0(x)\\
    & = \bm{\lambda_1}^\top  \log\bm{\lambda_1} - \bm{\lambda_1}^\top  \log\bm{\lambda_0},\\
    \intertext{Where $\bm{\lambda_1}$ and $\bm{\lambda_0}$ are the vectors stacking $\lambda_1(x)$ and $\lambda_0(x)$. Since the Fourier transform $\mathcal{F}(\bm{\lambda_1}) = \bm{\widehat{\lambda_1}}$ is an orthogonal operation:}
    D_{KL}(\lambda_1 \ || \ \lambda_0) &= \bm{\widehat{\lambda_1}}^\top  \bm{ \widehat{\log \lambda_1}} - \bm{\widehat{\lambda_1}}^\top  \bm{\widehat{\log \lambda_0}}\label{eq:kl_derive_1}
\end{align}
Here $\bm{\widehat{\log{\lambda_1}}}$ and $\bm{\widehat{\log{\lambda_1}}}$ are the Fourier coefficients of the logarithms of the respective $\lambda_1$ and $\lambda_0$ distributions. Although the Fourier coefficients $ \bm{\widehat{\lambda_1}}$ and $ \bm{\widehat{\lambda_0}}$ are directly available, since these are the normalised Fourier coefficients obtained with the normalisation in Section~\ref{sec:ensure_eq}, their logarithms are not directly available without first computing them. Fortunately, through the normalisation approach in Section~\ref{sec:ensure_eq}, the unnormalised likelihood distribution (or log likelihood) relates to the logarithm of the normalised distribution:
\begin{align}
    \log \lambda(x) &= \log\left( \frac{e^{\lambda'(x) - \max(\lambda')}}{ z}\right)\\
    & = \lambda'(x) - \operatorname{max}(\lambda') - \log z.
\end{align}

Again, using the orthogonality of the Fourier transform, we can simply add these additional terms to Eq.~\ref{eq:kl_derive_1} and replace $\widehat{\bm{\log \lambda_1 } }$ and $\widehat{\bm{\log \lambda_0}}$ with $\bm{\widehat{\lambda'_1}}$ and $\bm{\widehat{\lambda_0'}}$. As a result:
\begin{align}
    D_{KL}(\lambda_1 \ || \ \lambda_0) & =\bm{\widehat{\lambda_1}}^\top  \bm{\widehat{\lambda_1'}} - \operatorname{max}(\lambda_1') - \log z_1 - \bm{\widehat{\lambda_1}}^\top  \bm{\widehat{\lambda_0'}} + \operatorname{max}(\lambda_0') + \log z_0 \label{eq:final_kl_ap}
\end{align}

Since we now use the Fourier coefficients directly, the entire distribution is used to compute the KL-divergence, rather than only a finite number of sampled points. This should result in a more stable computation. However, this computation is still dependent on the normalisation process, which does involve sampling of the distribution. Consequently, computing the KL-divergence as described here can still result in minor inaccuracies.


\section{Additional Theoretical Details}

\subsection{Comparison with Alternative Partially Equivariant Approaches}\label{sec:comp_approaches}

In this section, we provide a more in depth comparison with some related approaches to learn approximate equivariance.
A first important difference with previous works is that we only support equivariance breaking over compact groups and, therefore, cannot break translation equivariance.
Hence, we structure this section in three parts: first we compare with methods considering group convolution, but restrict our consideration to compact groups (in this case, a GCNN is equivalent to an equivariant MLP in our framework); then, we compare specifically with RPP for the case of generic equivariant MLPs, since this method is closely related to ours; finally, we compare with methods that support steerable CNNs (where, however, we only break equivariance to the compact rotation and reflection subgroup).

\paragraph{Compact group GCNNs}
We first consider the special case of GCNNs. 
Recall that group convolution can be expressed as a linear operator with kernel $k: G \times G \to \mathbb{R}$:
\begin{align}
    \left[k \star f \right](h) := \int_{g\in G} k(h, g) f(g) dg
\end{align}
which is equivariant to $G$ iff the kernel can be written as a 1-argument kernel $k(h, g) = \hat{k}(h^{-1}g)$.
At a high level, besides the underlying implementations, both \citet{romero2022learning} and \citet{wangApprox} use the equivariant kernel $\hat{k}(h^{-1}g)$ weighted by a learnable component $w(g)$, i.e. $k(h, g) = \hat{k}(h^{-1}g) w(g)$ (in a more or less flexible way; e.g. in \cite{romero2022learning}, $w(g)$ is the learned probability distribution over $G$).
Similarly, \citet{van2022relaxing} break equivariance by using \textit{stationary filters}, i.e. by replacing the 1-argument kernel $\hat{k}(h^{-1}g)$ with a 2-arguments kernel $k(h^{-1}g, g)$ conditioned on the input location $g$ but initialised to be dependent only on the first argument $h^{-1}g$. 
While more flexible than the previous two approaches, stationary filters still follows the same principle, i.e. \textit{breaking equivariance by learning a dependency on the absolute input location} $g$.

Instead, we follow a different principle.
Ignoring the bandlimiting, our approach learns the unconstrained kernel $k: G \times G \to \mathbb{R}$ and a likelihood distribution $\lambda$ over $G$ and performs the projection to the equivariant subspace of 1-argument kernels via group averaging (Reynold's operator), weighted by $\lambda$, that is:
\begin{align}
    \bar{k}(h, g) = \int_e k(eh, eg) \lambda(e) de \ .
\end{align}
One can verify that $\bar{k}(h, g)$ is equivalent to a one argument kernel $\hat{k}(h^{-1}g)$ when $\lambda$ is a uniform distribution over $G$, i.e. it is $G$-equivariant.
In other words, our method learns the degree of weight sharing to apply on a learnable unconstrained kernel.

Additionally, we note that in \cite{romero2022learning}, a smaller support for the learnt probability distribution does not change the number and expressiveness of filters, but only the number of input locations in their field of view, thereby reducing the computational cost.
In our case, instead, the computational cost is constant, but the reduced support leads to more learnable weights used to parameterise the filters.
For example, if $\lambda$ is a uniform distribution over the subgroup $H < G$ and both $G$ and $H$ are finite groups, then the projected operator $\bar{k}$ becomes equivalent to $\frac{|G|}{|H|} \times \frac{|G|}{|H|}$ independent $H$-convolutions - with a total complexity equivalent to a single $G$-convolution.
Conversely, the method proposed by \citet{romero2022learning} would result in a single $H$-convolution.

\paragraph{Residual Pathway Priors}
RPP \citep{finzi2021residual} is more closely related to our approach, for the case of a generic compact-group equivariant MLP (not necessarily using the regular representation and, therefore, not necessarily equivalent to a GCNN).
Indeed, let $W$ be the weight matrix of a linear layer and $G$ the equivariance group of interest, such that an equivariant layer should satisfy $\rho_{out}(g) W = W \rho_{in}(g)$, that is $\operatorname{vec}(W) = (\rho_{in} \otimes \rho_{out})(g) \operatorname{vec}(W)$.
An RPP prior on the vectorised weights $\operatorname{vec}(W)$ has the form $\operatorname{vec}(W) \sim \mathcal{N}(0, (\sigma_a^2 + \sigma_b^2) QQ^T + \sigma_b^2 PP^T)$ as in Sec.~4 of \citep{finzi2021residual}, with $Q$ a basis for the $G$ equivariant subspace and $P$ a basis for its complement.
This formulation is comparable to our framework with the following \textbf{fixed likelihood function}:
\begin{align}
    \lambda(g) \propto \frac{1}{\sigma_b^2}\delta_e(g) + \left(\frac{1}{\sigma_a^2 + \sigma_b^2} - \frac{1}{\sigma_b^2}\right)
\end{align}
where $\delta_e(g)$ is a Dirac delta function centred at the identity $e\in G$, combined with an L2 weight decay regularisation term on the final projected weights; indeed, plugging this distribution in the averaging operator leads to:
\begin{align}
    \Pi_\lambda(\operatorname{vec}(W)) &= \int_G (\rho_{in} \otimes \rho_{out})(g) \operatorname{vec}(W) \lambda(g) dg \\
                                   &\propto \frac{1}{\sigma_b^2} \operatorname{vec}(W) + \left(\frac{1}{\sigma_a^2 + \sigma_b^2} - \frac{1}{\sigma_b^2}\right) \Pi_G(\operatorname{vec}(W))  \\
                                   &= \frac{1}{\sigma_b^2} P \alpha + \frac{1}{\sigma_a^2 + \sigma_b^2} Q \beta
\end{align}
for some weight vector $\alpha, \beta$.
An L2 weight decay term on $\Pi_\lambda(\operatorname{vec}(W))$ would to the same regularisation used in RPP.
However, in our experiments we typically used a weight decay regularisation directly on $\operatorname{vec}(W)$ rather than on $\Pi_\lambda(\operatorname{vec}(W))$; we did not explore the effect of this choice but this connection with RPP might be an interesting direction to further improve our method.

\paragraph{Steerable CNNs}
The RSteer method from \cite{wangApprox} can be written as (from their open source code):
\begin{align}
    \left[ k \star f \right] (x) = \int_y dy \left[\sum_b \kappa_b(y-x) w_b(y-x) \right] f(y)
\end{align}
where $\{ \kappa_b \}_b$ is the basis for equivariant steerable kernels.
Essentially, rather than linearly combining the basis elements with a set of learnable weights, the authors use scalar weight functions $w_b: \mathbb{R}^n \to \mathbb{R}$; since these scalar functions are not steerable - $w_b(y-x) \neq w_b(g(y-x))$ - this causes the equivariance breaking.
This can be thought as tweaking the underlying steerable basis, breaking its steerable property in a learnable way.
We do not currently see any direct connection with our method.

\subsection{Monotonically Decreasing Equivariance}\label{sec:monotonically}
In Section~\ref{sec:ensure_eq} we discussed employing KL-divergence as a soft constraint to ensure that the degree of equivariance is monotonically decreasing in the network. This choice was justified with the fact that once equivariance has been broken it cannot be recovered in a subsequent layer, and thus to accurately represent this a layer should never model a likelihood distribution that suggests it has re-gained equivariance that a previous layer has lost. However, in their experiments with Partial-GCNN, \citet{romero2022learning} find that a higher degree of equivariance in the last layer compared to previous layers often results in superior performance compared to when enforcing a monotonically decreasing equivariance by applying a ReLU activation over the differences in subsequent likelihood distributions. 

In our experiments, we observed that using such a regularisation generally resulted in improved performance and stability (Apx.~\ref{sec:result_kl}). Furthermore, with or without the employment of our regularisation, we have not necessarily observed that our models consistently tended to become significantly more uniformly equivariant in the final layer. However, this could be due to the employment of KL-divergence in favour of their approach, and the fact that we detach the Fourier coefficients of layer $n$ when computing the KL-divergence between layer $n$ and $n+1$, ensuring that backpropagation can only push layer $n+1$ to be a subset of layer $n$ while being unable to push layer $n$ to be close to layer $n+1$. This prevents the regularisation from overly reducing equivariance in earlier layers. Finally, it is important to note that since our constraint is a soft-constraint, our approach can still model such an increase in equivariance in later layers if required, as long as the KL-divergence is not weighted too strongly in the training objective.

\section{Additional Implementation Details}\label{ap:implement_details}
This section covers additional details regarding implementations of the theoretical procedures described in the paper. All code implementations are performed as an extension of the PyTorch \hyperlink{https://github.com/QUVA-Lab/escnn}{\texttt{escnn}} library~\cite{cesa2022ae(n)}.

\subsection{Non-linearites}\label{sec:nonlinearites} A GCNN layer maps an input in $\mathbb{R}^n \times GL(\mathbb{R}^{|H|})$ or $\mathbb{R}^n$ to regular features $GL(\mathbb{R}^{|H|})$. This allows them to use regular pointwise non-linearities acting on the scalar values without breaking equivariance. Instead, a steerable layer might use other feature types, some of which do not allow the use of these pointwise nonlinearities without breaking equivariance. 
Since non-linearities are essential to learning complex functions, alternative methods are required.

\paragraph{Fourier-based} Since an SCNN often uses irrep fields as intermediate features for infinitely large groups $H$, which describe the signal over the group $H$ in terms of its Fourier coefficients, one method to perform a non-linearity is to sample the Fourier coefficients for $N$ points through the \textit{inverse Fourier transform} (Def.~\ref{definition:inv_fourier})~\cite{NEURIPS2021_4bfbd52f,poulenard2021functional,cesa2022ae(n)}. The resulting features are akin to a regular representation $\rho: G \to GL(\mathbb{R}^N)$, although limited to $N\leq |H|$ permutation axes. These features can then be transformed using regular pointwise non-linearities, after which the Fourier transform is performed again to obtain the original feature field. Consider an irrep feature field $f_\text{in}: \mathbb{R}^n \to \mathbb{R}^c$. The Fourier-based non-linearity is defined as follows:
\begin{equation}
f_\text{out}(\bm{x}) = \mathcal{F}\left(\gamma\left(\mathcal{F}^{-1}\left(f_\text{in}\left(\bm{x}\right)\right)\right)\right) \quad \forall \bm{x} \in \mathbb{R}^n,
\end{equation}\label{eq:fourier_nonlinearity}
where $\gamma$ is some pointwise non-linearity function.

The Fourier transform and the inverse Fourier transform can be described using the $\operatorname{FT}\in \mathbb{R}^{c \times N}$ and $\operatorname{IFT}\in \mathbb{R}^{N \times c}$ matrices, respectively, where $\operatorname{IFT}=\operatorname{FT}^\dagger$\footnote{Here $\operatorname{FT}^\dagger$ is the pseudo-inverse of $\operatorname{FT}$}. Using these matrices, the Fourier-based activation function can be efficiently implemented as follows:
\begin{equation}
    f_\text{out}(\bm{x}) = \operatorname{FT}\gamma\left(\operatorname{IFT} f_{\text{in}}(\bm{x})\right).
\end{equation}

It is important to note that, depending on the chosen value for $N$, the method used for sampling the points, the group $H$ and the non-linearity $\gamma$, this method is only \textit{approximately} equivariant. This is due to the fact that activation functions like ReLU introduce sharp non-linearities in the signal, which result in high frequencies in the Fourier transform. In order to optimise equivariance, it is important that $N\ge c$, where $c$ is the size of the combined feature size of the irrep field, and that the $N$ points are sampled uniformly over the space of the group $H$. The latter can be complicated for certain groups, particularly 3D rotation groups such as $SO(3)$ and $O(3)$ due to the continuous and non-Euclidean space. For these groups, there are no methods to provide exact solutions for any arbitrary $N$, although methods pertaining to the Thomson problem are capable of providing approximately uniform sampling for such groups. 

Since it is expensive to sample a sufficiently large number of points over large three-dimensional groups, such as $O(3)$ and $SO(3)$, it is possible to use features of the quotient space $Q=H/K$ rather than the full space $H$ \citet{cesa2022ae(n),bekkers2023fast}, where $K\leq H$ is a subgroup of $H$. Although the exact definition of a quotient space is outside the scope of this paper, the quotient space $Q=H/K$ can be interpreted as the space of $H$ with the subspace $K$ factored out. We refer to~\cite{cesa2022ae(n)} for more details. The Fourier-based non-linearity then only applies the non-linearity only over this quotient space $Q=H/K$, reducing computational complexity.

\paragraph{Norm-based} Any non-linearity acting solely on the norm $||\cdot ||_n$ on features transforming through unitary representations of $H$ is $H$-equivariant. \textit{Norm-ReLUs}, used in~\cite{worrall2017harmonic,weiler2018learning}, are an example of such non-linearities. Here, the norms of the features $f(\bm{x})$ are transformed as follows: $||f_\text{out}(\bm{x})||_2 = \operatorname{ReLU}(||f_\text{in}(\bm{x})||_2)\quad \forall \bm{x} \in \mathbb{R}^n$. Alternatively,~\citet{weiler20183d} introduced \textit{Gated non-linearities}, which scale the norms of the feature vectors $f(\bm{x})$ using a learnable scalar field $s: \mathbb{R}^n \to \mathbb{R}$ through a sigmoid gate: $\frac{1}{1+e^{-s(\bm{x})}}$, resulting in: $||f_\text{out}(\bm{x})||_2 = (||f_\text{in}(\bm{x})||_2)\frac{1}{1+e^{-s(\bm{x})}}$. Although these types of non-linearities maintain perfect equivariance,~\citet{cesae(2)} have shown that they are generally outperformed by the Fourier-based non-linearities.

\subsection{Sharing Equivariance}\label{section:sharing_equivariance}
Algorithm~\ref{algo:algo} enables the learning of partially equivariant mappings in SCNNs, which can be applied to obtain corresponding weight projections that are also partially equivariant. Traditional CNNs consist of a hierarchical sequence of convolution layers, each specialised in modelling features at a unique scale. Therefore, it is straightforward for such CNNs to learn the degree of equivariance individually, resulting in different degrees of equivariance for each scale. Using the provides framework, it is also possible to create equivariant MLPs (E-MLPs) by taking a steerable basis $\mathcal{B}=\left\{Y_j^k: \mathbb{R}\to1\right\}^{k=0}_{j=0}$ and consider a space $X=\left\{0\right\}$. However, such a layered specialisation in scale does not translate well to MLPs. As such, it is perhaps more prudent to have a single, shared degree of equivariance across all layers in an E-MLP.

To implement this shared degree of equivariance, each layer receive a unique \texttt{layer\_id} by default, which can be overridden by the user at initialisation of the layer. If two, or more, layers receive the same id, then the parameters parameterising the degree of equivariance are shared between these layers, resulting in the sharing of the degree of equivariance between these layers.

This notion of shared equivariance also finds potential utility in other architectures or architectural elements, such as residual connections and auto-encoders. For instance, in a residual network, a skip-connection could share its degree of equivariance with the layers it bypasses. Similarly, in auto-encoders, it could be advantageous for the first layer of the encoder to share its degree of equivariance with the last layer of the decoder, and so forth.

\subsection{Interpretability losses}\label{section:implement_interpretability_loss} This section discusses the implementation details of the losses introduced in Section~\ref{sec:ensure_eq}, which are aimed at improving the interpretability of the likelihoods for the probabilistic approach. The following two paragraphs cover KL-divergence and \textit{alignment loss} respectively.
\paragraph{KL-divergence} The KL-divergence from Eq.~\ref{eq:final_kl} is defined over the likelihood distribution of two layers. The intended use is that the two layers are consecutive, with the likelihood of the first layer acting as reference. To account for varying architectures, for example a layer which is preceded by two different layers, the user can provide a set $\tau$, containing pairs of \texttt{layer\_id}'s. Subsequently, the total KL-divergence loss is computed as the mean average KL-divergence between the pairs:
\begin{align}
    \mathcal{L}_\text{KL} &= \frac{1}{|\tau|} \sum_{(m, n) \in \tau} D_{\text{KL}}(\lambda_{m}\mid \mid \lambda_{n})\\
    \intertext{from Eq.~\ref{eq:final_kl}}
    & = \frac{1}{|\tau|} \sum_{(m, n) \in \tau} \bm{\widehat{\lambda_m}}^\top  \bm{\widehat{\lambda_m'}} - \operatorname{max}(\lambda'_m) - \log z_m - \bm{\widehat{\lambda_m}}^\top  \bm{\widehat{\lambda'_n}} + \operatorname{max}(\lambda'_n) + \log z_n.
\end{align}
Where $\lambda_m'$ is the unnormalised likelihood distribution of the layer with \texttt{layer\_id} $m$, with corresponding Fourier coefficients $\bm{\widehat{\lambda_m'}}$, Fourier coefficients of the normalised distribution $\bm{\widehat{\lambda_m}}$ and the logarithmic term $\log z_m$. Since the aim of the KL-divergence is to incentivise the likelihood of layer $m$ to be a subset of the reference likelihood of layer $n$, the terms corresponding to layer $n$ are treated as constants and therefore do not contribute to the KL-divergence loss. As a result, the model can only alter the likelihood of layer $m$ to be closer to the likelihood of layer $n$, and not the other way around. Finally, we allow the user to pair a \texttt{layer\_id} $m$ with $n=\operatorname{None}$, rather than an existing \texttt{layer\_id} $n$. In this case $\lambda_n'$ (and by extension $\lambda_n$) is taken to be the uniform likelihood distribution and subsequently results in the likelihood of layer $m$ being pushed towards a uniform and therefore equivariant projection.

\paragraph{Alignment Loss}
The alignment error $D_{\text{align}}$ from Eq. \ref{eq:align_loss} prevents the misalignment by ascribing a nonzero error when the likelihood $\lambda(e)$ for the identity element $e \in H$ is not the maximum likelihood appearing in the entire likelihood distribution. To compute this error, we use the points sampled by the inverse Fourier transform used for the normalisation. As the sampling is a (near) uniform sampling over the group $H$, using these points should ensure that the likelihood is shifted in approximately the correct direction. However, for continuous rotational groups, particularly continuous rotational 3D groups, more precision may be required. Therefore, for such continuous groups, $100$ additional points are randomly sampled around the identity element $e$. 

For 2D rotation groups, we sample the rotation angles of these points from a normal distribution $\mathcal{N}(0, 0.2)$, where the standard deviation is $0.2$ radians. In contrast, for 3D rotation groups, the sampling procedure is more involved. We start with the quaternion representation of the identity element, represented as $\mathbf{q} = [1, 0, 0, 0]$. Random perturbations are then added to each component of this quaternion, drawn from a normal distribution $\mathcal{N}(0, 0.1)$. Formally, if $\mathbf{d} = [d_1, d_2, d_3, d_4]$ are the random deviations, the perturbed quaternion is given by:
\[
\mathbf{q}_{\text{perturbed}} = \mathbf{q} + \mathbf{d} = [1 + d_1, 0 + d_2, 0 + d_3, 0 + d_4]
\]
Subsequently, these randomly sampled quaternions are normalised to unit quaternions. In this manner, the uniformly sampled points serve to shift the likelihood in case the likelihood is strongly misaligned, and the additional points aid in shifting the likelihood distribution when it is nearly aligned.

Similarly to KL-divergence, the total alignment loss $\mathcal{L}_{\text{align}}$ is defined as the mean average alignment error of all likelihood distributions $\lambda_{i} \in \Lambda$, where $\Lambda$ is the set of learnt likelihood distributions.
\begin{align}
    \mathcal{L}_{\text{align}} = \frac{1}{|\Lambda|} \sum_{\lambda_i \in \Lambda} D_{\text{align}}(\lambda_i)
\end{align}

\section{Experimental Details}\label{ap:experiment_details}
This section contains an outline of our evaluation methodology used to obtain our results. First, in Section \ref{sec:datasets} we outline the three main classes of datasets, along with their variations. Section \ref{sec:model_architectures} contains descriptions of the model architectures used for the varying experimental setups. 
\subsection{Datasets}\label{sec:datasets}
    In this paper we evaluate our learnable partially equivariant SCNNs and E-MLPs on various datasets to assess the behaviour of our approaches, both in terms of task-specific performance and in terms of the learnt degrees of equivariance. 
    \subsubsection{Vectors}\label{sec:vectors_dataset}
    In our additional results we also evaluate our approach on equivariant MLPs. This can be achieved by taking an steerable basis $\mathcal{B}=\left\{Y_j^k: \mathbb{R}\to1\right\}^{k=0}_{j=0}$ and consider a space $X=\left\{0\right\}$.
    
    To assess the performance of learnable partial equivariance in E-MLPs, we employ a simple \texttt{Vectors} dataset. This dataset consists of $1000$ two-dimensional vectors. Each vector is generated with an independently and uniformly sampled angle between $0$ and $2\pi$ radians that represents the angle between the vector and the positive y-axis. Moreover, the $\ell^2$ norms of the vectors are between $0$ and $\sqrt{2}$. Importantly, the dataset is designed to be flexible in its training targets: a model can be trained to predict either the cosine of the angle or the norm of the vectors as binary regression targets, or it can be trained to predict both the cosine of the angle and the norm simultaneously as two regression targets. 

     While the prediction of the norm of a vector is a clear $O(2)$-invariant task, predicting the angle is inherently non-invariant. Specifically, when the cosine of the angle is considered, vectors that are opposite to each other produce the most significant difference in the regression target. This implies that an appropriate equivariance likelihood distribution would predominantly exhibit a frequency signal of one, reaching its minimum likelihood at rotations of $\pi$. Given the contrasting nature of these two tasks, one invariant and one non-invariant, this dataset serves as a valuable resource to juxtapose the performance of various model architectures.

    \subsubsection{DDMNIST}\label{ap:ddmnist}
    While the \texttt{Vectors} dataset focusses on E-MLPs, the main motivation for incorporating a learnable degree of equivariance is the varying degrees of scale in CNNs acting on planar or volumetric images. For planar images, we build the \texttt{DDMNIST} dataset, which is an adaptation of the popular MNIST dataset \cite{mnist}. Whereas the original MNIST dataset contains $60,000$ $28\times28$ training images of single handwritten digits numbered $0$ through $9$, our \texttt{DDMNIST} dataset contains $10,000$ $56\times 56$ images containing double-digit numbers ranging from $0$ through $99$. We created these images by sampling $100$ images of the two required digits from the original MNIST dataset for each of the $100$ double-digit numbers. Before combining the images of the two digits, they are both independently and randomly transformed by a random element of one of the pre-determined groups in Table~\ref{tab:double_mnist_groups} without altering the label. To keep interpolation artefacts consistent, before augmentation each digit is rotated by a random angle $\theta \in [0, 2\pi) $ and consequently rotated by $-\theta$. This introduces interpolation artefacts in each image, even if they are not transformed at all. Afterwards, the double-digit images are concatenated horizontally and padded so that the resulting image has a size of $56\times 56$ pixels.

    \begin{table}[h!]
    \centering
    \begin{tabular}{llll}
    \toprule
    \textbf{Group} & \textbf{Group Type} & \textbf{Symmetries} & \textbf{\# of elements} \\ \midrule\midrule
    $C_1$ & Cyclic/None & None & 1 \\
    $C_4$ & Cyclic & 90 Degree rotations & 4 \\
    $D_1$ & Dihedral & Horizontal reflection & 2 \\
    $D_4$ & Dihedral & Horizontal reflection + 90 degree rotations & 8 \\
    $SO(2)$ & Special Orthogonal & Continuous rotations in 2D & $\infty$ \\
    $O(2)$ & Orthogonal & Horizontal reflection + continuous rotations in 2D & $\infty$\\
    \bottomrule
    \end{tabular}
    \caption{Overview of available symmetries for our \texttt{DDMNIST} dataset.}
    \label{tab:double_mnist_groups}
    \end{table}

\begin{figure}[ht!]
        \centering
        \begin{subfigure}[b]{0.49\textwidth}
        \centering
            \includegraphics[width=0.46\linewidth]{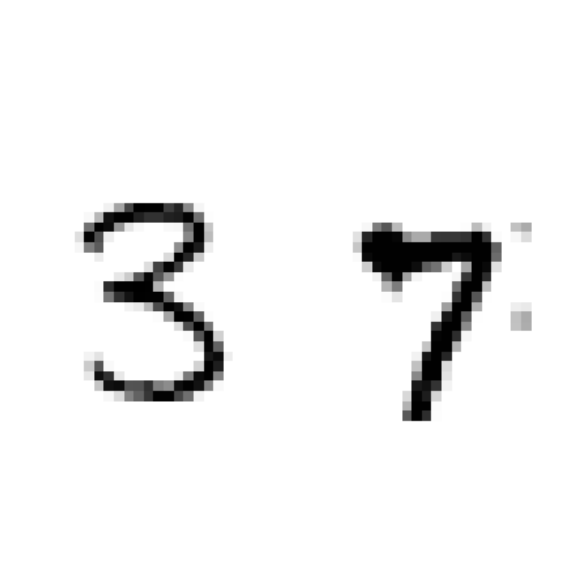}
            \caption{Original \textcolor{black}{unaugmented} number.}
            \label{fig:no_augment}
        \end{subfigure}
        \begin{subfigure}[b]{0.49\textwidth}
            \includegraphics[width=0.46\linewidth]{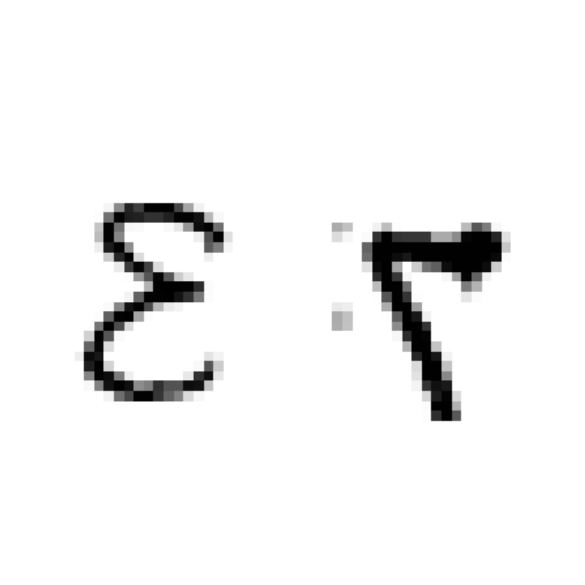}
           \includegraphics[width=0.46\linewidth]{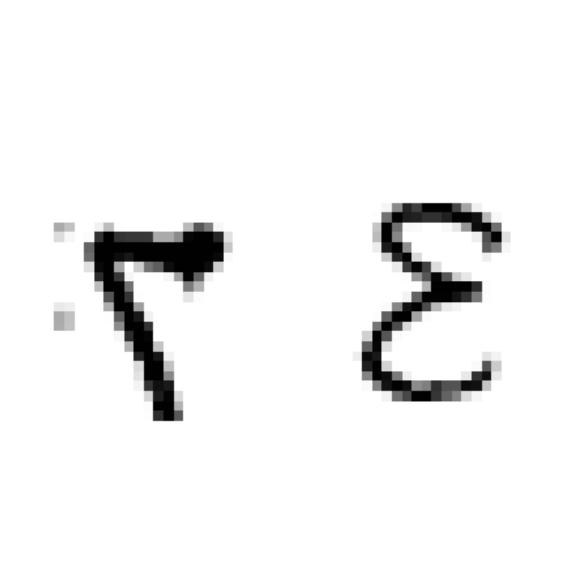}
           \caption{\textcolor{blue}{Local} vs \textcolor{red}{global} horizontal reflection.}
            \label{fig:vertical_reflection}
        \end{subfigure}
        \begin{subfigure}[b]{0.49\textwidth}
            \includegraphics[width=0.46\linewidth]{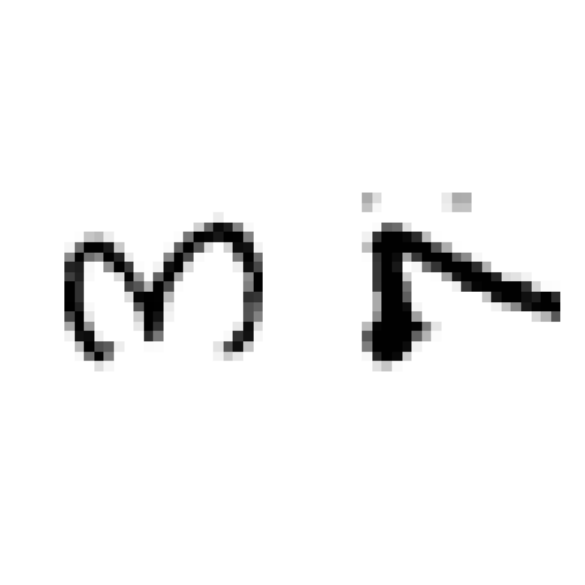}
           \includegraphics[width=0.46\linewidth]{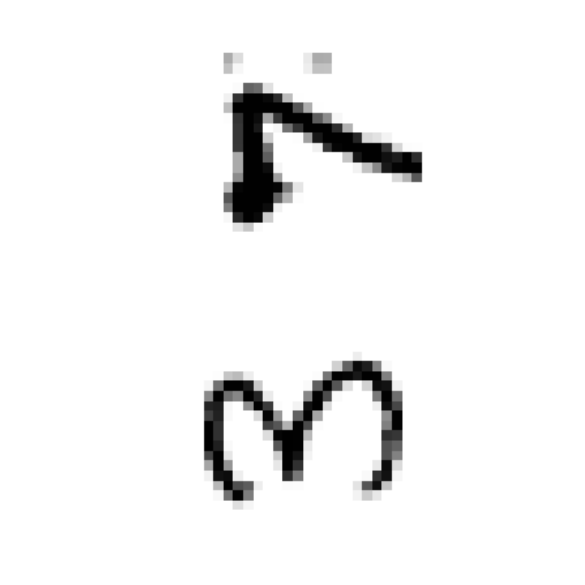}
           \caption{\textcolor{blue}{Local} vs \textcolor{red}{global} 90 degree rotation.}
           \label{fig:90_degree_rot}
        \end{subfigure}
        \begin{subfigure}[b]{0.49\textwidth}
            \includegraphics[width=0.46\linewidth]{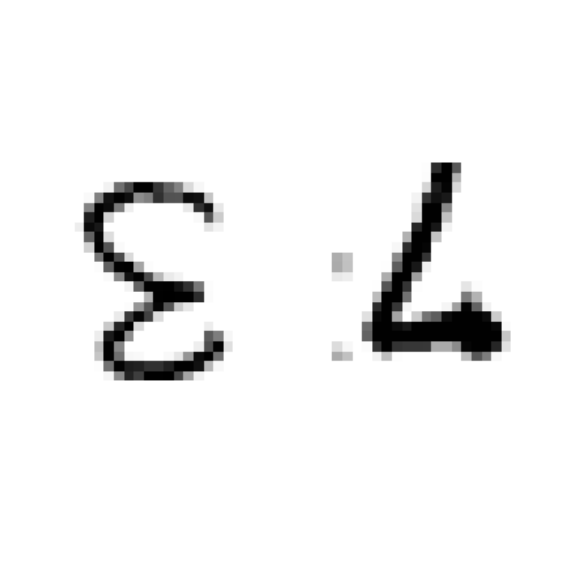}
           \includegraphics[width=0.46\linewidth]{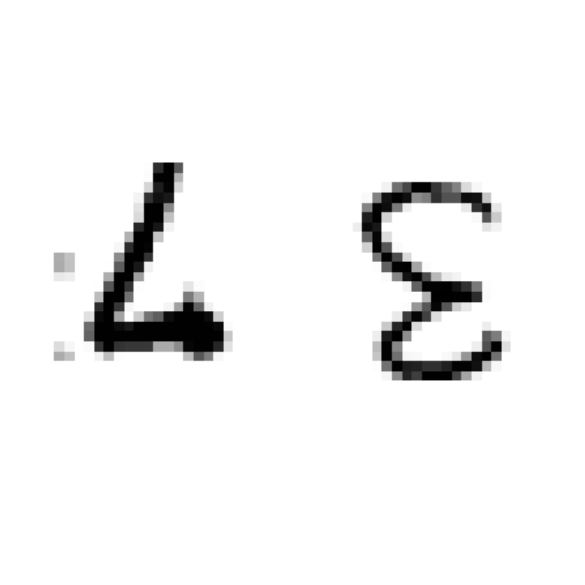}
           \caption{\textcolor{blue}{Local} vs \textcolor{red}{global} 180 degree rotation.}
           \label{fig:180_degree_rot}
        \end{subfigure}
        \caption{Examples of the effect of \textcolor{blue}{local} vs \textcolor{red}{global} augmentations on the double-digit number $37$. Figure~\subref{fig:no_augment} shows the original image. Figures~\subref{fig:vertical_reflection}, \subref{fig:90_degree_rot}, and \subref{fig:180_degree_rot} show the effect of applying a local (per-digit) compared to a global augmentation using a horizontal reflection, $90$ and $180$ degree rotation, respectively. In each case, it becomes apparent that while a local augmentation acting purely on the digit does not change the corresponding number, but applying the augmentation globally changes the number to $73$ in the case of figures \subref{fig:vertical_reflection} and \subref{fig:180_degree_rot}, resulting in incorrect equivariance. Conversely, \subref{fig:90_degree_rot} shows that a global rotation by 90 degrees results in a non-existing double-digit number and is hence extrinsic equivariance.} 
        \label{fig:double_mnist_symmetries}
    \end{figure}

    Although the resulting datasets contain clear symmetries in the individual digits, these symmetries do not always occur in the entire double-digit number. For instance, consider \texttt{DDMNIST} with $O(2)$ augmentations. The digits can be rotated and reflected individually without changing the label. However, applying a reflection on the whole number changes the corresponding label; for instance, $37$ becomes $73$. See Figure~\ref{fig:double_mnist_symmetries} for an example.  

    \subsubsection{Medical MNIST}
    To evaluate our approach on 3D groups, we use several volumetric datasets from the MedMNIST dataset collection~\cite{yang2023medmnist}. This collection contains challenging, diverse, and standardised biomedical datasets using real-world data. The different datasets seem to exhibit different types and degrees of symmetries \cite{kuipers2023regular}, making these datasets suitable for evaluating our approach. 
    
    MedMNIST provides several planar and volumetric biomedical datasets, where each sample is labelled and of size $28\times 28$ or $28\times 28 \times 28$. For the evaluation of 3D groups, we focus on the following three volumetric datasets. 
    \paragraph{OrganMNIST3D} The \texttt{OrganMNIST3D} dataset~\cite{organmnist, organmnist2} consists of volumetric CT scans of the following $11$ human organs that are also the classification targets: liver, right kidney, left kidney, right femur, left femur, bladder, heart, right lung, left lung, spleen, and pancreas. All samples are aligned to the abdominal view such that the sagittal, coronal and axial planes are aligned with the x, y, and z axes. This dataset is intriguing due to its wide variety of symmetric patterns. While some organs and structures display evident symmetries that an equivariant model might leverage, distinguishing between the left and right versions of certain organs can pose challenges for models that are rotationally or reflectionally equivariant.

    \paragraph{SynapseMNIST3D} \texttt{SynapseMNIST3D}~\cite{yang2023medmnist} focusses on 3D microscopy scans of neural synapses in adult rats to classify synapses as excitatory or inhabitory. These 3D scans are obtained by a multi-beam scanning electron microscope. 
    Historically, using microscopy, one of the main features used to classify a synapse as inhibitory or excitatory has been its structural symmetry; excitatory synapses are generally symmetric, while inhibitory synapses tend to be asymmetric\cite{synapse_symmetric}. However, there are other potential discriminating features~\cite{synapse_vague, synapse_vague2, synapse_vague3}. Although biomedically informed decisions are outside the scope of this paper, a model with a learnable and interpretable degree of equivariance can give some insight into which type of features the model prefers. This can in turn yield potentially valuable theoretical information for related fields.

    \paragraph{NoduleMNIST3D} The \texttt{NoduleMNIST3D} dataset~\cite{nodulemnist} contains thoracic CT scans of lung nodules that are categorised as low or high malignancy. International guidelines cite nodule size (and its growth rate) as the predominant indicator for malignancy levels \cite{nodule_size_1, nodule_size_2, nodule_size_3, nodule_size_4}. Given that size-based features remain invariant to any compact group, we employ this dataset primarily to assess how our approach performs on tasks that exhibit strong symmetry and, as a result, presumably demand minimal equivariance loss.

    \subsubsection{Smoke and JetFlow}\label{sec:simulation datasets}
    Finally, to provide a more comprehensive comparison between alternative partially equivariant approaches we employ the \texttt{Smoke} and \texttt{JetFlow} datasets used by~\citet{wangApprox}. They compare their RSteer and RGroup approach with various other approaches, allowing us to compare our models with their results.
    \paragraph{Smoke} Generated by PhiFlow~\cite{holl2020phiflow}, the synthetic $64\times64$ 2D \texttt{Smoke} dataset contains smoke simulations that allows for various types of equivariance breaking. In our case, we focuses on the partial rotation equivariances. In this setting, the dataset contains $40$ simulations with different inflow positions and buoyant forces of the smoke. The inflow location and buoyant forces exhibit $C_4$ rotation symmetry, but the buoyant factor varies based on the inflow position, thus breaking rotation equivariance. 
    \paragraph{JetFlow} The 2D JetFlow dataset~\cite{wangApprox} contains $62\times23$ real experimental turbulence velocity fields measured in NASA multi-stream jets. These fields are measured by 24 stations at different locations. 

\newpage
\subsection{Model Architectures}\label{sec:model_architectures}
    \begin{wrapfigure}[24]{r}{0.24\textwidth}  
      \begin{center}
          \includegraphics[width=0.8\linewidth]{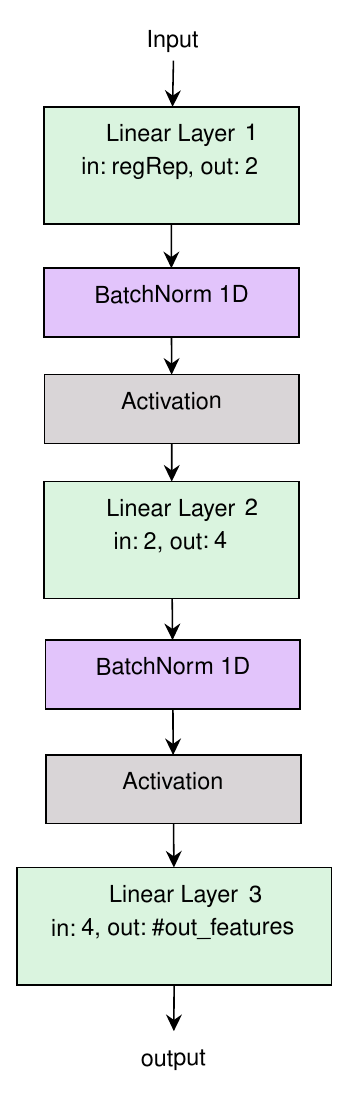}
      \end{center}
      \caption{The base structure of the MLP networks. In and out refer to the number of input and output features.}\label{fig:MLP}      
    \end{wrapfigure}
For \texttt{Vectors}, \texttt{DDMINST} and \texttt{MedMNIST}, we use separate base structures for the neural networks. While the base structure for each of the datasets remains mostly the same throughout our experiments, certain aspects, such as the type of convolution/linear layer, are determined by the individual experiment and method. By default, these base architectures are used for all approaches in our experiments. This includes the RPP approach~\cite{finzi2021residual}, which we include as one of the baselines, as RPPs are comparable to some degree to our approach and are also trivially applicable to SCNNs. By using the same base architecture, we ensure that each model has the same number of output features at each intermediate layer. For the \texttt{Smoke} and \texttt{JetFlow} datasets we employ the set-ups and implementations from~\citet{wangApprox}.

\paragraph{MLP} For \texttt{Vectors} we use a simple three-layer MLP, where the first two layers are followed by a 1D batch normalisation and a non-linearity. The complete structure can be found in Figure \ref{fig:MLP}. This structure has four main configurations, see Table \ref{tab:mlp_configs} for an overview. For each of the equivariant configurations, we use an irrep-field containing irreps up to frequency $4$ as intermediate features, and a trivial representation as output feature to ensure an invariant mapping. For the configurations with a learnt degree of equivariance, each layer learns an individual degree of equivariance by default, using a band-limit of $L=4$. 

In addition to our \textit{probabilistic} approach, we also provide experiments with a \textit{preliminary} approach that does not use a Fourier-based parameterisation in the appendix. This approach instead separately parameterises the $c^{jj'}$ matrices from Eq.~\ref{eq:prob_cnn_final} for each irrep-pair $\psi_j$ and $\psi_j'$, reducing the degree of weight sharing and consistency in the breaking of equivariance. This allows us to gain more insights into the potential advantages of our \textit{probabilistic} approach compared to a more \textit{naive} approach.

    \begin{figure}[ht!]
    \begin{subfigure}[b]{0.425\textwidth}
        \includegraphics[width=\linewidth, trim={0cm 0cm 4.5cm 0cm}, clip]{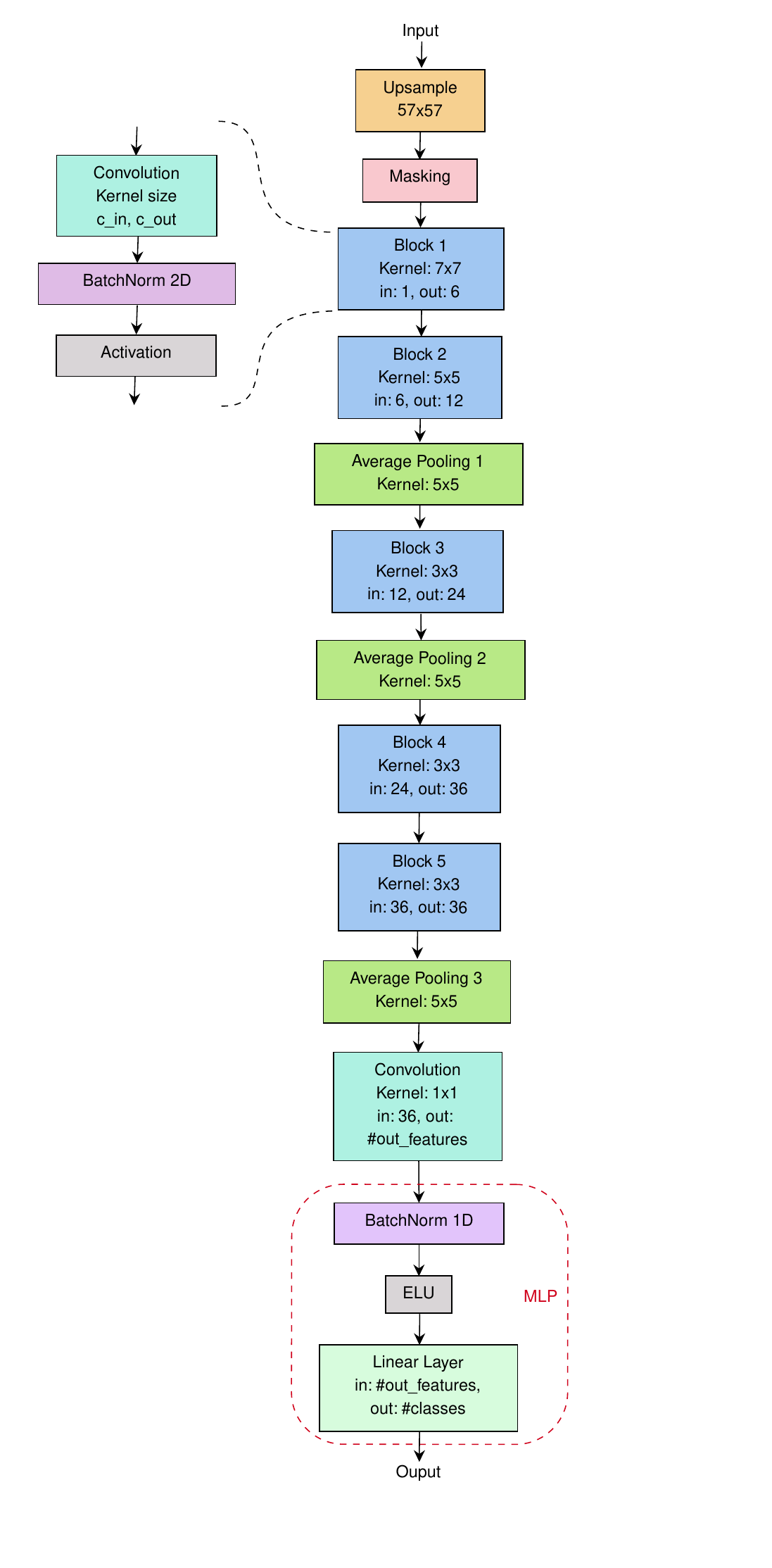}        
        \caption{Base 2D CNN configuration}
        \label{fig:2dcnn}
    \end{subfigure}
    \begin{subfigure}[b]{0.565\textwidth}
        \includegraphics[width=\linewidth]{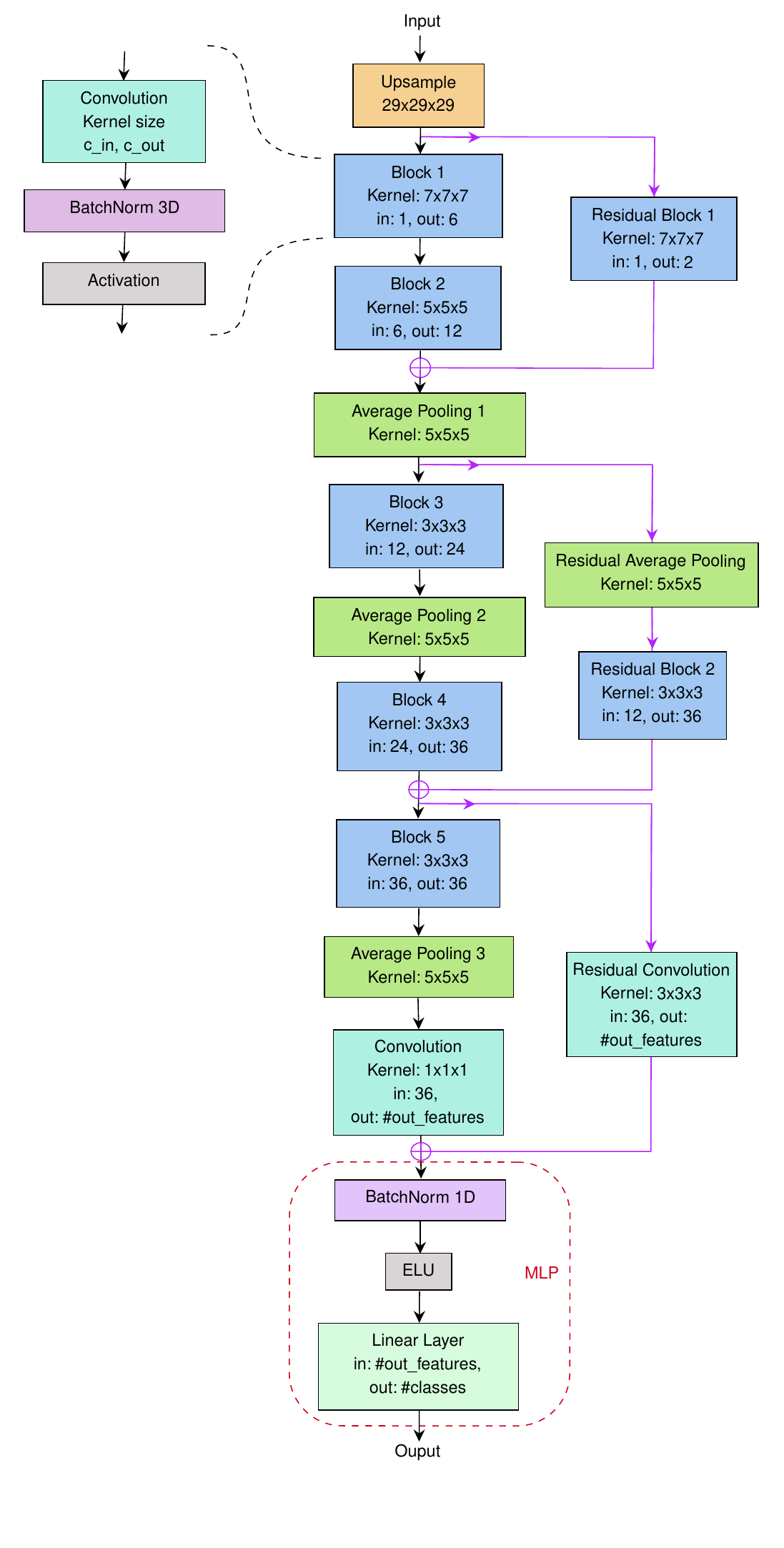}        
        \caption{Base 3D CNN configuration}
        \label{fig:3dcnn}
    \end{subfigure}
    \caption{Diagrams of our 2D and 3D CNN base configurations.}
    \end{figure}

\begin{table}[h]

\begin{minipage}[b]{0.7\linewidth}
\resizebox{\textwidth}{!}{%
\begin{tabular}{lll}
\toprule
\textbf{Configuration} & \textbf{Equivariance} & \textbf{Non-linearity} \\ \midrule\midrule
MLP & Non-equivariant & ELU \\
E-MLP & $O(2)$ & Gated/FourierELU \\
PE-MLP (Preliminary) (ours) & Partial $O(2)$  & Gated/FourierELU \\
PE-MLP (Probabilistic) (ours) & Partial $O(2)$ & Gated/FourierELU\\
\bottomrule
\end{tabular}}
\caption{Overview of the four main MLP architectures. FourierELU refers to a Fourier-based non-linearity using a pointwise ELU non-linearity acting on the sampled signal.}
\label{tab:mlp_configs}    
\end{minipage}
\end{table}
\newpage
\paragraph{2D CNN}

For the \texttt{DDMNIST} and its variations, we employ a CNN with five blocks as its primary structure. This is followed by a $1 \times 1$ convolution and a final classification layer. Notably, the $1\times 1$ convolution is executed on a $1\times 1$ spatial domain, resulting in translation \textit{invariance}.

Each block in this architecture comprises a convolution, 2D batch normalisation, and a non-linearity. See Figure \ref{fig:2dcnn} for a complete overview of the base architecture. It is worth noting that the $56\times 56$ images are initially upsampled to $57\times 57$. Subsequently, the input is masked with a circular mask with a radius of $28$ pixels. As a result, information is lost at the corners of the input, thereby enhancing rotational equivariance.
\begin{table}[ht!]
\centering
\resizebox{\textwidth}{!}{%
\begin{tabular}{llll}
\toprule
\textbf{Configuration} & \textbf{Description} & \textbf{Partial Equivariance} & \textbf{Non-linearity} \\ \midrule\midrule
CNN & CNN & N/A & ELU \\
SCNN & Steerable CNN & None \cite{cesa2022ae(n)} & Gated/FourierELU \\
RPP & SCNN + Residual Pathway Prior & RPP \cite{finzi2021residual} & Gated \\
R-SCNN & Restricted SCNN & Restriction \cite{cesa2022ae(n)} & Gated \\
P-SCNN (ours) & Partial/Probabilistic Steerable CNN & Ours & Gated\\
\bottomrule
\end{tabular}}
\caption{Overview of the main configurations of our base 2D CNN used for our \texttt{DDMNIST} experiments.}
\label{tab:2dcnn_table}
\end{table}

For our experiments, we use five primary configurations of the base structure. These configurations and their specifics, especially concerning the attainment of partial equivariance (if present), are detailed in Table~\ref{tab:2dcnn_table}. Here, the RPP configuration \cite{finzi2021residual} is derived by incorporating the SCNN configuration and appending residual conventional (non-equivariant) CNN connections. These connections span each individual block and conclude at the final pooling convolution, cumulatively resulting in six residual pathways for partial equivariance. \begin{wraptable}[14]{r}{5.5cm}
\begin{tabular}{l|c}
\toprule
\textbf{Group}   & \begin{tabular}[c]{@{}c@{}}\textbf{Feature Field} \\ \textbf{Bandlimiting}\end{tabular} \\ \midrule\midrule
$O(2)$  & 4                                                                     \\
$SO(2)$ & 4                                                                     \\
$D_4$   & 2                                                                     \\
$C_4$   & 2                                                                     \\ 
$D_1$   & 0                                                                     \\ 
$C_1$   & 0                                                                     \\\midrule
$SO(3)$ & 3                                                                     \\
$O(3)$  & 2                                                                    \\
\bottomrule
\end{tabular}
\caption{Maximum level of bandlimiting of the intermediate features for each group.}
\label{tab:maximum_bandlimiting}\end{wraptable}Additionally, for the R-SCNN configuration. This decision aligns with the theoretical absence of symmetries on a broader scale in \texttt{DDMNIST}. Through empirical analysis, this setup was deemed optimal when applied to the final two convolution layers.

For all configurations, whether fully or partially equivariant, we conduct experiments with various compact groups $H$. Each group has its unique maximum level of bandlimiting for the intermediate irrep feature fields, as detailed in Table~\ref{tab:maximum_bandlimiting}. In line with the MLP structure, we map to the trivial representation during the final $1\times1$ convolution to guarantee invariance by default. 

Table~\ref{tab:MAXGroupAppendixBandlimitTranspose} contains the maximum level of bandlimiting of the intermediate features for each group. Table~\ref{tab:MLPAppendixBandlimit} contains the level of bandlimiting for each of the three layers in the E-MLP-based configurations. In addition to the degrees of feature bandlimiting for each layer, Table~\ref{tab:SCNNAppendixBandlimitMin} presents the strides and padding for each of the layers for the SCNN-based configurations. Furthermore, we set the default band-limit for the likelihood to $L=2$.

\begin{table}[h!]
\centering
\begin{tabular}{c|cccccccc}
\toprule
    \textbf{Group} &  $C_1$ & $C_4$ & $SO(2)$ &$D_1$ &$D_4$ &$O(2)$ &$SO(3)$ &$O(3)$\\ \midrule
    \textbf{Maximum}&0&2&4&0&2&4&3&2\\
    \bottomrule
\end{tabular}
\caption{Maximum level of bandlimiting of the intermediate for each supported transformation group.}
\label{tab:MAXGroupAppendixBandlimitTranspose}
\end{table}

\begin{table}[h!]
\centering
\begin{tabular}{c|c}
\toprule
    \textbf{Layer} & \textbf{Features Bandlimiting} \\ 
    \midrule\midrule
    1& 4\\
    2& 3\\
    3& \textit{Trivial}\\
    \bottomrule
\end{tabular}
\caption{Bandlimiting of the intermediate features of the $O(2)$ E-MLP configurations. \lq\textit{Trivial}\rq \ means that the output is only a trivial representation.}
\label{tab:MLPAppendixBandlimit}
\end{table}

\begin{table}[h!]
\centering
\begin{tabular}{l|cccc}
\toprule
\multirow{2}{*}{\textbf{Layer}} & \multirow{2}{*}{\textbf{Features Bandlimiting}} & \multirow{2}{*}{\textbf{Stride}} & \multicolumn{2}{c}{\textbf{Padding}} \\
                  &                                        &   & \textbf{2D SCNN} & \textbf{3D SCNN} \\ \midrule \midrule
Block 1           & $\min{\left(2, L_{\text{max}}\right)}$ & 1 & 2       & 2       \\
Block 2           & $\min{\left(3, L_{\text{max}}\right)}$ & 1 & 2       & 2       \\
Avg. Pooling 1    & -                                      & 2 & 1       & 1       \\
Block 3           & $L_{\text{max}}$                       & 2 & 2       & 1       \\
Avg. Pooling 2    & -                                      & 2 & 1       & 1       \\
Block 4           & $L_{\text{max}}$                       & 2 & 0       & 2       \\
Block 5           & $\min{\left(2, L_{\text{max}}\right)}$ & 1 & 1       & 1       \\
Avg. Pooling 3    & -                                      & 1 & 1       & 1       \\
Convolution       & \textit{Trivial} or $L_{\text{max}}$   & 1 & 0       & 0       \\ \midrule
Res. Block 1      & $\min{\left(3, L_{\text{max}}\right)}$ & 1 & -       & 2       \\
Res. Avg. Pooling & -                                      & 2 & -       & 0       \\
Res. Block 2      & $L_{\text{max}}$                       & 1 & -       & 0       \\
Res. Convolution  & \textit{Trivial} or $L_{\text{max}}$   & 1 & -       & 0       \\ \bottomrule
\end{tabular}
\caption{Additional details for all layers in our SCNN-based configurations. Features bandlimiting denotes the degree of bandlimiting of the layer's output features. Here, $L_{\text{max}}$ refers to the maximum level of bandlimiting for each specific group (Table~\ref{tab:MAXGroupAppendixBandlimitTranspose}). The residual blocks in the last few rows are only applicable for our 3D SCNNs acting on \texttt{MedMNIST}, all other rows are applicable for all (2D and 3D) SCNN-based configurations. The output features of the final convolution and residual convolution layers are only trivial features or band-limited features up to $L_{\text{max}}$ depending on whether it is the base configuration with a structurally invariant mapping or the non-invariant mapping.}
\label{tab:SCNNAppendixBandlimitMin}
\end{table}

    \paragraph{3D CNNs}

    For \texttt{MedMNIST} we use a modified version of our base structure used for \texttt{DDMNIST} acting on $\mathbb{R}^3$ rather than $\mathbb{R}^2$. As the exact symmetries in these datasets are not known a-priori, we do not perform experiments using group restrictions. Otherwise, the base configurations are the same as in Table~\ref{tab:2dcnn_table}. A diagram of our modified base-structure can be found in Figure~\ref{fig:3dcnn}. Here, we add three residual connections, spanning two blocks at a time. Due to these residual connections, the residual non-equivariant connections for the RPP configuration now also span two blocks at a time instead of one. Additionally, we set the width of each layer to half of the other models for the RPP configuration. This is required to prevent overfitting, since the regular CNN layers of the RPP configuration become too large in order to commute with the steerable layers.

    Additionally, for our approach, rather than learning a separate likelihood distribution for each individual layer, the first block of each residual block --consisting of two sequential blocks and one residual block-- has a fixed non-learnable degree of equivariance, whereas the second block and the residual block share a degree of equivariance. As a result, we end up with three likelihood distributions rather than six. For each (partially) equivariant configuration, we perform experiments using $H=SO(3)$ and $H=O(3)$, using the maximum feature field bandlimiting from Table~\ref{tab:maximum_bandlimiting}. Similarly to the experiments on \texttt{DDMNIST}, we set the default bandlimiting of the likelihood distributions to $L=2$.

    \paragraph{Smoke/JetFlow CNNs} Finally, for the experiments on the \texttt{Smoke} and \texttt{JetFlow} experiments we employ the same network structures as employed by~\citet{wangApprox}. As these details are not available in the main paper, we rely on the corresponding code repository for the model architectures as well as the hyper-parameter tuning. For each of our models, we simply take the configuration as reported for their RSteer model in their code repository and directly port that to our approach without additional fine-tuning, besides for the $\alpha_{\text{KL}}$ and $\alpha_{\text{align}}$ regularisation weights. Tab.~\ref{tab:wang_models} contains an overview of the various approaches in their comparison, along with citation and explanation. All (partially) equivariant approaches from the main results in Sec.~\ref{sec:benchmarks_main} use $C_4$-equivariance with regular feature fields. The band-limiting experiments in Apx.~\ref{sec:result_bandlimit} use $SO(2)$-equivariance encoded via irrep fields up to a frequency of 2.

    \begin{table}[h!]
    \centering
    \resizebox{\linewidth}{!}{%
    \begin{tabular}{cc|c|c}
    \toprule
     & Model & Citation & Description \\ \midrule\midrule
    \multirow{3}{*}{\begin{tabular}[c]{@{}c@{}}(non) equivariant\\ approaches\end{tabular}} & MLP & - & Regular non-equivariant MLP \\ \cline{2-4} 
     & CNN & \citet{lecun1995convolutional} & Regular CNN \\ \cline{2-4} 
     & \texttt{e2cnn} & \citet{cesae(2)} & SCNNs created with 2D predecessor of \texttt{escnn} \\ \midrule
    \multirow{6}{*}{\begin{tabular}[c]{@{}c@{}}Partially equivariant \\ approaches\end{tabular}} & RPP & \citet{finzi2021residual} & Residual Pathway Priors \\ \cline{2-4} 
     & Combo & \citet{wangApprox} & Non-equivariant convolutions followed by equivariant convolutions \\ \cline{2-4} 
     & CLCNN & \citet{wangApprox} & \begin{tabular}[c]{@{}c@{}}Locally connected neural networks with equivariance \\ constraints imposed in the loss function\end{tabular} \\ \cline{2-4} 
     & Lift & \citet{wang2022equivariantQ} & \begin{tabular}[c]{@{}c@{}}Lift expansion; non-equivariant features are lifted to the\\  dimensionality of equivariant features and concatenated\end{tabular} \\ \cline{2-4} 
     & RGroup & \citet{wangApprox} & GCNNs with weights conditioned on the group element \\ \cline{2-4} 
     & RSteer & \citet{wangApprox} & SCNNs with weights conditioned on the group element \\ \bottomrule
    \end{tabular}}
    \caption{Overview of the various models used for the \texttt{Smoke} and \texttt{JetFlow} datasets employed by \citet{wangApprox}. RSteer and RGroup are part of their main contribution, whereas CLCNN, Lift and Combo are their alternative approaches.}\label{tab:wang_models}
    \end{table}

    As only the final configuration for \texttt{RSteer} is reported in their repository, we obtain the parameter counts by taking the \texttt{RSteer} configuration and porting that to any of the other approaches. The parameter counts we obtained in Sec.~\ref{sec:benchmarks_main} therefore do not necessarily represent the parameter counts corresponding to the models used in the original paper to obtain the results. To sanity check, we verify that the models obtain performance comparable to the results reported by \citet{wangApprox}. This was not the case for CLCNN, so we did not include it in our results.

\subsection{Equivariance Error}
To verify the validity of the learnt likelihood distributions, we include measured equivariance errors across the group. However, instead of simply computing the error,
we compensate for the differences in feature magnitude by normalising the relative error vectors based on the magnitude of the original output feature. Additionally, to obtain scalar values we calculate the $\ell^2$ norm over the feature vectors. Thus, for each layer $l$ and a batch of samples $B$, we compute the following:
\begin{align}
    \epsilon_l(h) = \frac{1}{|B|}\sum_{b \in B}\frac{||h \cdot l(b) - l(h \cdot b)||_2}{||l(b)||_2}, \quad \forall h\in H.
\end{align}

Since the errors can be non-zero for the fully invariant models due to interpolation artefacts or other sources of inherent partial equivariance, rather than reporting the equivariance errors directly we report the difference in equivariance errors between the fully invariant SCNNs/E-MLPs and our partially equivariant P-SCNNs/PE-MLPs. As a result, the reported errors indicate how our approach differs from the invariant setting. Furthermore, all reported errors are calculated using one batch of data.

\subsection{Training details}
    All models undergo training five times, each with a different, pre-determined seed. Training is performed on an NVIDIA A100 40GB GPU using the Adam optimiser~\cite{kingma2014adam}. For our \textit{probabilistic} PE-MLP and our P-SCNN we use the following objective function:
    \begin{equation}
        \mathcal{L}_\text{total} = \mathcal{L}_{\text{task}} + \alpha_{\text{align}} \mathcal{L}_{\text{align}} + \alpha_{\text{KL}} \mathcal{L}_{\text{KL}}
    \end{equation}
    Where $\mathcal{L}_{\text{task}}$ is the task-specific loss function that is used for all other models, and $\alpha_{\text{align}}$ and $\alpha_{\text{KL}}$ are tunable weighting factors for the alignment loss and KL-divergence. We refer to Table~\ref{tab:training_details} for more specific training details on the individual datasets, including these weighting factors.
    \begin{table}[h!]
    \resizebox{\textwidth}{!}{%
    \begin{tabular}{c|ccccccc}
    \toprule
    \textbf{Dataset} & \texttt{Vectors} & \texttt{DoubleMNIST} & \texttt{OrganMNIST3D} & \texttt{SynapseMNIST3D} & \texttt{NoduleMNIST3D} & \texttt{Smoke} & \texttt{JetFlow} \\ \midrule\midrule
    \textbf{Learning rate} & $5 \times 10^{-5}$ & $5 \times 10^{-4}$ & $5 \times 10^{-5}$ & $1 \times 10^{-5}$ & $1 \times 10^{-5}$ & $1\times 10^{-3}$ & $1\times 10^{-3}$ \\
    \textbf{batch size} & 1024 & 256 & 32 & 32 & 32 & 16 & 16 \\
    \textbf{\# of epochs} & 100 & 50 & 100 & 100 & 100 & 100& 100\\
    \textbf{Loss function} & MSE & Cross-entropy & Cross-entropy & Cross-entropy & Cross-entropy & MSE & MSE\\
    \textbf{Evaluated Epoch} & Last & Best on validate & Last & Last & Last & Best on validate & Best on validate \\
    $\bm{\alpha_{\text{align}}}$ & 5 & 5 & 5 & 5 & 5 & 5 & 5 \\
    $\bm{\alpha_{\text{KL}}}$ & 25 & 3 & 1 & 1 & 1 & 3 & 3\\
    \bottomrule
    \end{tabular}}
    \caption{Training details for the datasets employed in our experiments.}
    \label{tab:training_details}
    \end{table}

    Furthermore, all RPP models are trained with a prior variance of $\sigma^2=10^5$ on the equivariant weights and $\sigma^2=10^3$ on the non-equivariant weights. Empirically, we found these settings to be the most optimal. These priors are comparable to those used for some experiments in \cite{finzi2021residual}.

\section{Additional Experiments}\label{ap:additional_experiments}
In this section we cover additional experiments. In these experiments, we include \textit{preliminary} approach to learning the degree of equivariance. In contrast to our main approach, this approach does not parameterise the degree of equivariance through a likelihood distribution. Instead, for each irrep pair ($\psi_j$, $\psi_{j'}$) we construct the projection matrix $c^{jj'}$ as in the fully equivariant state. Then we directly parameterise this projection matrix for each irrep pair ($\psi_j$, $\psi_{j'}$). As a result, this approach has significantly less weight sharing, and therefore a reduced consistency in terms of breaking equivariance. 

In Section~\ref{sec:results_prelim} we cover the aforementioned comparison between the preliminary approach and our main approach. Section~\ref{ap:med_result} contains the confusion matrices obtained on \texttt{OrganMNIST3D} from the \texttt{MedMNIST} collection. Section~\ref{ap:inspect_LL} provides additional likelihood distributions, including ones on our 3D models. In Section~\ref{sec:results_shared} we compare the concept of shared versus layer-wise equivariance in E-MLPs. Section~\ref{sec:results_regularisation} covers our experiments regarding KL-divergence and alignment loss, as well as additional results from our bandlimiting experiments. Here we show that these two terms are essential for interpretable likelihood distributions. Finally, in Section~\ref{sec:result_compet} we combine the tuning of bandlimiting and the use of stronger (less equivariant) non-linearities to maximise performance. 

\subsection{Preliminary Approach compared to Probabilistic Approach}\label{sec:results_prelim}
To assess the influence of weight sharing we compare the performance of the baselines to our \textit{preliminary} and \textit{probabilistic} approach on the simple \texttt{Vectors} dataset. For the \textit{preliminary} approach, we include two configurations; one where the coefficients are initialised according to described above, and one with a small amount of added noise to the matrices that are initialised as zeros. Table~\ref{tab:vector} provides an overview of the results. In the following two paragraphs, we discuss the performance on the angle and norm regression tasks separately.

\begin{table}[h!]
    \centering
    \begin{tabular}{ll|cc}
    \toprule
    \textbf{Model} & \textbf{Non-linearity} & \textbf{Angle}                 & \textbf{Norm}                  \\ \midrule \midrule
    MLP   & ELU           & $0.169\ $\myfontsize{$ ( 0.188)$}  & $0.058\ $\myfontsize{$ ( 0.032)$}  \\ \midrule
    \multirow{2}{*}{E-MLP }                      & Gated & $1.486\ $\myfontsize{$ ( 0.044)$} & $0.080\ $\myfontsize{$ ( 0.002)$}  \\
          & FourierELU    & $0.643 \ $\myfontsize{$ ( 0.176)$} & $0.004 \ $\myfontsize{$ ( 0.006)$} \\ \midrule
    \multirow{2}{*}{Ours (Preliminary)}         & Gated & $1.496\ $\myfontsize{$ ( 0.010)$} & $0.035\ $\myfontsize{$ ( 0.018)$}  \\
          & FourierELU    & $0.264 \ $\myfontsize{$ ( 0.345)$} & $0.019 \ $\myfontsize{$ ( 0.032)$} \\ \midrule
    \multirow{2}{*}{Ours (Preliminary + Noise)} & Gated & $0.064\ $\myfontsize{$ ( 0.022)$} & $\underline{0.033}\ $\myfontsize{$ ( 0.016)$}  \\
          & FourierELU    & $0.258 \ $\myfontsize{$ ( 0.316)$} & $0.020 \ $\myfontsize{$ ( 0.040)$} \\ \midrule
    \multirow{2}{*}{Ours (Probabilistic)}       & Gated & $\bm{\underline{0.046}}\ $\myfontsize{$ ( 0.004)$} & $0.052\ $\myfontsize{$ ( 0.008 )$} \\
          & FourierELU    & $\underline{0.128} \ $\myfontsize{$ ( 0.103)$} & $\underline{\bm{0.001}} \ $\myfontsize{$ ( 0.001)$}
          \\ \bottomrule
    \end{tabular}
    \caption{Test MSE scores of our approaches and the baseline models on \texttt{Vectors} norm and angle regression tasks. \textbf{Bold} indicates the lowest MSE error for the specific task. For each non-linearity, \underline{underline} indicates the lowest MSE error for this specific non-linearity on for the given task. Standard deviations over 5 runs are denoted in parentheses.}
    \label{tab:vector}
    \end{table}

\paragraph{Angle} We observe that, in terms of the non-$O(2)$-invariant angle regression task, the fully $O(2)$ invariant E-MLP is significantly outperformed by most of the other models, regardless of non-linearity. It is notable that the performance improves considerably when using the FourierELU non-linearity. This could partially be from the approximate equivariance of this non-linearity. Furthermore, we note that omitting the addition of noise to our \textit{preliminary} approach results in similar performance compared to the E-MLP under the Gated non-linearity. However, the addition of noise results in a significant increase in performance. Otherwise, our \textit{probabilistic} configuration and our \textit{preliminary} configuration with added noise both achieve significantly lower errors compared to the E-MLP, while also outperforming the non-equivariant MLP. Finally, using the Gated non-linearity results in a lower error for these two configurations.

Upon further inspection, we find that, under the Gated-nonlinearity, employing the \textit{preliminary} approach without added noise ensures that the $c^{lJ}$ matrices for non-matching irreps $\psi_l$ and $\psi_J$ remain zero matrices throughout training. Consequently, the model exhibits invariance for the duration of the training process. To unravel the cause, it is vital to consider the difference between FourierELU and Gated non-linearities. The Gated non-linearity, due to its perfect equivariance, lacks any initial mappings between non-matching input and output irreps $\psi_l$ and $\psi_J$. As a result, with the $c^{lJ}$ matrices initialised to zero, the associated weights remain inactive and are, in effect, zero. Therefore, these weights do not influence the output, and since there are multiple sequential layers that are initialised in this manner, these weights cannot be updated through backpropagation.

This behaviour does not manifest with the \textit{approximately} equivariant FourierELU non-linearity, which induces a degree of feature mixing across the irreps through the inverse Fourier transform to generate the sampled signal. Thus, weights that define mappings between non-matching irreps are utilised at initialisation, allowing backpropagation to update these weights.

The \textit{probabilistic} approach, intriguingly, does not encounter this issue. This is a result of the additional weight sharing; within this framework, the Fourier coefficient of the trivial representation contributes to mappings for both matching and non-matching irreps. This allows for the possibility of non-zero mappings between non-matching irreps after updates to this Fourier coefficient. Then, through normalisation, the other Fourier coefficients become slightly non-zero, freeing them up for backpropagation. Moreover, our method of normalising the likelihood distribution inadvertently introduces minor rounding errors, giving rise to slight, non-zero weights for non-trivial Fourier coefficients at initialisation.
\newpage
\paragraph{Norm} In the $O(2)$ invariant norm regression task, the performance gaps between the models are much smaller in general, suggesting that breaking equivariance is not as beneficial in this task. In fact, under the FourierELU non-linearity --which is comparable to the MLP's regular ELU-- the E-MLP outperforms the regular MLP. Although we do note that using our preliminary approach with Gated non-linearity or our probabilistic approach with either non-linearity does result in an increase in terms of performance.

\subsection{Benchmarks on Biomedical Image Classification}\label{ap:med_result}
Here we provide additional confusion matrices for our results on \texttt{MedMNIST}
\begin{figure}[H]
\begin{minipage}[b]{0.92\linewidth}
    \centering
    \begin{subfigure}[b]{0.49\textwidth}
        \includegraphics[width=\textwidth]{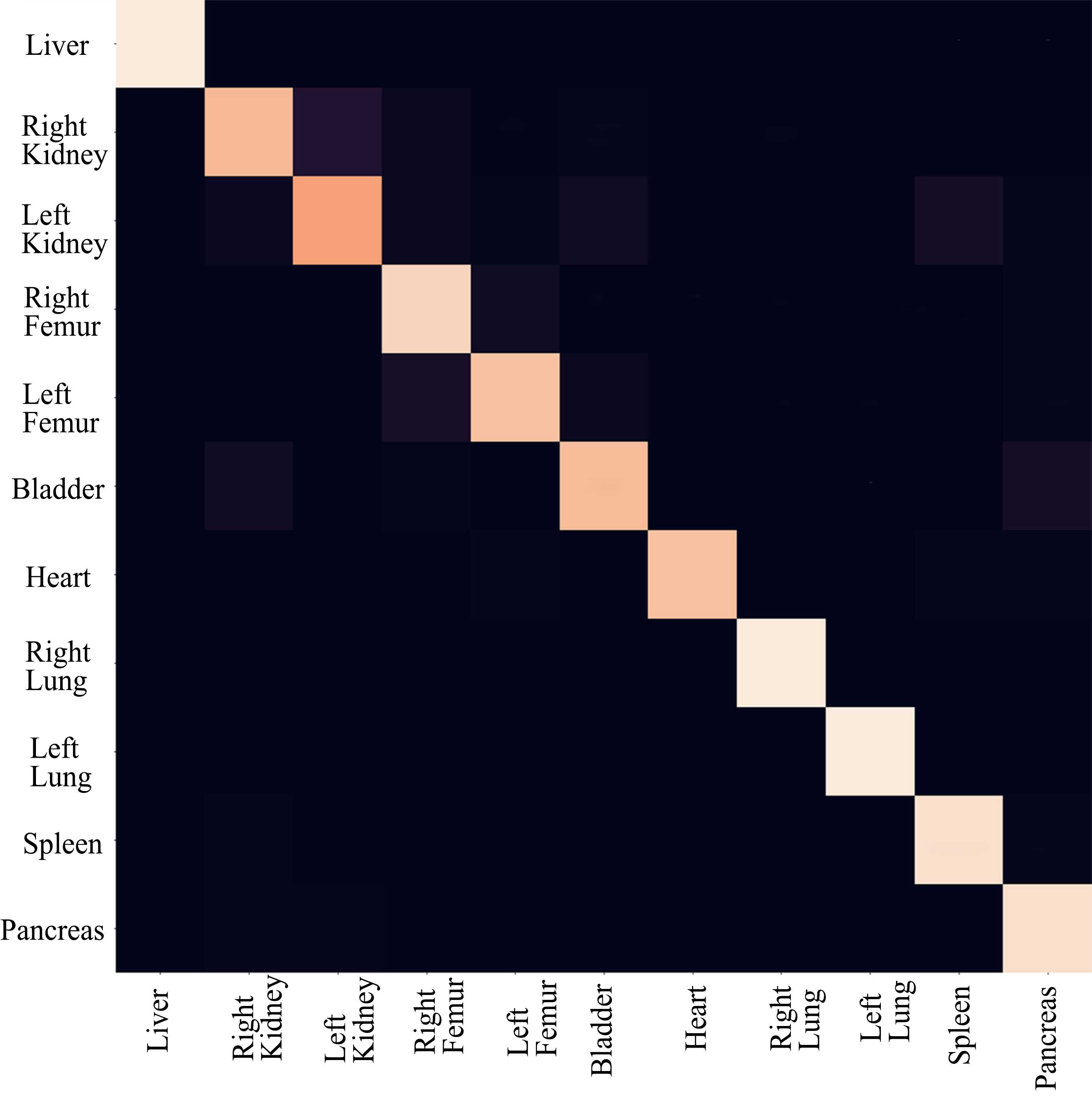}
    \caption{CNN}\label{fig:confuse_organ_cnn}    
    \end{subfigure}
    \begin{subfigure}[b]{0.49\textwidth}
        \includegraphics[width=\textwidth]{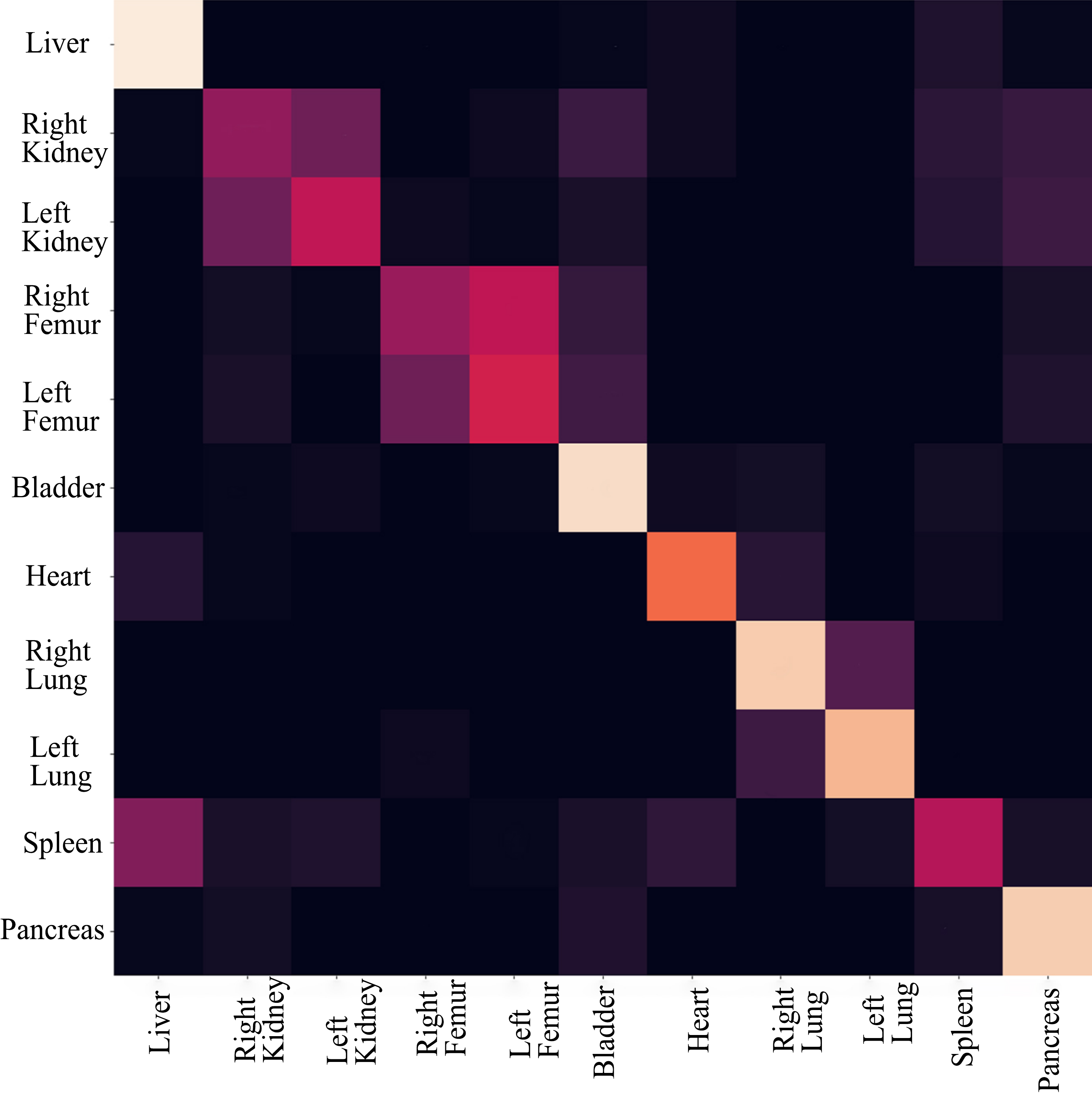}
    \caption{$O(3)$ SCNN}\label{fig:confuse_organ_o3}
    \end{subfigure}
    \begin{subfigure}[b]{0.49\textwidth}
        \includegraphics[width=\textwidth]{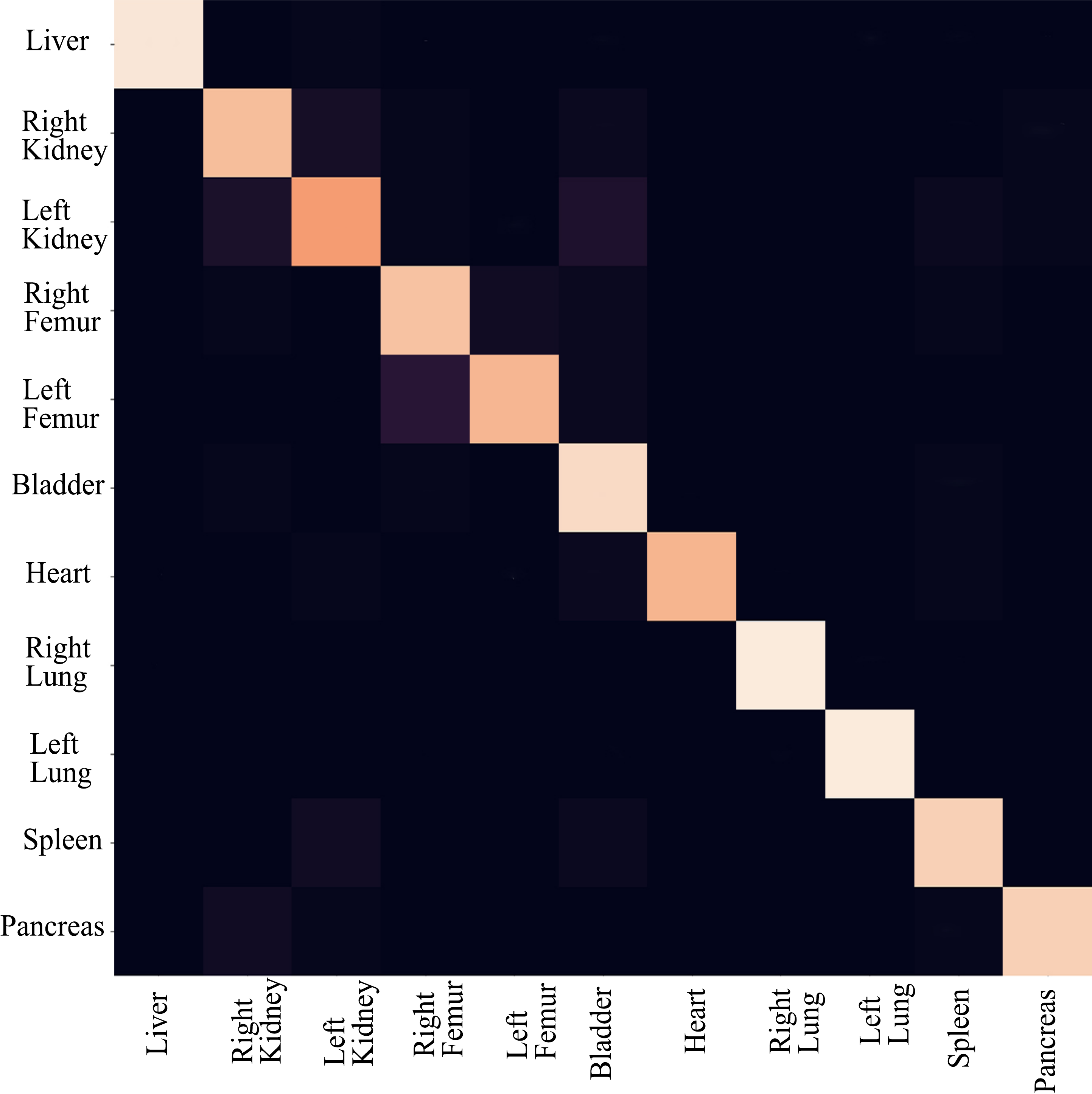}
        \caption{$O(3)$ P-SCNN (ours)}\label{fig:confuse_organ_ours}
    \end{subfigure}
        \begin{subfigure}[b]{0.49\textwidth}
        \includegraphics[width=\textwidth]{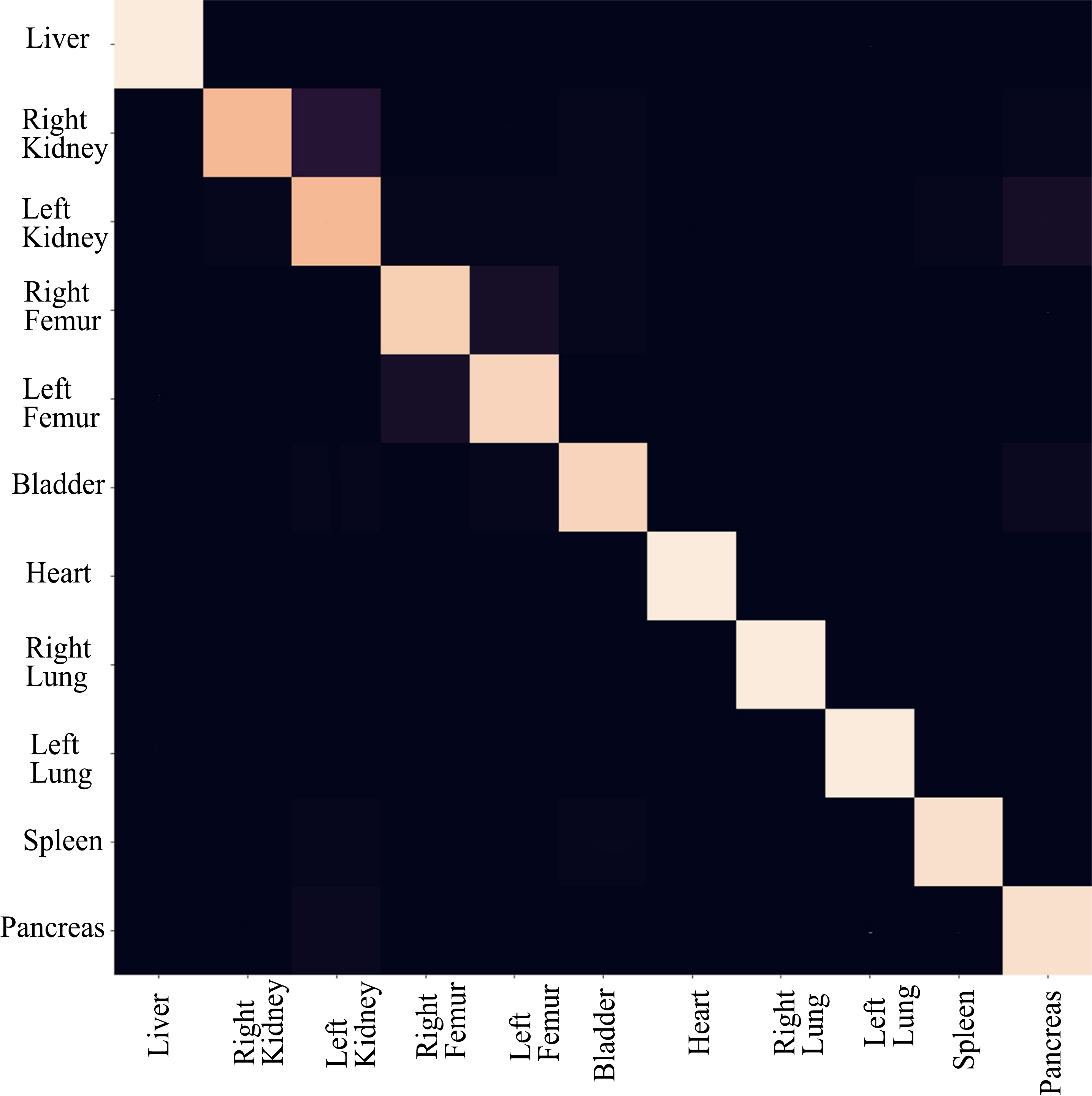}
        \caption{$O(3)$ RPP}\label{fig:confuse_organ_rpp}
    \end{subfigure}
    \end{minipage}
    \hspace{0cm}
    \begin{minipage}[b]{0.01\linewidth}
    \begin{subfigure}[b]{\textwidth}
        \resizebox{\width}{16cm}{\includegraphics[scale=0.25]{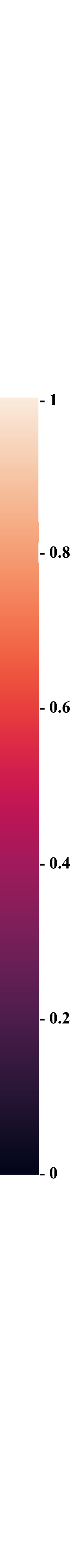}}
    \label{fig:scale_2}
    \end{subfigure}
    \end{minipage}
    \caption{Confusion matrices obtained on OrganMNIST3D symmetries using a regular CNN, an $O(3)$ SCNN, $O(3)$ RPP and our $O(3)$ P-SCNN. }\label{fig:confusion_organ}
\end{figure}

The confusion matrices in Figure~\ref{fig:confusion_organ} provide some insight into this behaviour on \texttt{OrganMNIST3D}. Here, the CNN's confusion matrix in Figure~\ref{fig:confuse_organ_cnn} shows some, albeit minor, increased confusion for the left and right versions of the kidney and femur. These confusions are significantly worsened for the $O(3)$ SCNN in Figure~\ref{fig:confuse_organ_o3}, showing poor performance for any of the organs with left and right versions, in addition to the confusion of other organs such as the liver and the spleen. Our $O(3)$ P-SCNN model shows a significant improvement over the SCNN in Figure~\ref{fig:confuse_organ_ours}, eliminating most missclassifications, although still showing some additional density for the left and right versions for the femur and kidney compared to the CNN. Finally, the $O(3)$ RPP from Figure~\ref{fig:confuse_organ_rpp} manages to show an additional slight improvement for these specific organs.

\subsection{Inspecting the Learnt Likelihood distributions}\label{ap:inspect_LL}
Here we present additional learnt likelihood distributions for the datasets. We often refer to specific elements of \( O(2) \) to distinguish between different types of transformations. To simplify the notation, we use colour-coding: \textcolor{blue}{blue} signifies a basic rotation, such as a rotation by \( \pi \) denoted as \(\textcolor{blue}{\pi}\), while \textcolor{red}{red} indicates a composite transformation involving a horizontal reflection followed by a rotation, exemplified by a horizontal reflection and a \( \frac{\pi}{2} \) rotation denoted as \(\textcolor{red}{\frac{\pi}{2}}\). Therefore, this section is best viewed in colour.

\paragraph{\texttt{Vectors}} 

\begin{figure}[h!]
        \centering
        \begin{subfigure}{0.4\textwidth}
        \includegraphics[width=\linewidth, trim={22cm 0cm 19.5cm 1.2cm},clip]{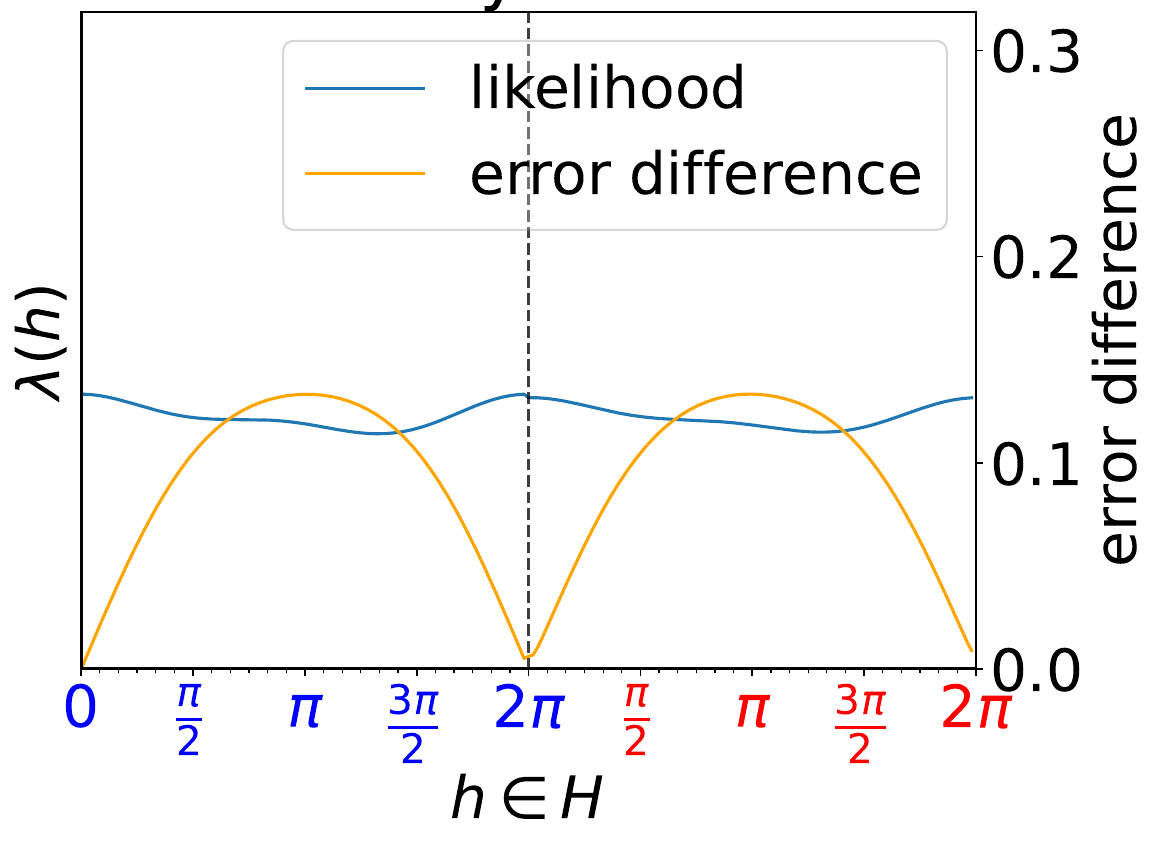}
        \caption{Angle + FourierELU}\label{fig:eq_angle_fourier}                
        \end{subfigure}
        \begin{subfigure}{0.4\textwidth}
        \includegraphics[width=\linewidth, trim={21.5cm 0cm 19.5cm 1.2cm},clip]
        {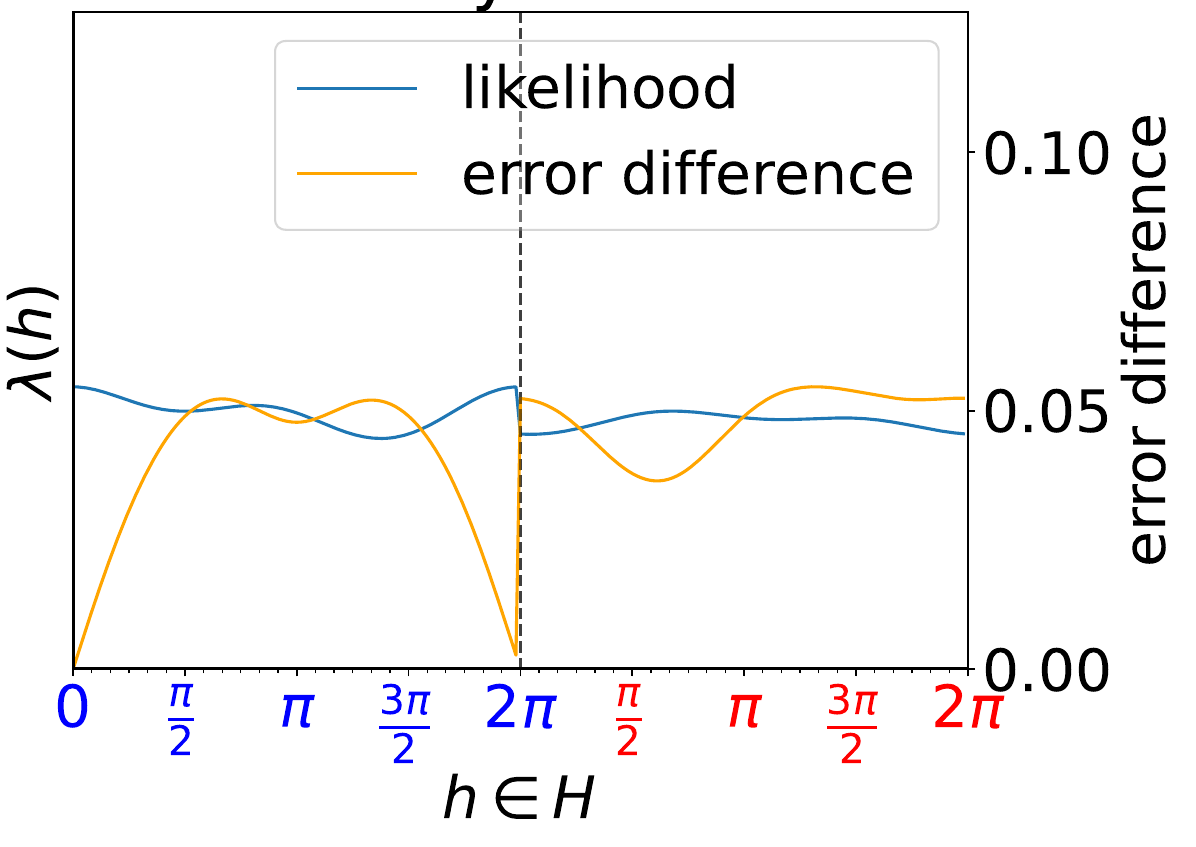}
        \caption{Angle + Gated}\label{fig:eq_angle_Gated}        
        \end{subfigure}
        \begin{subfigure}{0.4\textwidth}
        \includegraphics[width=\linewidth, trim={22cm 0cm 19.6cm 1.2cm},clip]{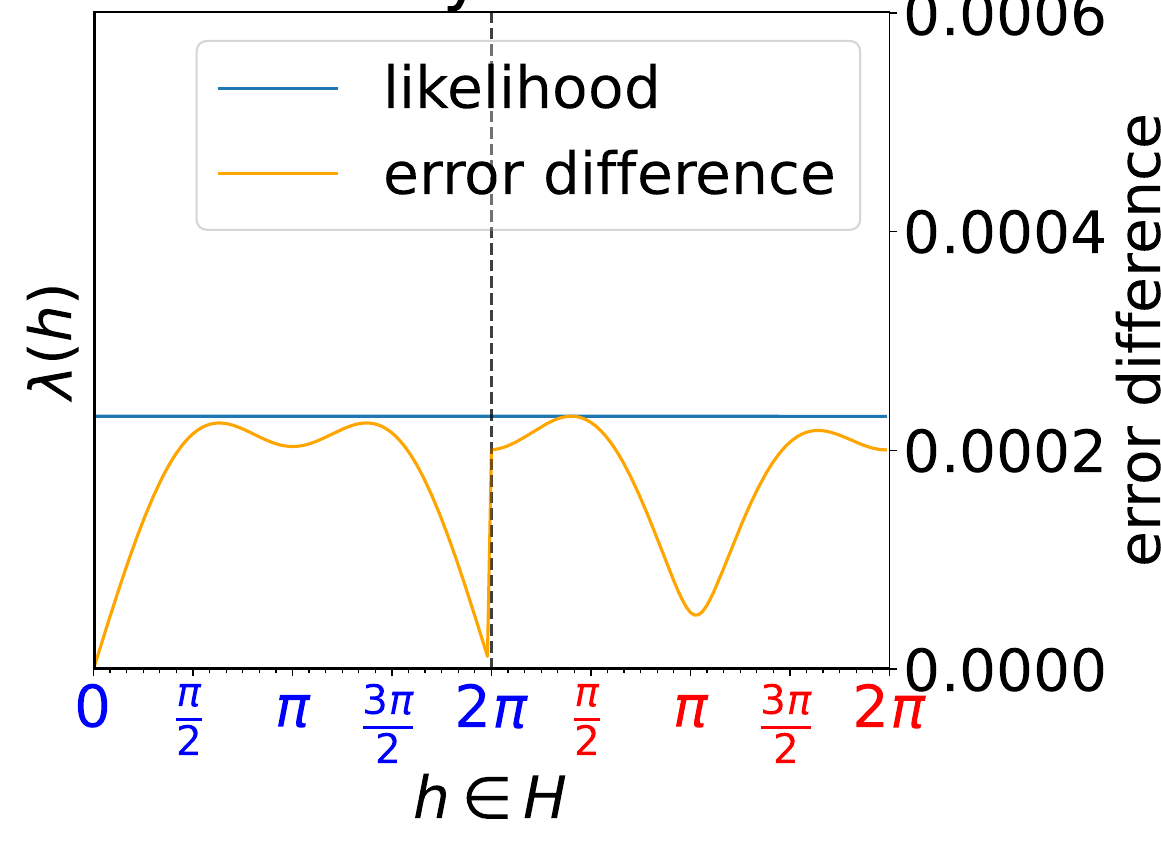}
        \caption{Norm + FourierELU}\label{fig:eq_norm_fourier}                
        \end{subfigure}
        \begin{subfigure}{0.4\textwidth}
        \includegraphics[width=\linewidth, trim={22.2cm 0cm 19.5cm 1.2cm},clip]
        {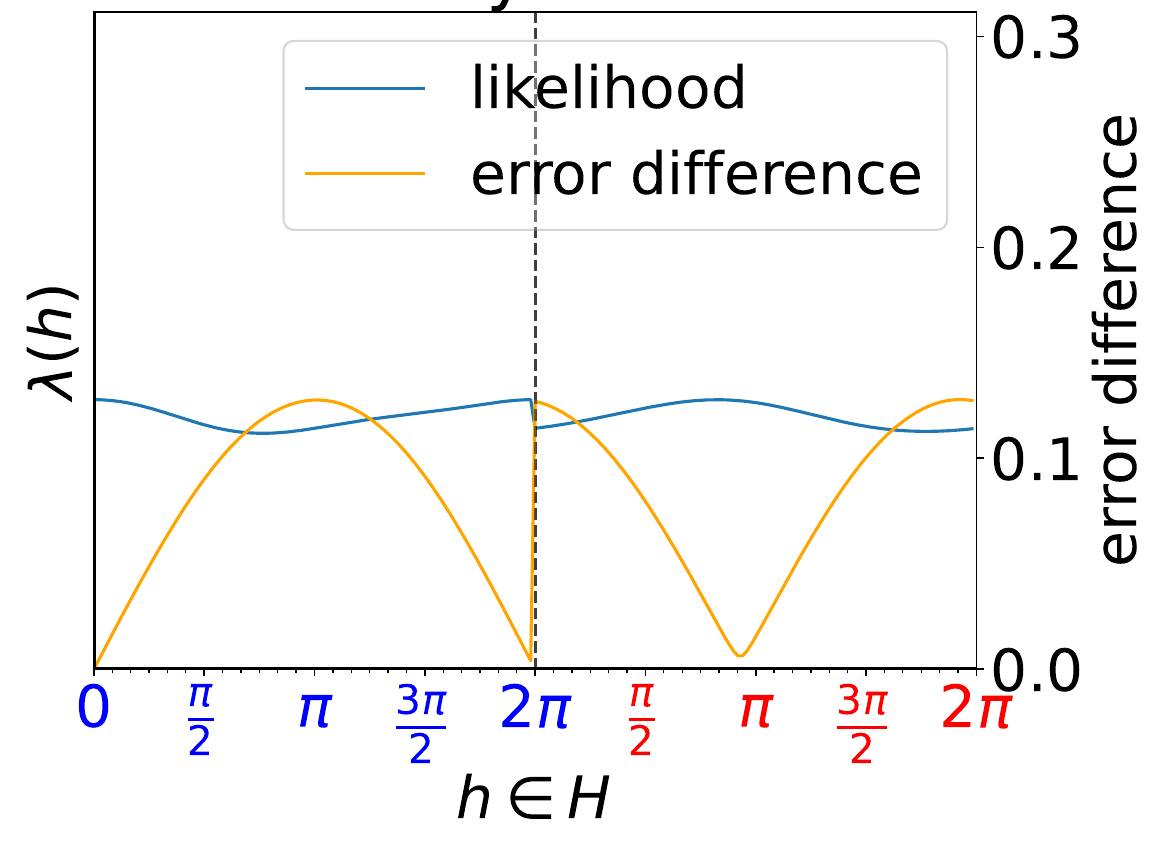}
        \caption{Norm + Gated}\label{fig:eq_norm_Gated}        
        \end{subfigure}
        \caption{Learnt likelihood $\lambda$ and error difference for the second layer in our \textit{probabilistic} PE-MLP trained on angle or norm regression, with Gated and FourierELU non-linearities. The error difference is calculated against an E-MLP. The dotted line marks the transition between the non-reflective and $O(2)$ reflection domain. Note that the scale of the equivariance error varies between the plots.}\label{fig:eq_vectors}
    \end{figure}
    
    Figure~\ref{fig:eq_vectors} displays the learnt likelihood distributions and equivariance errors for the second layer of our \textit{probabilistic} $O(2)$ PE-MLP utilising Gated and FourierELU non-linearities. These metrics are defined over the group $O(2)$, which is divided into two domains with a dotted line in our visualisations: the first for rotation elements and the second for elements comprising a horizontal reflection followed by a rotation. It should be noted that the scales differ between subfigures due to substantial variations in equivariance errors.

    An analysis of Figure~\ref{fig:eq_vectors} reveals that a decline in likelihood is typically correlated with an increase in the equivariance error. Additionally, the equivariance error exhibits frequencies comparable to those observed in the likelihood distribution. Likewise, a more uniform likelihood distribution leads to reduced equivariance errors, as highlighted in Figure~\ref{fig:eq_norm_fourier}.

    In the task of non-invariant angle prediction, both Gated and FourierELU non-linearities maintain a similar degree of partial equivariance in the rotational domain. Both break equivariance for all but the identity elements, with the lowest likelihood and the highest error occurring for rotations close to an angle of \(\textcolor{blue}{\pi}\), which is to be expected, as rotating a vector by $\pi$ yields a vector in the opposite direction. However, the Gated non-linearity exhibits reduced equivariance in the reflective domain when compared to its FourierELU counterpart. 
    
    For the invariant norm prediction task, the FourierELU non-linearity features an almost uniform likelihood distribution, with low equivariance errors as a result. Conversely, the Gated non-linearity has a less uniform distribution, displaying a signal with a frequency of one and consequently higher equivariance errors.

        \begin{figure}[h!]
        \centering
        \begin{subfigure}{\textwidth}
        \centering
        \includegraphics[width=0.9\linewidth, trim={0cm 0 20.5cm 0},clip]
        {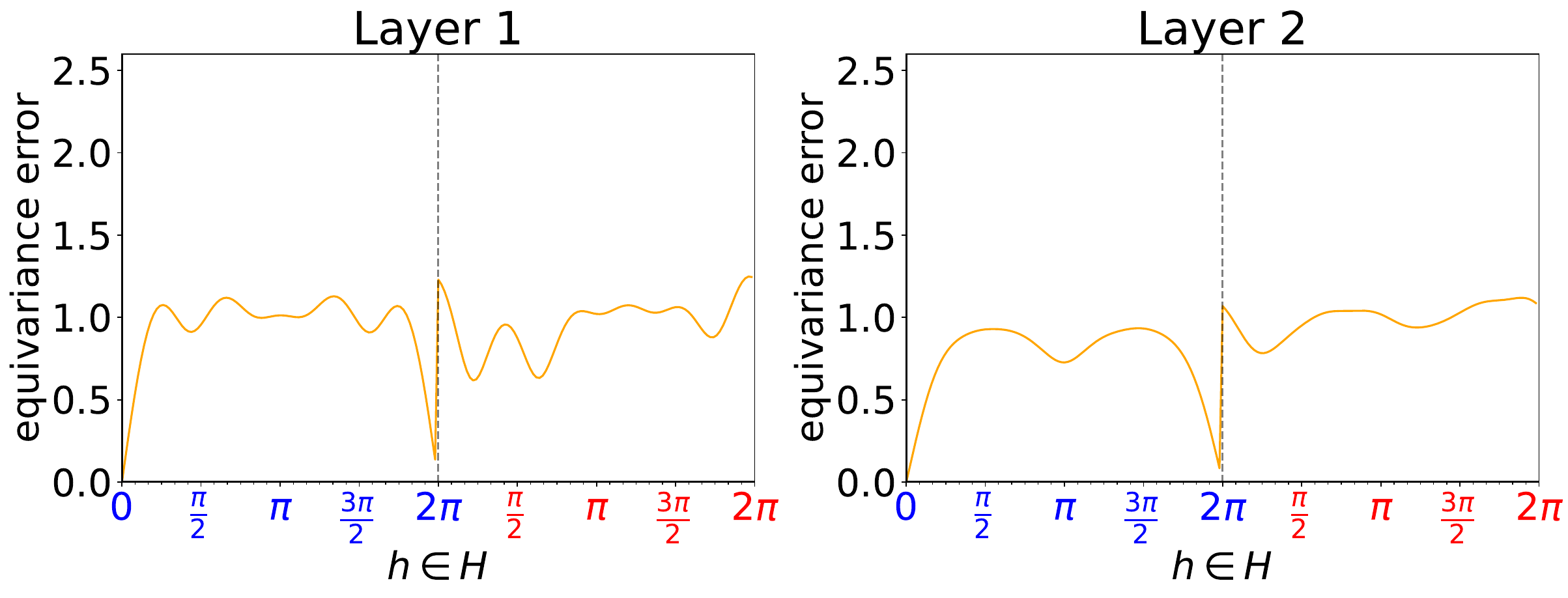}
        \caption{Preliminary approach on angle regression task.}
        \label{fig:error_prelim_angle}        
        \end{subfigure}
        \begin{subfigure}{\textwidth}
        \centering
        \includegraphics[width=0.9\linewidth, trim={0cm 0 20.5cm 0},clip]
        {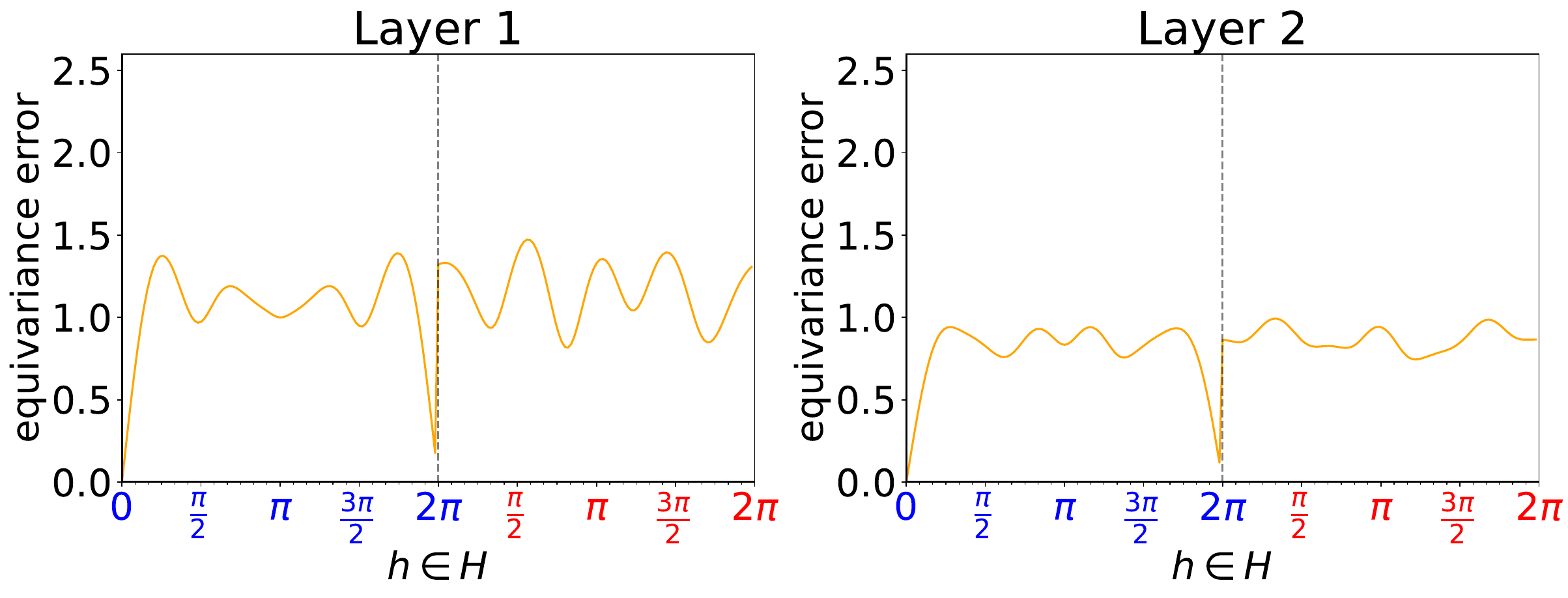}
        \caption{Preliminary approach on norm regression task.}
        \label{fig:error_prelim_norm}
        \end{subfigure}        
        \caption{Measured equivariance error of the first two layers in our \textit{preliminary}  PE-MLP with added noise trained on angle or norm regression with Gated non-linearity. The error difference is calculated against an E-MLP. The dotted line marks the transition between the non-reflective and $O(2)$ reflection domain. Note that the scale of the equivariance error varies between the plots.}\label{fig:error_preliminary}
        
    \end{figure}
    
    Finally, the equivariance errors in the \textit{probabilistic} approach manifest low-frequency signals, whereas those in the \textit{preliminary} approach, as seen in Figure~\ref{fig:error_preliminary}, contain high-frequency signals across both regression tasks in the first layer and in the second layer for the norm regression task. Furthermore, ignoring high-frequency fluctuations, the equivariance errors also seem to remain relatively consistent throughout the group $H$.
    
    \textit{These results suggest that the \textit{probabilistic} approach results in a more consistent and representative breaking of equivariance in comparison to the \textit{preliminary approach}, with easy interpretability as an additional advantage. Due to these results, we only consider the probabilistic approach in future sections.}

\begin{figure}[h!]
    \centering
    \begin{subfigure}{\textwidth}
    \includegraphics[width=\linewidth]{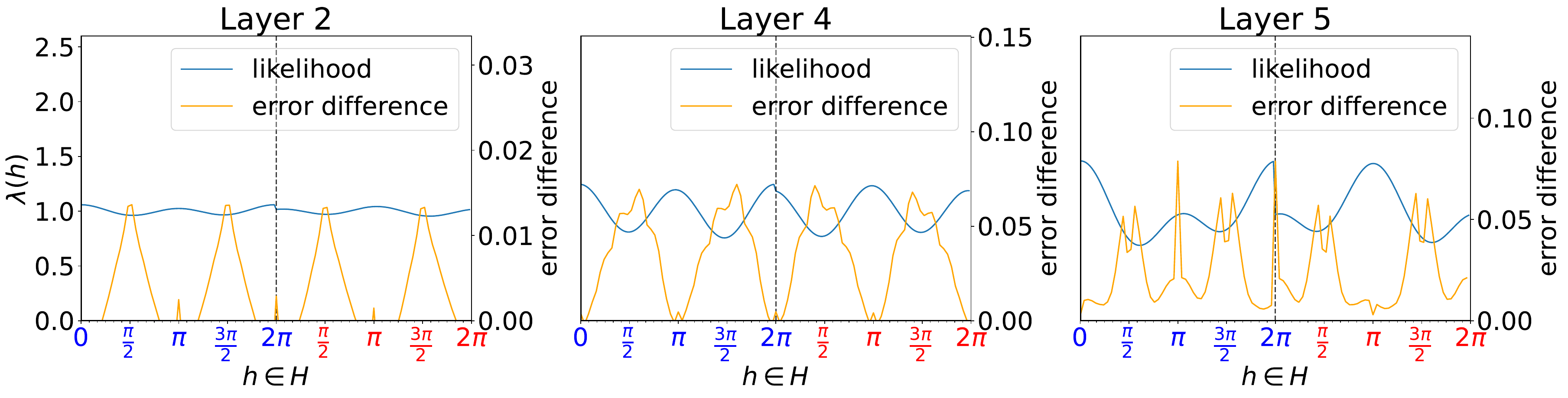}
    \caption{$O(2)$ symmetries}
    \label{fig:o2_on_o2_ap}        
    \end{subfigure}
    \begin{subfigure}{\textwidth}
    \includegraphics[width=\linewidth]{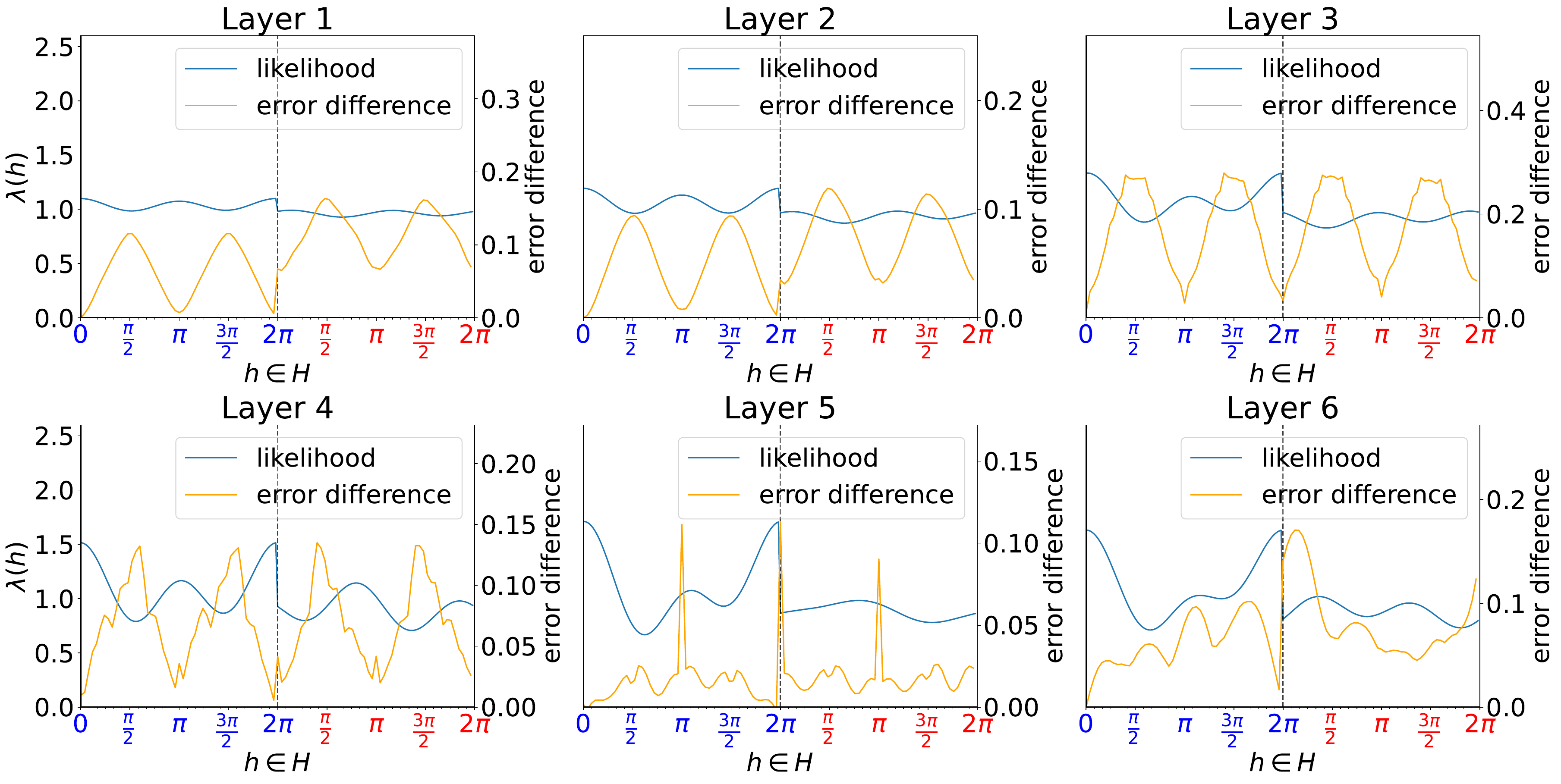}
    \caption{$C_1$ symmetries (i.e. no symmetries)}
    \label{fig:o2_on_C_1_ap}   
    \end{subfigure}
    \caption{Learnt equivariance likelihood $\lambda$ and error difference for layers $1$ through $6$ in an $O(2)$ P-SCNN trained on Double MNIST, with and without $O(2)$ symmetries. Error difference is calculated against an $O(2)$ SCNN. Dotted line marks the $O(2)$ reflection domain transition. Note that the scale of the equivariance error varies between the plots.}\label{fig:double_error_vs_learnt_ap}
\end{figure}

\paragraph{\texttt{Double MNIST}} For our \texttt{Double MNIST} dataset we compare the learnt likelihood distributions and resulting equivariance errors for each layer of our $O(2)$ P-SCNN, trained on \texttt{Double MNIST} with $O(2)$ or $C_1$ symmetries. We present the results in Figure~\ref{fig:double_error_vs_learnt_ap}. Note that earlier layers correspond to smaller-scale features and later layers correspond to larger-scale features.

Like the results from \texttt{Vectors}, we observe that a lower likelihood results in a higher equivariance error. However, while the frequencies of the errors seem to match the frequencies of the likelihoods in the first three layers, the last three layers exhibit high frequencies in their equivariance errors. This becomes especially apparent in the fifth layer \(h=\textcolor{blue}{\pi}\), indicating that this might be the result of interpolation artefacts.

\begin{figure}[h!]
    \centering
    \begin{subfigure}{\textwidth}
        \includegraphics[width=0.3\linewidth, trim={0cm, 2.5cm, 0cm, 2.5cm}, clip]{figures/og.pdf}
        \includegraphics[width=0.3\linewidth, trim={0cm, 2.5cm, 0cm, 2.5cm}, clip]{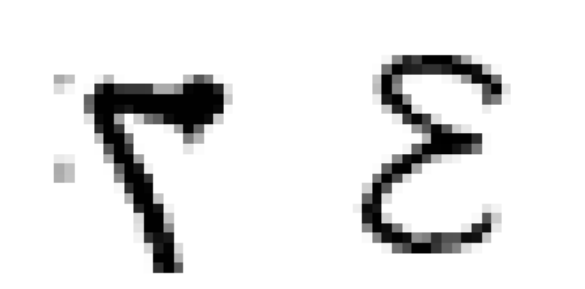}
        \includegraphics[width=0.3\linewidth, trim={0cm, 2.5cm, 0cm, 2.5cm}, clip]{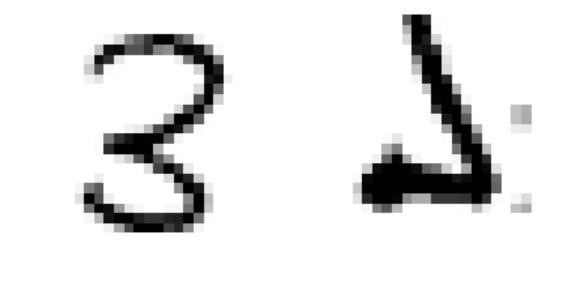}
        \caption{Global augmentations.}
    \end{subfigure}
    \begin{subfigure}{\textwidth}
    \includegraphics[width=0.3\linewidth, trim={0cm, 2.5cm, 0cm, 2.5cm}, clip]{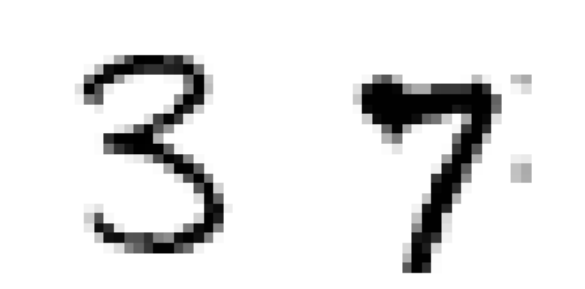}
    \includegraphics[width=0.3\linewidth, trim={0cm, 2.5cm, 0cm, 2.5cm}, clip]{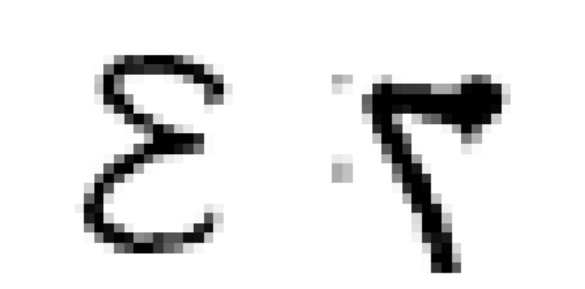}
    \includegraphics[width=0.3\linewidth, trim={0cm, 2.5cm, 0cm, 2.5cm}, clip]{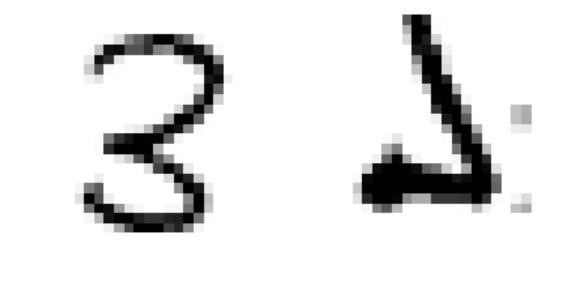}
    \caption{Local augmentations.}
    \end{subfigure}
    \caption{Illustration comparing global and local (digit-wise) augmentations on images of double-digit numbers. Each row presents a case, showing, from left to right: the original image, a horizontal reflection denoted by \(h=\textcolor{red}{0}\), and a rotation by \(h=\textcolor{red}{\pi}\). It should be noted that although the images in the middle column differ -- with the image in the bottom row denoting a different number—- the images in the last column are identical.}
    \label{fig:show_rotations}
\end{figure}

In Figure~\ref{fig:o2_on_o2_ap}, the model trained on $O(2)$ symmetries remains relatively equivariant in its first two layers. However, in layers three and four, it subsequently loses equivariance for $\frac{\pi}{2}$ and $\frac{3\pi}{2}$ rotations in both the reflective and non-reflective domains, corresponding to instances of extrinsic equivariance. Subsequently, layers 5 and 6 show a decrease in equivariance with respect to \(h=\textcolor{blue}{\pi}\) and \(h=\textcolor{red}{0}\). At a large scale, these two symmetries both result in a digit-reversal, and are thus instances of incorrect equivariance, with the former also turning the individual digits upside down. \textit{Interestingly, the model maintains a high degree of equivariance with respect to \(h=\textcolor{red}{\pi}\), while applying this symmetry at a large scale corresponds to a rotation by $\pi$ on the individual digits, as the digits are swapped twice. Therefore, this is indeed a case of correct equivariance, even though we did not initially consider this symmetry.} See Figure~\ref{fig:show_rotations} for a visualisation.

Comparatively, the model trained on $C_1$ symmetries in Figure~\ref{fig:o2_on_C_1_ap} breaks equivariance much more rapidly throughout the network compared to the previous model, with higher equivariance errors as a result. Most notable, reflection equivariance is significantly reduced in layer two rather than layer five, and the final two layers do not maintain equivariance with respect to \(h=\textcolor{red}{\pi}\). This is to be expected, as there are no rotation or reflection augmentations under $C_1$ symmetries, and therefore these transformations are cases of extrinsic equivariance.
 
    \paragraph{\texttt{MedMNIST}} 
While 2D rotations can be described using a single angle, making them easy to visualise, 3D rotations require at least three parameters and can be represented in multiple ways —such as Euler angles, quaternions, or rotation matrices— complicating their intuitive understanding and visualisation. However, similar to how elements of 2D rotations can be interpreted as elements on a circle, or a $1$-sphere, the quaternion representation of 3D rotations can be interpreted as elements on a $3$-sphere, which is a sphere living in $4$D space. As this is still difficult to visualise, we aim for a different approach. 

\begin{figure}[h!]
    \begin{subfigure}[b]{0.9\textwidth}
        \includegraphics[width=0.195\linewidth]{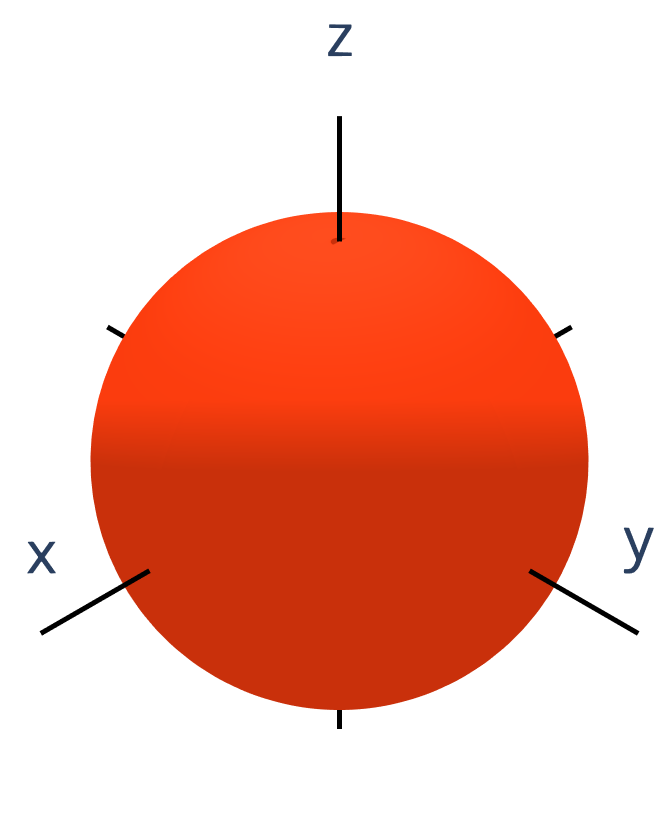}
        \includegraphics[width=0.195\linewidth]{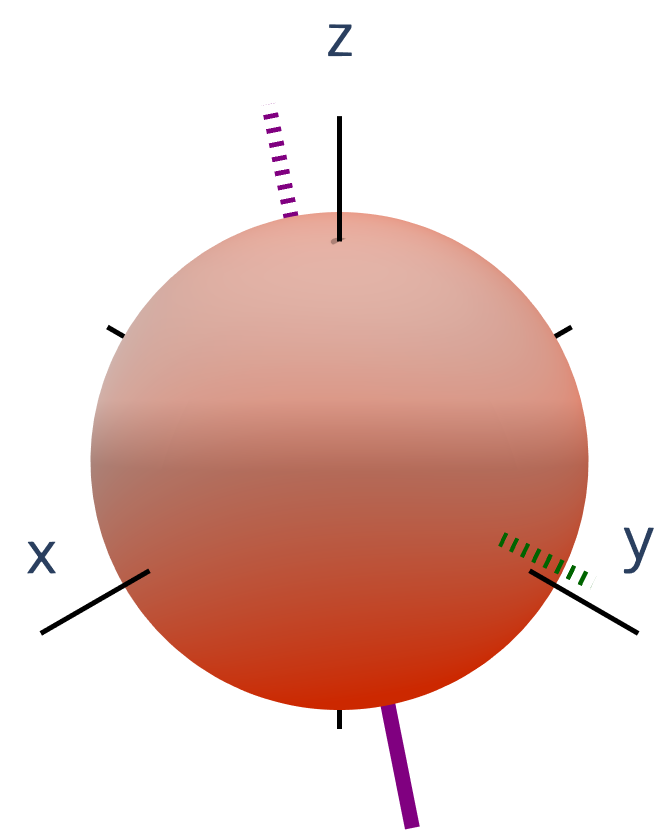}
        \includegraphics[width=0.195\linewidth]{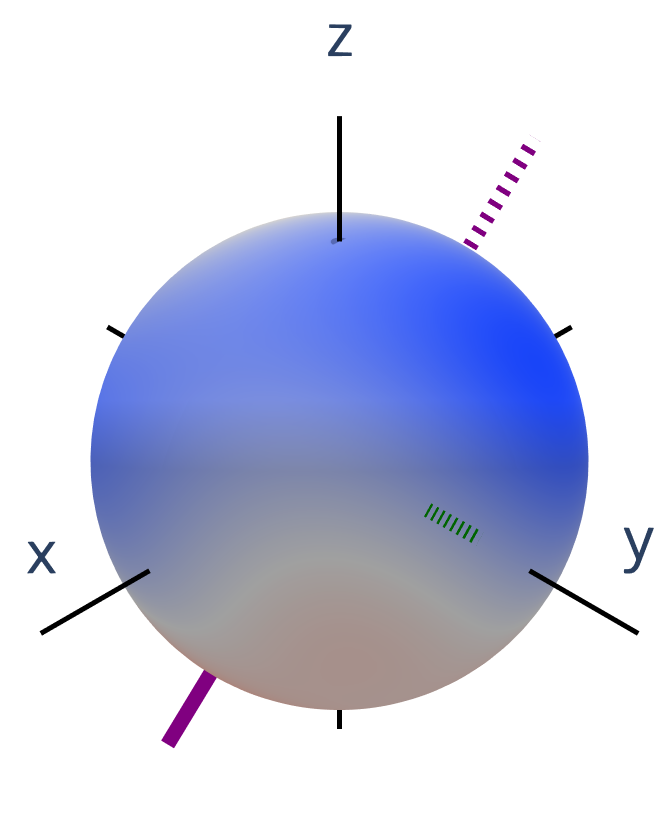}
        \includegraphics[width=0.195\linewidth]{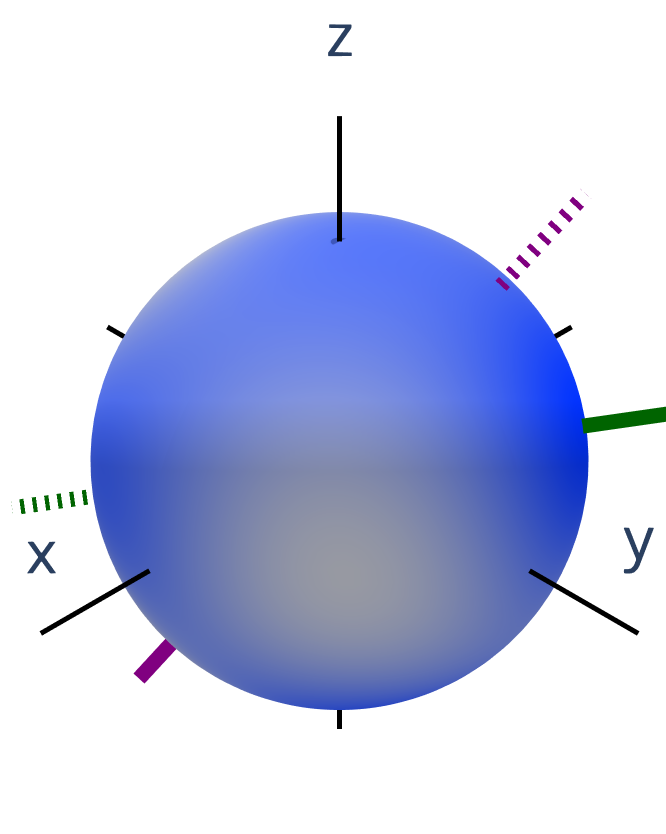}
        \includegraphics[width=0.195\linewidth]{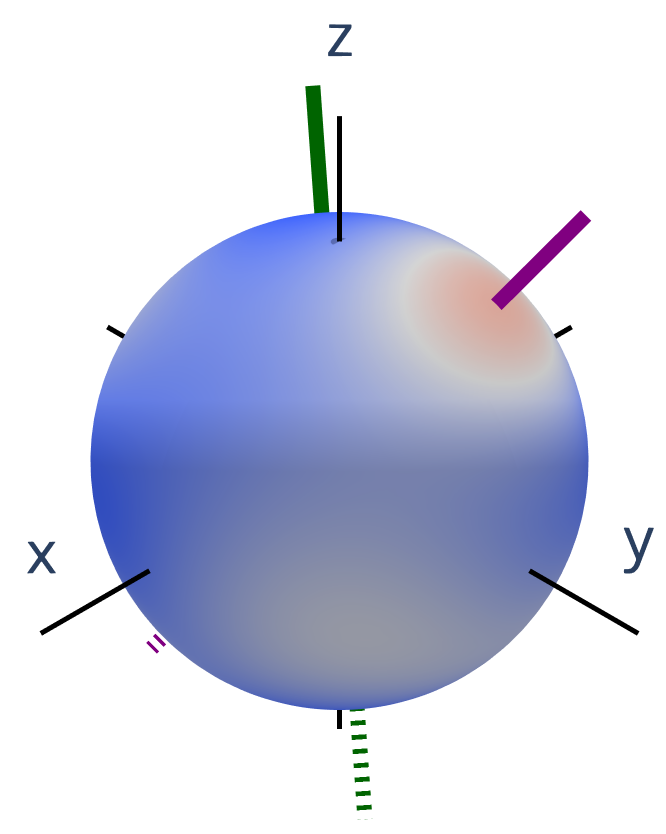}
        \includegraphics[width=0.195\linewidth]{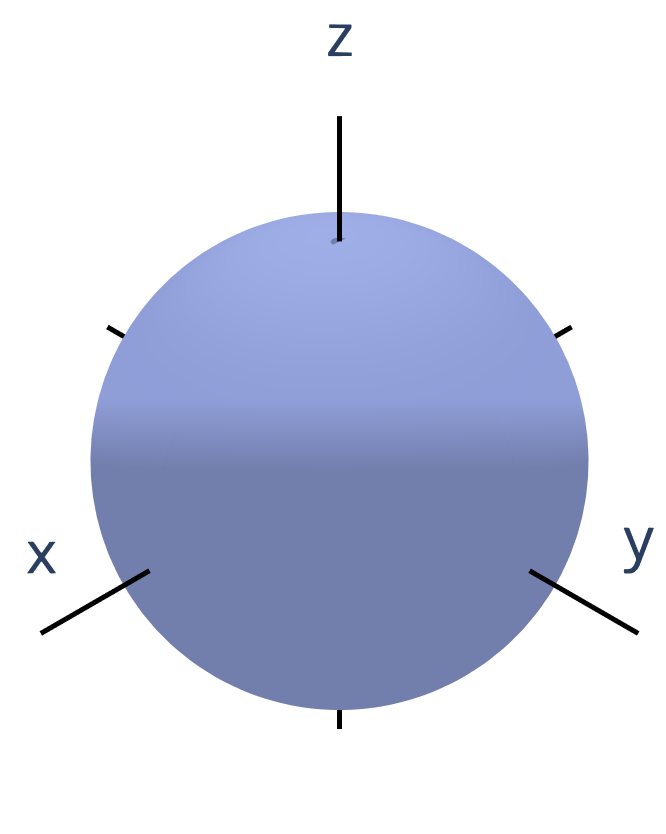}
        \includegraphics[width=0.195\linewidth]{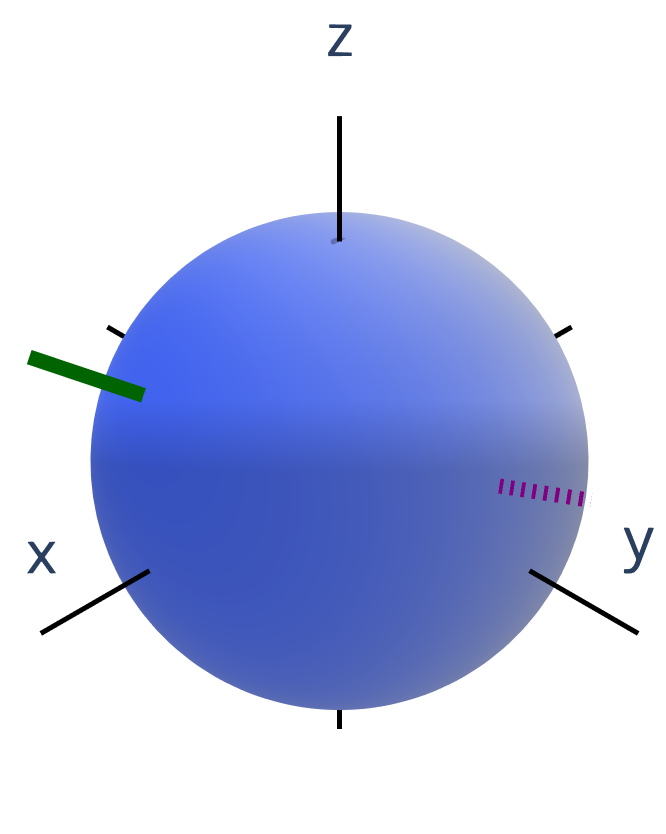}
        \includegraphics[width=0.195\linewidth]{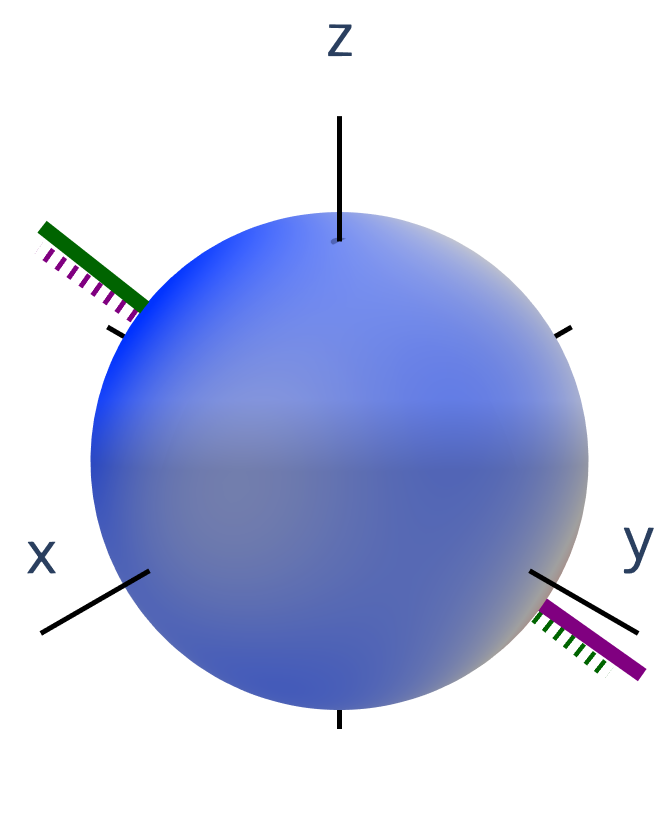}
        \includegraphics[width=0.195\linewidth]{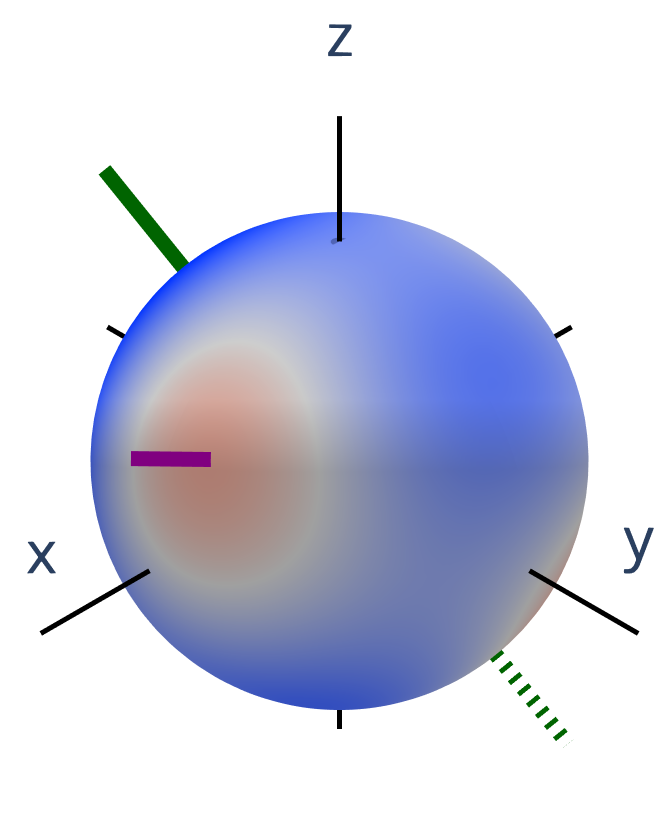}
        \includegraphics[width=0.195\linewidth]{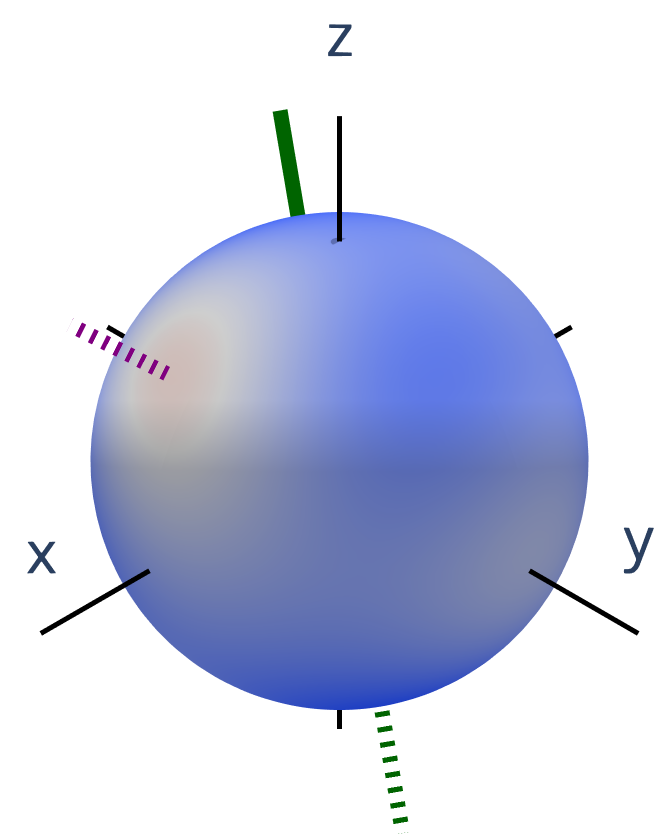}
        \caption{\texttt{OrganMNIST3D}}
        \includegraphics[width=0.195\linewidth]{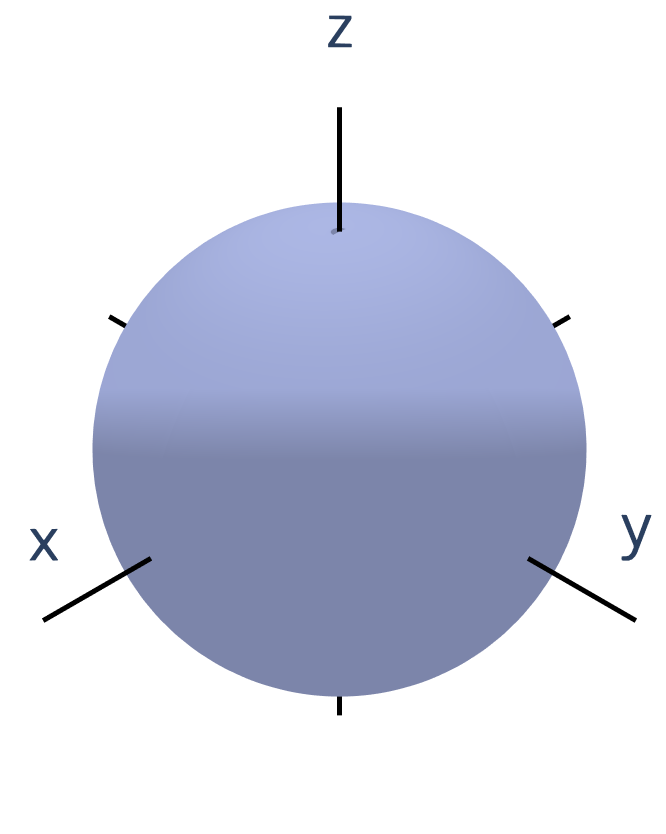}
        \includegraphics[width=0.195\linewidth]{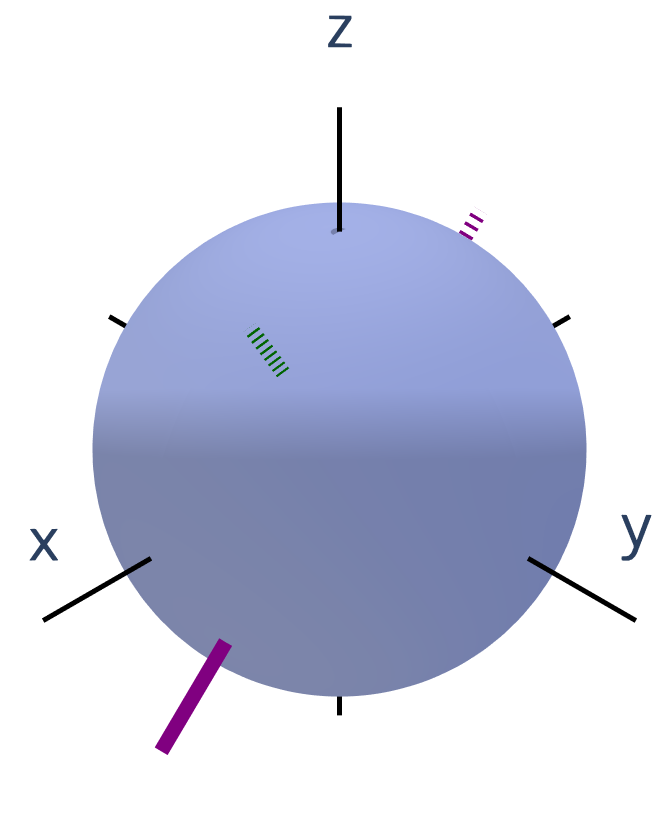}
        \includegraphics[width=0.195\linewidth]{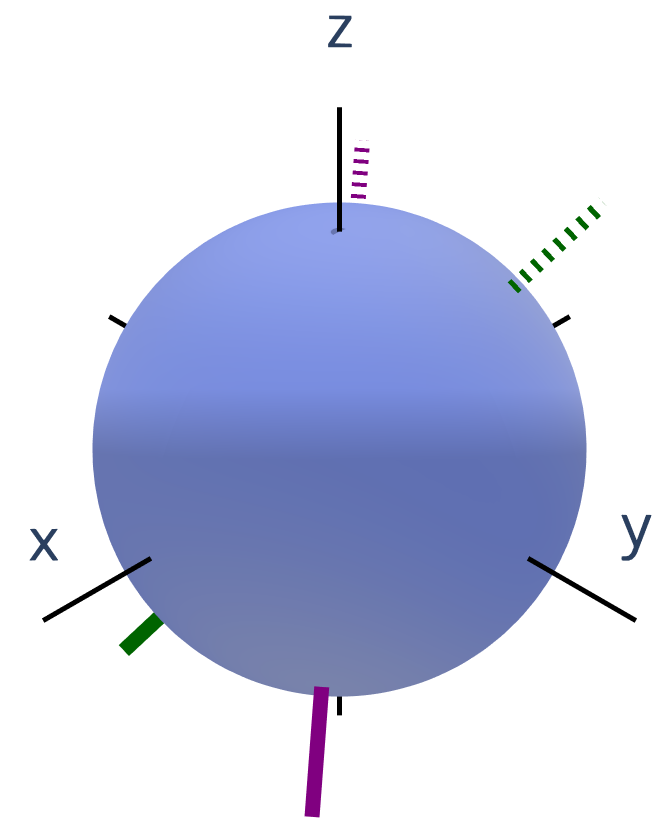}
        \includegraphics[width=0.195\linewidth]{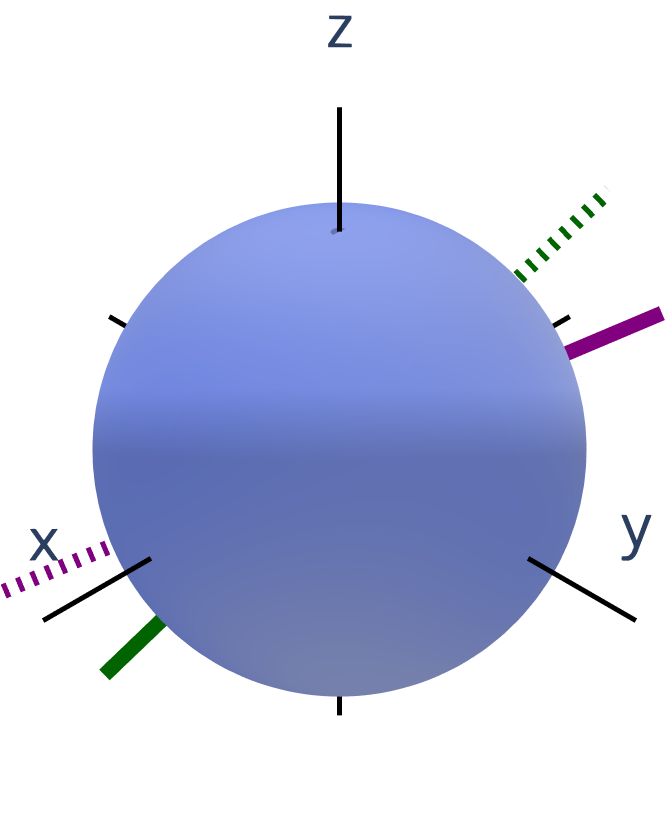}
        \includegraphics[width=0.195\linewidth]{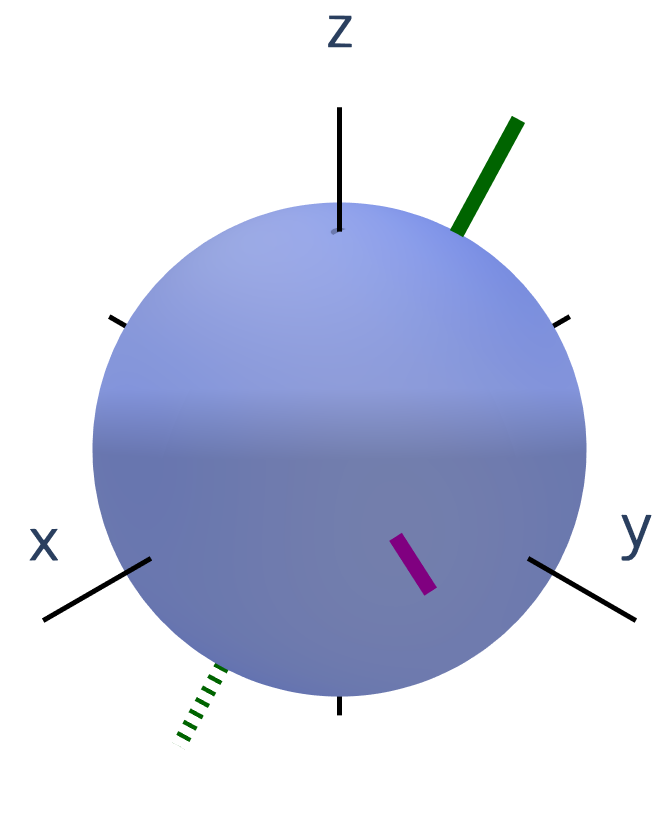}
        \includegraphics[width=0.195\linewidth]{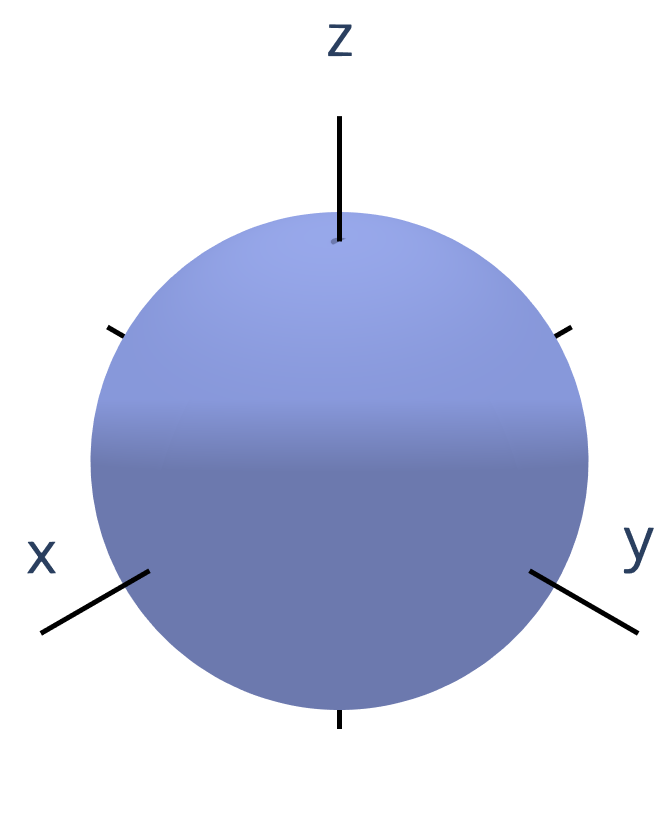}
        \includegraphics[width=0.195\linewidth]{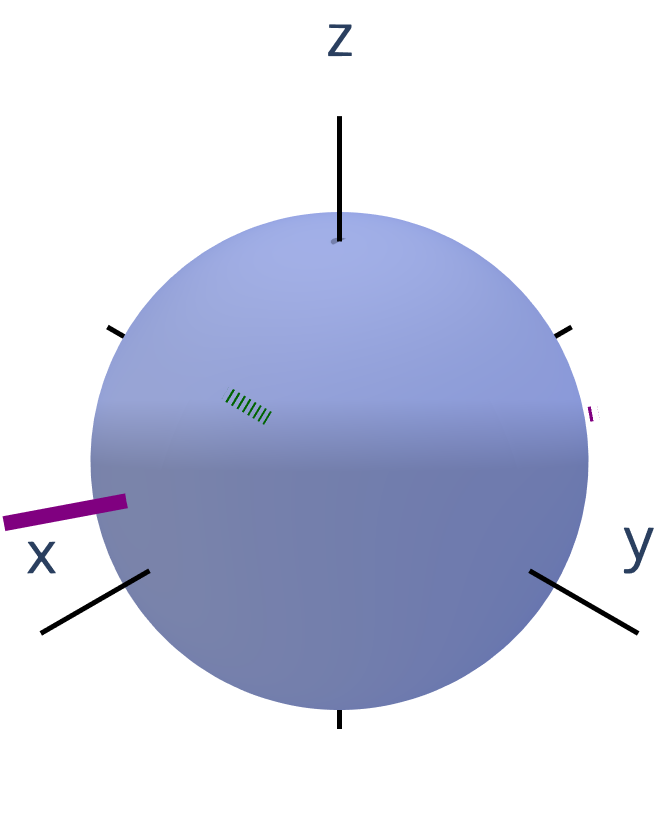}
        \includegraphics[width=0.195\linewidth]{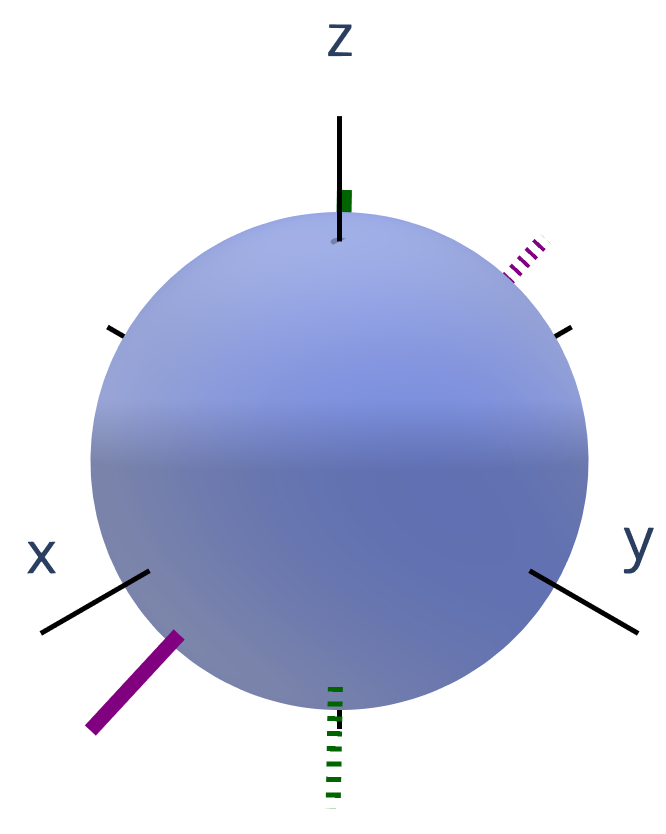}
        \includegraphics[width=0.195\linewidth]{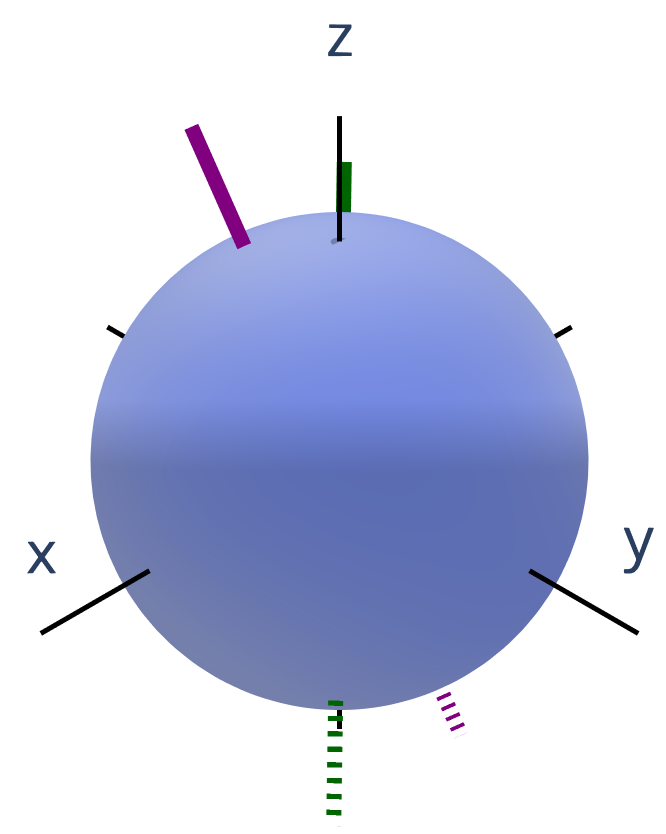}
        \includegraphics[width=0.195\linewidth]{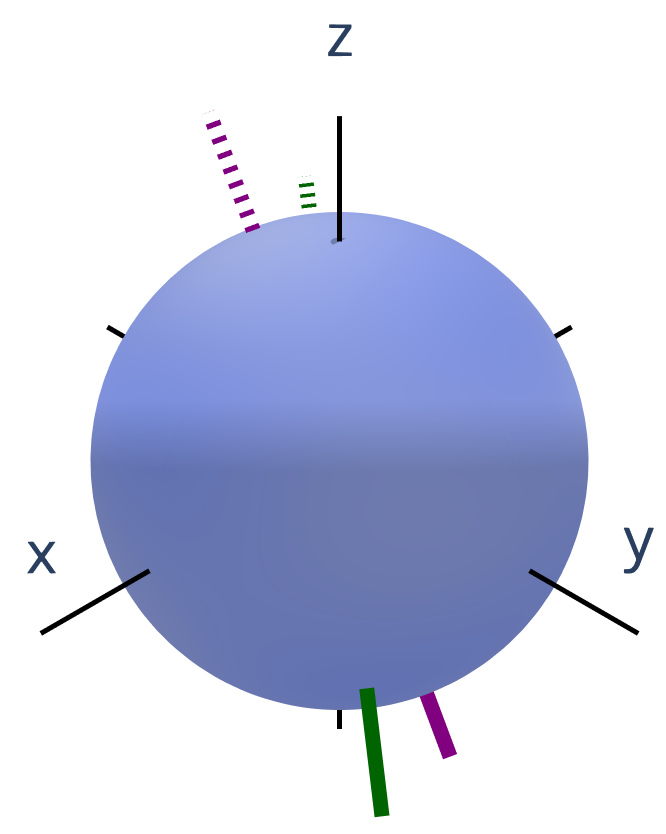}
        \caption{\texttt{SynapseMNIST3D}}
    \end{subfigure}
    \begin{subfigure}[b]{0.05\textwidth}
        \resizebox{\width}{17cm}{\includegraphics[scale=0.7]{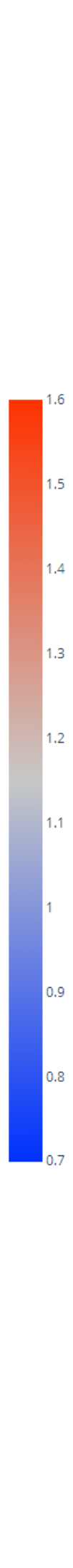}}
    \end{subfigure}
    \caption{Heatmap of the learnt $O(3)$ likelihood distribution from the second residual block for \texttt{OrganMNIST3D} and \texttt{SynapseMNIST3D} visualised as $2$-sphere slices of the $3$-ball using the axis-angle representation. A higher value corresponds to a higher degree of equivariance. For each dataset, the first row contains the non-reflective domain and the second row the reflective domain. The columns correspond to rotations over the Euler vectors by increments of $\frac{\pi}{4}$ up to $\pi$. The first column indicates a rotation of $0$ radians and is thus uniform. The solid green and purple lines point in the directions of the respective lowest and highest likelihood for each sphere, with dotted lines denoting their opposite directions to improve the visualisation in case of occlusions. Each $2$-sphere is sampled at 10,000 points.} \label{fig:eq_med_mnist} 
    \end{figure}

    For our parameterisation, we choose an axis-angle representation, where each rotation is described by a 3D unit vector, or Euler vector, and a single rotation between $0$ and $\pi$ around this vector. Since the collection of all 3D unit vectors naturally forms a $2$-sphere, the addition of a rotation vector makes this representation similar to a $3$-ball. We generate our $2$-sphere slices by taking increments of $\frac{\pi}{2}$ over the rotation angle. As a rotation over $\theta$ over an Euler vector is equal to a rotation of $-\theta$ over the opposite vector, we only need to parameterise the rotation angle up to $\pi$. Figure~\ref{fig:eq_med_mnist} shows the resulting likelihood distributions of the first residual block of our $O(3)$ P-SCNN trained on \texttt{OrganMNIST3D} and \texttt{SynapseMNIST3D}.

    Here, it becomes apparent that the model trained on \texttt{OrganMNIST3D} shows a non-uniform degree of equivariance in the non-reflective and reflective domains. Specifically, the model breaks equivariance considerably for rotations of \( \frac{\pi}{2} \) or higher for most Euler vectors in the non-reflective domain, and for any rotation in the reflective domain. However, there are exceptions where certain rotation axes result in significantly higher likelihoods for the same degree of rotation. For example, a rotation by \( \pi \) along the vector located between the positive y and z axes yields a higher likelihood. Likewise, a reflection followed by a \( \frac{3\pi}{4} \) rotation along an axis between the positive x and z axes generates a markedly elevated likelihood compared to a standalone reflection.

    On the contrary, the likelihood for \texttt{SynapseMNIST3D} is almost entirely uniform over the entire group, containing only a small decrease for non-identity elements. \textit{These results suggest that the task defined in \texttt{OrganMNIST3D} is indeed non-symmetric, whereas the task in \texttt{SynapseMNIST3D} is mostly symmetric.}

    \subsection{Shared Equivariance compared to Individual Equivariance}\label{sec:results_shared}

\begin{figure}[h!]
    \centering
    \begin{subfigure}{0.49\textwidth}
        \includegraphics[width=\linewidth, trim={0cm, 0cm, 0cm, 1cm}, clip]{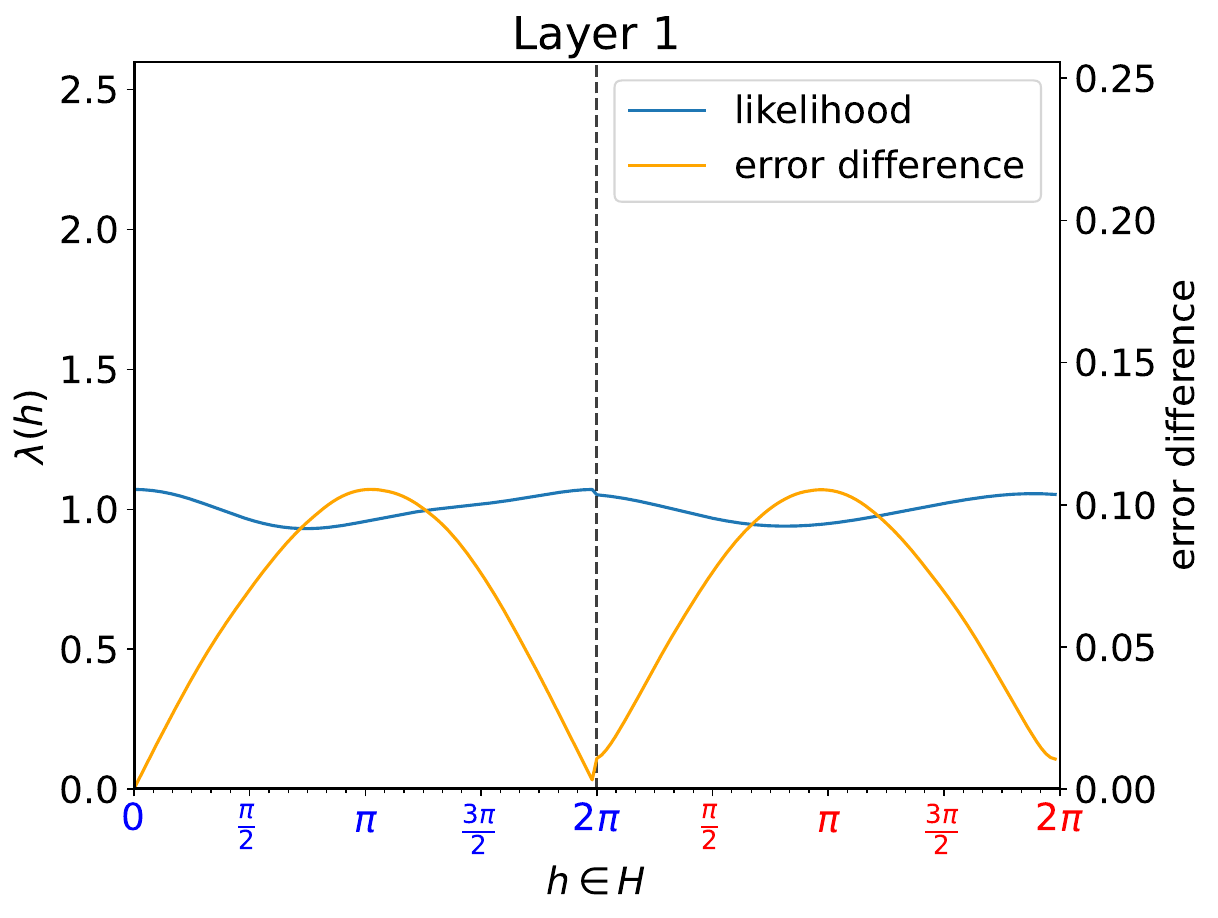}
        \caption{Angle}\label{fig:angle_oneq}        
    \end{subfigure}    
        \begin{subfigure}{0.49\textwidth}
        \includegraphics[width=\linewidth, trim={0cm, 0cm, 0cm, 1cm}, clip]{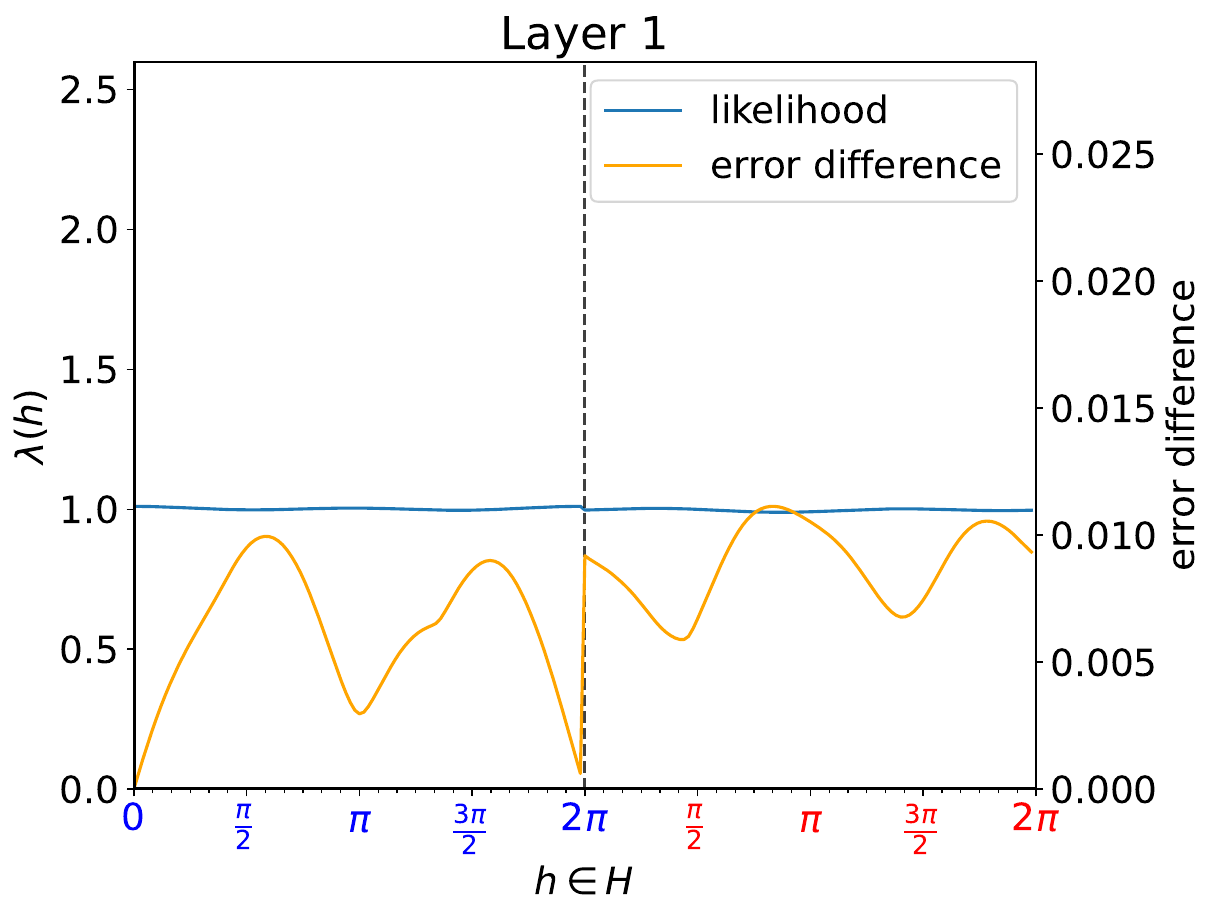}
        \caption{Norm}\label{fig:norm_oneq}
    \end{subfigure}
    \caption{Resulting likelihood distributions and equivariance errors of our PE-MLP using a shared degree of equivariance across the layers. Note that the scale of the equivariance error varies between the plots.}\label{fig:one_eq}
\end{figure}

In the previous experiments, each layer, or block of layers, has learnt its own likelihood distribution. However, as discussed in Section~\ref{section:sharing_equivariance}, since MLPs lack the concept that each layer models features of its own scale, it might be beneficial to share the degree of equivariance between MLP layers. To explore this, we compare PE-MLPs parameterised by a single shared equivariance with PE-MLPs with layer-wise likelihoods from Section~\ref{sec:results_prelim}, both using the Gated non-linearity. Table~\ref{tab:shared_eq} contains the MSE regression errors for both configurations. Figure~\ref{fig:one_eq} shows the likelihoods along with the equivariance errors for the models parameterised by a single likelihood distribution. 
In terms of regression loss, both parameterisations achieve comparable performance across both tasks, with an individual parameterisation outperforming the shared parameterisation slightly on the norm regression task. However, comparing Figure~\ref{fig:one_eq} to Figure~\ref{fig:eq_angle_Gated} and~\ref{fig:eq_norm_Gated} reveals that the likelihoods are more uniform under the layer-wise parameterisation, particularly for the norm regression task, which more closely represents the invariant nature of this task. Furthermore, for the angle regression task, the likelihood distribution only appears to exhibit a frequency one signal, whereas the individual parameterisation in Figure~\ref{fig:eq_angle_Gated} also contains higher frequencies. This more closely follows our expectations regarding the symmetries of the tasks in this dataset described in Section~\ref{sec:vectors_dataset}.

\begin{table}[h!]
    \centering
    \begin{tabular}{l|cc}
    \toprule
     \textbf{Equivariance} & \textbf{Angle}                 & \textbf{Norm}                  \\ \midrule\midrule
     Shared & $\bm{0.044}\ $\myfontsize{$ ( 0.003)$} & $0.064\ $\myfontsize{$ ( 0.016 )$} \\
    Individual    & $0.0046 \ $\myfontsize{$ ( 0.005)$} & $\bm{0.052} \ $\myfontsize{$ ( 0.008)$}\\
    \bottomrule
    \end{tabular}
    \caption{Comparison of MSE test scores between models parameterised by shared and individual degrees of equivariance on \texttt{Vectors}. For both regression tasks, the lowest error is  \textbf{bold}. Standard deviations over 5 runs are denoted in parentheses.}\label{tab:shared_eq}
\end{table}

\textit{These findings suggest that sharing a single degree of equivariance between subsequent linear layers (MLP) is preferable to learning distinct degrees for each. This approach yields more representative likelihood distributions.}

\subsection{Investigating the Effect of Regularisation Methods}\label{sec:results_regularisation}
In Section~\ref{sec:ensure_eq}, we proposed two error terms that serve as regularisation components for the training objective function. One is the alignment loss, responsible for aligning the likelihood properly with the identity element, while the other is the KL-divergence, aimed at discouraging the model from inaccurately recovering lost equivariance in layers that follow. In the previous experiments, both regularisation terms were enabled. Therefore, we analyse the effect of separately disabling both these regularisation methods on the quantitative and qualitative results in Sections~\ref{sec:result_align}~and~\ref{sec:result_kl}. 

Additionally, in Section~\ref{sec:bandlimit}, we proposed to bandlimit the likelihood distribution as further method of regularisation. As described in Section~\ref{sec:model_architectures}, all previous SCNN experiments used a bandlimit of $L=2$. Section~\ref{sec:result_bandlimit} explores the effect of higher and lower levels of bandlimiting on the performance and degrees of equivariance.
    \newpage
    \subsubsection{Alignment Loss}\label{sec:result_align}
\begin{figure}[h!]
    \centering
    \begin{subfigure}{0.485\textwidth}
    \includegraphics[width=\linewidth]{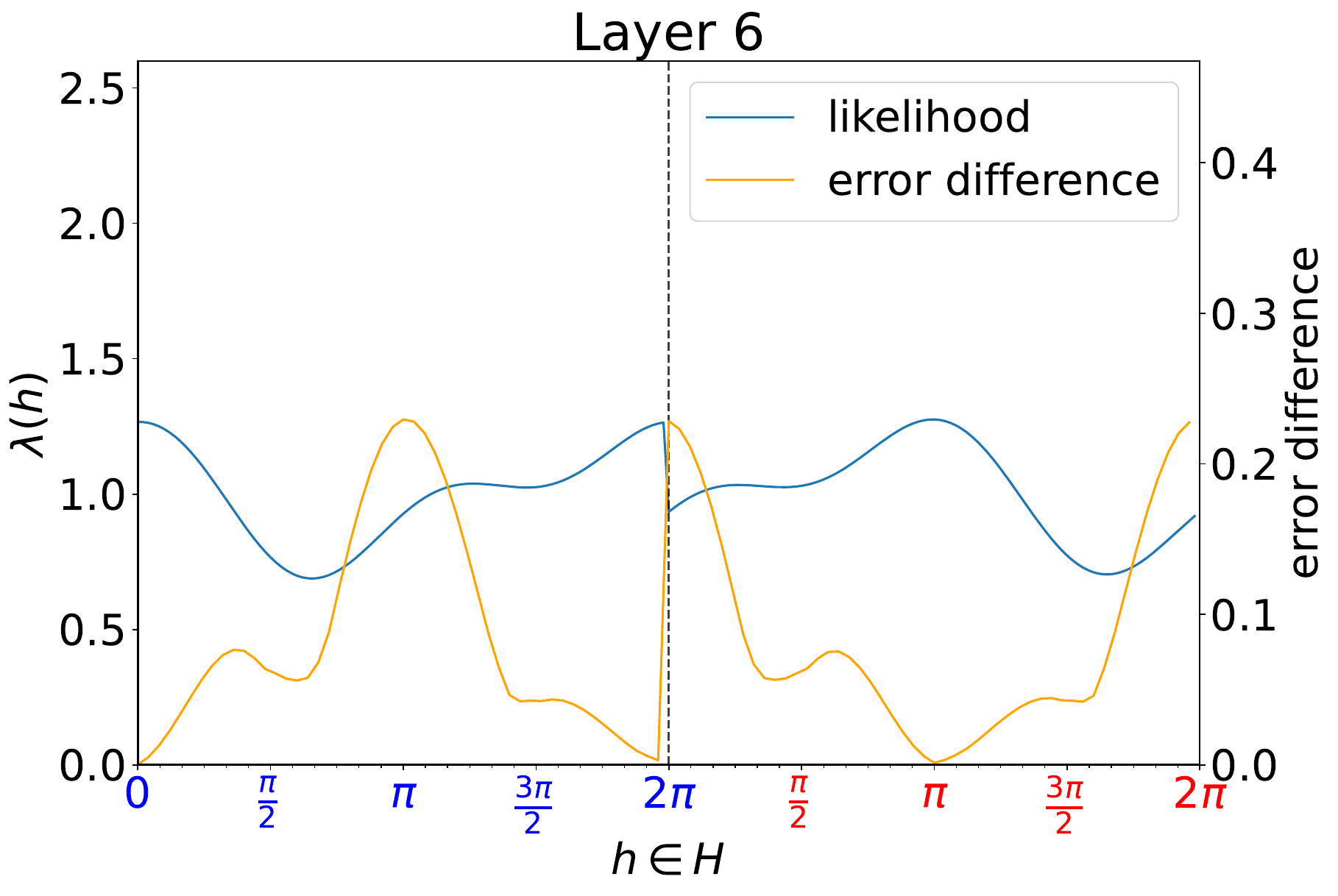}
    \caption{With alignment loss}
    \end{subfigure}    
    \begin{subfigure}{0.485\textwidth}
    \includegraphics[width=\linewidth]{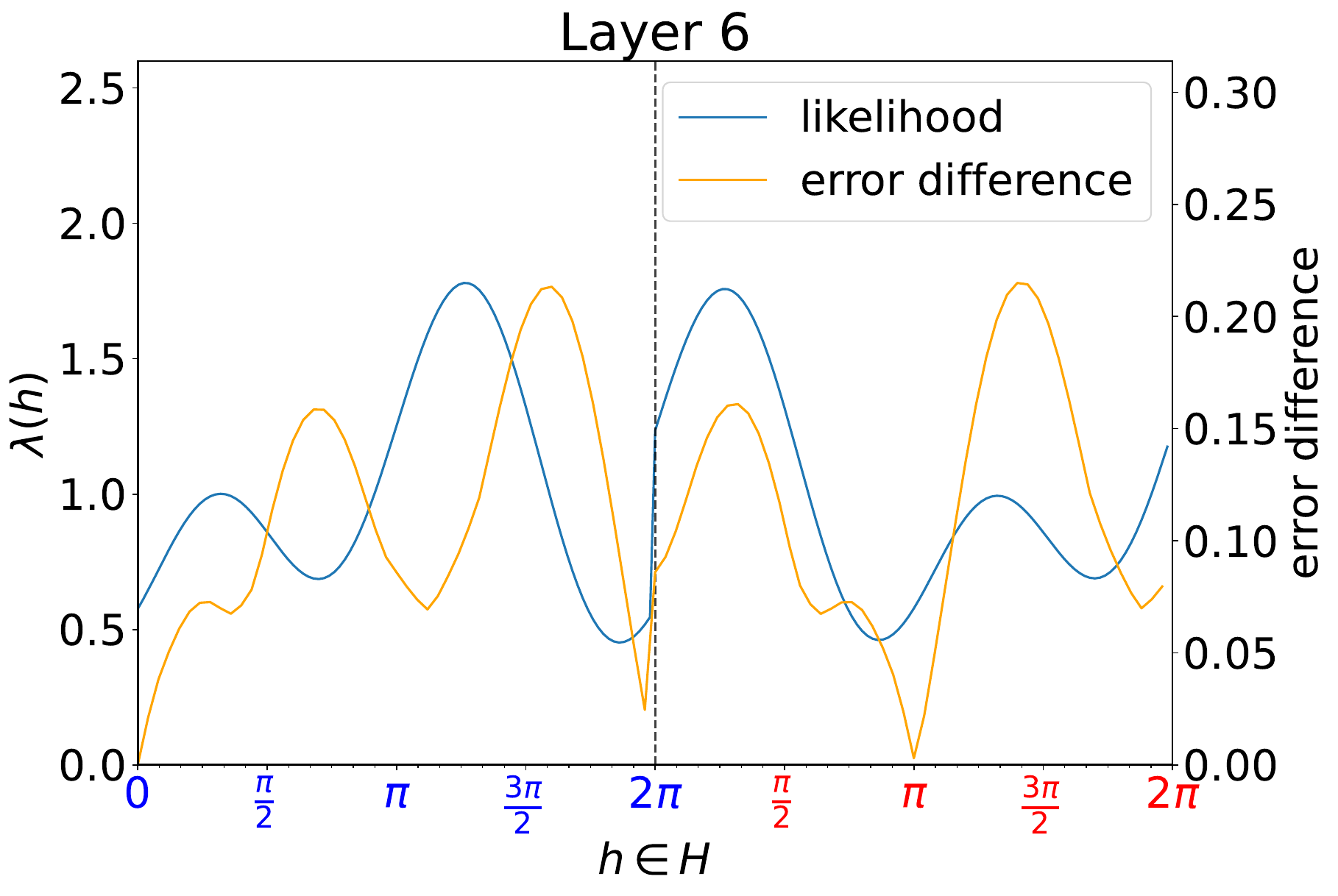}
    \caption{Without alignment loss}
    \end{subfigure}
    \caption{Comparison of the likelihood learnt in layer $6$ of an $O(2)$ P-SCNN on \texttt{Double MNIST} with $O(2)$ symmetries. Note that the scale of the equivariance error varies between the plots.}\label{fig:compare_align_likelihood}
\end{figure}

    Quantitative results for both \texttt{Vectors} and \texttt{Double MNIST} with $O(2)$ symmetries, comparing scenarios with alignment loss either enabled or disabled, are presented in Tables~\ref{tab:compare_align_vector} and \ref{tab:compare_align_mnist}, respectively. Furthermore, Figure~\ref{fig:compare_align_likelihood} provides a comparison of the resulting likelihoods and equivariance errors for the sixth layer of our $O(2)$ P-SCNN. 

    \begin{table}[h!]
    \begin{minipage}{.49\linewidth}
      \centering
        \begin{tabular}{c|cc}
        \toprule
        \textbf{Alignment Loss} & \textbf{Angle} & \textbf{Norm} \\ \midrule\midrule
        \cmark & $\bm{0.044} \ $\myfontsize{$( 0.002)$} & $\bm{0.052} \ $\myfontsize{$( 0.008)$} \\
        \xmark & $0.064\ $\myfontsize{$( 0.028)$} & $0.062 \ $\myfontsize{$( 0.018)$}\\
        \bottomrule
    \end{tabular}
    \caption{Comparing the effect of alignment loss on our PE-MLP in terms of MSE regression loss on \texttt{Vectors}. For both regression tasks, the lowest error is \textbf{bold}. Standard deviations over 5 runs are denoted in parentheses.}\label{tab:compare_align_vector}
    \end{minipage}%
    \hfill
    \begin{minipage}{.49\linewidth}
      \centering
        
    \begin{tabular}{c|c}
    \toprule
    \textbf{Alignment Loss} & \textbf{Accuracy} \\ \midrule\midrule
    \cmark & $\bm{0.819} \ $\myfontsize{$( 0.010)$} \\
    \xmark & $0.805\ $\myfontsize{$( 0.015)$}\\
        \bottomrule
    \end{tabular}
    \caption{Comparing the effect of alignment loss on our $O(2)$ P-SCNN in terms of classification accuracy on \texttt{Double MNIST} with $O(2)$ symmetries. The higest accuracy is \textbf{bold}. Standard deviations over 5 runs are denoted in parentheses.}\label{tab:compare_align_mnist}
    \end{minipage} 
\end{table}
    
    For both datasets, Tables~\ref{tab:compare_align_vector} and \ref{tab:compare_align_mnist} show a moderate increase in performance from the addition of alignment loss in the training objective. More importantly, Figure~\ref{fig:compare_align_likelihood} reveals that the likelihood distribution is not properly aligned when the alignment loss from the loss function is omitted. Here, the group element of around \(h=\textcolor{blue}{\frac{5\pi}{4}}\) yields the maximum likelihood rather than the required identity element \(h=\textcolor{blue}{0}=e\). As a result, the measured equivariance error does not match the likelihood distribution, with some high equivariance errors occurring for group elements ascribed with high likelihoods, such as \(h=\textcolor{red}{\frac{\pi}{4}}\), and vice-versa. Upon closer inspection, it appears that the entire likelihood distribution is indeed shifted by $\frac{5\pi}{4}$ across the entire distribution, since performing a shift of $-\frac{5\pi}{4}=\frac{3\pi}{4}$ aligns the peak at \(h=\textcolor{blue}{\frac{5\pi}{4}}\) to the identity \(h=\textcolor{blue}{0}=e\), in addition to aligning the peak at \(h=\textcolor{red}{\frac{1\pi}{4}}\) to \(h=\textcolor{red}{\pi}\). 

    \textit{As such, excluding alignment loss from the objective leads to misaligned and thus uninterpretable likelihood distributions, indicating that is critical for interpretability of the learnt degree of equivariance.}
    \newpage
    \subsubsection{KL-Divergence}\label{sec:result_kl}
\begin{figure}[h!]
    \centering
    \begin{subfigure}{0.98\textwidth}
    \includegraphics[width=\linewidth]{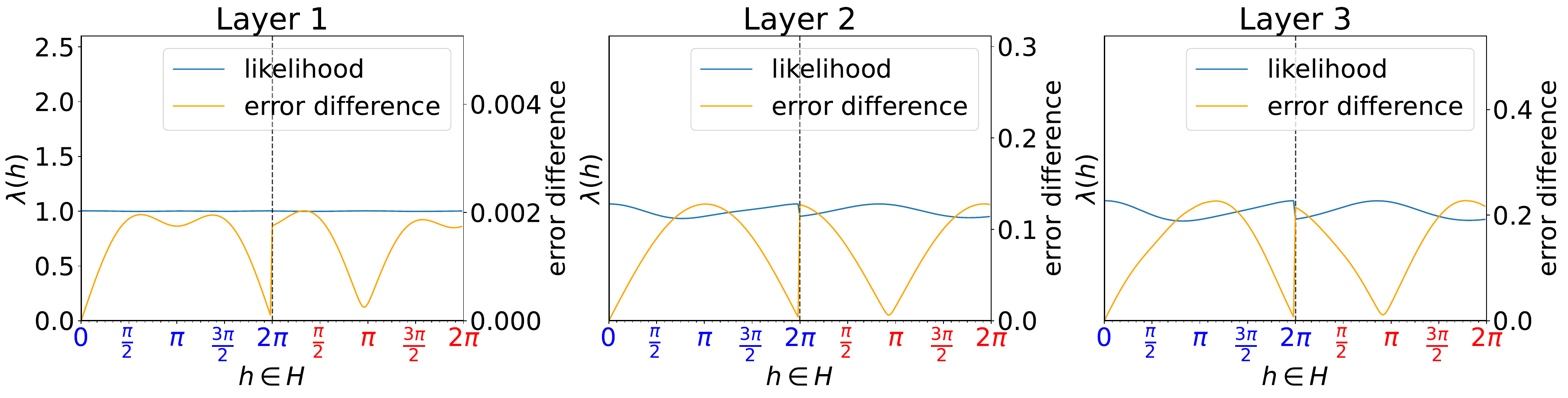}
    \caption{With KL regularisation}
    \end{subfigure}    
    \begin{subfigure}{0.98\textwidth}
    \includegraphics[width=\linewidth]{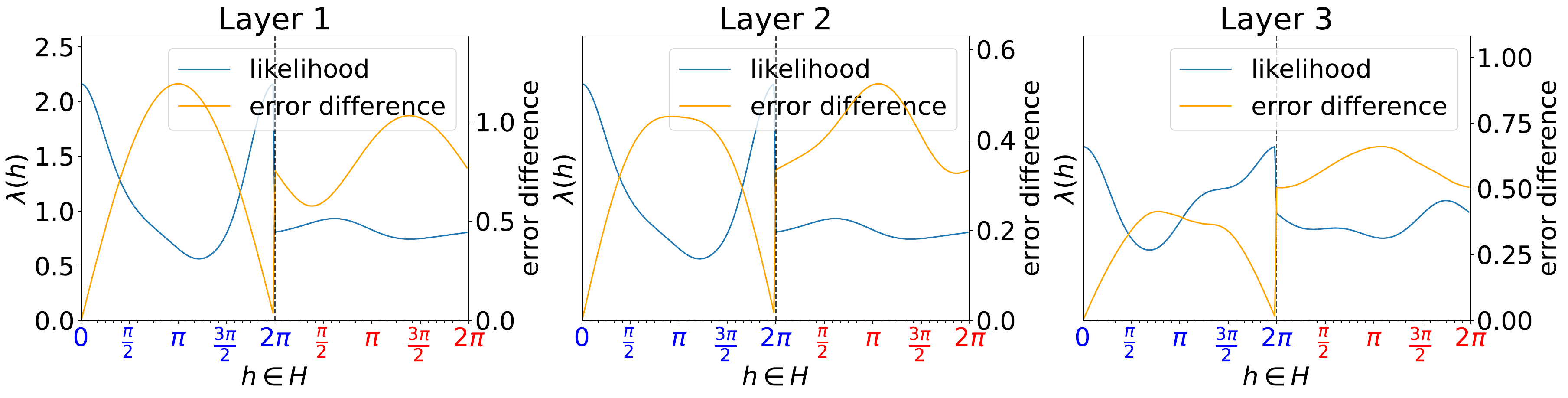}
    \caption{Without KL regularisation}
    \end{subfigure}
    \caption{PE-MLP layer-wise likelihoods with and without KL-divergence on \texttt{Vector}'s norm regression task. Note that the scale of the equivariance error varies between the plots.}\label{fig:compare_kl}
\end{figure}

\begin{figure}[h!]
    \centering
    \begin{subfigure}{0.98\textwidth}
    \includegraphics[width=\linewidth]{figures/app_results/O2_on_O2_BL_2.pdf}
    \caption{With KL regularisation.}
    \end{subfigure}    
    \begin{subfigure}{0.98\textwidth}
    \includegraphics[width=\linewidth]{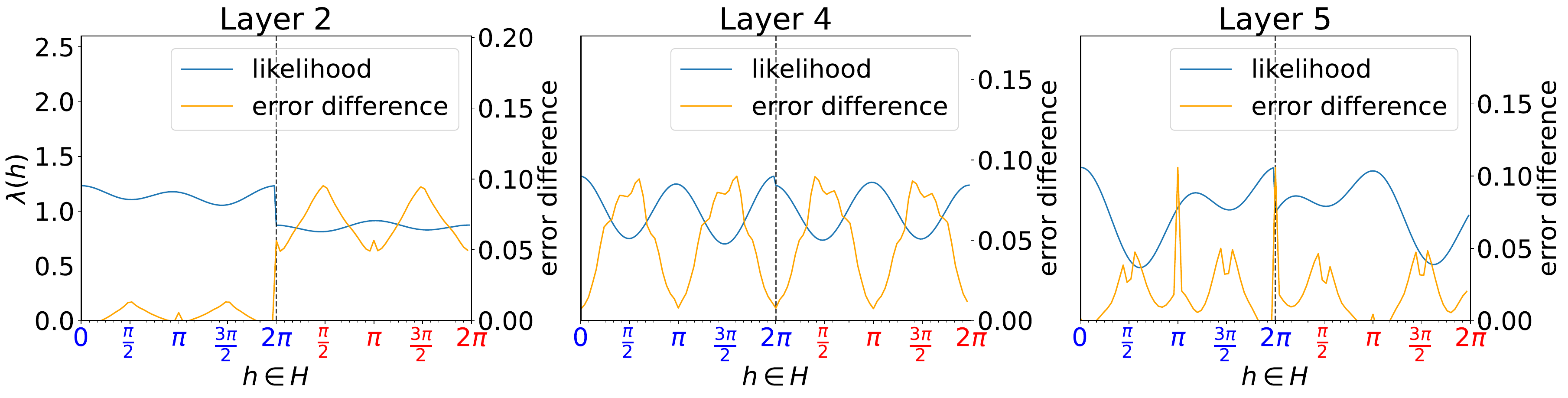}
    \caption{Without KL regularisation.}
    \end{subfigure}
    \caption{P-SCNN likelihoods for layer $2$, $4$ and $5$ with and without KL-divergence on \texttt{Double MNIST} with $O(2)$ symmetries. Note that the scale of the equivariance error varies between the plots.}\label{fig:compare_kl_double}
\end{figure}

    Similarly to the previous section, we present our quantitative results for \texttt{Vectors} and \texttt{Double MNIST} with $O(2)$ symmetries in Tables~\ref{tab:kl_vector}~and~\ref{tab:kl_double} respectively. To inspect the effect of KL-divergence on subsequent layers, we present the resulting likelihood distributions for various layers in our PE-MLP trained on norm regression and $O(2)$ PS-SCNN in Figures~\ref{fig:compare_kl} and~\ref{fig:compare_kl_double} respectively. 

    \begin{table}[h!]
    \begin{minipage}{.48\linewidth}
      \centering
        \begin{tabular}{c|cc}
        \toprule
        \textbf{KL-divergence} & \textbf{Angle} & \textbf{Norm} \\ \midrule\midrule
        \cmark & $\bm{0.044} \ $\myfontsize{$( 0.002)$} & $0.055 \ $\myfontsize{$( 0.011)$} \\
        \xmark & $0.061\ $\myfontsize{$( 0.001)$} & $\bm{0.003} \ $\myfontsize{$( 0.002)$}\\
        \bottomrule
    \end{tabular}
    \caption{Comparing the effect of KL-divergence on our PE-MLP in terms of MSE regression loss on \texttt{Vectors}.}\label{tab:kl_vector}
    \end{minipage}%
    \hspace{1cm}
    \begin{minipage}{.48\linewidth}
      \centering        
    \begin{tabular}{c|c}
    \toprule
    \textbf{KL-divergence} & \textbf{Accuracy} \\ \midrule\midrule
    \cmark & $\bm{0.819} \ $\myfontsize{$( 0.010)$} \\
    \xmark & $0.797\ $\myfontsize{$( 0.015)$}\\
    \bottomrule
    \end{tabular}
    \caption{Comparing the effect of KL-divergence on our $O(2)$ P-SCNN in terms of classification accuracy on \texttt{Double MNIST} with $O(2)$ symmetries.}\label{tab:kl_double}
    \end{minipage}
    \end{table}

    From Tables~\ref{tab:kl_vector}~and~\ref{tab:kl_double}, we observe that both \texttt{Vector}'s angle regression and \texttt{Double MNIST} number classification tasks benefit from incorporating KL-divergence in the objective function. However, \texttt{Vector}'s norm regression task sees a significant performance reduction from the KL-divergence regularisation. Figure~\ref{fig:compare_kl} shows that removing the KL-divergence from the loss function results in significantly less uniform likelihood distributions. In fact, each layer essentially entirely loses equivariance with respect to all non-identity group elements. However, as the regression loss has improved and the norm prediction is an invariant feature, the results suggest that the entire model is still invariant, despite the loss of equivariance. Furthermore, the unregularised setting shows that previously lost equivariance is incorrectly regained in the third layer at around \(h=\textcolor{red}{\frac{3\pi}{2}}\), which does not occur in the regularised setting.

\begin{wrapfigure}[15]{r}{0.56\textwidth}  
      \begin{center}
          \includegraphics[width=0.4\textwidth, trim={0.1cm 0cm 0.1cm 0.25cm},clip]{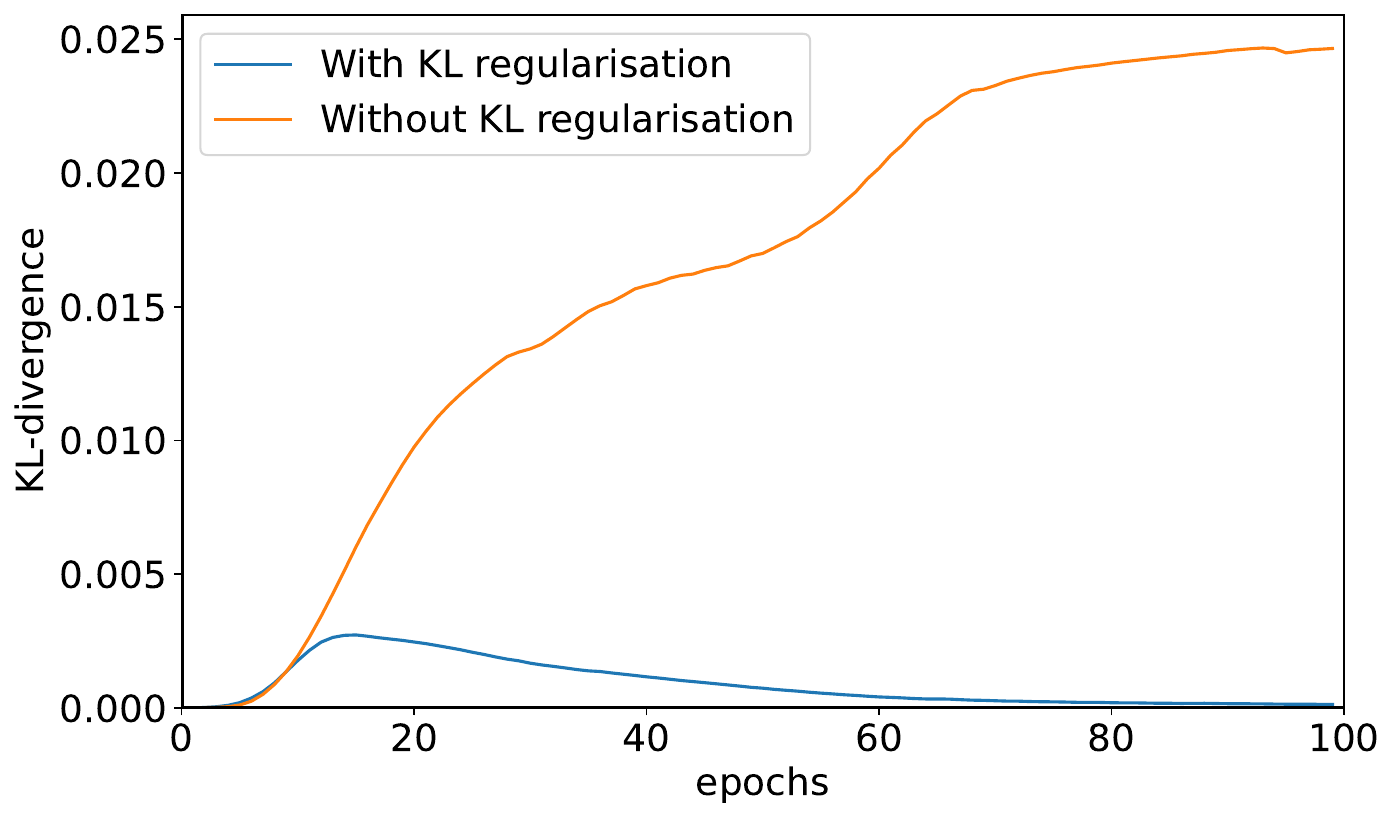}
      \end{center}
      \caption{Development of KL-divergence during training with and without KL-regularisation.}
      \label{fig:kl_epochs}
    \end{wrapfigure}

    To further investigate this behaviour, we visualise the development of the KL-divergence for each epoch during training on \texttt{Vectors}' norm regression task in Figure~\ref{fig:kl_epochs}. Here, we observe that, in both settings, the KL-divergence rises rapidly at the start of training. However, in the regularised setting, the KL-divergence starts to decrease around epoch $15$, while in the unregularised setting the KL-divergence continues to increase steadily. This might be a result of the additional parameters introduced from the breaking of equivariance. To learn the task quickly at the beginning of training, it might be most beneficial to reduce equivariance to gain access to more parameters. However, with the help of the regularisation term, the model finds a way to use fewer parameters and still accurately compute the norm at a later stage. Another potential reason is that while computing the norm is an invariant task, given the available non-linearity and network structure, the most efficient way to compute the norm might be through non-invariant intermediate features that result in an invariant feature when combined. Both of these potential causes are supported by the fact that the error is lower when KL-regularisation is disabled.

    For \texttt{Double MNIST}, Figure~\ref{fig:compare_kl_double} shows that without KL regularisation our $O(2)$ P-SCNN similarly suffers from incorrectly regaining previously lost equivariance. In the second layer, equivariance is lost for the entire reflective domain. However, in the fourth layer equivariance for \(h=\textcolor{red}{0}\) and \(h=\textcolor{red}{\pi}\) has been regained. Subsequently, the fifth layer models a relatively high likelihood for \(h=\textcolor{blue}{\frac{3\pi}{2}}\) and \(h=\textcolor{red}{\frac{\pi}{2}}\), whereas the fourth layer models a significantly lower likelihood for these group elements.

    \textit{Overall, these results indicate that while KL-regularisation might result in a performance reduction on certain datasets, it makes the learnt likelihood distributions more representative of the task.}
    \newpage
    \subsubsection{Bandlimiting}\label{sec:result_bandlimit}
\begin{figure}[h!] 
      \centering
          \includegraphics[width=\textwidth,trim={0cm 19cm 0cm 0cm},clip]{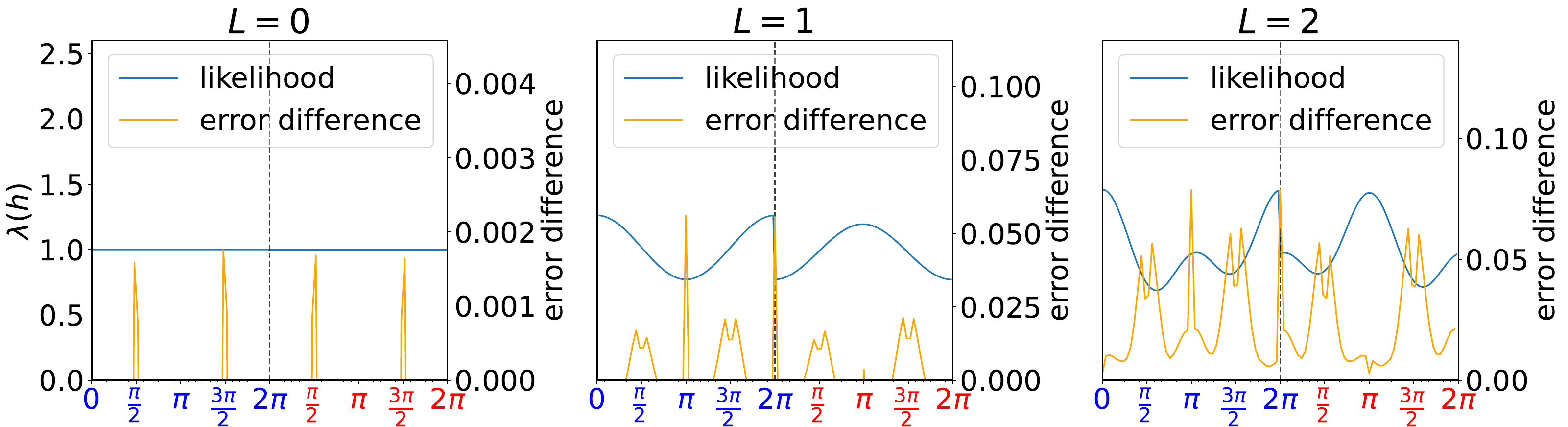}
        \hspace*{3cm}\includegraphics[width=\textwidth,trim={0cm 0cm 0cm 17.5cm}, clip]{figures/app_results/all_plots.pdf}
      \caption{Likelihoods of the fifth layer of our $O(2)$ P-SCNN trained on $O(2)$ \texttt{Double MNIST} using a bandlimit of $L=0$ through $L=4$. Note that the scale of the equivariance error varies between the plots.}
      \label{fig:bandlimit_1_and_2}
    \end{figure}

    \begin{wraptable}[22]{r}{8cm}\resizebox{8cm}{!}{\begin{tabular}{lc|cc}
    \toprule
     &  & \multicolumn{2}{c}{\textbf{Symmetries}} \\
    \begin{tabular}[c]{@{}c@{}}\textbf{Network}\\ \textbf{Group}\end{tabular} & \textbf{L} & $\bm{SO(2)}$ & $\bm{O(2)}$ \\ \midrule\midrule
    \multirow{5}{*}{$SO(2)$} & None & $0.474 \ $\myfontsize{$( 0.016)$} & $0.403 \ $\myfontsize{$( 0.021)$} \\ \cmidrule{2-4} 
     & 1 & $0.883 \ $\myfontsize{$( 0.007)$} & $0.794 \ $\myfontsize{$( 0.011)$} \\
     & 2 & $0.901 \ $\myfontsize{$( 0.005)$} & $\bm{\underline{0.823}} \ $\myfontsize{$( 0.005)$} \\
     & 3 & $\bm{\underline{0.908}} \ $\myfontsize{$( 0.006)$} & $0.821 \ $\myfontsize{$( 0.002)$} \\
     & 4 & $0.904 \ $\myfontsize{$( 0.004)$} & $0.820 \ $\myfontsize{$( 0.013)$} \\ \midrule
    \multirow{6}{*}{$O(2)$} & None & $0.391 \ $\myfontsize{$( 0.012)$} & $0.399 \ $\myfontsize{$( 0.014)$} \\ \cmidrule{2-4} 
     & 0 & $0.469 \ $\myfontsize{$( 0.010)$} & $0.402 \ $\myfontsize{$( 0.003)$} \\
     & 1 & $\underline{0.894} \ $\myfontsize{$( 0.011)$} & $0.780 \ $\myfontsize{$( 0.009)$} \\
     & 2 & $\underline{0.894} \ $\myfontsize{$( 0.004)$} & $\underline{0.819} \ $\myfontsize{$( 0.010)$} \\
     & 3 & $0.889 \ $\myfontsize{$( 0.013)$} & $0.817 \ $\myfontsize{$( 0.007)$} \\
     & 4 & $0.891 \ $\myfontsize{$( 0.006)$} & $\underline{0.819} \ $\myfontsize{$( 0.018)$}\\
     \bottomrule
    \end{tabular}}
    \caption{\texttt{Double MNIST} test accuracies using various levels of bandlimiting for our $SO(2)$ and $O(2)$ P-SCNNs. For each symmetry, the highest accuracy is \textbf{bold}, and the highest for each network group within this type of symmetry is \underline{underlined}. Standard deviations over 5 runs are denoted in parentheses.}
    \label{tab:double_bandlimit}  \end{wraptable}
    In previous experiments, we used a bandlimit of $L=2$ for the likelihood distributions of our partially equivariant models. By choosing a higher or lower level of bandlimiting, the flexibility of the likelihood distributions can be altered. In this section, we experiment with multiple levels of bandlimiting on \texttt{Double MNIST}, the subsets of \texttt{MedMNIST} and \texttt{Smoke}. We present the results in Tables~\ref{tab:double_bandlimit}~and~\ref{tab:bandlimit_med}. We also show the learnt likelihood distributions for the fifth layer of our $O(2)$ P-SCNN trained on $O(2)$ \texttt{Double MNIST} for a bandlimit of $L=0$ through $L=4$ in Figure~\ref{fig:bandlimit_1_and_2}. Furthermore, due to the rapid increase in complexity, the highest level of bandlimiting for $O(3)$ is set at $L=2$.
    
    \paragraph{\texttt{Double MNIST}}
    Table~\ref{tab:double_bandlimit} shows that \texttt{Double MNIST} with $SO(2)$ symmetries requires a bandlimit of $L\geq 1$, with generally minimal differences between higher levels of bandlimiting, and a moderate improvement from $L=1$ to $L=2$ for $H=SO(2)$. A bandlimit of $L=0$, which only allows reducing reflective equivariance, yields a performance improvement compared to the fully $O(2)$-invariant model, resulting in effectively the same performance as using a fully $SO(2)$-invariant model. However, it is significantly outperformed by higher levels of bandlimiting. 
    
    Under $O(2)$ symmetries, there is a more substantial performance benefit going from $L=1$ to $L=2$. Figure~\ref{fig:bandlimit_1_and_2} shows that, using a bandlimit of $L=1$, the model seems to struggle to describe low likelihoods for \(h=\textcolor{red}{\frac{\pi}{2}}\) and \(h=\textcolor{red}{\frac{3\pi}{2}}\), while maintaining a high likelihood for \(h=\textcolor{red}{\frac{\pi}{2}}\). This problem does not occur under $L=2$. Furthermore, a bandlimit of $L=0$ no longer yields a substantial improvement over a fully $O(2)$-invariant model, which is supported by the uniform likelihood distribution in Figure~\ref{fig:bandlimit_1_and_2}. 
    
    For both symmetries, we observe that a bandlimit of $L=4$ does not offer a substantial advantage over a bandlimit of $L=2$ in terms of performance. Furthermore, the corresponding learnt likelihood distribution in Figure~\ref{fig:bandlimit_1_and_2} shows a significantly more complex and, therefore, less easily interpretable likelihood distribution. 

    \paragraph{\texttt{MedMNIST}}
    The results in Table~\ref{tab:bandlimit_med} reveal that for \texttt{OrganMNIST3D} and \texttt{SynapseMNIST3D} a bandlimit of $L=2$ (or $L=3$ when available) obtains the highest performance. Using a bandlimit of $L=4$ yields a performance reduction of around 2 percentage points, indicating minor overfitting. Moreover, \texttt{NoduleMNIST3D} sees a minimal performance difference for the various levels of bandlimiting, with only a bandlimit of $L=0$ using $H=O(3)$ yielding a small performance improvement compared to the other models. This further indicates that this dataset is highly symmetric.
    \begin{table}[h!]
    \centering
    \begin{tabular}{lc|ccc}
    \toprule
    \begin{tabular}[c]{@{}c@{}}\textbf{Network}\\ \textbf{Group}\end{tabular} & \textbf{L} & \textbf{Synapse} & \textbf{Nodule} & \textbf{Organ} \\ \midrule\midrule
    \multirow{5}{*}{$SO(3)$} & None & $0.738 \ $\myfontsize{$( 0.009)$} & $\underline{0.873} \ $\myfontsize{$( 0.005)$} & $0.607 \ $\myfontsize{$( 0.006)$} \\ \cmidrule{2-5} 
     & 1 & $0.753 \ $\myfontsize{$( 0.018)$} & $\underline{0.873} \ $\myfontsize{$( 0.007)$} & $0.862 \ $\myfontsize{$( 0.012)$} \\
     & 2 & $0.770 \ $\myfontsize{$( 0.030)$} & $0.871 \ $\myfontsize{$( 0.001)$} & $\underline{0.902} \ $\myfontsize{$( 0.006)$} \\
     & 3 & $\bm{\underline{0.771}} \ $\myfontsize{$( 0.030)$} & $0.872 \ $\myfontsize{$( 0.010)$} & $\underline{0.902} \ $\myfontsize{$( 0.002)$} \\
     & 4 & $0.750 \ $\myfontsize{$( 0.006)$} & $0.867 \ $\myfontsize{$( 0.009)$} & $0.896 \ $\myfontsize{$( 0.003)$} \\ \midrule
    \multirow{4}{*}{$O(3)$} & None & $0.743 \ $\myfontsize{$( 0.004)$} & $0.868 \ $\myfontsize{$( 0.009)$} & $0.592 \ $\myfontsize{$( 0.008)$} \\ \cmidrule{2-5} 
     & 0 & $0.737 \ $\myfontsize{$( 0.008)$} & $\bm{\underline{0.879}} \ $\myfontsize{$( 0.009)$} & $0.636 \ $\myfontsize{$( 0.003)$} \\
     & 1 & $0.756 \ $\myfontsize{$( 0.022)$} & $0.872 \ $\myfontsize{$( 0.006)$} & $0.875 \ $\myfontsize{$( 0.005)$} \\
     & 2 & $\underline{0.769} \ $\myfontsize{$( 0.013)$} & $0.873 \ $\myfontsize{$( 0.002)$} & $\bm{\underline{0.905}} \ $\myfontsize{$( 0.004)$}\\
     \bottomrule
    \end{tabular}
    \caption{\texttt{MedMNIST} test accuracies using various levels of bandlimiting for our $SO(2)$ and $O(2)$ P-SCNNs. For each type of symmetry, the highest accuracy is \textbf{bold}, and the highest for each network group within this type of symmetry is \underline{underlined}. Standard deviations over 5 runs are denoted in parentheses.}
    \label{tab:bandlimit_med}
    \end{table}

    \paragraph{\texttt{Smoke}} While the main experiments on the \texttt{Smoke} dataset are performed using (partial) $C_4$ equivariance, which does not require band-limiting in our approach, we perform additional experiments experiments using $SO(2)$ equivariance using both our approach and the RSteer approach from~\citet{wangApprox} by changing the feature type to irrep fields up to frequency $2$. The results can be found in Tab.~\ref{tab:smoke_bandlimit_params}. Here we present the task-specific performance, as well as the number of parameters for each configuration. Additionally, we also report the number of parameters for a set-up using larger $5\times 5$ kernels instead of the default $3\times 3$. In terms task-performance we observe that for our approach, as long as $L>1$, the performance is nearly identical to RSteer for the various levels of band-limiting, while outperforming RPP. Furthermore, from the number of parameters we see that for the default set-up with $3\times 3$ kernels our approach without band-limiting ($L=6$) achieves a comparable number of parameters with RPP and RSteer, with noticeably lower parameter counts for $L\leq 2$. However, when using $5\times5$ kernels our approach uses significantly fewer parameters compared to both RPP and RSteer regardless of the band-limit $L$.

    \begin{table}[h!]
    \centering
    \begin{tabular}{cc|ccc|c}
     &  & \multicolumn{2}{c}{RMSE} & Params (M) using & Hypothetical params (M) \\
    \textbf{Model} & L & Future & Domain & default $3\times 3$ kernels & using $5\times 5$ kernels \\ \hline
    RPP & - & 0.81 \myfontsize{$( 0.01)$} & 0.70 \myfontsize{$( 0.04)$} & 6.43 & 17.24 \\
    RSteer & - & \textbf{0.78} \myfontsize{$( 0.01)$} & \textbf{0.58} \myfontsize{$( 0.00)$} & 5.57 & 27.75 \\ \hline
    \multirow{4}{*}{Ours} & 1 & 0.85 \myfontsize{$( 0.01)$} & 0.63 \myfontsize{$( 0.00)$} & 2.80 & 5.05 \\
     & 2 &\textbf{0.78} \myfontsize{$( 0.01)$} & \textbf{0.58} \myfontsize{$( 0.02)$} & 4.30 & 7.84 \\
     & 4 & 0.79 \myfontsize{$( 0.02)$} & 0.61 \myfontsize{$( 0.03)$} & 5.95 & 10.83 \\
     & 6 & \textbf{0.78} \myfontsize{$( 0.01)$} & 0.59 \myfontsize{$( 0.03)$} & 6.36 & 11.65
    \end{tabular}
    \caption{\texttt{Smoke} RMSE scores and parameter counts (in millions, M) comparing $SO(2)$ equivariant RSteer with our $SO(2)$-PSCNN using various levels of band-limiting. Note that a band-limit of $L=6$ equates to performing no band-limiting at all. Standard deviations over 5 runs are denoted in parentheses.}
    \label{tab:smoke_bandlimit_params}
    \end{table}    

    \textit{Overall, these results show that in addition to the computational benefits, performing a higher degree of bandlimiting can act as a regularisation method, potentially improving performance and improving the interpretability of the likelihood distributions. Furthermore, band-limiting is not necessarily required to obtain a reasonable parameter space compared to other partially equivariant approaches.}

    \subsection{Competitive Results on Image Classification Tasks}\label{sec:result_compet}
    In previous sections, all (partially) equivariant CNNs used a trivial output representation at the last convolution layer to ensure a structurally invariant architecture. This equivariance could only be broken by breaking the equivariance of the individual layers. Furthermore, while Fourier-based non-linearities allow for more complexity, we opted for Gated non-linearities to prevent models from breaking equivariance through the non-linearity.

In this section, we aim to obtain higher performance by removing these restrictions, in addition to providing insight into the behaviour of our approach when there are other potential sources of partial equivariance. To achieve this, rather than limiting the output of the final convolution layer to a trivial field, we use an irrep field consisting of irreps up to frequency 2. This allows the subsequent MLP unrestricted freedom in mixing the resulting irrep features. Furthermore, we replace the Gated non-linearity in all applicable models with a Fourier-based non-linearity. For the 2D groups we use the FourierELU non-linearity. As discussed in Section~\ref{sec:nonlinearites}, applying a Fourier-based activation function to 3D groups such as $SO(3)$ and $O(3)$ is computationally expensive. Therefore, we use quotient FourierELU non-linearities, with features over the quotient space $Q=SO(3)/SO(2)$ and $Q=O(3)/O(2)$, i.e. the $2$-sphere. Here, both subgroups $K\leq H$ are defined as rotations (and reflections) around the z-axis. Since these spaces are smaller than the respective $SO(3)$ and $O(3)$ groups, these features require fewer channels and allow for a more efficient architecture. As a result, this also enables us to experiment with higher levels of bandlimiting of the likelihood distribution for $O(3)$.

The results are summarised in Table~\ref{tab:qfourier}, covering both \texttt{Double MNIST} with \( O(2) \) symmetries and \texttt{OrganMNIST3D}. We observe noteworthy improvements in fully equivariant SCNNs across both datasets as a result of our modifications. Specifically, our approach yields minor improvements for \( O(3) \) on \texttt{OrganMNIST3D} and \( SO(2) \) on \texttt{Double MNIST}. Moreover, more substantial performance gains are observed for \( SO(3) \) and \( O(2) \) on \texttt{OrganMNIST3D}, as well as for \( SO(2) \) on \texttt{Double MNIST}. Interestingly, all RPP models exhibit diminished performance after these modifications, particularly in the case of \texttt{Double MNIST}. This could potentially be attributed to the inherently unconstrained nature of RPP's residual CNN connections that already allow a high degree of flexibility in the base configuration. Consequently, our \( SO(3) \) P-SCNN with \( L=4 \) now matches the performance of the baseline RPP featuring a structurally invariant mapping and Gated non-linearity, while our \( O(2) \) P-SCNN with \( L=3 \) extends its lead over RPP models.

\begin{table}[h!]
\centering
\begin{tabular}{lcc|cccc}
\toprule
\begin{tabular}[c]{@{}c@{}}\textbf{Network}\\ \textbf{Group}\end{tabular} & \begin{tabular}[c]{@{}c@{}}\textbf{Partial}\\ \textbf{Equivariance}\end{tabular} & \textbf{L} & \multicolumn{2}{c}{\textbf{\texttt{OrganMNIST3D}}} & \multicolumn{2}{c}{\begin{tabular}[c]{@{}c@{}}\textbf{\texttt{Double MNIST} with}\\ $\bm{O(2)}$ \textbf{Symmetries}\end{tabular}}  \\ 
& & & \textbf{FourierELU} & \textbf{Gated} & \textbf{FourierELU} & \textbf{Gated} \\\midrule\midrule
CNN & N/A & N/A & \multicolumn{2}{c}{$\underline{0.921} \ $\myfontsize{$( 0.003)$}} & \multicolumn{2}{c}{$\underline{0.649} \ $\myfontsize{$( 0.019)$}}  \\ \midrule
\multirow{5}{*}{$SO(n)$} & None & N/A & $0.879 \ $\myfontsize{$( 0.007)$}  & $0.607 \ $\myfontsize{$( 0.006)$}  & $0.842 \ $\myfontsize{$( 0.007)$} & $0.403 \ $\myfontsize{$( 0.021)$} \\
 & RPP & N/A & $0.930 \ $\myfontsize{$( 0.011)$} & $\underline{0.936} \ $\myfontsize{$( 0.002)$}  & $0.617 \ $\myfontsize{$( 0.043)$} & $0.779 \ $\myfontsize{$( 0.021)$} \\ \cmidrule{2-7} 
 & \multirow{3}{*}{Ours} & 2 & $0.935 \ $\myfontsize{$( 0.003)$} & $0.902 \ $\myfontsize{$( 0.006)$} & $0.852 \ $\myfontsize{$( 0.009)$} & $\bm{\underline{0.823}} \ $\myfontsize{$( 0.005)$} \\
 &  & 3 & $0.932 \ $\myfontsize{$( 0.003)$} & $0.902 \ $\myfontsize{$( 0.002)$} &  $0.853 \ $\myfontsize{$( 0.016)$} & $0.821 \ $\myfontsize{$( 0.002)$} \\
 &  & 4 & $\bm{\underline{0.941}} \ $\myfontsize{$( 0.007)$} & $0.896 \ $\myfontsize{$( 0.003)$} & $\underline{0.855} \ $\myfontsize{$( 0.004)$} & $0.820 \ $\myfontsize{$( 0.013)$} \\ \midrule
\multirow{6}{*}{$O(n)$} & None & N/A & $0.821 \ $\myfontsize{$( 0.005)$} & $0.592 \ $\myfontsize{$( 0.008)$} & $0.860 \ $\myfontsize{$( 0.005)$} & $0.399 \ $\myfontsize{$( 0.014)$} \\
 & RPP & N/A & $\underline{0.936} \ $\myfontsize{$( 0.004)$} & $\bm{\underline{0.940}} \ $\myfontsize{$( 0.006)$} & $0.677 \ $\myfontsize{$( 0.037)$} &  $0.772 \ $\myfontsize{$( 0.019)$}\\ \cmidrule{2-7} 
 & \multirow{3}{*}{Ours} & 2 & $0.911 \ $\myfontsize{$( 0.007)$} & $0.905 \ $\myfontsize{$( 0.004)$} & $0.869 \ $\myfontsize{$( 0.005)$} & $\underline{0.819} \ $\myfontsize{$( 0.010)$} \\
 &  & 3 & $0.920 \ $\myfontsize{$( 0.008)$} & - & $\bm{\underline{0.885}} \ $\myfontsize{$( 0.003)$} &  $0.817 \ $\myfontsize{$( 0.007)$} \\
 &  & 4 & $0.911 \ $\myfontsize{$( 0.003)$} & - & $0.876 \ $\myfontsize{$( 0.006)$} &  $\underline{0.819} \ $\myfontsize{$( 0.018)$}\\
 \bottomrule
\end{tabular}
\caption{Test accuracies on \texttt{OrganMNIST3D} and \texttt{DoubleMNIST} comparing the performance of our baseline configurations (Gated) with the structurally non-invariant configurations using a Fourier based non-linearity. For each column, \textbf{bold} indicates the highest accuracy and \underline{underline} denotes the highest accuracy for the given network group. Standard deviations over 5 runs are denoted in parentheses.}
\label{tab:qfourier}
\end{table}

\end{document}